%% file: main.tex
\documentclass[letterpaper, 10 pt, journal, twoside]{IEEEtran}

\ifCLASSINFOpdf
\else
\fi

\usepackage{graphics} 
\usepackage{epsfig} 
\usepackage{times} 
\usepackage{amsmath} 
\usepackage{amssymb}  

\newcommand{\Ours}{DGFusion}

\usepackage{multirow}

\newcommand{\PAR}[1]{\noindent{\bf #1}}

\usepackage{pgfplots}
\usepackage{tikz}
\pgfplotsset{width=10cm,compat=1.8}
\usetikzlibrary{pgfplots.polar}
\usepackage{pgfplotstable}

\usepackage{subcaption}
\usepackage{booktabs}

\makeatletter
\let\NAT@parse\undefined
\makeatother

\usepackage[pagebackref=true,breaklinks=true,colorlinks,bookmarks=false]{hyperref}

\usepackage[capitalize]{cleveref}
\crefname{section}{Sec.}{Secs.}
\Crefname{section}{Section}{Sections}
\Crefname{table}{Table}{Tables}
\crefname{table}{Tab.}{Tabs.}

\usepackage{enumitem}
\setlist[enumerate]{leftmargin=*, labelsep=0.3em, labelwidth=1em, itemsep=0em, parsep=0pt, topsep=0.5ex}
\setlist[itemize]{leftmargin=*, labelsep=0.3em, labelwidth=1em, itemsep=0em, parsep=0pt, topsep=0.5ex}

\begin{document}
\title{\Ours{}: Depth-Guided Sensor Fusion\\
for Robust Semantic Perception}

\author{
Tim Br\"odermannn$^{1}$,
Christos Sakaridis$^{1}$,
Luigi Piccinelli$^{1}$,
Wim Abbeloos$^{2}$, and Luc Van Gool$^{1,3}$%
\thanks{This paper has been accepted for publication in \emph{IEEE Robotics and Automation Letters}.
The final published version is available at https://doi.org/10.1109/LRA.2026.3656789}
\thanks{© 2024 IEEE.  Personal use of this material is permitted.  Permission from IEEE must be obtained for all other uses, in any current or future media, including reprinting/republishing this material for advertising or promotional purposes, creating new collective works, for resale or redistribution to servers or lists, or reuse of any copyrighted component of this work in other works.
This work was supported by Toyota Motor Europe via the research project TRACE-Z\"urich.
}
\thanks{$^{1}$Tim Br{\"o}dermann, Christos Sakaridis, Luigi Piccinelli, and Luc Van Gool are with Computer Vision Laboratory, ETH Zurich, 8057 Zurich, Switzerland  {\tt\footnotesize\{timbr, csakarid, lpiccinelli, vangool\}@ethz.ch}.
$^{2}$ Wim Abbeloos is with Toyota Motor Europe, {\tt\footnotesize\{wim.abbeloos\}@toyota-europe.com}.
$^{3}$Luc Van Gool is with INSAIT, Sofia University St.\ Kliment Ohridski, Bulgaria, {\tt\footnotesize\{luc.vangool\}@insait.ai}.
}%
\thanks{Digital Object Identifier (DOI): 10.1109/LRA.2026.3656789}
}

\markboth{IEEE Robotics and Automation Letters. Preprint Version. Accepted January, 2026}
{Br{\"o}dermann \MakeLowercase{\textit{et al.}}: DGFusion: Depth-Guided Sensor Fusion}

\maketitle

\begin{abstract}
Robust semantic perception for autonomous vehicles relies on effectively combining multiple sensors with complementary strengths and weaknesses. State-of-the-art sensor fusion approaches to semantic perception often treat sensor data uniformly across the spatial extent of the input, which hinders performance when faced with challenging conditions. By contrast, we propose a novel depth-guided multimodal fusion method that upgrades condition-aware fusion by integrating depth information. Our network, \Ours{}, poses multimodal segmentation as a multi-task problem, utilizing the lidar measurements, which are typically available in outdoor sensor suites, both as one of the model's inputs and as ground truth for learning depth. 
Our corresponding auxiliary depth head helps to learn depth-aware features, which are encoded into spatially varying local depth tokens that condition our attentive cross-modal fusion.
Together with a global condition token, these local depth tokens dynamically adapt sensor fusion to the spatially varying reliability of each sensor across the scene, which largely depends on depth.
In addition, we propose a robust loss for our depth, which is essential for learning from lidar inputs that are typically sparse and noisy in adverse conditions.
Our method achieves state-of-the-art panoptic and semantic segmentation performance on the challenging MUSES and DeLiVER datasets. 
Code and models are available at \url{https://github.com/timbroed/DGFusion}
\end{abstract}


\IEEEpeerreviewmaketitle

\section{Introduction}
\label{sec:intro}

\IEEEPARstart{R}{obust}
segmentation for automated driving systems demands reliable sensor fusion under any environmental condition. Conventional semantic perception pipelines that rely on a single sensor struggle in domains such as fog, rain, or low light, where the performance of individual sensors degrades. Prior work~\cite{cafuser} has shown that conditioning cross-modal fusion on a \emph{global} Condition Token (CT) that encodes the scene-wide environmental condition can effectively guide fusion by dynamically adapting sensor contributions. 
However, while this global approach models the overall scene condition, it overlooks the spatial variability of the effect of this condition on the sensor measurements, which largely depends on the local depth.

\begin{figure}[t]
  \centering
  \includegraphics[width=\linewidth,
  ]{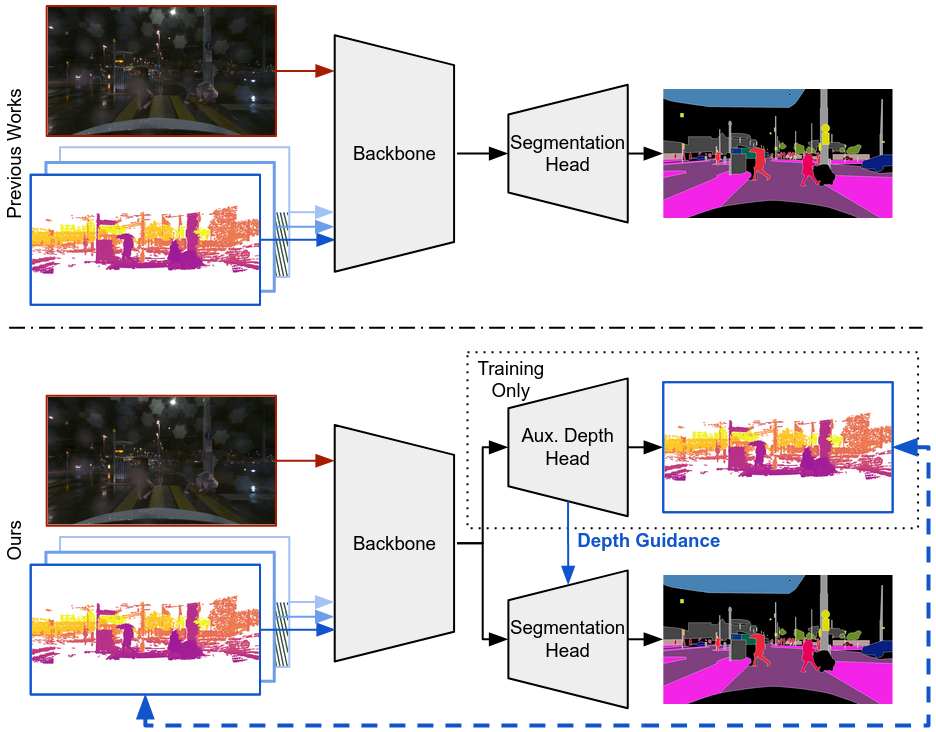}
  \caption{\textbf{Intuition on \Ours{}}. Unlike previous sensor fusion works that use lidar only as an input, we additionally utilize this readily available modality for depth supervision to create a multi-tasking setup, hinging on the well-known benefits of depth estimation for semantic perception.
  }
  \label{fig:teaser}
\end{figure}

In practice, depth is a key factor affecting the \emph{variable reliability} of each sensor in each part of the scene. For instance, lidars provide accurate returns for nearby objects but degrade rapidly at longer distances, e.g.\ in fog or snowfall~\cite{hahner2021fog,hahner2022lidar}. Radars, on the other hand, operate well at long distances even in adverse weather but suffer from strong noise due to multi-path effects, whereas cameras are reliable only in well-lit parts of the scene, which are in turn related to the 3D structure of the scene. In other words, environmental conditions introduce spatial variations in the signal-to-noise ratio of each sensor, which call for a fine-grained local adaptivity of the fusion strategy.
Consequently, incorporating explicit depth cues into the fusion mechanism for semantic perception could enable such spatial adaptivity to sensor reliability, which would in turn enhance segmentation robustness in challenging environments. This hypothesis is reinforced by several works on RGB-only semantics and depth multi-tasking~\cite{bruggemann2021exploring,rahman2024semi}, which improve segmentation performance over semantics-only learning. Despite this promise for depth-informed fusion, most state-of-the-art multi-sensor segmentation methods~\cite{broedermann2023hrfuser,zhang2023delivering,li2024stitchfusion,jia2024geminifusion} treat lidar solely as an \emph{input} modality, overlooking 
its potential for improving segmentation via multi-tasking by leveraging 
its pixel-aligned sparse depth measurements as 
\emph{ground truth} for learning depth together with semantics.

We fill in this gap and propose \Ours{}, a depth-guided multimodal fusion network which upgrades condition-aware sensor fusion for semantic perception with explicit local depth guidance (cf.~\cref{fig:teaser}).
Building on CAFuser~\cite{cafuser}, we adopt its shared backbone with modality-specific feature adapters and its global CT that encodes scene-wide environmental conditions.
Beyond this, in \Ours{} we introduce an \emph{auxiliary depth prediction head}
that leverages lidar's pixel-aligned sparse depth measurements as noisy supervision 
to enrich local features with implicit depth cues to guide each local-window cross-attention fusion. This dual encoding of global environmental conditions and spatially-varying depth cues enables the network to make more precise decisions based on which sensor to rely more on for each feature. By incorporating depth cues, the fusion mechanism can adaptively adjust the weighting of each modality per image region, favoring the per-case more reliable sensor. Moreover, since these noisy depth measurements are available as one of our network's inputs, i.e.\ the lidar, depth supervision comes at \emph{no extra cost} compared to existing fusion pipelines. The synergy between the global condition and local depth representation is especially beneficial in adverse conditions, which have been found~\cite{seeing_through_fog} to cause pronounced spatial variations in sensor reliability.

Besides, training our model on adverse conditions poses additional challenges. 
For example, in fog or snowfall the lidar measurements, which 
provide pixel-aligned depth supervision, become inherently noisy~\cite{hahner2021fog,hahner2022lidar}, making the direct integration of standard depth estimation approaches and losses in our multi-tasking model infeasible. 
In order to get a reliable training signal across all conditions, we carefully design our depth loss to be robust to sparse and noisy ground truth by combining an outlier-robust $L_1$ loss with strategies commonly used in monocular depth estimation. These strategies consist of an edge-aware smoothness loss to penalize depth gradients in uniform regions of the RGB image and a novel panoptic-edge-aware smoothness loss that better captures depth discontinuities between instances of the same semantic classes compared to previous semantic-edge-aware smoothness losses.

Casting the \emph{training} of \Ours{} as a multi-tasking problem leverages the synergy between depth and semantics with minimal changes to the \emph{inference} pipeline. Extensive experiments on MUSES~\cite{MUSES} and DeLiVER~\cite{zhang2023delivering} demonstrate that \Ours{} sets the new state of the art in multi-sensor semantic perception across diverse conditions.

Our key contributions can be summarized as follows:
\begin{itemize}
    \item  We propose \Ours{}, a novel depth-guided multi-sensor \emph{fusion} network for semantic perception that reformulates the problem \emph{as multi-tasking of semantics and depth}, at no extra annotation or inference cost.
    \item We leverage the internal depth features 
    to perform a spatially-variant conditioning of our cross-modal attentive fusion on local depth tokens. Combined with a global CT, this makes our fusion \emph{locally adaptive to} the effect of this \emph{condition} on the sensor at the respective feature location.
    \item We design an \emph{outlier-resistant depth loss} on sparse, noisy lidar ground truth, incorporating edge-aware and panoptic-aware smoothness priors for reliable training. 
    \item We extensively validate and ablate \Ours{} on two widely used outdoor datasets, MUSES and DeLiVER, and set the new state of the art on multi-sensor segmentation.
\end{itemize}

\begin{figure*}[t]
  \vspace{0.5em}
  \centering
  \includegraphics[width=\textwidth,
  ]{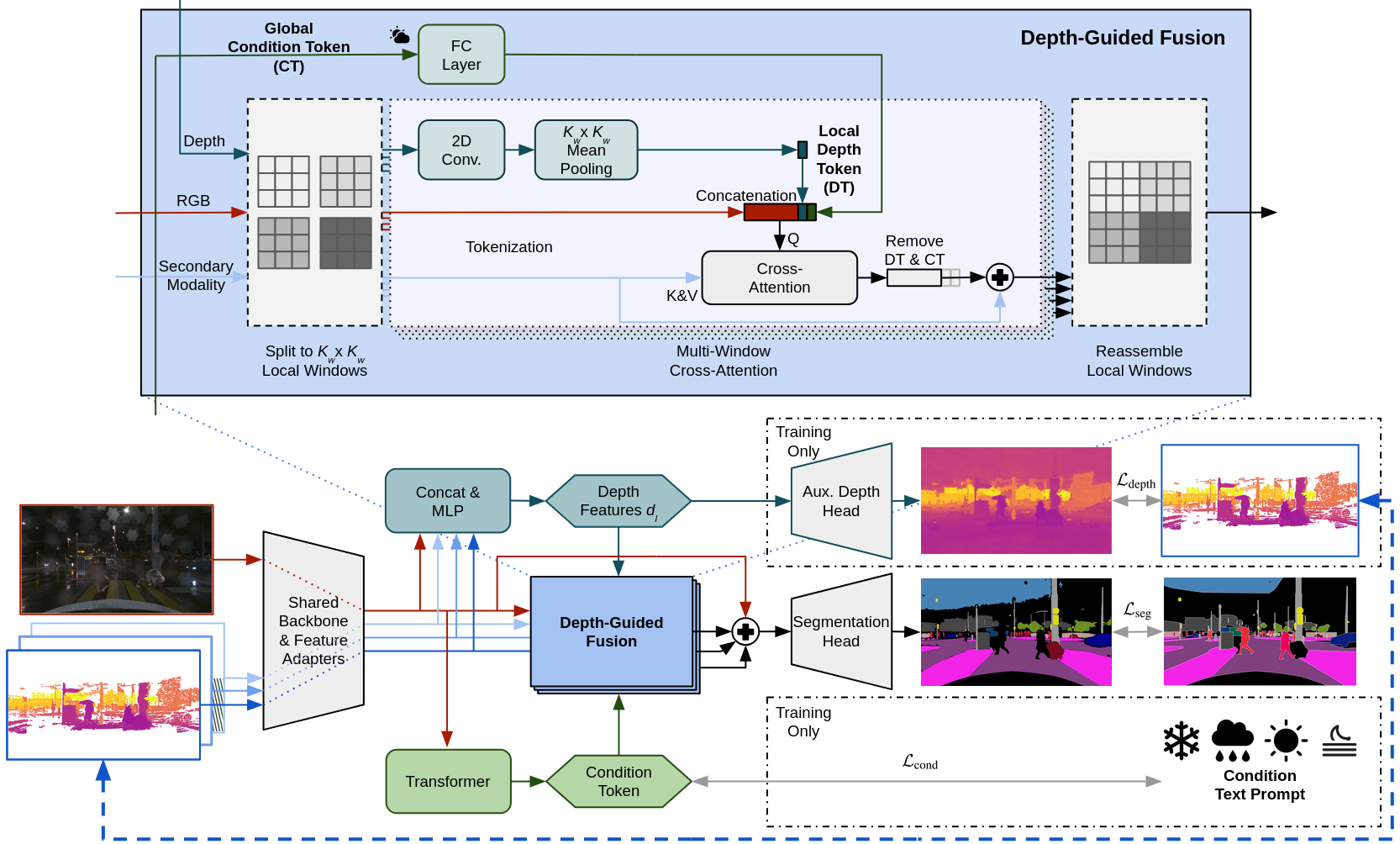}
  \caption{\textbf{\Ours{} overview}.
  We process all input modalities
  with a shared backbone with individual feature adapters. Their outputs are split into three branches: the depth estimation branch at the top, the segmentation branch in the middle, and the condition representation branch at the bottom. In our multi-task setup, the sparse and noisy lidar serves as supervision for the auxiliary depth head, enabling the network to learn depth-informed features that improve semantic representations. 
  Both the depth and the condition branch send additional features into the Depth-Guided Fusion modules. In these modules, features from the RGB input, the respective secondary modality, and the depth are divided into local windows. Each depth window is processed to extract a local Depth Token (DT), which is concatenated with the RGB tokens and the Condition Token (CT) to form the set of queries for cross-attention. After fusion, the DT and CT are removed, and the windows are reassembled, yielding enriched features that are fed to the segmentation head to produce the final segmentation prediction.
  }
  \label{fig:overview}
\end{figure*}

\section{Related Work}
\label{sec:related}

\noindent{}\textbf{Semantic perception} encompasses tasks like semantic~\cite{Cityscapes}, panoptic~\cite{kirillov2019panoptic}, and instance~\cite{he2017mask} segmentation, which are pivotal for robotics applications such as automated driving~\cite{does:vision:matter:for:action}. Recent mask-based networks have advanced the state of the art~\cite{li2023mask}, and unifying these tasks within a single model has proven beneficial~\cite{jain2023oneformer}. In contrast to these RGB-only methods, our work focuses on \emph{multimodal} semantic perception, leveraging sensor fusion to enhance robustness in challenging conditions where RGB data is unreliable.

\noindent{}\textbf{Sensor fusion} is central in practical vision applications, as different sensors feature complementary characteristics. The availability of large multimodal datasets, from early ones like KITTI~\cite{kitti} to recent ones focusing on adverse conditions~\cite{zhang2023delivering,seeing_through_fog,MUSES}, has driven the development of fusion methods. Architectures have evolved from two-modality systems~\cite{chaturvedi_pay} to modular, attention-based networks that handle an arbitrary number of inputs~\cite{broedermann2023hrfuser,zhang2023delivering}. Recent approaches have integrated large-scale pre-trained models~\cite{li2024stitchfusion}, explored modality-agnostic designs with shared backbones~\cite{centeringthevalue}, and combined intra-modal with inter-modal attention~\cite{jia2024geminifusion}. The MUSES method~\cite{MUSES} extends~\cite{cheng2022masked} to the multimodal setting, solving panoptic and semantic segmentation via local-window cross-attention. However, all these works fuse sensors \emph{uniformly}, lacking explicit adaptation to the conditions that affect each image region \emph{non-uniformly}, largely depending on \emph{depth}.

\noindent{}\textbf{Condition-aware fusion} utilizes information about the environmental conditions to guide perception. \cite{chaturvedi_pay} addresses object detection by implicitly assigning larger weights to the modality that yields better detection features at late fusion. Knowledge distillation from vision-language models is used in~\cite{learningmodalityagnostic} for learning modality-agnostic representations. The most closely related work to ours in this regard is~\cite{cafuser}, which guides cross-modal attention with a \emph{global} camera-based CT that is learned contrastively based on training-time supervision with image-level verbal condition descriptions.
In contrast, we aim at a more fine-grained, \emph{pixel-level adaptivity} to the environmental condition, as its effects on sensor data vary non-uniformly with \emph{depth}.

\noindent{}\textbf{Semantics and depth} are strongly correlated tasks, and it is well-established that they benefit from being learned jointly in a multi-task setup~\cite{bruggemann2021exploring,rahman2024semi}; cf.~\cite{vandenhende2021multitask} for a comprehensive overview of such semantics-depth multi-tasking methods.
Prior work has leveraged this synergy, with semantics guiding monocular depth estimation~\cite{chen2019towards,zhu2020edge}. Similarly, depth supervision has benefited from edge-aware smoothness losses based on image gradients~\cite{godard2017unsupervised} and semantics-aware losses that use semantic annotations~\cite{li2024semantic}, which we extend to use more fine-grained \emph{panoptic} annotations.
In contrast to previous works, we explore the use of \emph{depth} in \emph{multimodal} settings as a readily available \emph{input} signal, e.g.\ from sparse lidar measurements in typical outdoor multi-sensor suites, in order to benefit semantics. While the aforementioned multi-sensor semantic perception works utilize such depth information only as input to the fusion architecture, we instead leverage it as a reference \emph{output} during training, reframing sensor fusion as a multi-task problem where depth completion serves as an \emph{auxiliary} task to learn informative, localized features for semantic perception. The depth features learned from this auxiliary task are leveraged as a \textit{spatially varying} conditioning in our attentive fusion together with global condition information. This combination allows \Ours{} to adaptively weigh modalities not only per-input, but also \emph{per-region}, based on how depth modulates the effect of the environmental condition on each sensor.

\section{\Ours}
\label{sec:method}

In this section, we present \Ours{}, our depth-guided sensor fusion network for robust segmentation. We first describe our architecture (\cref{sub:sec:architecture}), which integrates an auxiliary depth head and a depth-guided fusion module. Next, we detail our multi-task loss design (\cref{sub:sec:loss:design}) that improves depth supervision with noisy lidar ground truth. Finally, we outline key implementation details (\cref{subsec:implementation}).

\subsection{Architecture}
\label{sub:sec:architecture}

Our goal is to achieve robust multimodal segmentation in all weather conditions.
We build on CAFuser~\cite{cafuser}, which processes all modalities through a shared backbone with lightweight modality-specific feature adapters to align heterogeneous inputs into a common feature space, and uses global environmental condition encodings to guide the sensor fusion for the OneFormer~\cite{jain2023oneformer} segmentation head.
Following CAFuser, we project the secondary modalities (lidar, radar, and event camera) as 3-channel images onto the RGB camera plane and dilate them with a kernel of size $K_\text{sensor}$ to compensate for sparsity. We adopt CAFuser's shared backbone with feature adapters and its global condition token (CT).
In contrast to CAFuser, we treat lidar not only as an input modality but also as sparse, noisy ground-truth depth for multi-task learning, which poses additional challenges in adverse conditions where lidar measurements degrade.
With our carefully designed robust losses (cf.~\cref{sub:sec:loss:design}), this enables our network to learn depth-informed features that provide additional supervision for semantics and enable \emph{locally adaptive} depth-guided fusion. 

In \cref{fig:overview}, our complete \Ours{} architecture is depicted. On the left, an arbitrary number of sensor modalities are processed by the shared backbone with modality-specific feature adapters, resulting in per-level feature maps. These features are then split into three branches: the depth estimation branch (top), the segmentation branch (center), and the condition representation branch adopted from CAFuser (bottom). The depth branch uses raw, undilated lidar ground truth to supervise depth predictions and induce depth-informed features via a lightweight fusion module, while the condition branch generates a global CT from the RGB input. Both branches provide extra cues to the central depth-guided fusion module.

\PAR{Depth estimation branch.} 
At each feature level $l \in \{1,2,3,4\}$, we fuse all modalities with a light-weight fusion module to produce a ``depth'' feature pyramid $\mathbf{d}_l$. Given RGB features $\mathbf{F}_{\text{rgb}}^l$ and secondary modality features $\mathbf{F}_{\text{sec}}^{l,m}$ for modality $m \in \{1, \ldots, M\}$, we compute
\begin{equation}
    \mathbf{d}_l = \text{MLP}([\mathbf{F}_{\text{rgb}}^l, \mathbf{F}_{\text{sec}}^{l,1}, \ldots, \mathbf{F}_{\text{sec}}^{l,M}]) + \mathbf{F}_{\text{rgb}}^l,
\end{equation}
where $[\cdot]$ denotes concatenation and MLP is a 2-layer perceptron with a bottleneck. The residual connection from RGB features ensures stable gradient flow.
The resulting depth features are passed to our auxiliary depth head and supervised by our depth loss described in~\cref{sub:sec:loss:design}.
Note that we use \textit{all} input modalities to exploit all available information and to fully utilize their complementary strengths.

We use a Semantic FPN~\cite{kirillov2019panoptic} decoder as our auxiliary depth head, which upsamples each depth feature $\mathbf{d}_l$ to 1/4 resolution and aggregates them:
\begin{equation}
    \hat{D} = \sum_{l=1}^{4} \text{Upsample}(\mathbf{d}_l),
\end{equation}
where Upsample consists of multiple convolutions and bilinear upsampling steps. This structure fuses coarse and fine information efficiently, enabling robust depth estimation.
At inference, we drop the auxiliary depth head but keep the depth features $\mathbf{d}_l$ for guiding fusion in the semantic branch.

\PAR{Condition representation branch.}
We adopt CAFuser's~\cite{cafuser} strategy to produce a global CT. We flatten the highest-level RGB features and process them with a lightweight Transformer composed of 2 encoder and 2 decoder layers:
\begin{equation}
    \mathbf{t}_c = \text{Transformer}(\text{Flatten}(\mathbf{F}_{\text{rgb}}^4)),
\end{equation}
supervised during training with a verbo-visual contrastive loss~\cite{cafuser,jain2023oneformer,bruggemann2023contrastive} that leverages text prompts describing the environmental conditions in detail.

\PAR{Depth-guided fusion.}
We use the depth features $\mathbf{d}_l$ to guide sensor fusion in our main semantic branch. Our Depth-Guided Fusion module utilizes parallel multi-window cross-attention fusion~\cite{broedermann2023hrfuser} to fuse in parallel each secondary modality via cross-attention with the primary camera modality within $K_w\times K_w$ local windows. This fusion design is efficient and scales well with the number of input modalities.
For each level $l$ and secondary modality $m$, we partition RGB features $\mathbf{F}_{\text{rgb}}^l$, secondary features $\mathbf{F}_{\text{sec}}^{l,m}$, and depth features $\mathbf{d}_l$ into $I_l$ windows of size $K_w\times K_w$. For each window $i \in \{1, \ldots, I_l\}$, we extract a \emph{local Depth Token (DT)}:
\begin{equation}
    \mathbf{t}_d^{l,i} = \text{Pool}_{\text{mean}}(\text{Conv}(\mathbf{d}_l^i)),
\end{equation}
which encodes local depth context.
We map the CT to the feature dimensionality of level $l$ with a fully connected layer:
\begin{equation}
    \mathbf{t}_c^l = \text{FC}(\mathbf{t}_c).
\end{equation}
We then form query features by concatenating RGB tokens with the global CT $\mathbf{t}_c^l$ and the local DTs $\mathbf{t}_d^{l,i}$ into $\mathbf{F}_q^{l,i} = [\mathbf{F}_{\text{rgb}}^{l,i}, \mathbf{t}_c^l, \mathbf{t}_d^{l,i}]$. Depth-guided fusion is performed locally by letting the condition- and depth-aware RGB features cross-attend to each secondary modality $m$ in window $i$:
\begin{equation}
    \label{eq:dgfusion:ca}
    \mathbf{F}_{\text{fused}}^{l,m,i} = \mathbf{F}_q^{l,i} + \text{Softmax}(\mathbf{Q}_{l,i}\mathbf{K}_{l,m,i}^T)\mathbf{V}_{l,m,i},
\end{equation}
where the queries are computed as linear projections of the enriched RGB features $\mathbf{Q}_{l,i} = \mathbf{W}_q^{l,i}\mathbf{F}_q^{l,i}$, and respectively for the keys and values and the secondary modality features via $\mathbf{K}_{l,m,i} = \mathbf{W}_k^{l,m,i}\mathbf{F}_{\text{sec}}^{l,m,i}$ and $\mathbf{V}_{l,m,i} = \mathbf{W}_v^{l,m,i}\mathbf{F}_{\text{sec}}^{l,m,i}$.
After the cross-attention layer of \eqref{eq:dgfusion:ca}, we remove the last two tokens from $\mathbf{F}_{\text{fused}}^{l,m,i}$, which correspond to the CT and the local DT, thus aligning the spatial dimensions of the resulting features $\hat{\mathbf{F}}_{\text{fused}}^{l,m,i}$ back to those of $\mathbf{F}_{\text{rgb}}^{l,i}$. We reassemble windows and sum over all modalities with a residual connection:
\begin{equation}
    \mathbf{F}_{\text{out}}^l = \sum_{m=1}^{M} \text{Reassemble}\left(\{\hat{\mathbf{F}}_{\text{fused}}^{l,m,i}\}_{i=1}^{I_l}\right) + \mathbf{F}_{rgb}^l.
\end{equation}
The resulting feature pyramid $\{\mathbf{F}_{\text{out}}^l\}_{l=1}^4$ is passed to the segmentation head, which outputs the panoptic prediction.

This design ensures that the predicted depth assists the multi-sensor semantic fusion while imposing minimal changes to the overall architecture. Crucially, the depth head is only active at training time, allowing the network to learn richer feature representations via multi-task supervision. At inference, we do not require the depth head, as the Depth-Guided Fusion module only needs the depth features that are computed from the sensor inputs. Each local-window cross-attentive fusion is \emph{jointly} guided both by the global condition context encoded in the CT and the localized depth cues encoded in the respective local DT, which together model well the local effect of the present condition on the reliability of each sensor and adapt fusion properly.

\subsection{Losses}
\label{sub:sec:loss:design}

Our training pipeline casts semantic perception and depth estimation as a multi-task problem by utilizing the raw and undilated lidar projections as ground truth for the auxiliary depth head. We need to carefully design the loss to obtain effective supervision of the depth head, as this ground truth is both sparse and noisy, especially in adverse weather.
Specifically for the depth branch, we combine three weighted loss terms into a single objective: a robust $L_1$ term, an RGB-edge-aware smoothness term, and a novel panoptic-edge-aware smoothness loss.

\PAR{Outlier-robust $L_1$ loss.}
Lidar returns degrade in adverse conditions, making the raw depth measurements noisy and unreliable. To address this, we compute an $L_1$ loss on logarithmic depth values. The $L_1$ loss is inherently more robust to outliers than $L_2$, and using the logarithmic scale better handles depth variations across orders of magnitude~\cite{rahman2024semi,piccinelli2024unidepth,piccinelli2025unidepthv2}. 
Let $\hat{D}_p$ be the predicted depth at pixel $p$ and $D_p$ the corresponding raw lidar reference depth.
Since raw lidar depths are often severely corrupted in adverse conditions, directly supervising depth predictions using all available measurements can degrade learning (cf.~\cref{sub:sec:ablations}). To address this, we restrict the loss computation to a more reliable subset of pixels, i.e.\ those pixels below the $\tau$-quantile of depth errors.
We first define the absolute log-error at pixel $p$ as
\begin{equation}
    r_p = \left|\log\left(\hat{D}_p\right) - \log\left(D_p\right)\right|.
\end{equation}
We define the set of valid pixels used for supervision as
\begin{equation}
    P_{\tau} = \left\{ p \in P_l: r_p \leq \operatorname{Quantile}_{\tau}\bigl(\{r_q : q \in P_l\}\bigr) \right\},
\end{equation}
where $P_l$ denotes the set of pixels containing a lidar measurement and $\operatorname{Quantile}_{\tau}$ returns the $\tau$-quantile of the error values. Our outlier-robust $L_1$ loss is then computed as
\begin{equation}
    \mathcal{L}_{\text{logL1}} ~=~ \frac{1}{|P_{\tau}|}\textstyle\sum_{p\in P_{\tau}} r_p.
\end{equation}
This ``$\tau$-filtering'' requires no additional preprocessing and renders the training robust to lidar noise, as our network leverages the noisy lidar measurements both as input for the fusion and reference for the depth.

\PAR{Edge-aware smoothness loss.}
To guide depth prediction in regions lacking reliable lidar supervision, we employ an edge-aware smoothness term popular in monocular depth estimation~\cite{godard2017unsupervised}. Specifically, we weigh depth gradients with the local intensity gradients in the RGB image, thereby encouraging smooth depth predictions while preserving discontinuities at actual object boundaries:
\begin{equation}
\label{eq:imsmooth}
    \mathcal{L}_{\text{es}}
    =\!
    \frac{1}{|P|} \sum_{p \in P}
    \left(\bigl|\partial_x \hat{D}_p\bigr|
    \,e^{-\left | \partial_x I_p\right |}
    ~+~
    \bigl|\partial_y \hat{D}_p\bigr|
    \,e^{-\left | \partial_y I_p\right |}\right),
\end{equation}
where $I$ is the image intensity, $\partial_x \hat{D}$ and $\partial_x I$ are approximate derivatives in the $x$-direction (resp.\ for $y$), and $P$ is the set of all pixels in the RGB image. This formulation penalizes large depth gradients in regions with homogeneous intensity while permitting sharper transitions where the intensities vary sharply. As our method uses lidar both as input and output, $\mathcal{L}_{\text{es}}$ ensures that we do not learn only an identity mapping from lidar to depth. Instead, we encourage the full depth map to remain geometrically regular, i.e.\ smooth, where no lidar supervision is available, thus forcing the network to predict plausible depths across the image and 
to
learn better 
features for guiding our fusion.

\PAR{Panoptic-edge-aware smoothness loss.}
In our multi-task setup, we leverage ground-truth panoptic segmentation labels to guide depth estimation. In depth maps, sharp discontinuities often coincide with object boundaries, which include both semantic class boundaries and instance-level boundaries within the same class. Previous semantic-aware smoothness losses~\cite{chen2019towards} only handle class transitions, e.g.\ between ``road'' and ``car'', while several scenes also contain multiple adjacent objects of the same class, e.g.\ two adjacent ``cars'', whose shared boundaries are equally informative about depth transitions. To overcome this limitation, we extend the semantic-edge formulation to a \emph{panoptic-edge} one, ensuring that depth gradients respect both class and instance boundaries. In particular, we compute a binary mask based on the panoptic labels to determine where depth gradients should be penalized. More formally, let
\begin{equation}
    \text{B}^x_p ~=~ 
    \begin{cases}
    0, & \text{if } S(p) \neq S(p + \Delta{}x),\\
    1, & \text{otherwise},
    \end{cases}
    \quad    
\end{equation}
\begin{equation}
    \quad
    \text{V}^x_p ~=~
    \begin{cases}
    1, & \text{if } S(p) \neq \emptyset \text{ and } S(p + \Delta{}x) \neq \emptyset,\\
    0, & \text{otherwise}.
    \end{cases}    
\end{equation}
Here, $\Delta{}x$ is a one-pixel shift in the $x$-direction and $S$ are the panoptic labels. After dilating $\text{B}^x$ by a $k\times{}k$ kernel (with $k=3$) to account for imperfect edge localization, we define
\begin{equation}
    w^x_p ~=~ \text{B}^x_p
    \,\text{V}^x_p
\end{equation}
and define $w^y_p$ similarly for the $y$-direction. Our panoptic-edge-aware smoothness loss is then defined as
\begin{equation}
\label{eq:panedge_smooth}
    \mathcal{L}_{\text{pes}}
    ~=~
    \frac{1}{|P|}
    \textstyle\sum_{p \in P}
    \left(w^x_p\,\left|\partial_x\hat{D}_p\right| + w^y_p\,\left|\partial_y\hat{D}_p\right|\right).
\end{equation}
In this formulation, $w^x_p$ and $w^y_p$ are 0 (i) around panoptic boundaries, allowing depth discontinuities, and (ii) at unlabeled regions, where variations in depth are not penalized. This promotes smooth depth estimates in object \emph{interior}. 
Thus, our loss in~\eqref{eq:panedge_smooth} guides depth estimation to smoothly fill in areas between valid lidar pixels supervised via $\mathcal{L}_{\text{logL1}}$, while allowing sharp transitions between adjacent objects.

\PAR{Total loss.} 
Finally, the outlier-robust depth loss is weighted together with our smoothness losses as
\begin{equation}
    \mathcal{L}_{\text{depth}} =
    \lambda_{\text{L1}} \,\mathcal{L}_{\text{logL1}} + \lambda_{\text{es}}\mathcal{L}_{\text{es}} + \lambda_{\text{pes}}\mathcal{L}_{\text{pes}},
\end{equation}
where $\lambda_{\text{L1}}$,  $\lambda_{\text{es}}$, and $\lambda_{\text{pes}}$ are hyperparameters that control the relative strength of each term.
Our total multi-task loss is
\begin{equation} 
\label{eq:total_loss} 
    \mathcal{L}_{\text{total}} = \lambda_{\text{seg}}\mathcal{L}_{\text{seg}} + \lambda_{\text{cond}}\mathcal{L}_{\text{cond}} + \lambda_{\text{depth}}\mathcal{L}_{\text{depth}},
\end{equation} 
where $\mathcal{L}_{\text{seg}}$ denotes the full segmentation loss used in OneFormer~\cite{jain2023oneformer}, which supervises our predicted semantics, $\mathcal{L}_{\text{cond}}$ supervises the CT embedding following CAFuser~\cite{cafuser}, and $\mathcal{L}_{\text{depth}}$ supervises our predicted depth.

\begin{table*}[ht]
\vspace{0.5em}
  \caption{Comparison of panoptic segmentation methods on the MUSES test set. C:~Camera, L: Lidar, R: Radar, E: Events.}
  \label{table:sota:pan:seg}
  \centering
  \setlength\tabcolsep{4pt}
  \scriptsize
  \begin{tabular*}{\textwidth}{l @{\extracolsep{\fill}} ccccccccccccc}
  \toprule
  \textbf{Method} & \textbf{Modalities} & \textbf{Backbone} & \textbf{Day} & \textbf{Night} & \textbf{Clear} & \textbf{Fog} & \textbf{Rain} & \textbf{Snow} & \textbf{Things} & \textbf{Stuff} & \textbf{SQ} & \textbf{RQ} & \textbf{PQ} $\uparrow$ \\
  \midrule
  Mask2Former~\cite{cheng2022masked} & C   & Swin-T   & 49.35 & 39.38& 48.84 & 46.48 & 45.39 & 45.1  &  31.29  &  58.23  &  78.34  &  58.01  & 46.89  \\
  MaskDINO~\cite{li2023mask}         & C   & Swin-T   & 51.85 & 42.73 & 54.05 & 46.20 & 46.23 & 48.54 &   38.64   &   57.29   &   80.51   &   60.28  &   49.44   \\ 
  OneFormer~\cite{jain2023oneformer}   & C   & Swin-T   & 57.55 & 47.83 & 58.33 & 53.68 & 53.43 & 53.77 &  43.54  &  63.69  &  80.93  &  66.97  &  55.21  \\ 
  HRFuser~\cite{broedermann2023hrfuser} & CLRE & HRFuser-T & 44.60 & 40.01 & 47.03 & 43.59 & 42.69 & 40.62 &  28.33  &  55.23  &  78.41  &  54.34  &  43.90  \\
  MUSES~\cite{MUSES}                  & CLRE & 4xSwin-T & 54.06 & 49.73 & 55.28 & 50.34 &53.77 & 50.51 &  39.94  &  63.53  &  81.06  &  65.02  & 53.60  \\
  CAFuser-CAA~\cite{cafuser}  & CLRE & Swin-T   & 59.93 &56.24 & 61.16& 56.41 & 59.38 & 57.88 & 	48.03 & 67.64  & 81.80  & 71.61   & 59.38  \\
  CAFuser~\cite{cafuser}              & CLRE & Swin-T   & 59.49 & 57.34 & 61.36 & 57.52 & 59.63 & 57.2 & 48.42  & 67.90 & 82.03  & 71.75  & 59.70 \\
  \midrule
  \Ours{}~(\textbf{Ours})                           & CLRE & Swin-T   & \textbf{60.94} & \textbf{58.97} & \textbf{62.16} & \textbf{58.86} & \textbf{61.26 }& \textbf{59.77} & \textbf{49.68 } & \textbf{ 69.28 }  & \textbf{82.34 }  & \textbf{73.07 }  & \textbf{61.03} \\
  \bottomrule
  \end{tabular*}
\end{table*}

\subsection{Implementation Details}
\label{subsec:implementation}

We use a Swin-T backbone~\cite{liu2021swin} pre-trained on ImageNet~\cite{deng2009imagenet} and follow the training protocol of CAFuser~\cite{cafuser}. \Ours{} is trained with a batch size of 8 for 180k iterations on MUSES and 200k on DeLiVER. For the depth branch, we set $\tau=0.8$, $\lambda_{\text{L1}}=0.9$, $\lambda_{\text{es}}=0.05$, $\lambda_{\text{pes}}=0.05$, and $\lambda_{\text{depth}}=1$. We use the OneFormer head~\cite{jain2023oneformer} for our semantic branch, which enables semantic, instance, and panoptic segmentation via a single model, and we adopt the default settings for the segmentation and CT losses from OneFormer and CAFuser, respectively. All input modalities are normalized over the entire dataset and the data augmentations from CAFuser are applied. Our network is optimized using AdamW with a poly learning rate scheduler.

\begin{table}
  \caption{Results of semantic segmentation methods on the MUSES test set. C:~Camera, L: Lidar, R: Radar, E: Events.}
  \label{table:sota:sem:seg}
  \centering
  \setlength\tabcolsep{4pt}
  \scriptsize
  \begin{tabular*}{\linewidth}{l @{\extracolsep{\fill}} ccc}
  \toprule
  \textbf{Method} & \textbf{Modalities} & \textbf{Backbone} & \textbf{mIoU} $\uparrow$ \\
   \midrule
   Mask2Former~\cite{cheng2022masked} & C & Swin-T & 70.7  \\
   SegFormer~\cite{xie2021segformer} & C & MiT-B2 &  72.5  \\ 
   OneFormer~\cite{jain2023oneformer} & C & Swin-T  &  72.8  \\ 
   CMNeXt~\cite{zhang2023delivering} & CLRE & MiT-B2 &  72.1\\ 
   GeminiFusion~\cite{jia2024geminifusion} & CLRE & MiT-B2 & 75.3 \\ 
   CAFuser-CAA~\cite{cafuser} & CLRE & Swin-T & 78.5 \\
   CAFuser~\cite{cafuser} & CLRE & Swin-T & 78.2 \\
   \midrule
   \Ours{}~(\textbf{Ours}) & CLRE & Swin-T  & \textbf{79.5}\\ 
  \bottomrule  
  \hline
  \end{tabular*}
\end{table}

\section{Experiments}

We evaluate \Ours{} on two challenging multi-sensor datasets with a focus on adverse conditions in driving scenes: MUSES and DeLiVER. MUSES~\cite{MUSES} is a real-world dataset that comprises 2500 multi-sensor scenes captured under eight environmental conditions (rain, snow, fog, and clear weather, each in day/night) and containing high-quality human-annotated panoptic and semantic ground truth. Each sample includes RGB camera, lidar, radar, and event data, all of which we project onto the RGB plane with the help of the official SDK. We perform all our ablations on this challenging real-world dataset. Due to computational constraints, the hyperparameter ablation in \cref{table:abl:hyperparams} was performed on half of the training data.
DeLiVER~\cite{zhang2023delivering} is a synthetic dataset, featuring 7885 scenes recorded under adverse conditions (cloudy, foggy, night, rainy) and corner-case sensor failures (e.g.\ motion blur, lidar jitter) with semantic ground truth. Each scene provides RGB camera, lidar, and event camera modalities as well as a pixel-perfect depth map of the scenes. As this depth map is not a direct sensor recording and not realistic to obtain in this quality in any real-world scenario, we argue that one should treat it rather as an additional ground truth. Hence, we use the depth map as the dense supervision for the depth head of our method. As previous methods treat it as an input, we compare to such methods on both scenarios, with (CLDE) and without (CLE) the depth map as an input.

\subsection{Comparison with The State of The Art}

\Ours{} sets the new state of the art (SOTA) in multi-sensor panoptic and semantic segmentation compared to prior methods. As presented in~\cref{table:sota:pan:seg}, we outperform previous methods on MUSES across all metrics, including a variety of adverse weather and lighting conditions. We set the new SOTA with 61.03\% PQ (+1.33\%) and improvements are most pronounced in adverse conditions. Compared with the previous SOTA method CAFuser~\cite{cafuser}, all adverse conditions show a substantial improvement: 1.34\% PQ in \emph{Fog}, 1.63\% PQ in \emph{Rain} and even +2.57\% PQ in \emph{Snow} conditions, compared to a moderate +0.8\% PQ in \emph{Clear} conditions.
Further, a \emph{Camera + Lidar} setup achieves 60.19\% PQ, exceeding CAFuser-CL by +1.49\%.
These findings confirm
our hypothesis
that fusing both global condition cues and local depth features is crucial for robust performance under adverse 
conditions.

\begin{table}[t]
  \caption{Comparison of semantic segmentation methods on the DeLiVER test set. C: RGB Camera, L: Lidar, D: Depth, E: Events.}
  \label{table:sota:deliver}
  \centering
  \setlength\tabcolsep{4pt}
  \scriptsize
  \begin{tabular}{lccc}
    \toprule
    \multirow{2}{*}{\textbf{Method}} & \multirow{2}{*}{\textbf{Backbone}} & \multicolumn{2}{c}{\textbf{mIoU-test} $\uparrow$} \\
    \cmidrule(lr){3-4}
     &  &  \textbf{CLE} & \textbf{CLDE}  \\
    \midrule
    CMNeXt~\cite{zhang2023delivering}    & MiT-B2     & 50.3  & 53.0 \\
    StitchFusion~\cite{li2024stitchfusion} & MiT-B2     & 50.8  & 53.4 \\
    GeminiFusion~\cite{jia2024geminifusion}   & MiT-B2   & 50.5  & 54.5 \\
    CAFuser-CAA~\cite{cafuser}           & Swin-T     & 51.2 & 55.2 \\
    CAFuser~\cite{cafuser}              & Swin-T     & 51.3  & 55.6 \\
    \midrule
    \Ours{}~(\textbf{Ours})              & Swin-T   & \textbf{51.6} 
    & \textbf{56.7} \\ 
    \bottomrule
  \end{tabular}
\end{table}

As the OneFormer segmentation head can predict both panoptic and semantic segmentation, we also compare \Ours{} to SOTA multimodal semantic segmentation methods on the MUSES dataset.~\cref{table:sota:sem:seg} shows that our method significantly outperforms both camera-only baselines and prior multimodal approaches, achieving 79.5\% mIoU (+ 1.0\%). Notably, we surpass other fusion-based methods, such as CMNeXt~\cite{zhang2023delivering} and CAFuser~\cite{cafuser}, highlighting the effectiveness of our multi-task setup for improving semantic segmentation in adverse-weather driving scenes.

\cref{table:sota:deliver} shows our semantic segmentation results on DeLiVER for two input settings: CLE (camera, lidar, events) and CLDE (camera, lidar, depth, events). \Ours{} outperforms prior fusion methods in both cases, reaching 51.6\% mIoU (+0.3\%) for CLE and 56.7\% mIoU (+1.1\%) for CLDE. Notably, using depth as an auxiliary training signal boosts performance even further in the CLDE setting, where 
perfect depth is already available as an additional sensor. This highlights our system’s ability to exploit even ``ground-truth'' depth in a more principled way, resulting in more reliable multimodal segmentation under challenging weather and sensor-failure conditions represented in DeLiVER.

\input{figures/qualitative_results}

\subsection{Ablations}
\label{sub:sec:ablations}

We perform ablation studies to investigate the impact of each proposed architectural component in~\cref{table:abl:modules}, supervised with the complete loss described in \cref{sub:sec:loss:design}. Starting with the baseline CAFuser (ID 1), adding an auxiliary depth head (ID 2) significantly boosts performance by +0.64\% PQ. 
Notably, this auxiliary depth head is used only during training and does not alter the inference architecture at all, indicating that leveraging existing lidar inputs as auxiliary supervision provides a substantial benefit.
When further adding our proposed depth-guided fusion module (ID 3), performance improves by an additional +0.69\% PQ. We also observe a synergy of CAFuser's global CT and our local DT, as removing either CT (ID 4) or DT (ID 2) leads to performance drops of 0.65\% PQ and 0.69\% PQ, respectively. 
These results confirm that localized depth cues and condition awareness mutually reinforce robust multimodal segmentation, demonstrating the individual contributions and complementary strengths of our proposed architectural components. 

\begin{table}[t]
  \caption{Ablation of architectural components on the MUSES test set.}
  \label{table:abl:modules}
  \centering
  \setlength\tabcolsep{2pt}
  \scriptsize
  \resizebox{\linewidth}{!}{%
  \begin{tabular*}{\linewidth}{l @{\extracolsep{\fill}} ccccc}
  \toprule
  \textbf{ID} 
  & \textbf{CT} 
  & \textbf{Aux. Depth Head} 
  & \textbf{DT} 
  & \textbf{mIoU} $\uparrow$ 
  & \textbf{PQ} $\uparrow$ \\
  \midrule
  1 & \checkmark & -  &  -  & 78.20 & 59.70\\ 
  2 & \checkmark & \checkmark & - & 78.63  & 60.34 \\ 
  3~(\textbf{Ours}) & \checkmark & \checkmark & \checkmark & \textbf{79.50 } & \textbf{61.03}\\ 
  \midrule     
  4 & - & \checkmark & \checkmark & 79.19 & 60.38 \\ 
  \bottomrule  
  \end{tabular*}}
\end{table}

\begin{table}[t]
  \caption{Loss design ablations on the MUSES test set.}
  \label{table:abl:losses}
  \centering
  \setlength\tabcolsep{4pt}
  \scriptsize
  \begin{tabular*}{\linewidth}{l @{\extracolsep{\fill}} ccc c}
    \toprule
    \textbf{ID} 
    & \boldmath{$\mathcal{L}_{\text{es}}$}\&\boldmath{$\mathcal{L}_{\text{pes}}$}
    & \textbf{$\tau$-filtering}
    & \textbf{mIoU} $\uparrow$ 
    & \textbf{PQ} $\uparrow$ \\
    \midrule
    1 &  - & -  & 78.87 & 59.84 \\ 
    2 &  \checkmark & -  & 78.94   & 59.98  \\ 
    3 &  - & \checkmark  & 78.50  & 60.04  \\ 
    4 (\textbf{Ours})~  & \checkmark  & \checkmark  &\textbf{79.50}& \textbf{61.03} \\
    \bottomrule
  \end{tabular*}
\end{table}

We also investigate our depth loss design from~\cref{sub:sec:loss:design}, with results presented in~\cref{table:abl:losses}. The baseline (ID 1), which is trained only with a basic $L_1$ loss as depth supervision, achieves moderate performance.
Introducing our smoothness terms $\mathcal{L}_{\text{es}}$ and $\mathcal{L}_{\text{pes}}$ (ID 2) improves PQ slightly by +0.14\%, indicating their benefit in regularizing sparse lidar depth supervision.
Finally, combining both our smoothness terms and $\tau$-quantile filtering (ID 4) results in the best overall performance, further improving PQ by +1.05\%. Our smoothness losses shine especially at the presence of $\tau$-filtering, as removing the former (from ID 4 to ID 3) reduces PQ by -0.99\%.
These experiments confirm the importance of explicitly addressing sparse and noisy lidar returns through outlier filtering and smoothness constraints for more robust depth guidance and hence improved panoptic segmentation.

\begin{table}[t]
  \caption{Hyperparameter ablation on the MUSES val. set.}
  \label{table:abl:hyperparams}
  \centering
  \setlength\tabcolsep{4pt}
  \scriptsize
  \begin{tabular}{l @{\extracolsep{\fill}} ccc}
    \toprule
    \textbf{Parameter}~ & \textbf{×0.5} & \textbf{Default} & \textbf{×2} \\
    \midrule
    $\lambda_{\text{L1}}$ & 54.15 & \textbf{55.34} & 55.03 \\
    $\lambda_{\text{es}}$ & 54.16 & \textbf{55.34} & 55.22 \\
    $\lambda_{\text{pes}}$ & 55.20 & \textbf{55.34} & 54.26 \\
    $\tau$ & 54.58 & \textbf{55.34} & 54.96 \\
    $\lambda_{\text{depth}}$ & 54.10 & \textbf{55.34} & 54.97 \\
    \bottomrule
  \end{tabular}
\end{table}

To demonstrate the robustness of our approach, we investigate the sensitivity of our method to variations in the hyperparameters introduced by our work. As shown in~\cref{table:abl:hyperparams}, we evaluate performance across a range of hyperparameter values by scaling each parameter by $0.5\times$ and $2\times$ relative to our default settings. All five introduced hyperparameters ($\lambda_{\text{L1}}$, $\lambda_{\text{es}}$, $\lambda_{\text{pes}}$, $\tau$, $\lambda_{\text{depth}}$) exhibit stable performance on the validation set with $\leq1.24$\% variation in PQ across these scaling factors. This stability demonstrates that our method does not require tedious hyperparameter tuning for each new deployment scenario, as evidenced by our cross-dataset generalization where the same hyperparameter values achieve SOTA performance on both MUSES (real-world) and DeLiVER (synthetic) without any dataset-specific tuning.

\begin{table}[t]
  \caption{Computational efficiency on MUSES test set.}
  \label{table:efficiency}
  \centering
  \scriptsize
  \begin{tabular}{l @{\extracolsep{\fill}} cccc}
    \toprule
    \textbf{Method}
    & \textbf{Parameters}
    & \textbf{FPS}
    & \textbf{GFLOPs}
    & \textbf{PQ} $\uparrow$ \\
    \midrule
    CAFuser~\cite{cafuser} & 77.68M & 7.04 & 349.3 & 59.70 \\
    \Ours{}~(\textbf{Ours}) & 79.45M & 6.83 & 358.1 & \textbf{61.03} \\
    \bottomrule
  \end{tabular}
\end{table}

Finally, in~\cref{table:efficiency}, we investigate the computational efficiency of our method. \Ours{} adds only 2.3\% to the parameter count at test time and runs at 6.83 FPS on an A6000 GPU. Hence, our significant +1.33\% PQ gain comes at only marginal computational cost. This demonstrates that our depth-guided fusion approach achieves substantial performance improvements while maintaining practical efficiency for real-world deployment.

\subsection{Qualitative Results}

In \cref{fig:muses_all}, we present qualitative prediction results on MUSES under various conditions, demonstrating the strengths of our approach compared to existing methods. Under the clear day scenario, we accurately capture the tram in the distance and all four pedestrians on the left sidewalk, whereas other baselines overlook some of these persons. In the second example, 
we correctly identify wall regions on the side of the bridge in front and detect the bicycle on the left.
In the foggy night scene, we are the only method that detects both the cyclist and the bicycle despite the very low visibility. 
Finally, in the snow daytime scene, \Ours{} is the only method that predicts at least one of the two pedestrians on the left sidewalk, though it misses the second person, highlighting a limitation in detecting heavily occluded objects. Our method 
occasionally confuses similar classes, such as ``wall'' and ``fence'' 
(last row). Nevertheless, these examples highlight how our depth-guided method improves 
detection under all environmental conditions.

\section{Conclusion}
\label{sec:conclusion}

We have presented \Ours{}, a novel depth-guided sensor fusion network for robust semantic perception. Our approach reformulates the multi-sensor fusion task as a multi-task problem by integrating an auxiliary depth head and directly leveraging noisy lidar input as additional supervision. In addition, we propose a depth-guided fusion module that combines local depth tokens with global condition tokens to adaptively guide sensor fusion in any condition. Extensive evaluations on both the real-world MUSES dataset and the synthetic DeLiVER dataset show that \Ours{} outperforms previous state-of-the-art methods in panoptic and semantic segmentation, particularly in adverse conditions such as night, fog, rain, and snow. Moreover, our ablation studies confirm the essential contribution of our multi-task loss design and robust depth supervision, demonstrating the value of leveraging noisy lidar depth signals without incurring extra annotation costs. These advancements position \Ours{} as a promising solution for enhancing semantic perception in autonomous driving and robotics under challenging environmental scenarios.

{\small
\bibliographystyle{IEEEtran}
\bibliography{IEEEabrv,main}
}

\appendix

\begin{table*}[tb]
  \caption{Class-wise PQ results on the MUSES test split.}
  \label{table:supp:class:PQ}
  \centering
  \setlength\tabcolsep{4pt}
  \footnotesize
  \resizebox{\textwidth}{!}{
  \begin{tabular}{@{\extracolsep{\fill}} lccccccccccccccccccc@{}}
  \toprule
    Method & \rotatebox[origin=c]{90}{road} & \rotatebox[origin=c]{90}{sidewalk} & \rotatebox[origin=c]{90}{building} & \rotatebox[origin=c]{90}{wall} & \rotatebox[origin=c]{90}{fence} & \rotatebox[origin=c]{90}{pole} & \rotatebox[origin=c]{90}{traffic light} & \rotatebox[origin=c]{90}{traffic sign} & \rotatebox[origin=c]{90}{vegetation} & \rotatebox[origin=c]{90}{terrain} & \rotatebox[origin=c]{90}{sky}  & \rotatebox[origin=c]{90}{person} & \rotatebox[origin=c]{90}{rider} & \rotatebox[origin=c]{90}{car} & \rotatebox[origin=c]{90}{truck} & \rotatebox[origin=c]{90}{bus} & \rotatebox[origin=c]{90}{train} & \rotatebox[origin=c]{90}{motorcycle} & \rotatebox[origin=c]{90}{bicycle} \\
  \midrule
  Mask2Former& 95.12 & 72.08 & 76.59 & 38.35 & 28.96 & 39.85 & 37.54 & 48.73 & 69.73 & 49.60 & 83.95 & 34.03 & 19.31 & 55.70 & 28.59 & 42.58 & 38.21 & 14.88 & 17.06 \\
  MaskDINO         & 95.22 & 69.06 & 77.87 & 31.32 & 32.81 & 44.31 & 37.52 & 44.56 & 70.81 & 42.23 & 84.51 & 46.88 & 27.65 & 62.22 & 35.41 & 41.41 & 38.07 & 30.89 & 26.61 \\
  OneFormer   & 95.95 & 75.68 & 81.29 & 44.13 & 37.30 & 47.80 & 48.21 & 56.87 & 72.37 & 52.53 & 88.50 & 47.07 & 29.92 & 64.41 & 40.94 & 62.03 & 47.37 & 31.14 & 25.44 \\
  HRFuser & 94.35 & 69.15 & 71.73 & 34.71 & 24.43 & 35.61 & 30.04 & 51.23 & 67.62 & 46.49 & 82.16 & 33.50 & 18.89 & 56.49 & 26.87 & 31.53 & 29.73 & 19.94 & 9.70 \\
  MUSES                  & 95.86 & 76.42 & 80.40 & 47.43 & 37.56 & 45.57 & 42.83 & 60.23 & 75.01 & 53.83 & 83.69 & 42.21 & 33.51 & 63.65 & 40.01 & 45.09 & 42.65 & 28.52 & 23.88 \\
  CAFuser-CAA  & 96.14 & 78.29 & 83.12 & 51.40 & 42.48 & 49.35 & 53.79 & 64.07 & 78.65 & 56.09 & 90.65 & \textbf{51.79} & 39.53 & 66.33 & 46.88 & 58.99 & 52.51 & 39.07 & \textbf{29.18} \\
  CAFuser             & 96.36 & 79.52 & 84.38 & 50.87 & 43.55 & 49.84 & 50.82 & 63.18 & 79.40 & 56.90 & 92.11 & 50.51 & 39.05 & 65.69 & 45.95 & 62.23 & \textbf{54.34} & \textbf{40.63} & 28.96 \\
  \Ours{}~(\textbf{Ours})              & \textbf{96.63} & \textbf{80.80} & \textbf{84.51} & \textbf{53.60} & \textbf{43.56} & \textbf{52.54} & \textbf{54.39} & \textbf{65.56} & \textbf{79.41} & \textbf{57.22} & \textbf{93.63} & 51.74 & \textbf{42.17} & \textbf{67.13} & \textbf{49.33} & \textbf{64.07} & 53.26 & 39.41 & 28.36 \\
  \bottomrule
  \end{tabular}
  }
\end{table*}

\section*{Detailed Class-Level Results}

We present detailed class-level panoptic quality results on the MUSES test split in~\Cref{table:supp:class:PQ}. Our method consistently achieves superior performance across most classes. This highlights our model's robust adaptation across varied semantic categories compared to other strong multimodal fusion methods, reflecting the effectiveness of our depth-guided fusion strategy.

\begin{figure*}[tbp]
\centering
\begin{tabular}{@{}c@{\hspace{0.03cm}}
                c@{\hspace{0.03cm}}
                c@{\hspace{0.03cm}}
                c@{\hspace{0.03cm}}
                c@{}}
\subfloat{\scriptsize RGB} &
\subfloat{\scriptsize Lidar} &
\subfloat{\scriptsize ID 1 of Tab. 5 ($\lambda_{\text{es}}=0; \lambda_{\text{pes}}=0; \tau=1$) } &
\subfloat{\scriptsize \Ours{}-depth \textbf{(Ours)}}\\
\vspace{-0.1cm}

\includegraphics[width=0.24\textwidth]{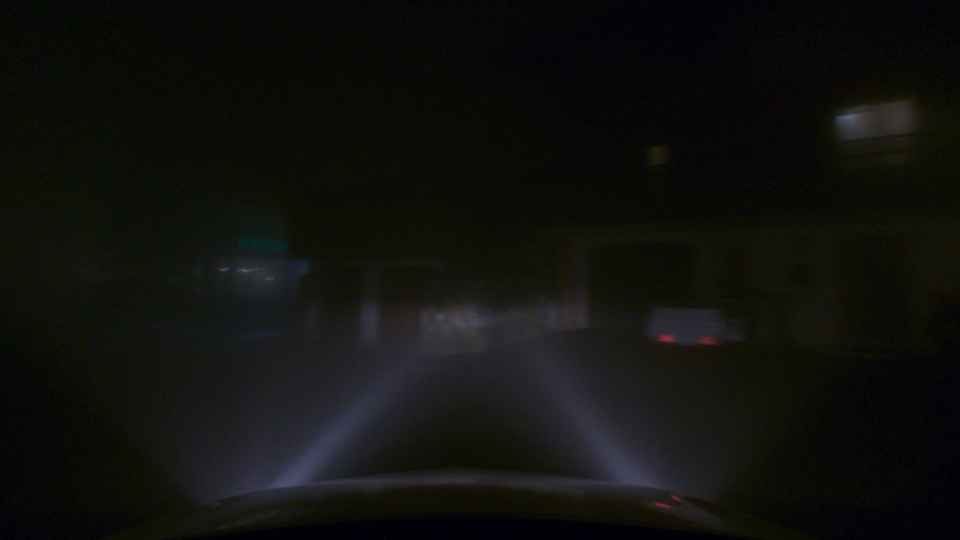} &
\includegraphics[width=0.24\textwidth]{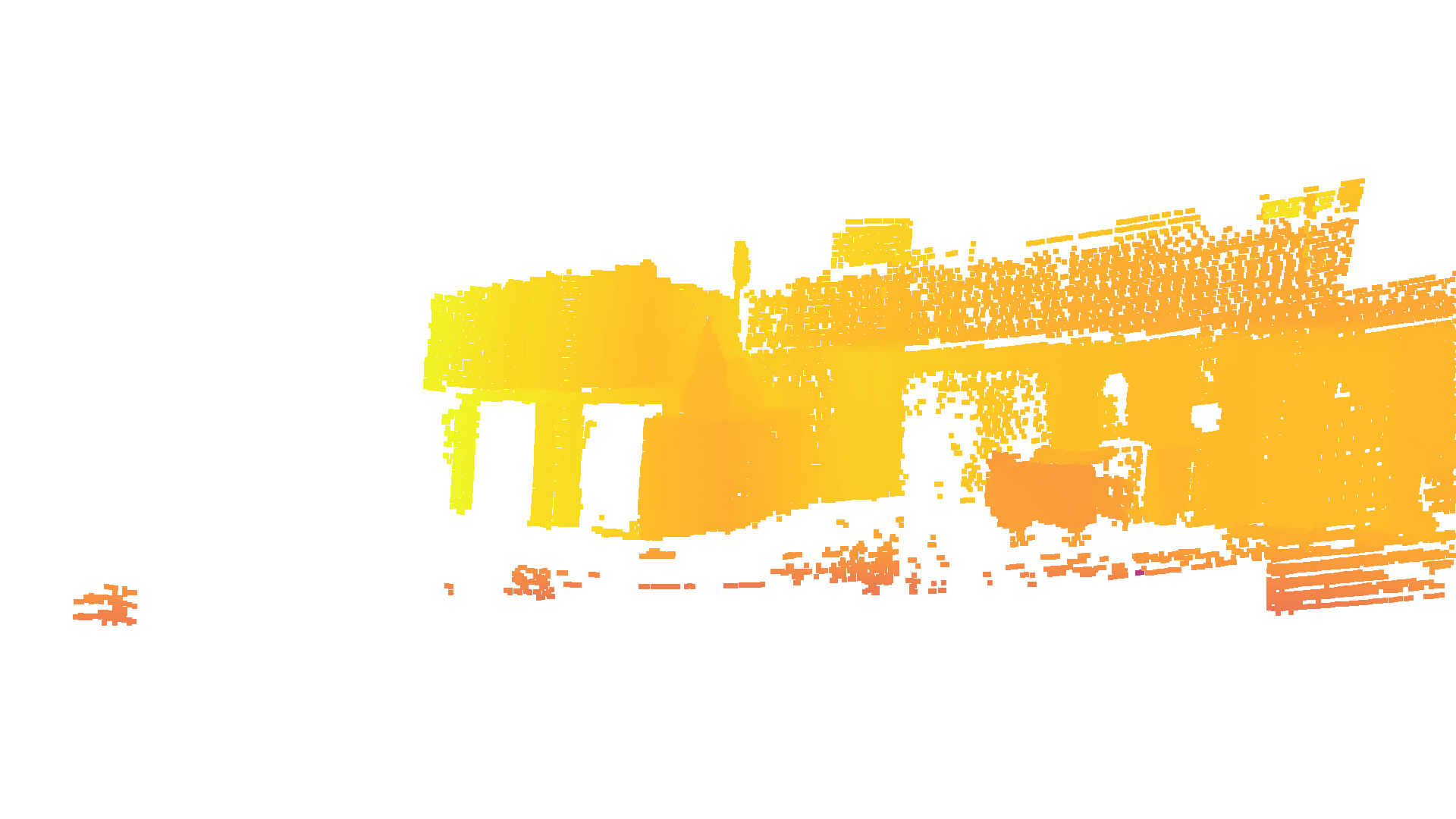}  &
\includegraphics[width=0.24\textwidth]{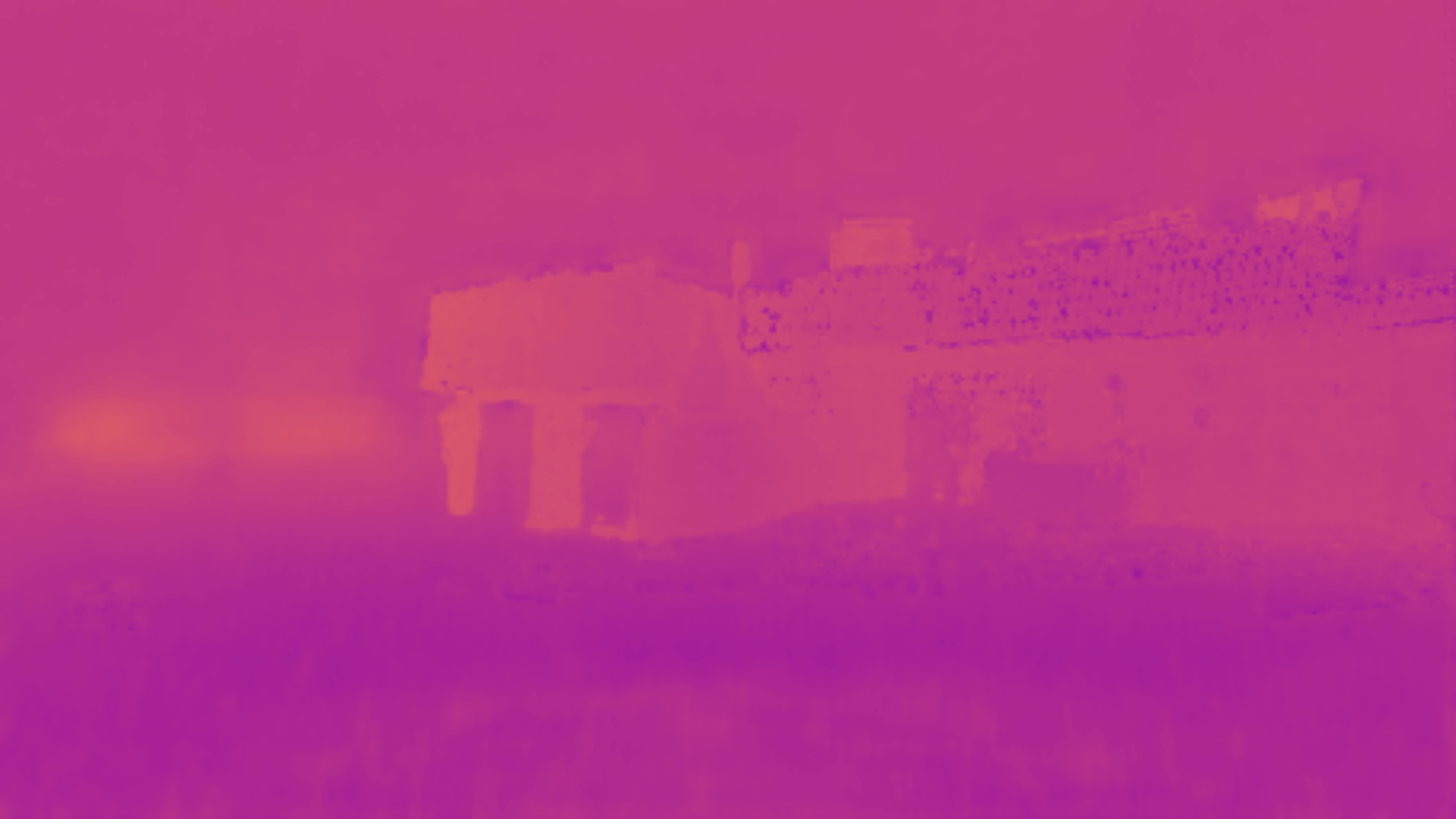} &
\includegraphics[width=0.24\textwidth]{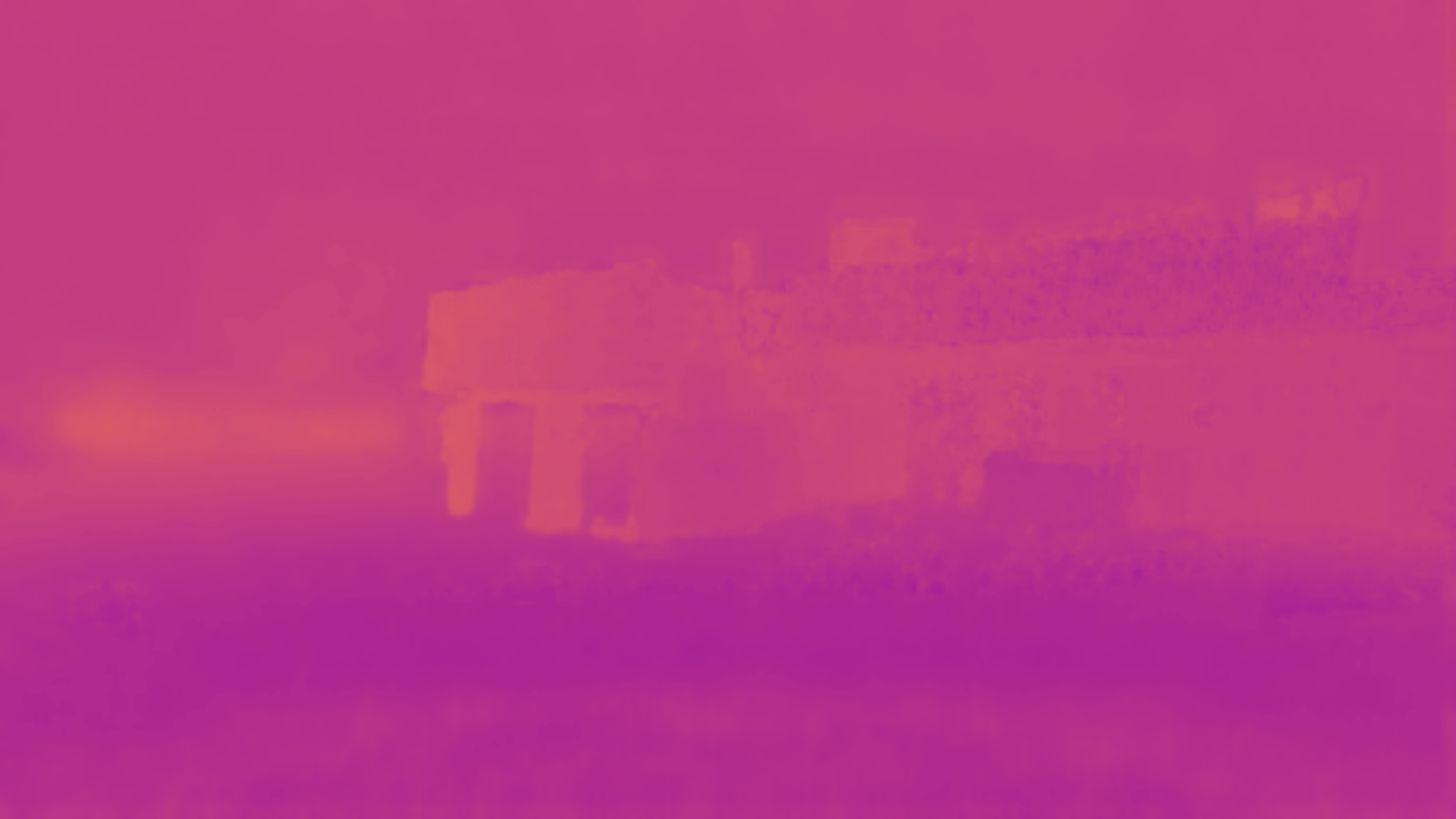}
\\
\vspace{-0.1cm}

\includegraphics[width=0.24\textwidth]{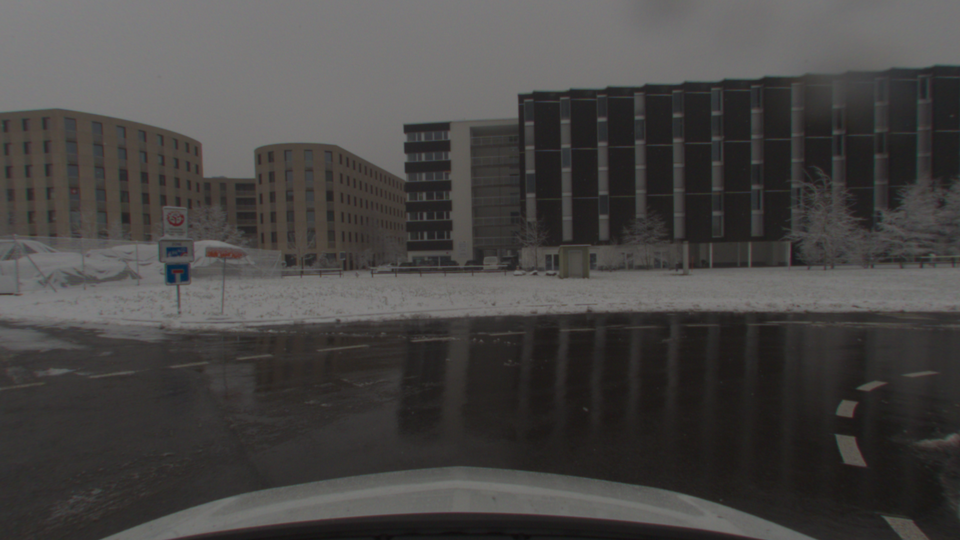} &
\includegraphics[width=0.24\textwidth]{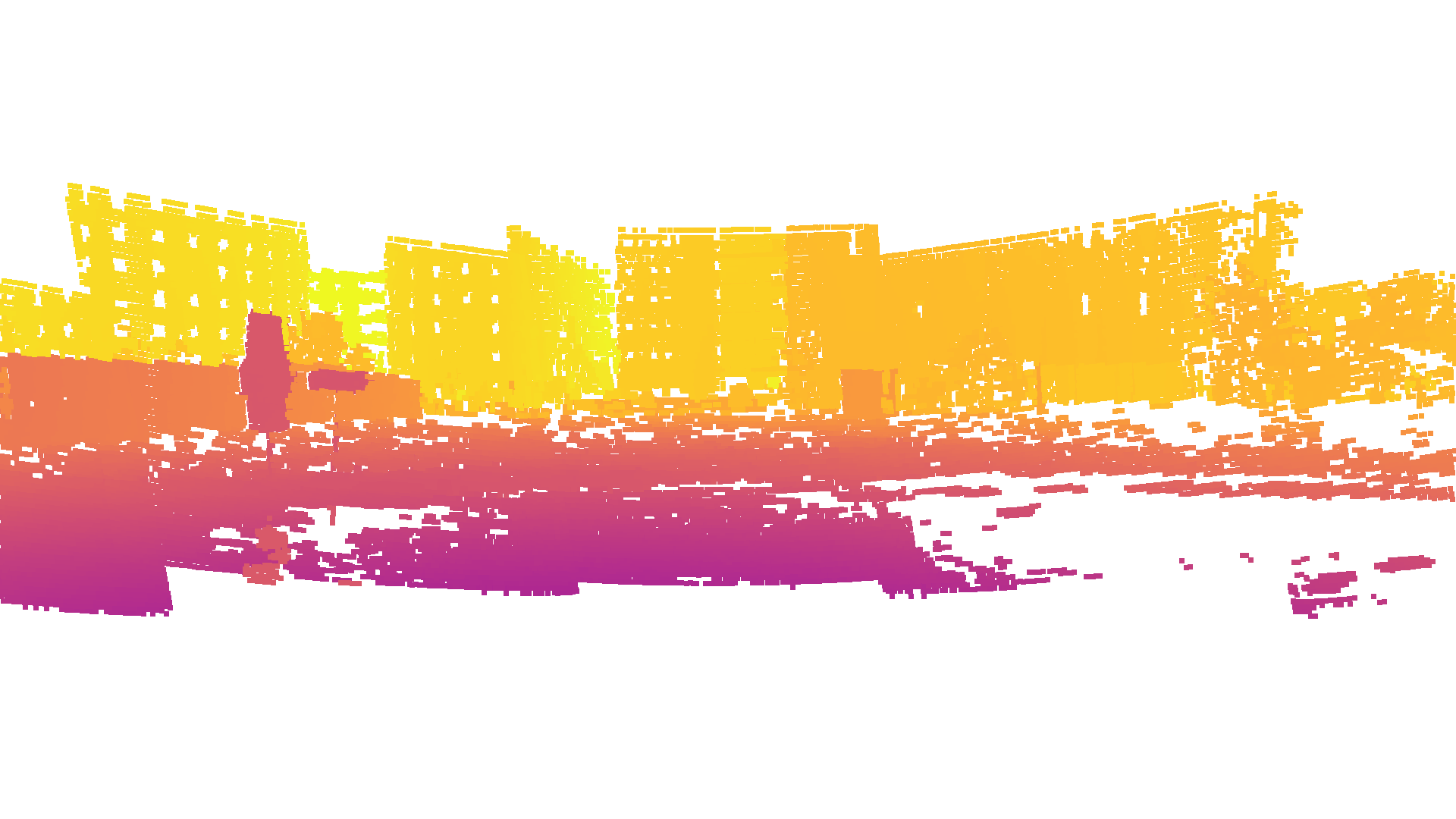}  &
\includegraphics[width=0.24\textwidth]{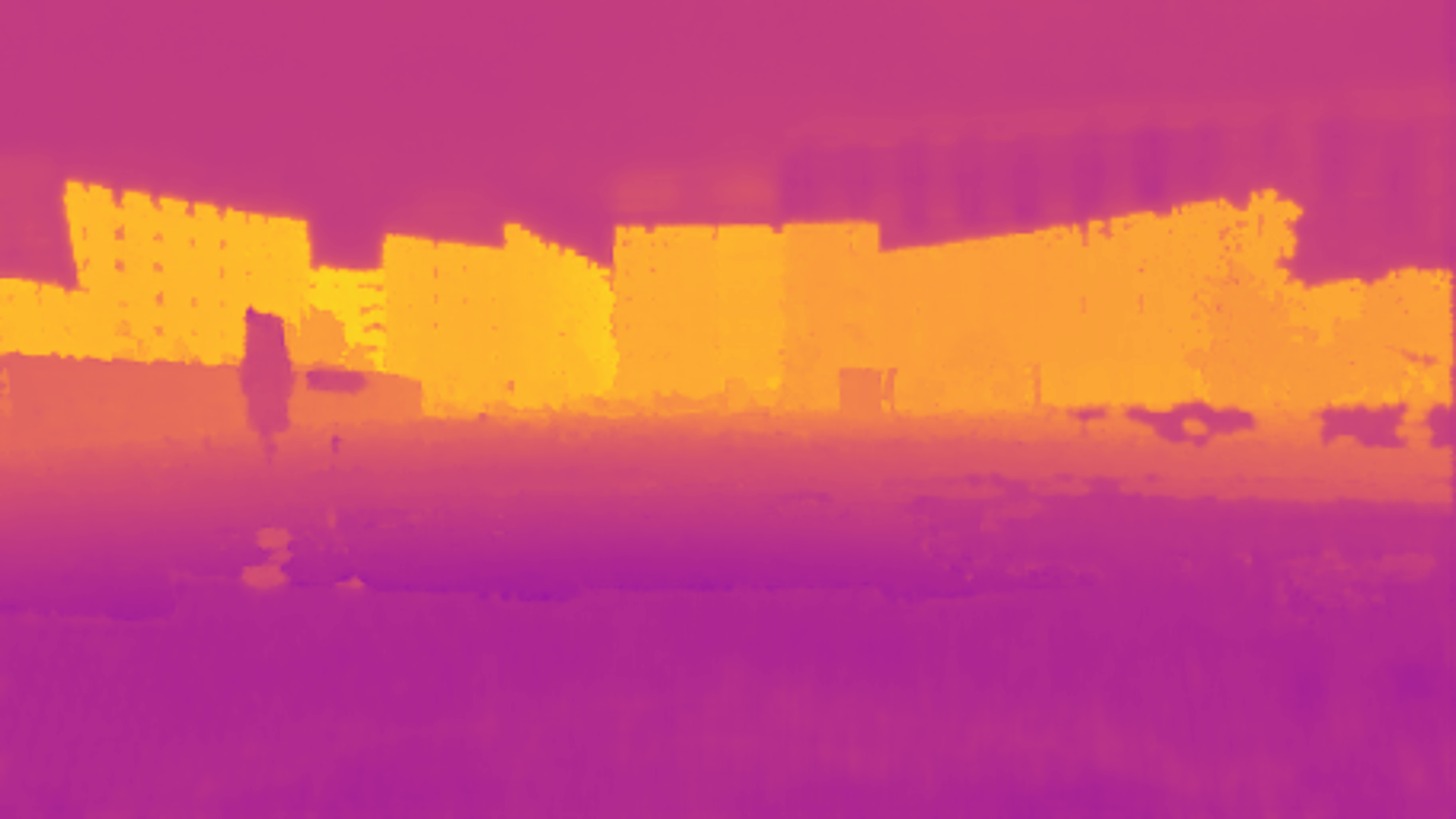} &
\includegraphics[width=0.24\textwidth]{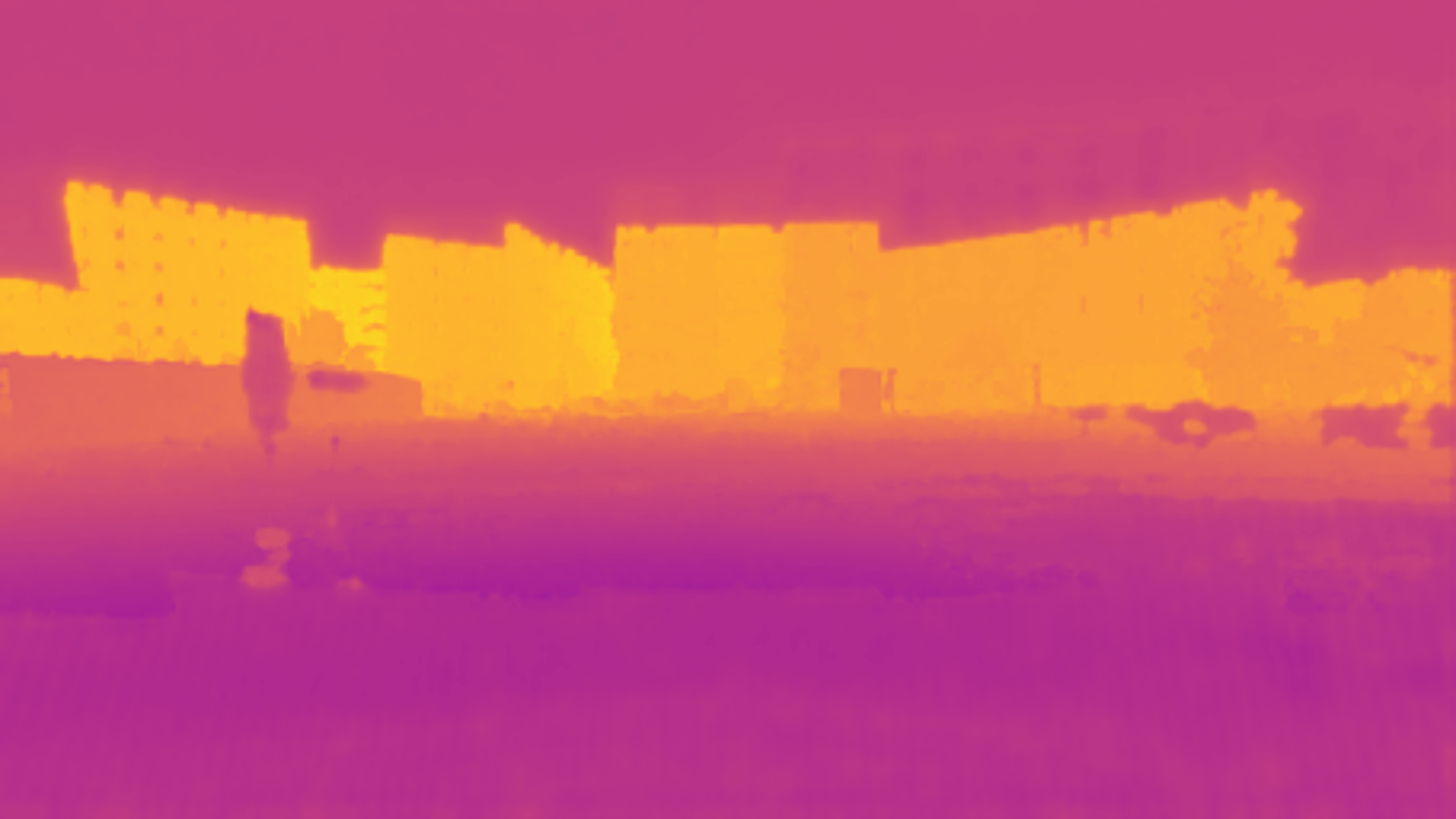}
\\
\vspace{-0.1cm}

\includegraphics[width=0.24\textwidth]{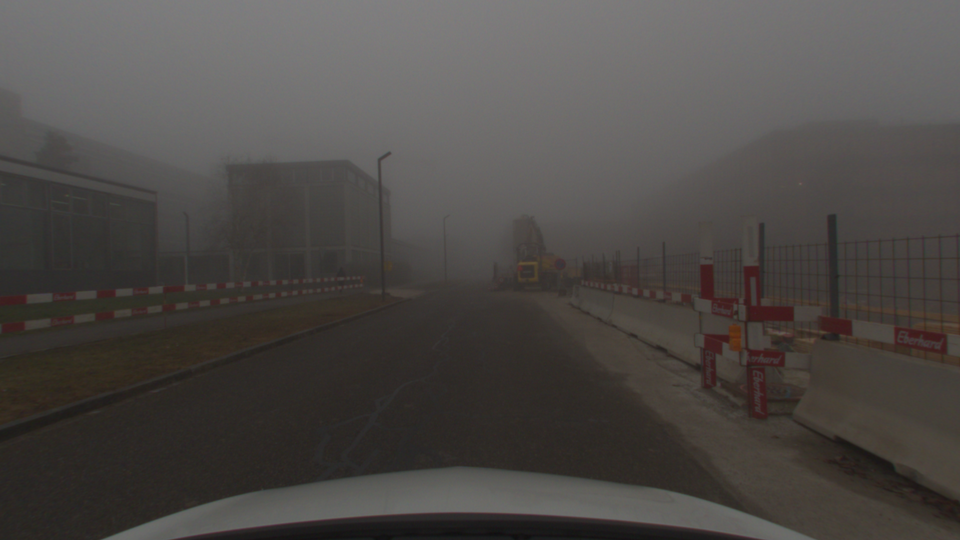} &
\includegraphics[width=0.24\textwidth]{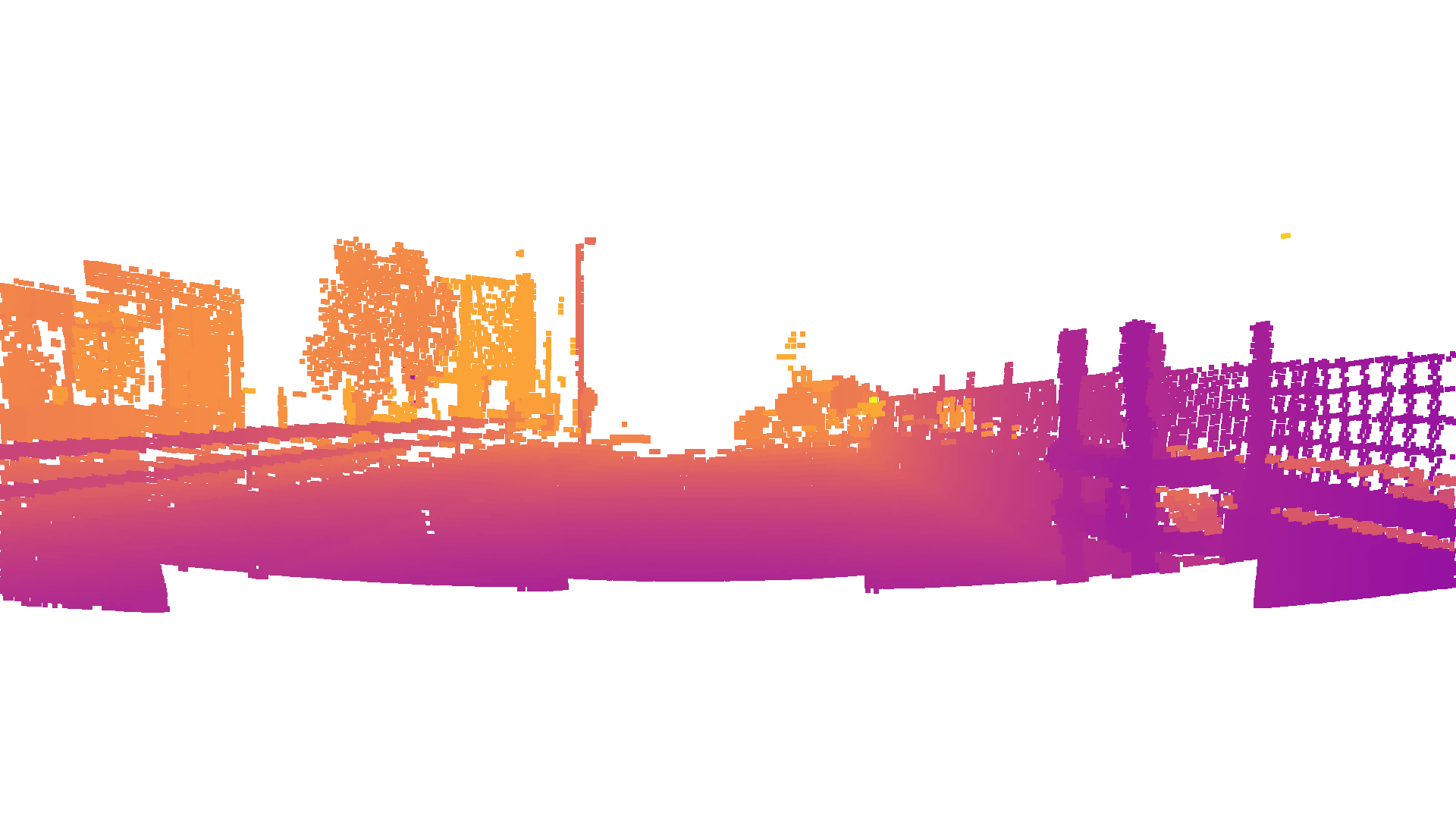}  &
\includegraphics[width=0.24\textwidth]{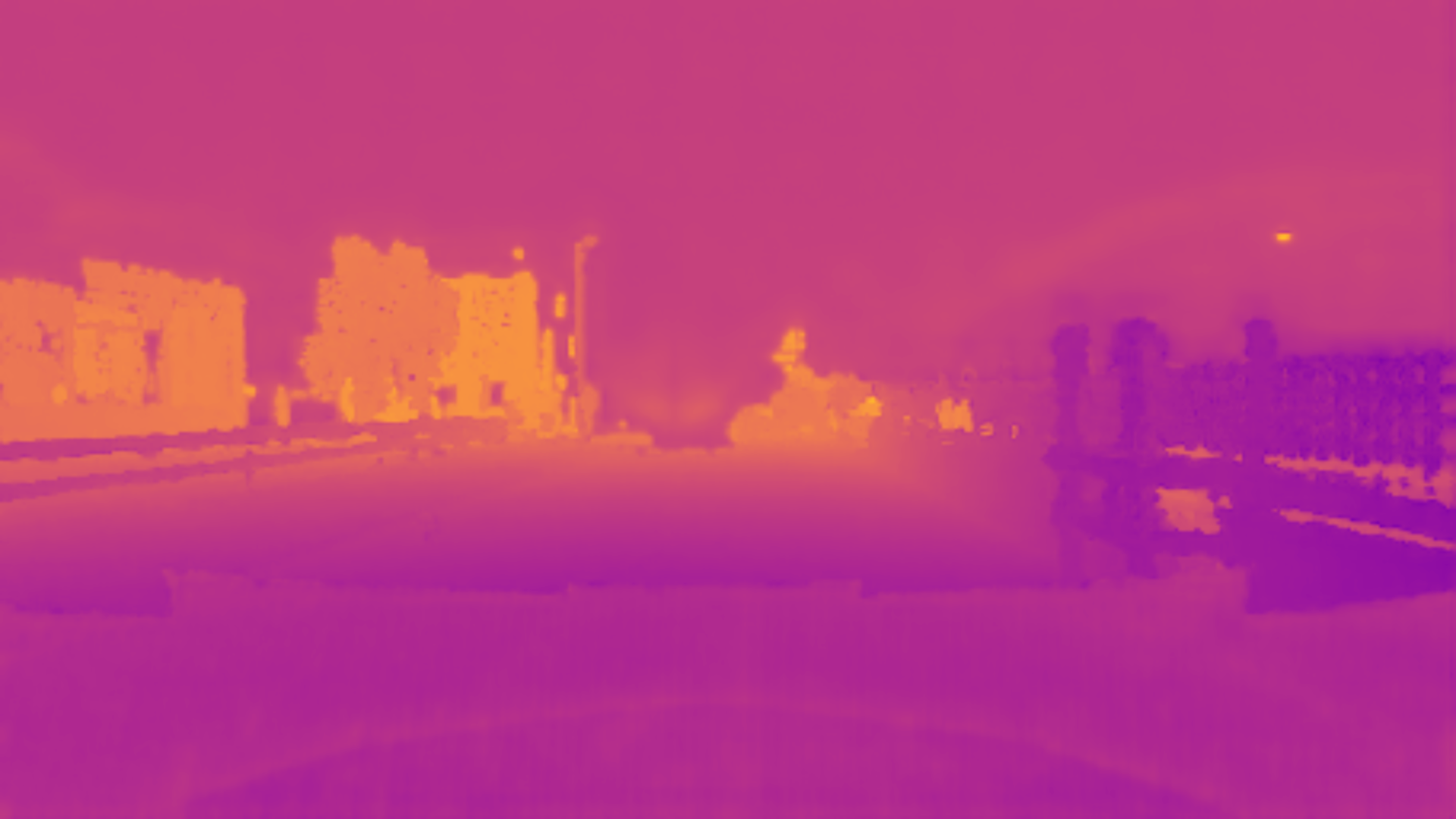} &
\includegraphics[width=0.24\textwidth]{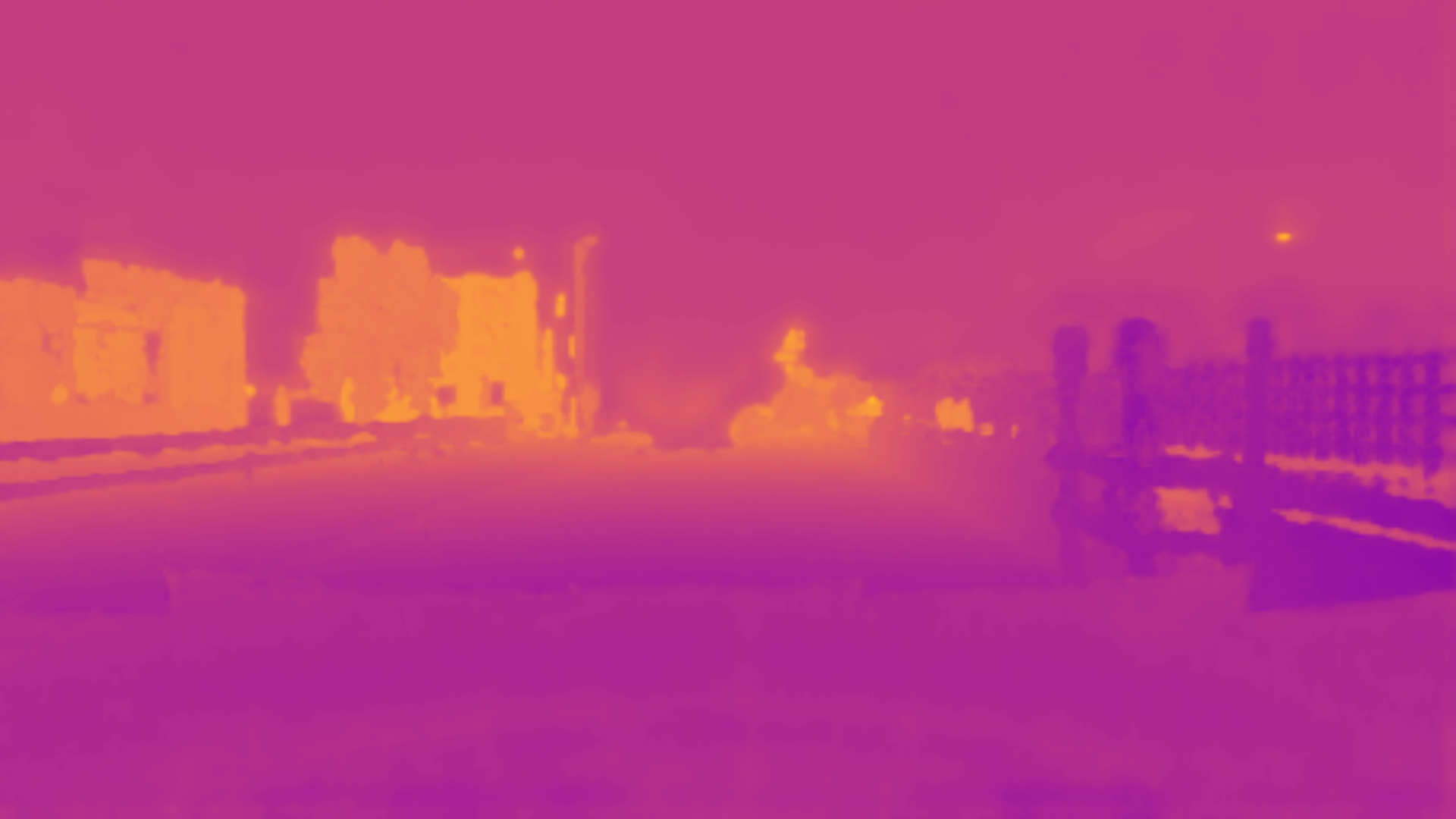}
\\
\vspace{-0.1cm}

\includegraphics[width=0.24\textwidth]{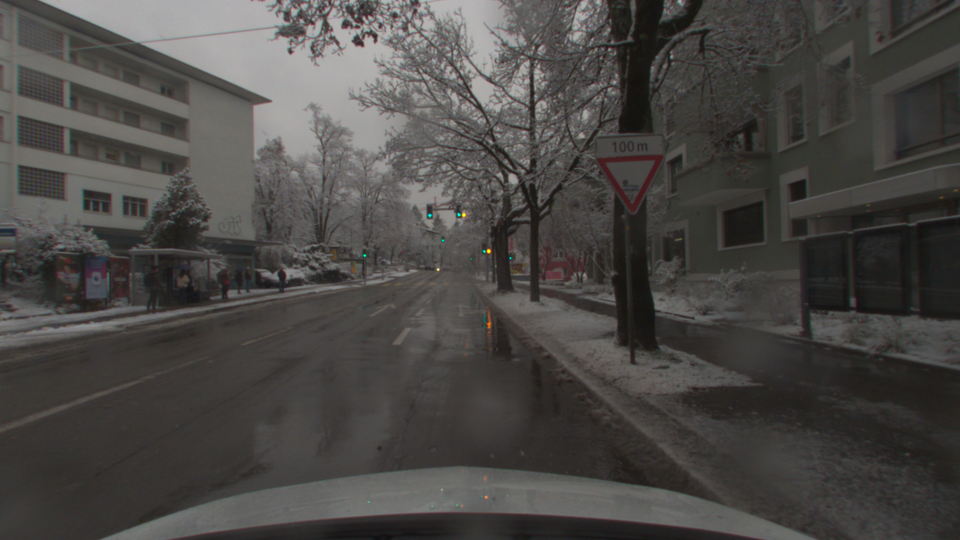} &
\includegraphics[width=0.24\textwidth]{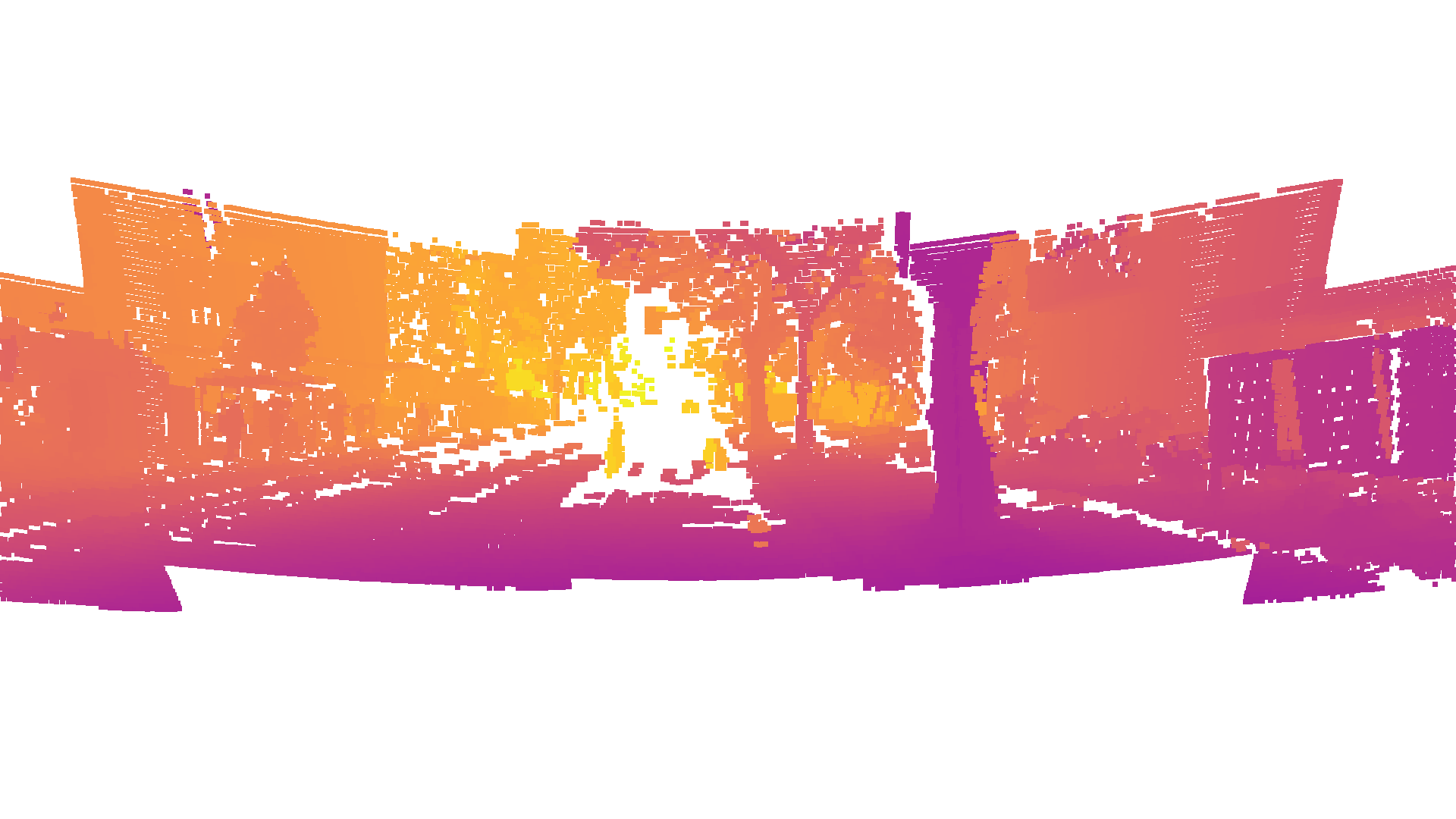}  &
\includegraphics[width=0.24\textwidth]{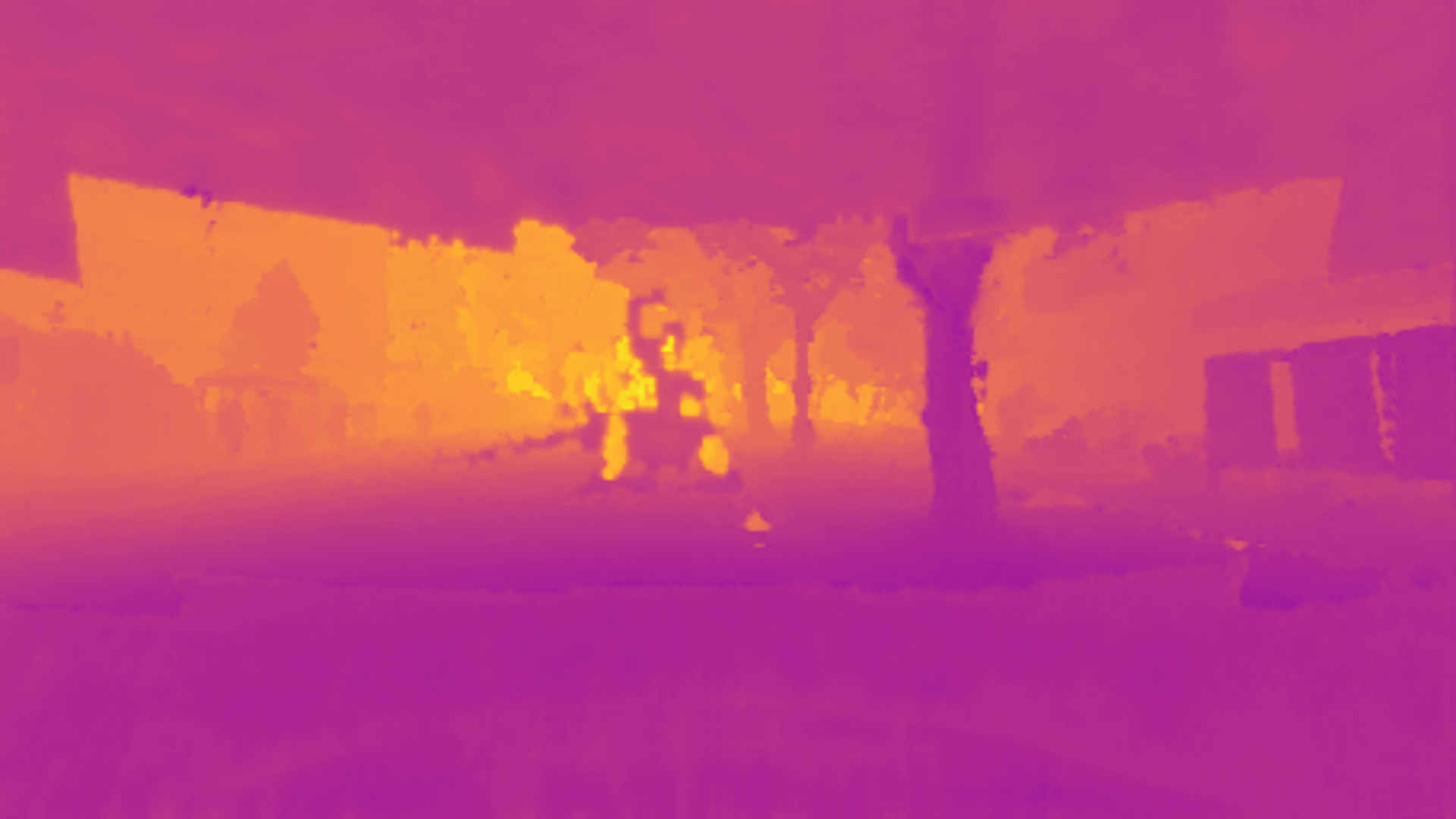} &
\includegraphics[width=0.24\textwidth]{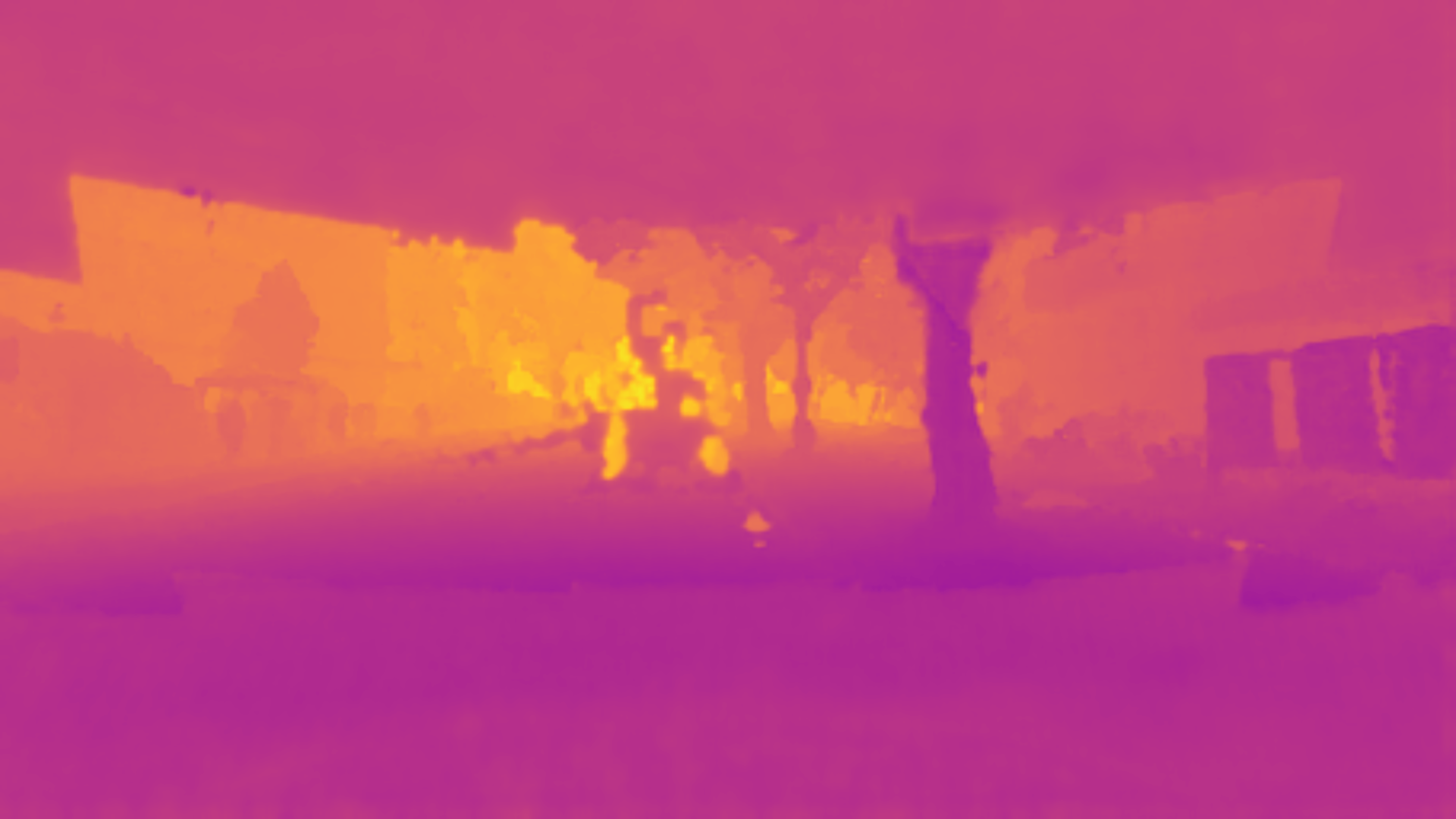}
\\
\vspace{-0.1cm}

\end{tabular}
\caption{
Visual study 
Qualitative ablation on our loss design on the MUSES dataset. Best viewed on screen at full zoom.}
\label{fig:supl:depth_muses_ablation}
\end{figure*}

\section*{Visual Ablation on Our Loss Design}

In ~\Cref{fig:supl:depth_muses_ablation}, we visually explore the effect of our loss design, which we investigate quantitatively in the main paper. The results illustrate that our proposed loss enhances the depth estimation quality by targeting the challenges of using noisy lidar as ground truth. Specifically, structural artifacts such as holes in building facades are effectively reduced, and boundaries are notably cleaner. Thereby demonstrating the strength of our method in handling challenging visual scenarios and providing visually superior depth predictions.

\section*{Further Qualitative Results}
\begin{figure*}[tbp]
\centering
\begin{tabular}{@{}c@{\hspace{0.03cm}}
                c@{\hspace{0.03cm}}
                c@{\hspace{0.03cm}}
                c@{\hspace{0.03cm}}
                c@{\hspace{0.03cm}}
                c@{\hspace{0.03cm}}
                c@{\hspace{0.03cm}}
                c@{\hspace{0.03cm}}
                c@{}
                c@{}}
\subfloat{\scriptsize RGB} &
\subfloat{\scriptsize Lidar} &
\subfloat{\scriptsize Radar} &
\subfloat{\scriptsize Events} &
\subfloat{\scriptsize Ours-depth} &
\subfloat{\scriptsize GT} &
\subfloat{\scriptsize MUSES} &
\subfloat{\scriptsize OneFormer} &
\subfloat{\scriptsize CAFuser} &
\subfloat{\scriptsize \Ours{}~(\textbf{Ours})} \\
\vspace{-0.1cm}

\includegraphics[width=0.10\textwidth]{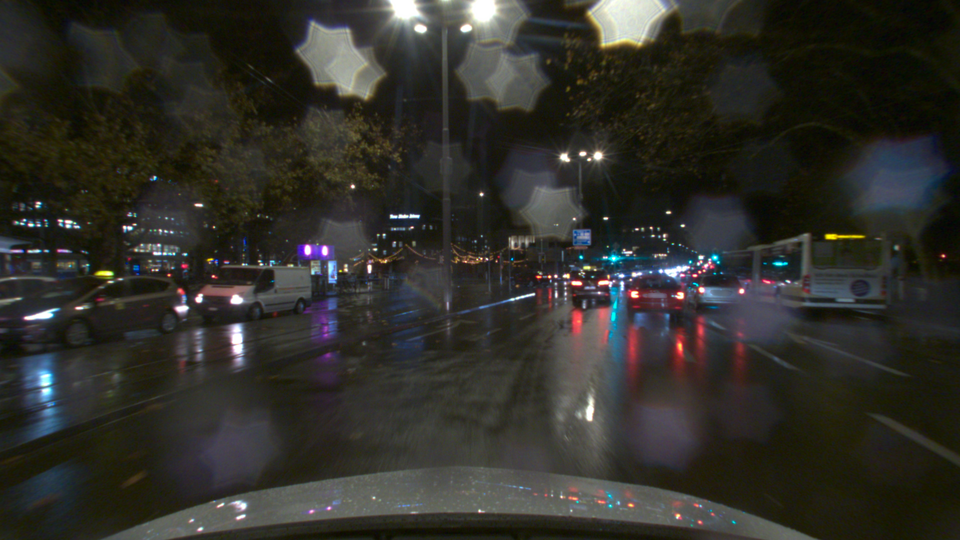} &
\includegraphics[width=0.10\textwidth]{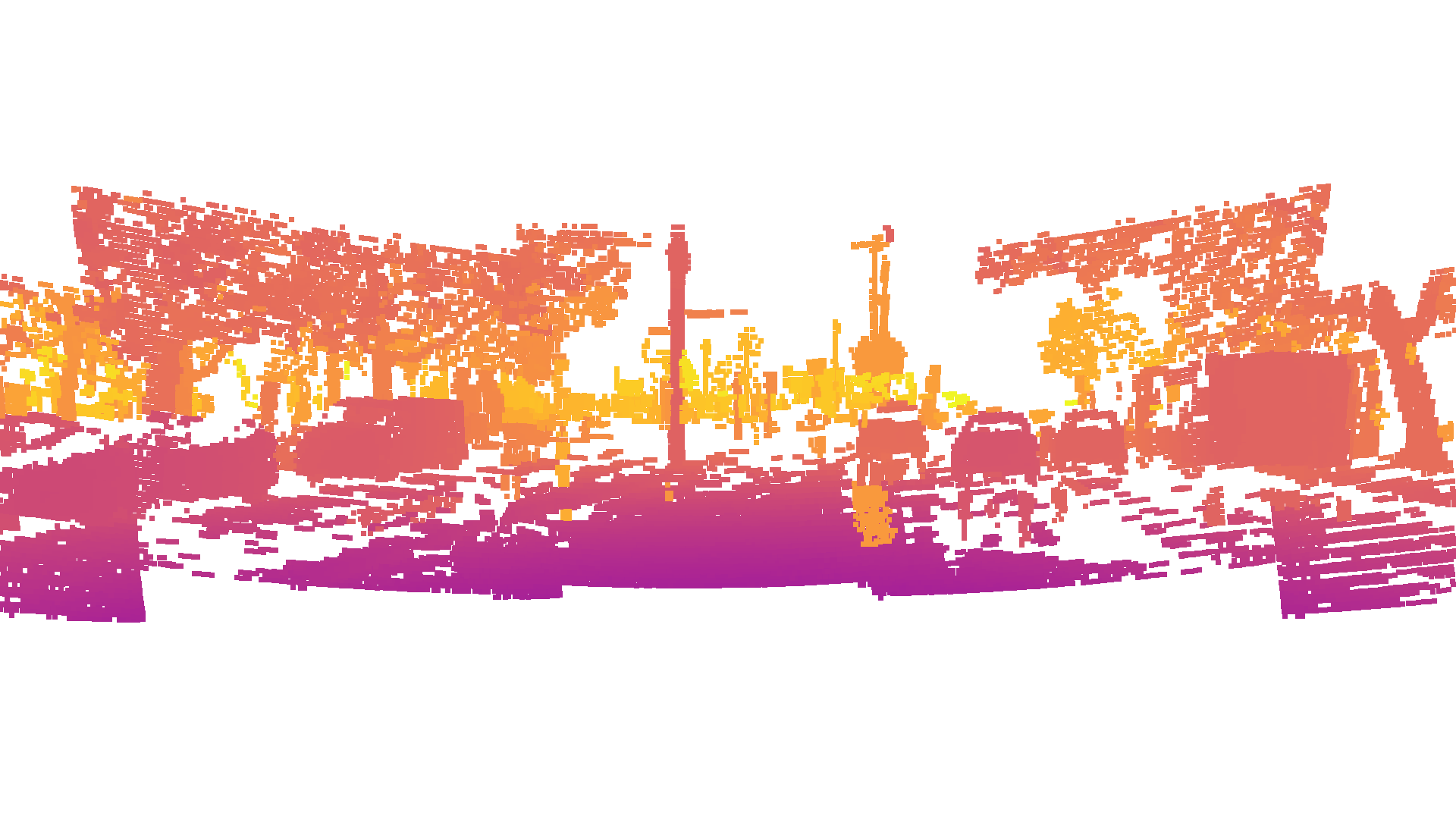} &
\includegraphics[width=0.10\textwidth]{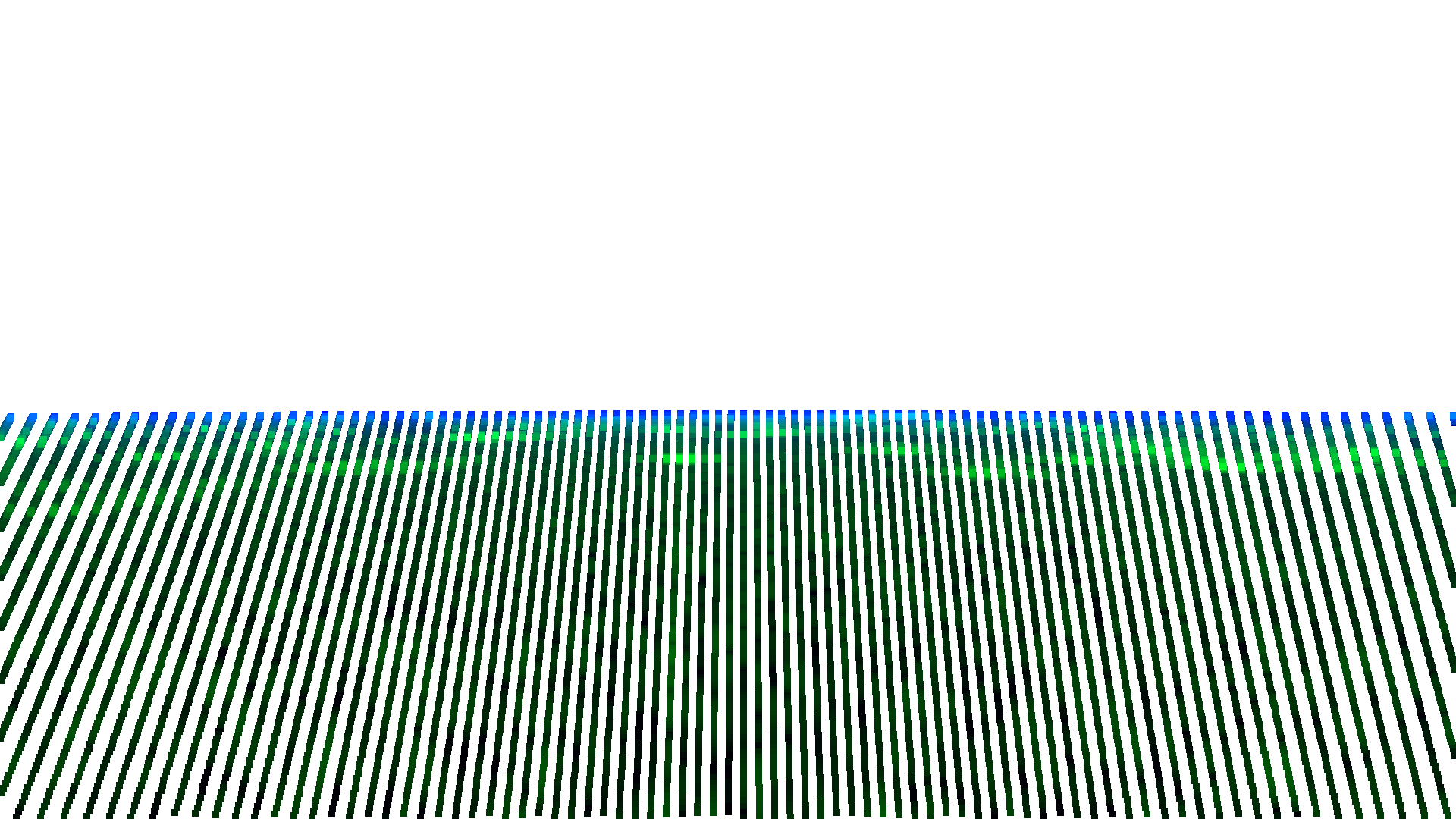} &
\includegraphics[width=0.10\textwidth]{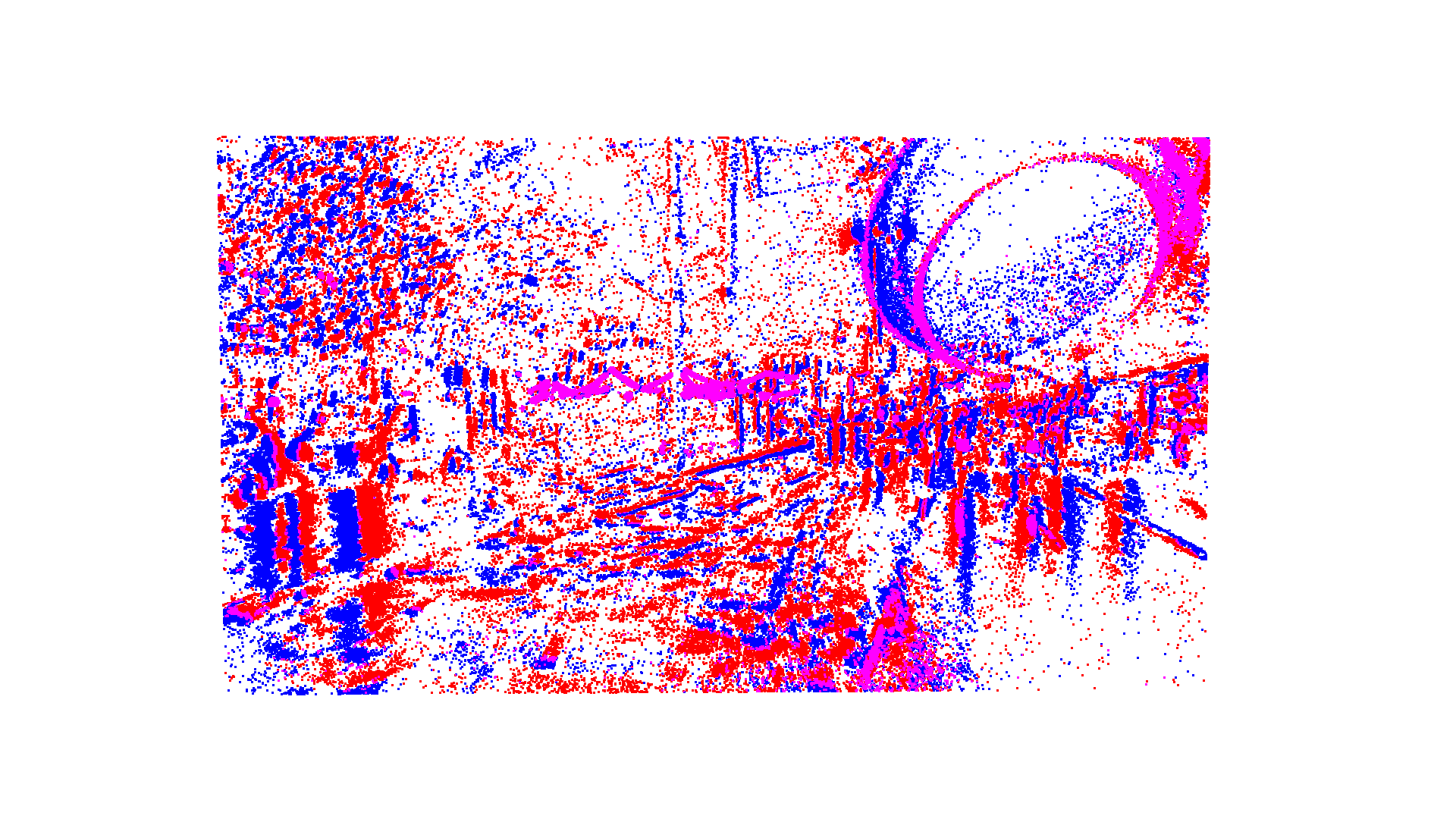} &
\includegraphics[width=0.10\textwidth]{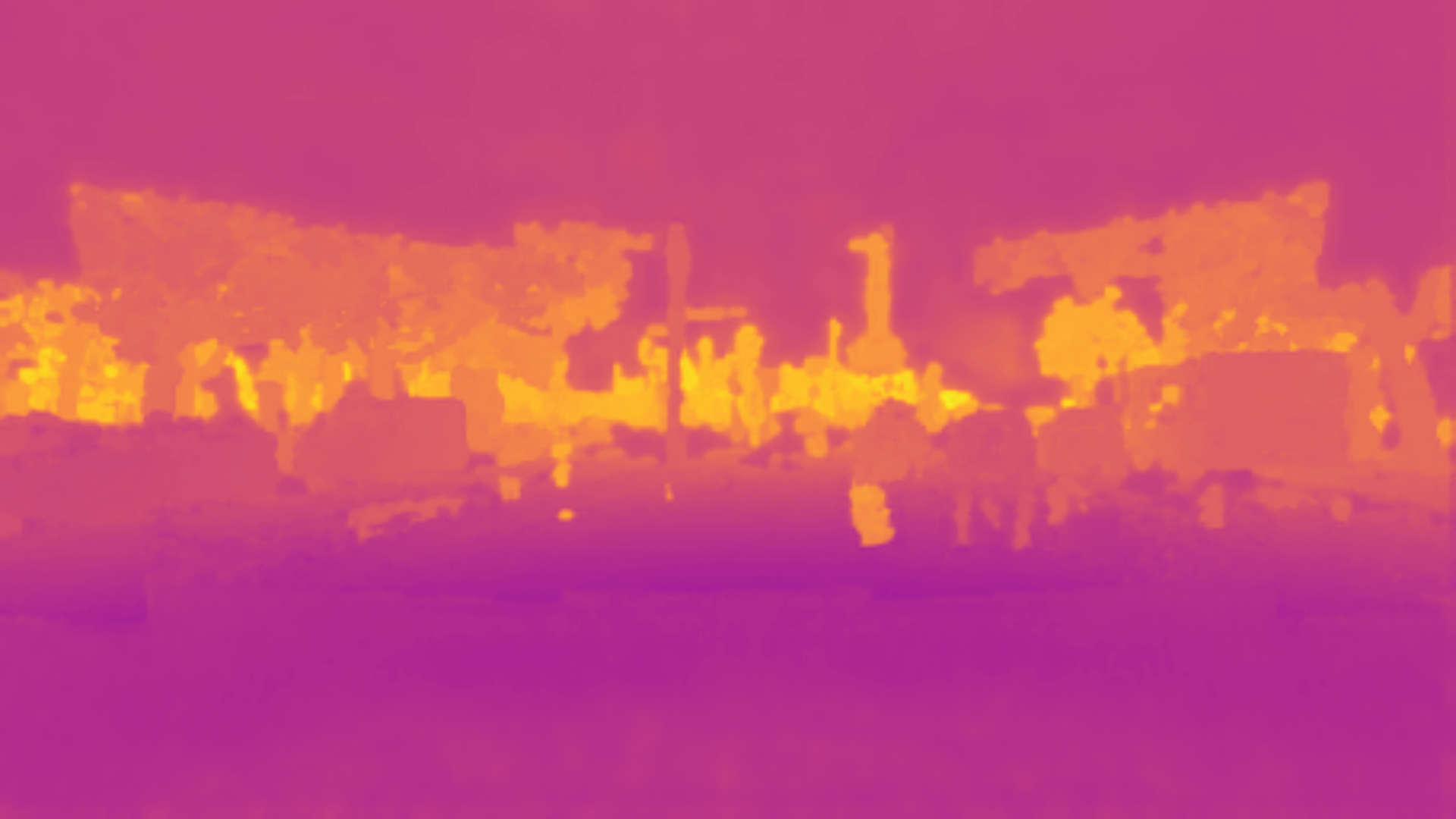} &
\includegraphics[width=0.10\textwidth]{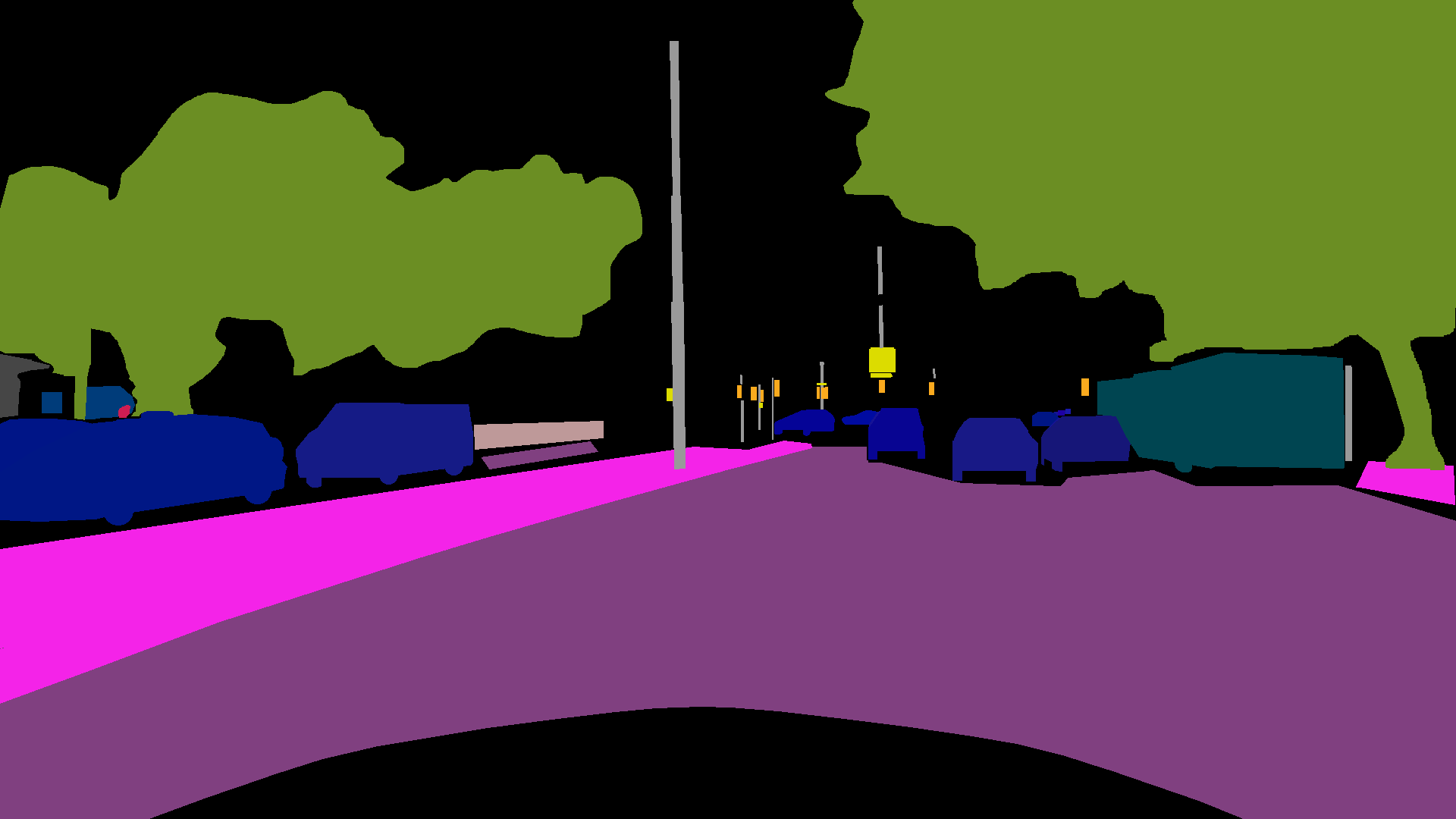} &
\includegraphics[width=0.10\textwidth]{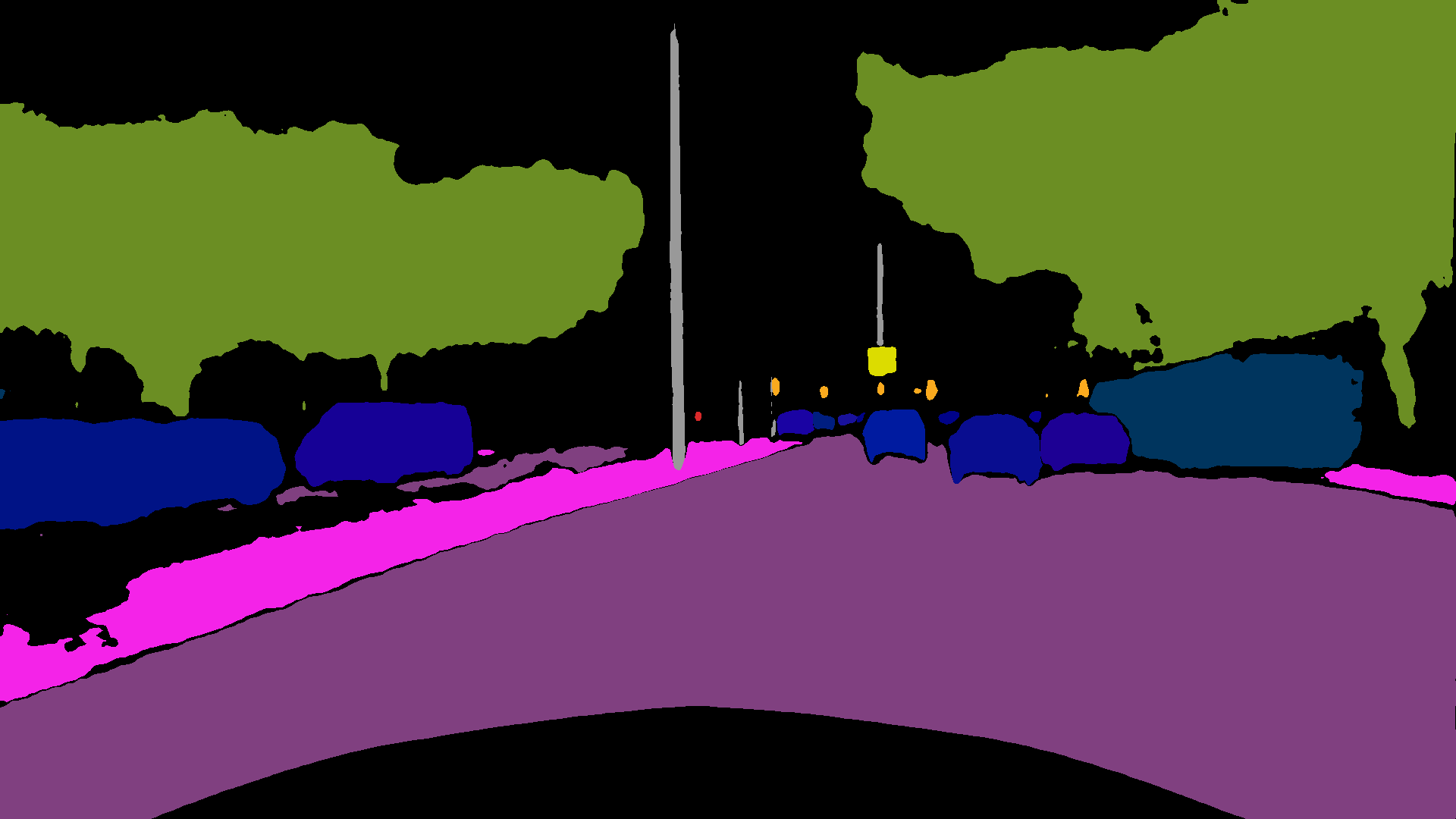} &
\includegraphics[width=0.10\textwidth]{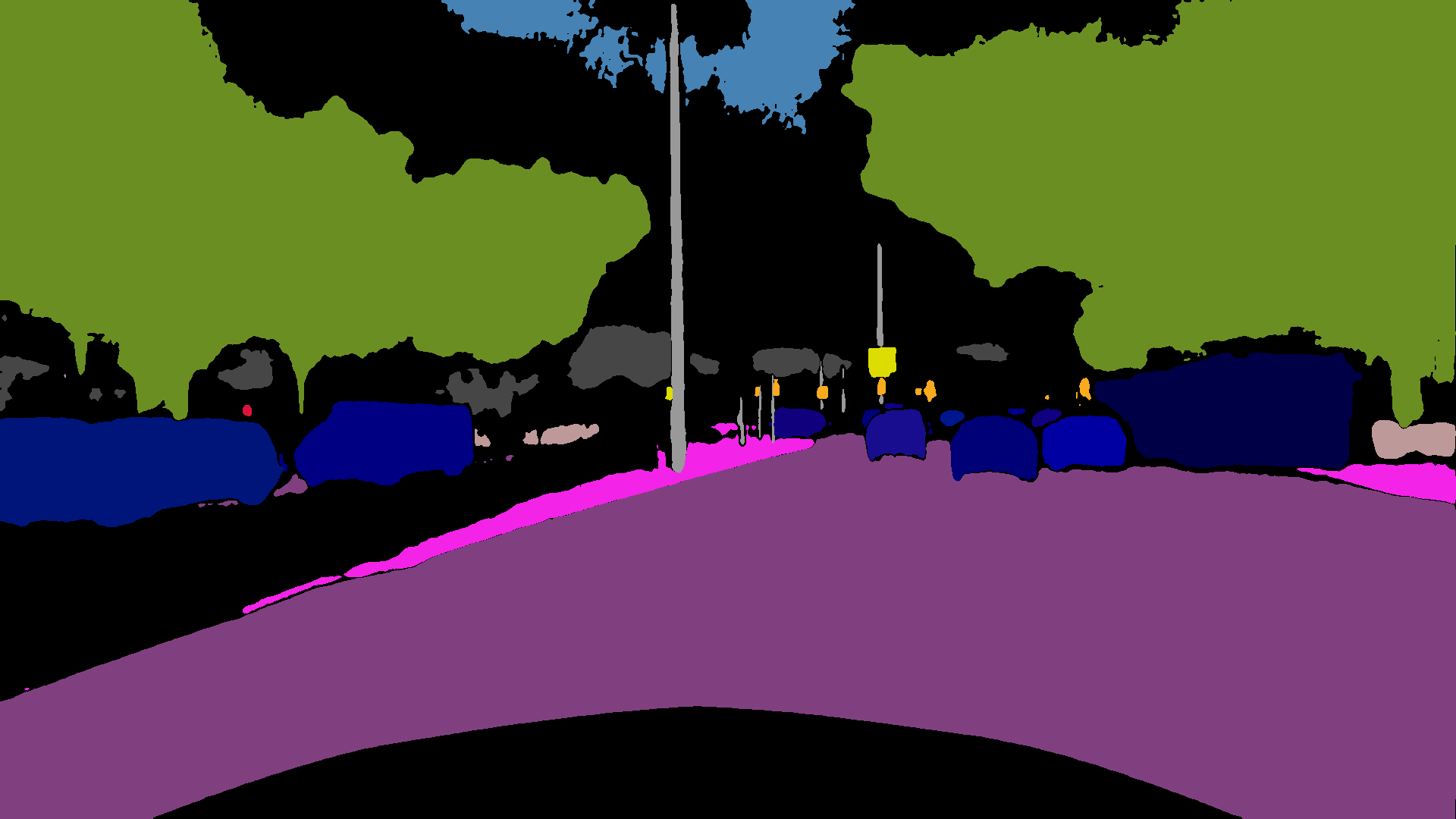} &
\includegraphics[width=0.10\textwidth]{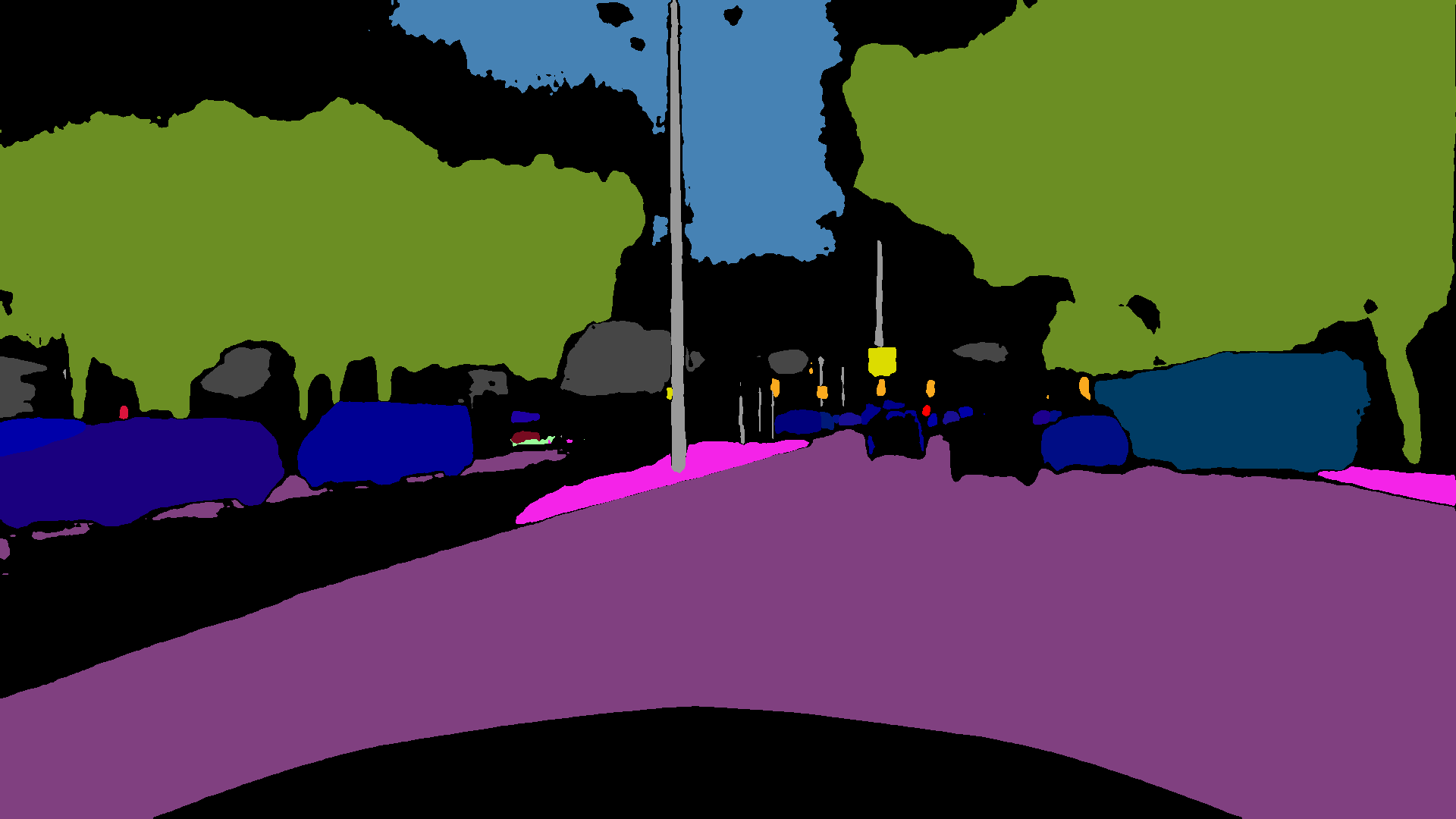} &
\includegraphics[width=0.10\textwidth]{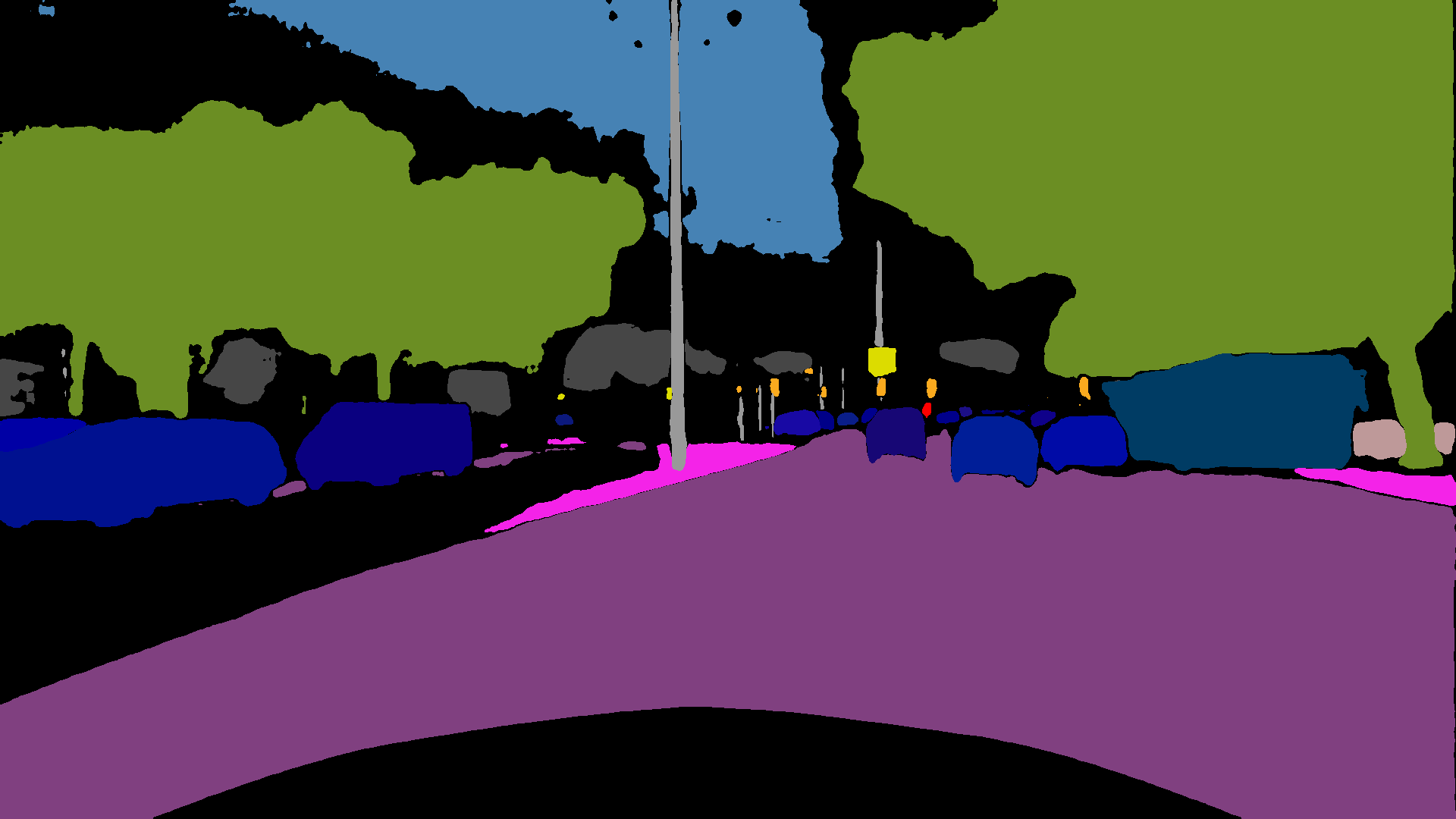} \\
\vspace{-0.1cm}

\includegraphics[width=0.10\textwidth]{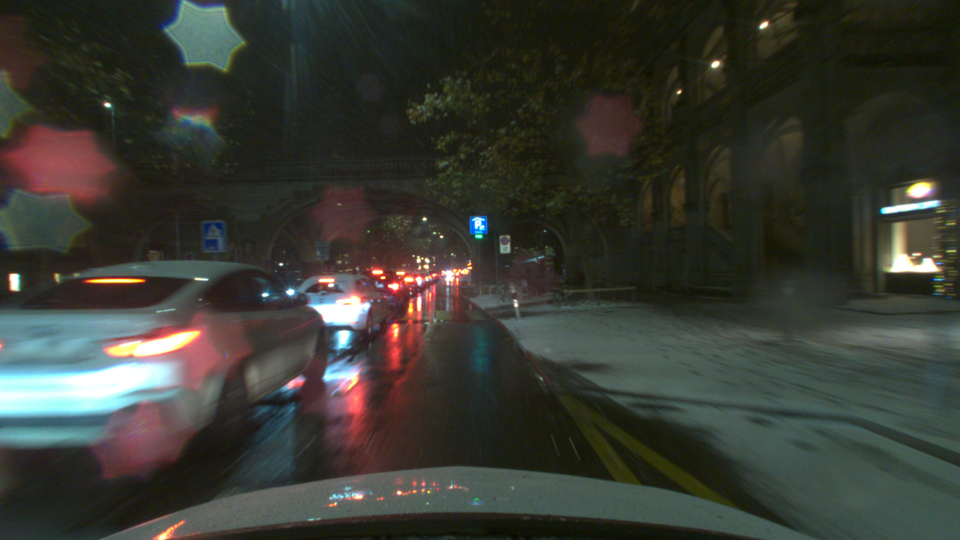} &
\includegraphics[width=0.10\textwidth]{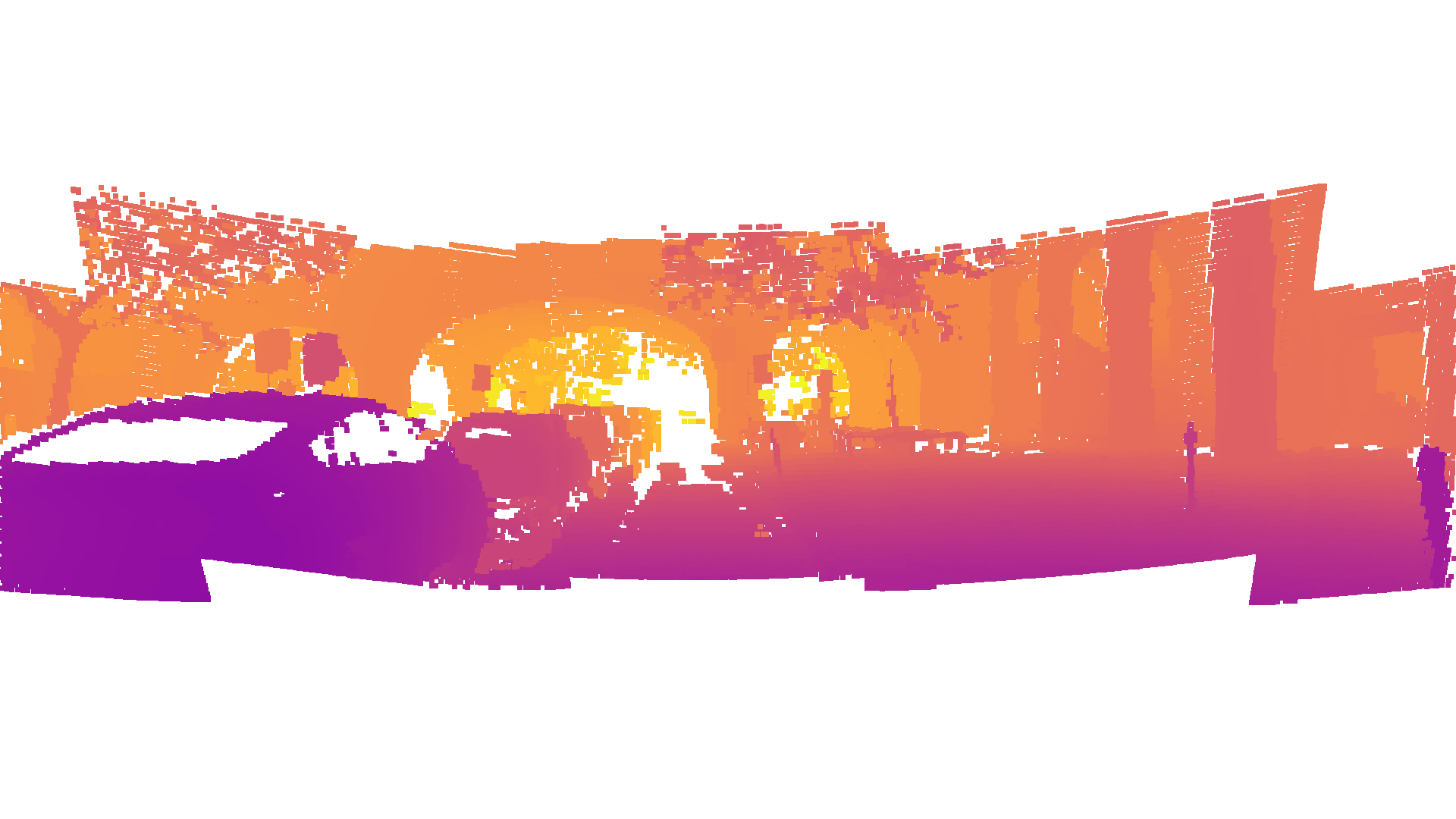} &
\includegraphics[width=0.10\textwidth]{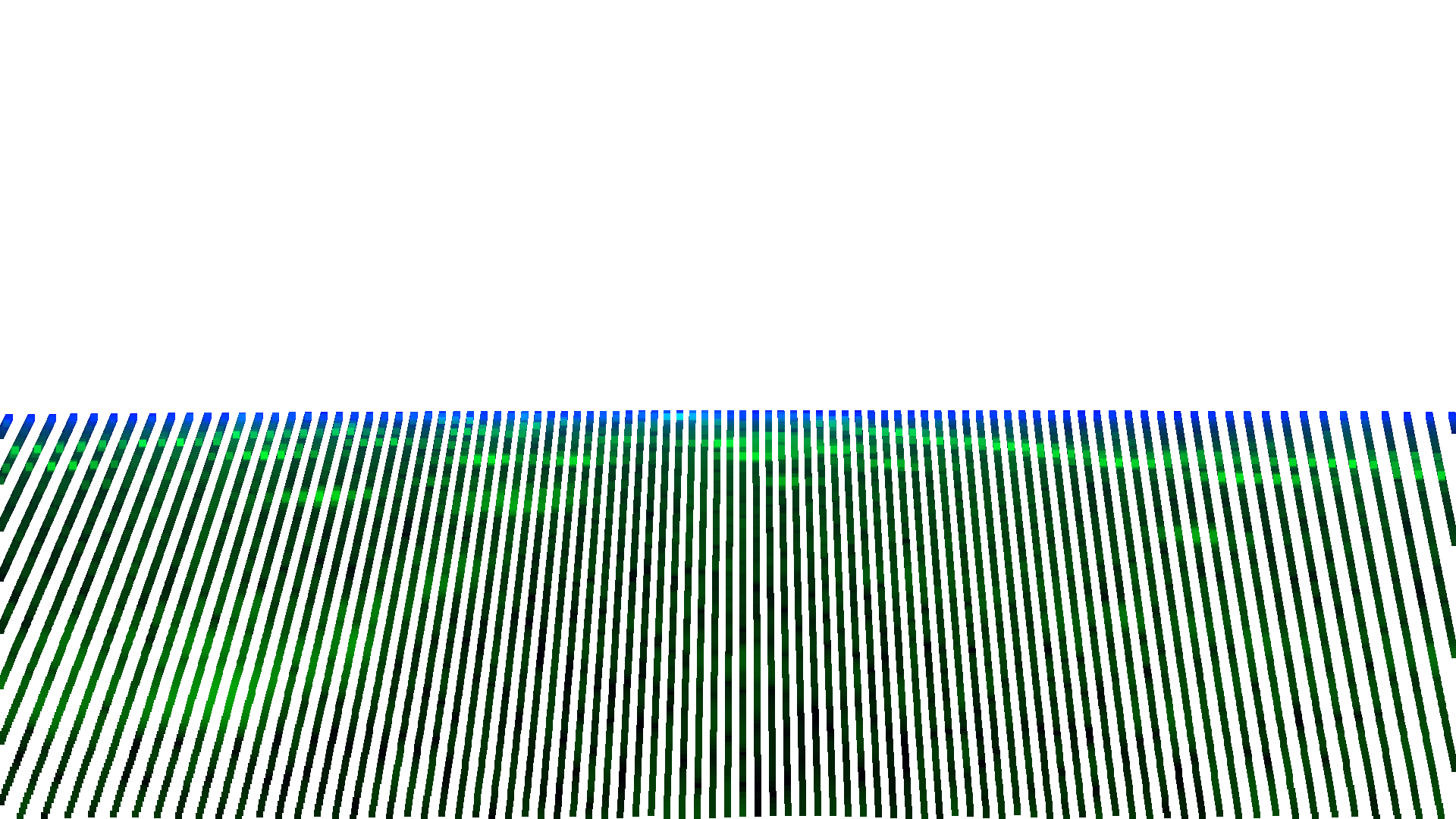} &
\includegraphics[width=0.10\textwidth]{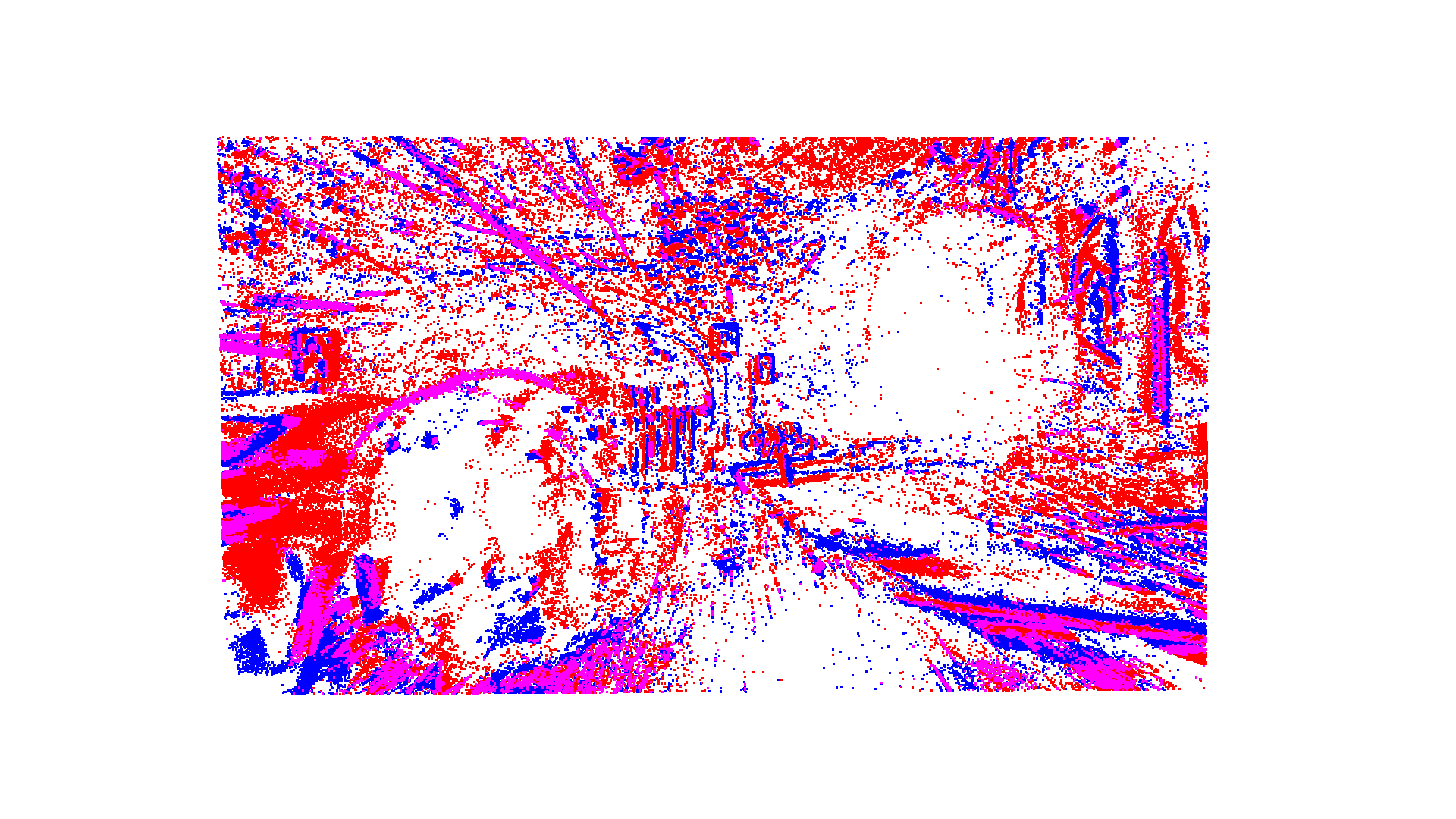} &
\includegraphics[width=0.10\textwidth]{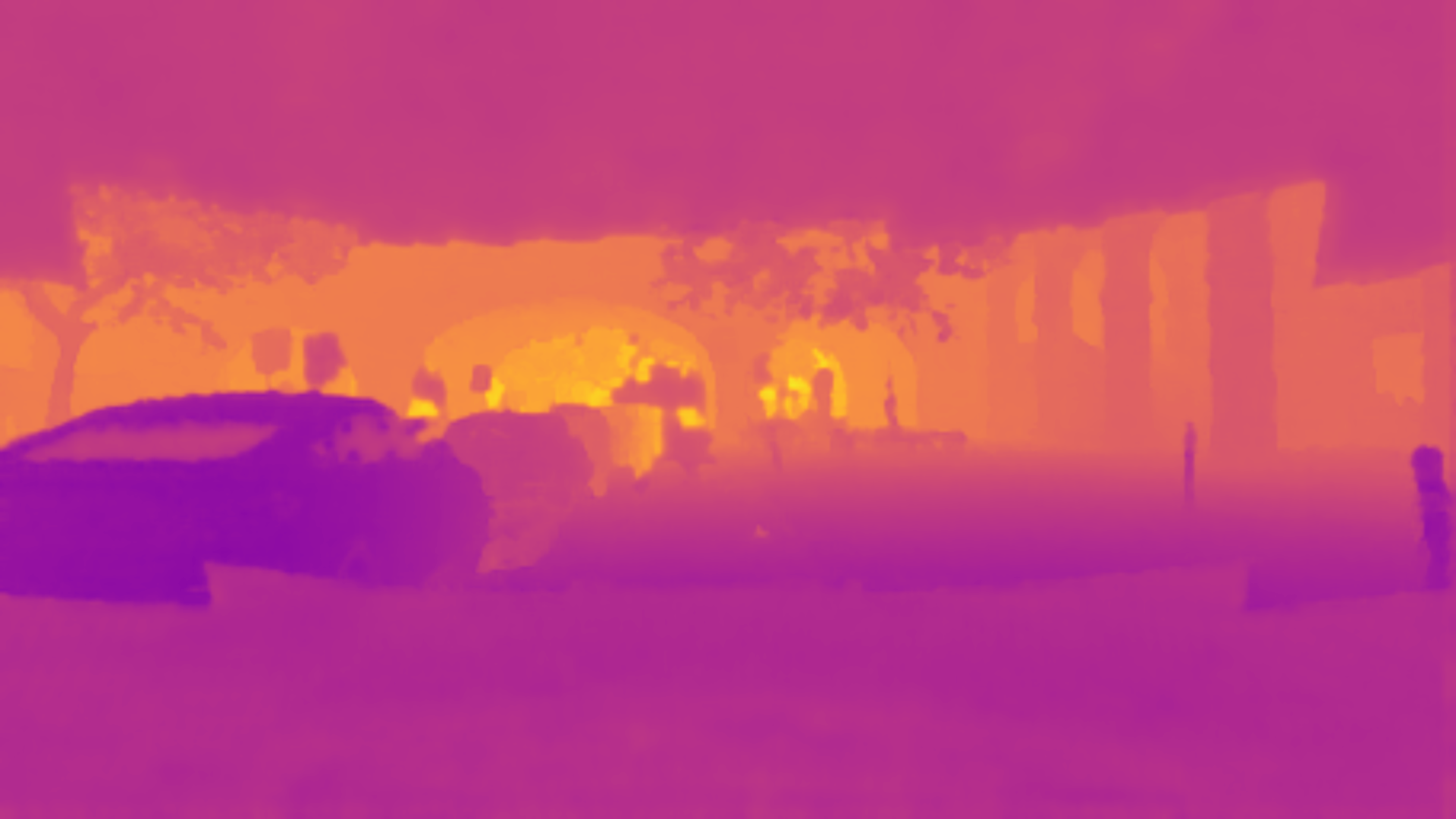} &
\includegraphics[width=0.10\textwidth]{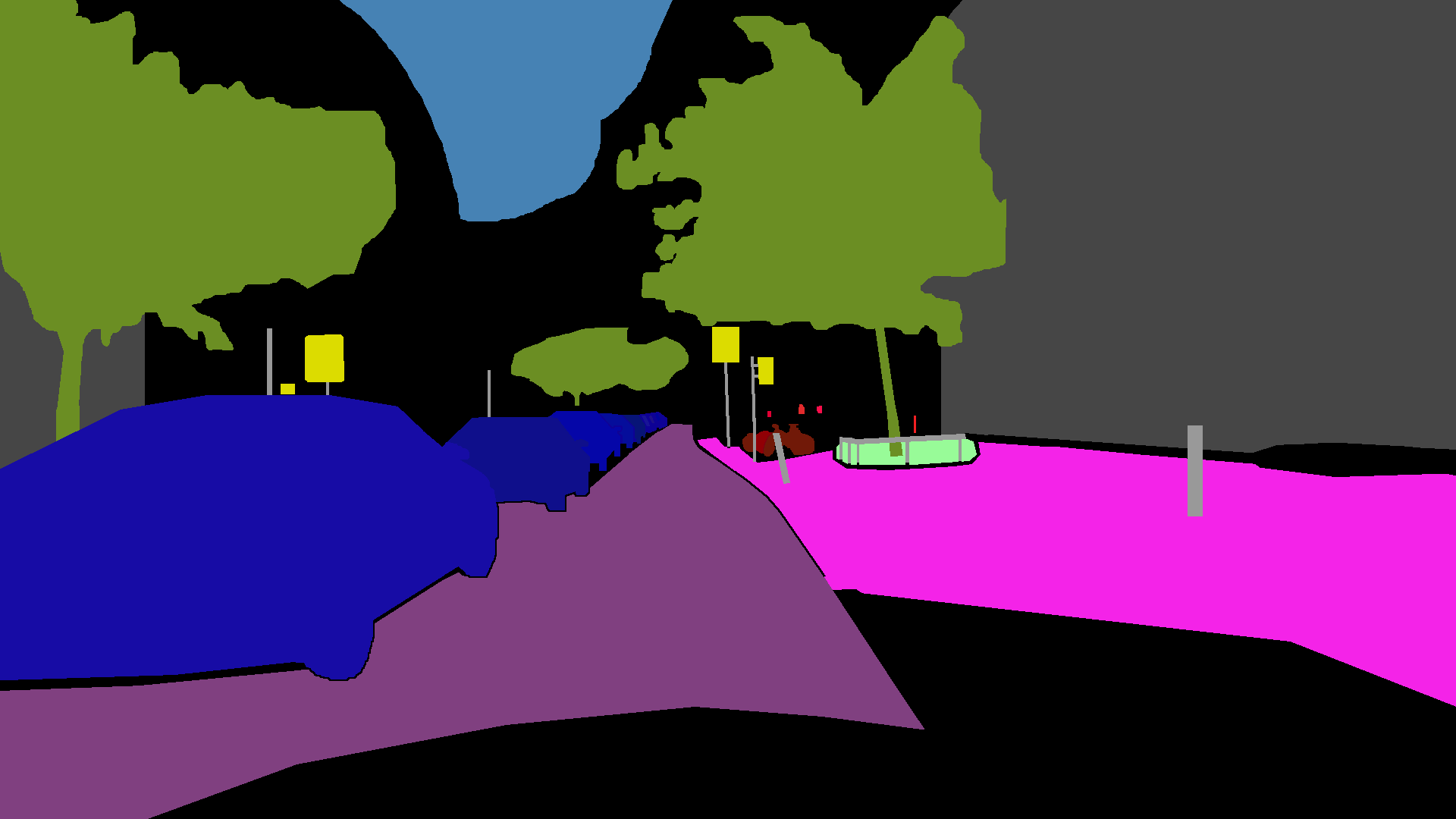} &
\includegraphics[width=0.10\textwidth]{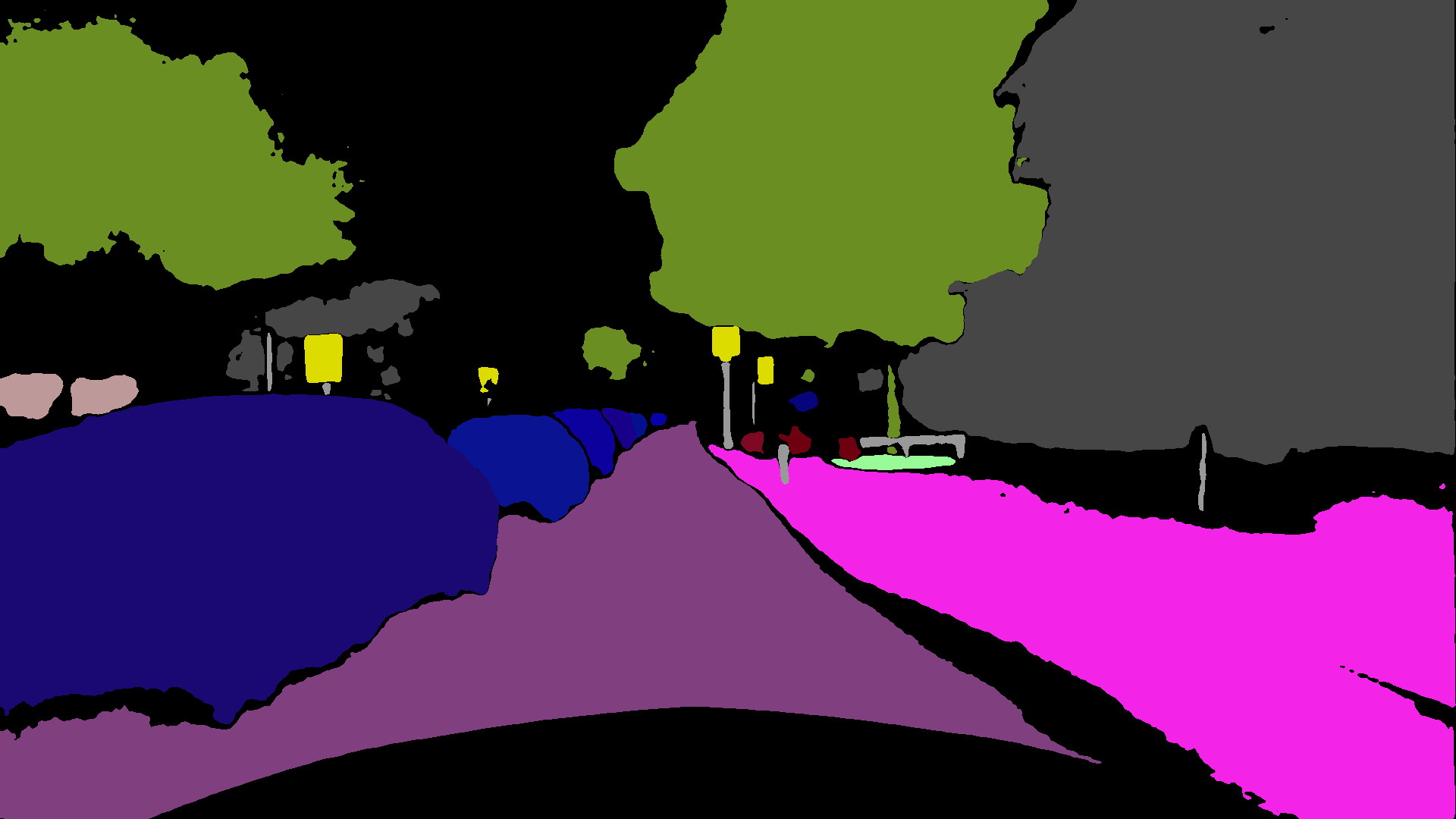} &
\includegraphics[width=0.10\textwidth]{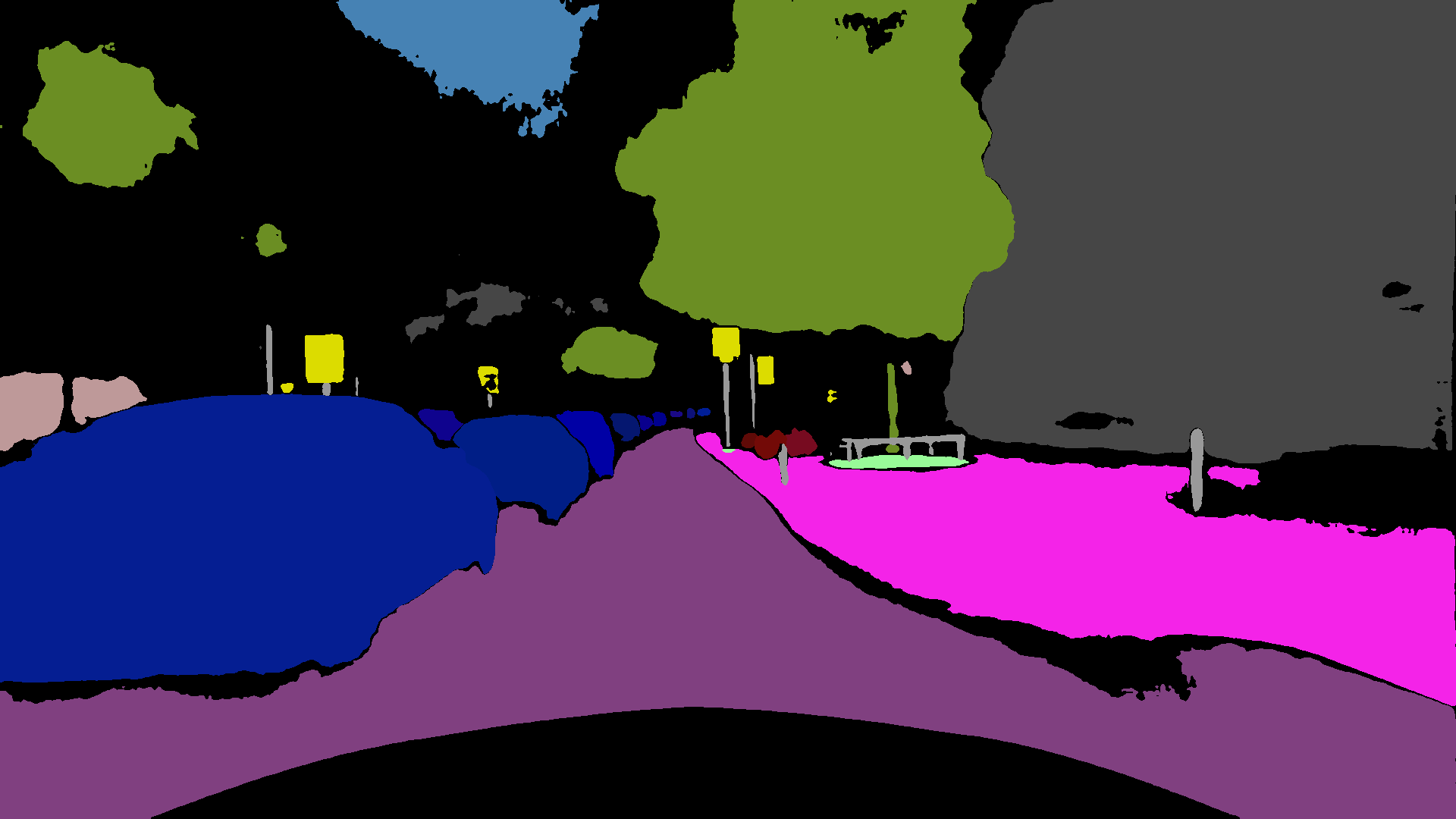} &
\includegraphics[width=0.10\textwidth]{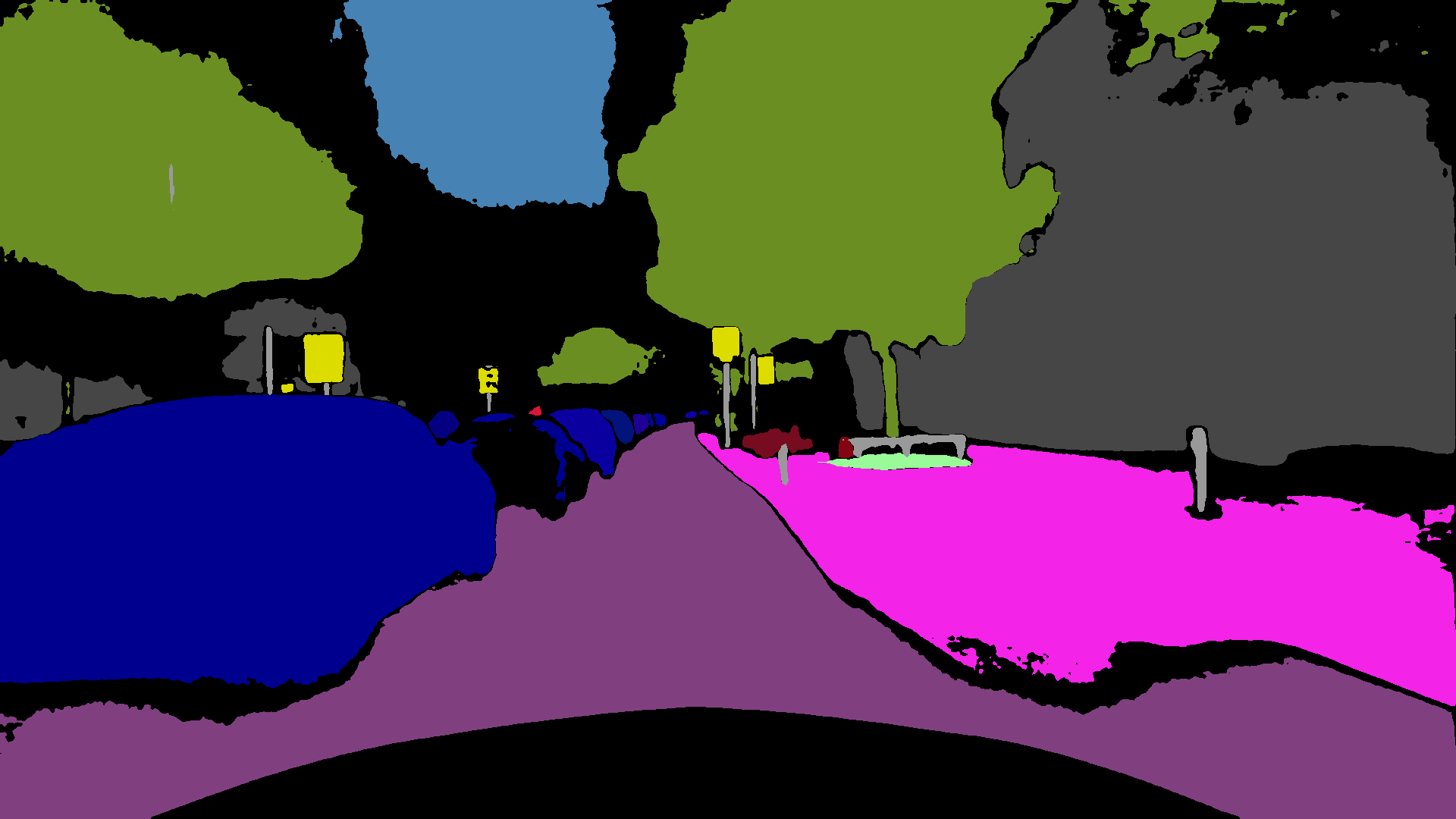} &
\includegraphics[width=0.10\textwidth]{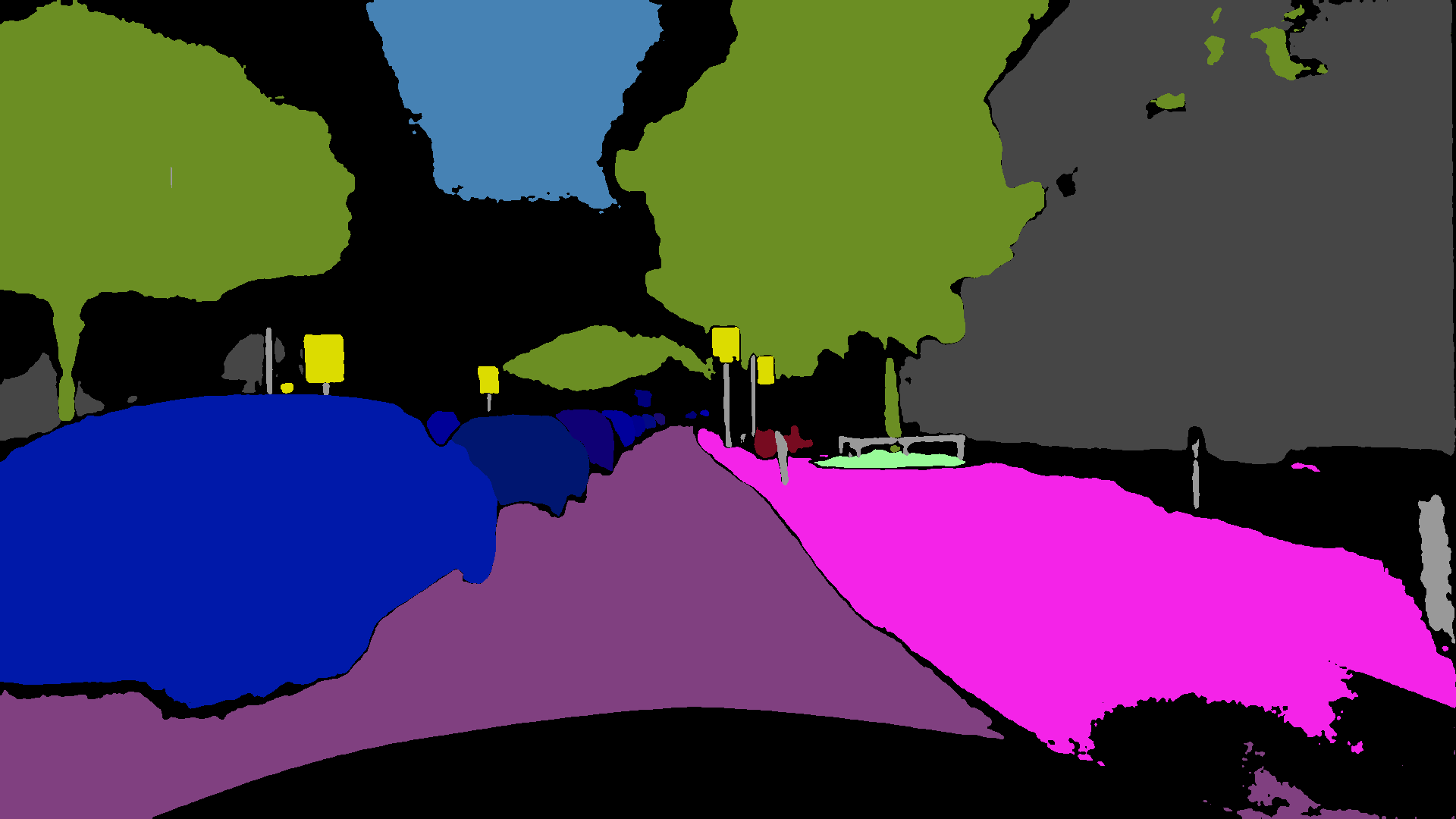} \\
\vspace{-0.1cm}

\includegraphics[width=0.10\textwidth]{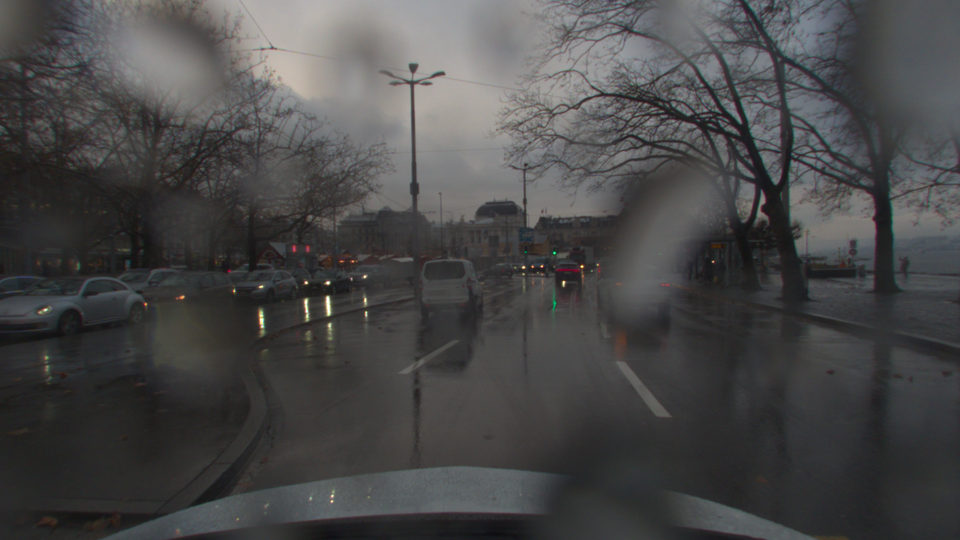} &
\includegraphics[width=0.10\textwidth]{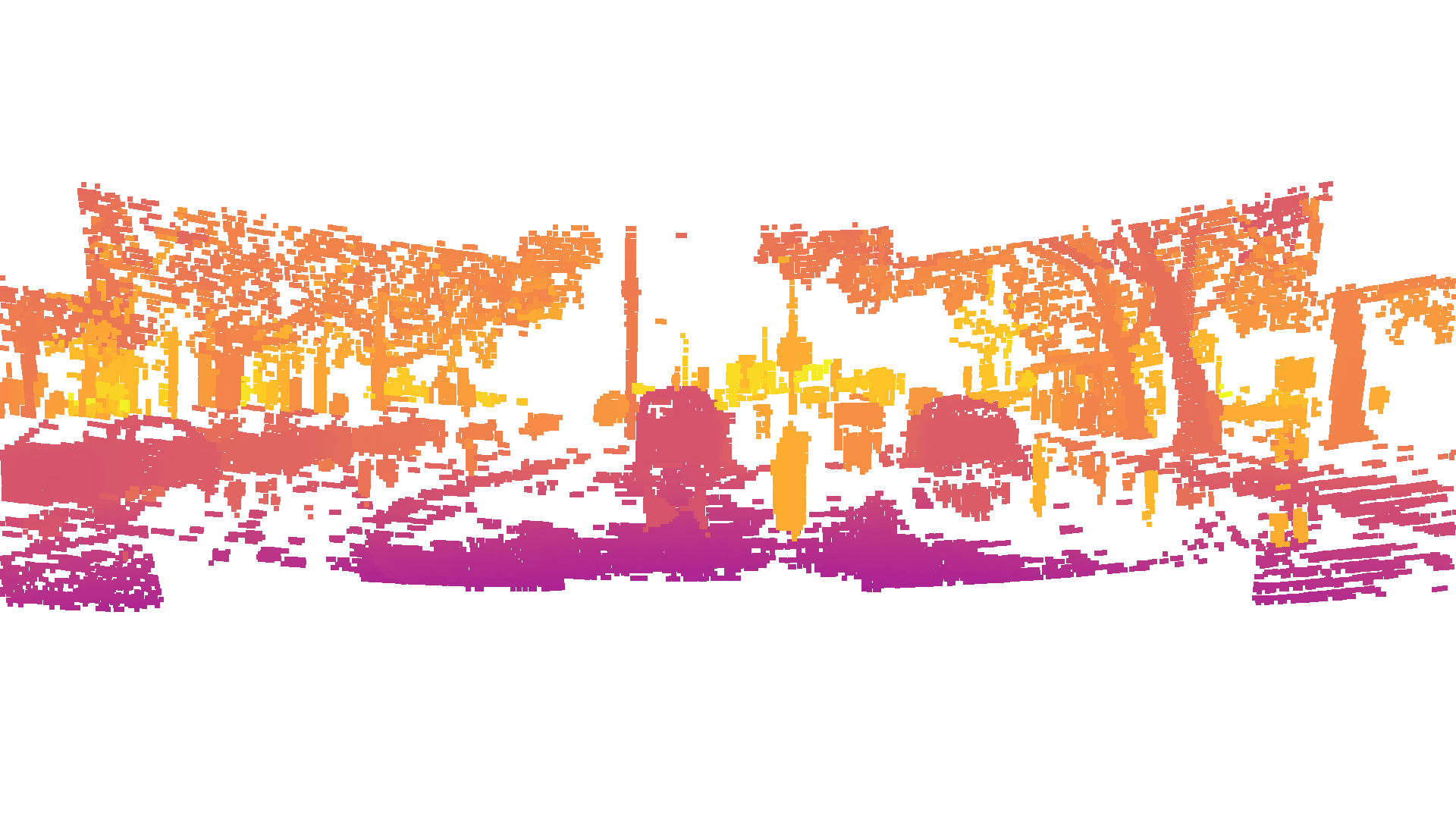} &
\includegraphics[width=0.10\textwidth]{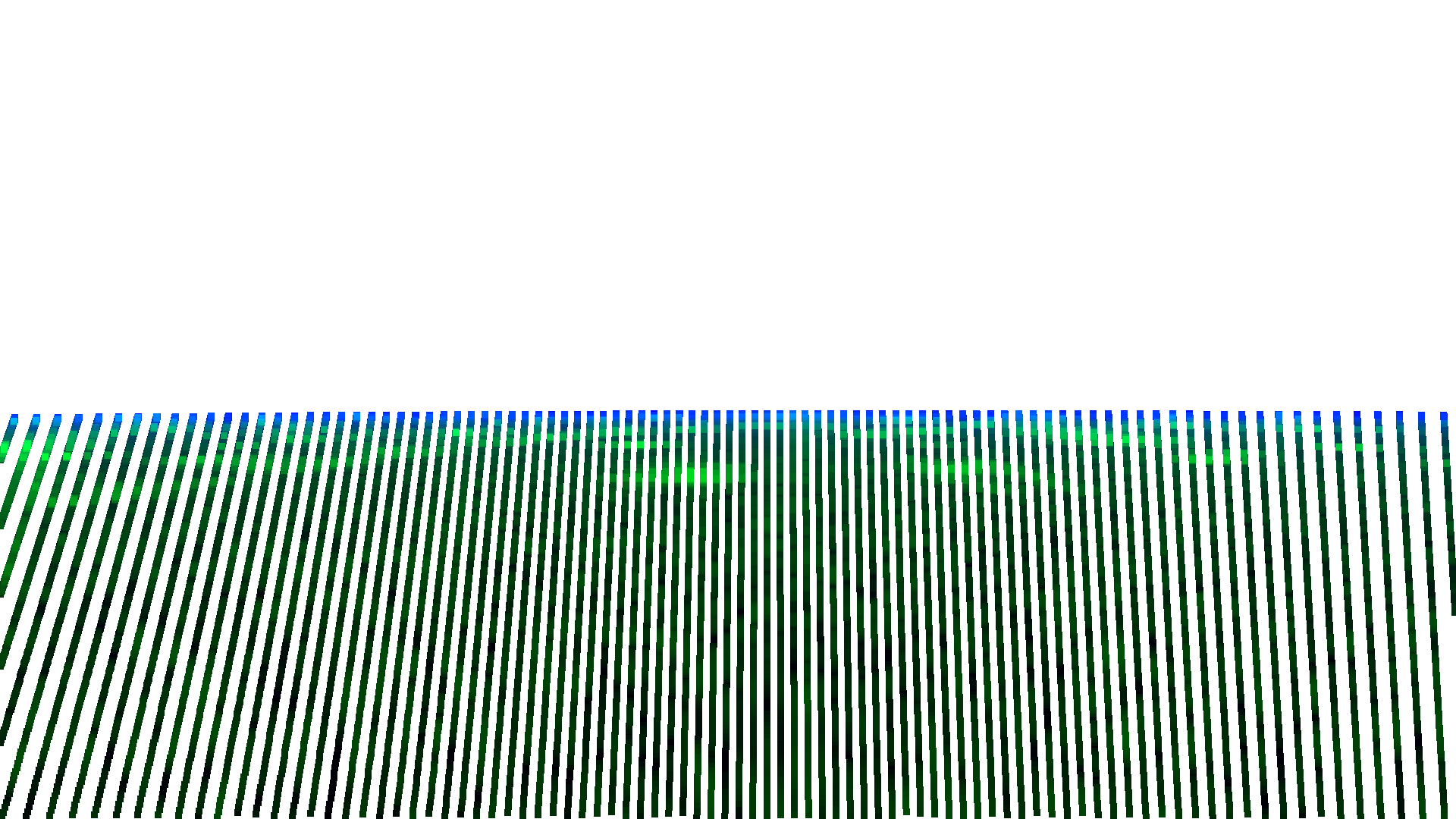} &
\includegraphics[width=0.10\textwidth]{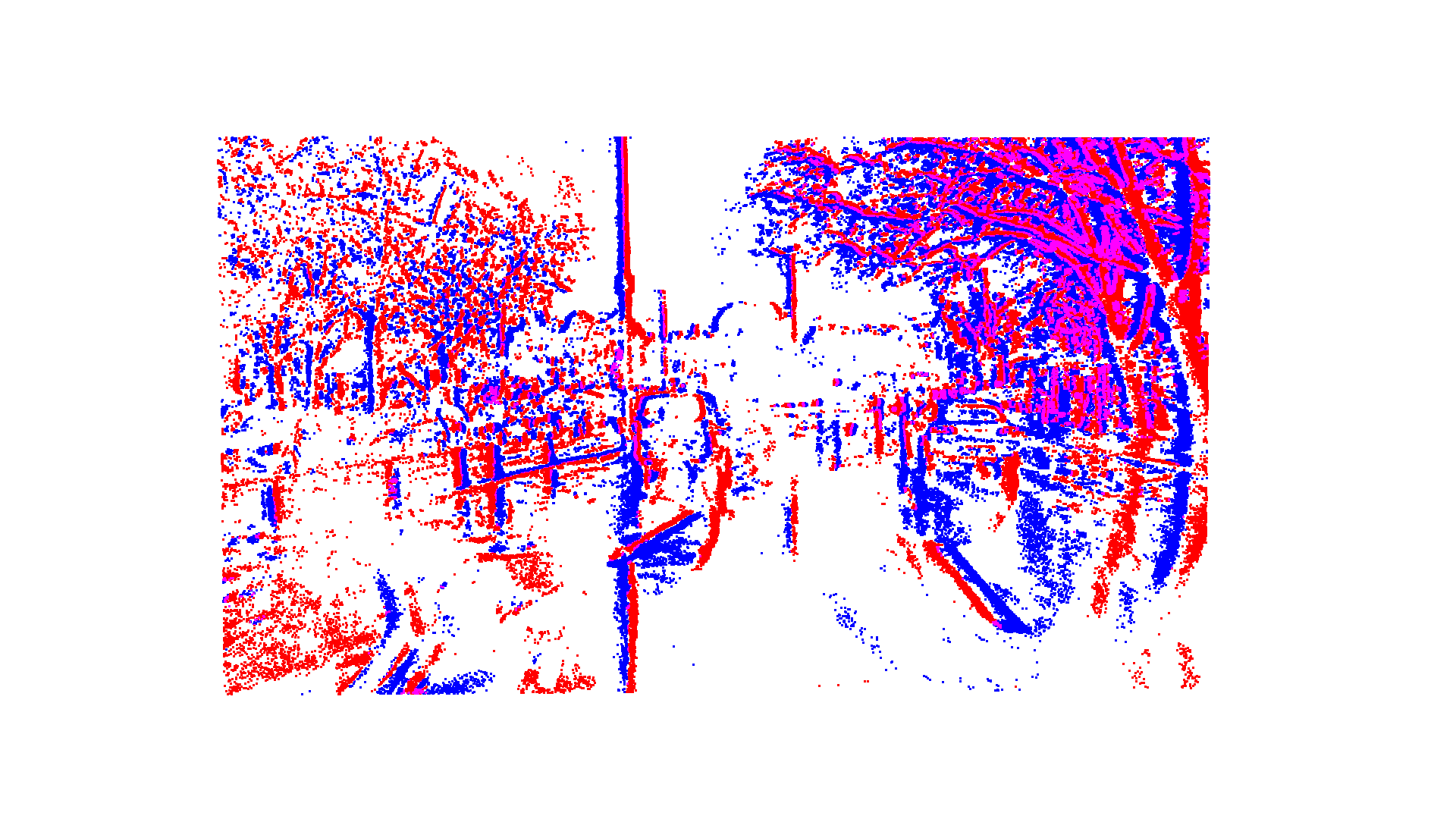} &
\includegraphics[width=0.10\textwidth]{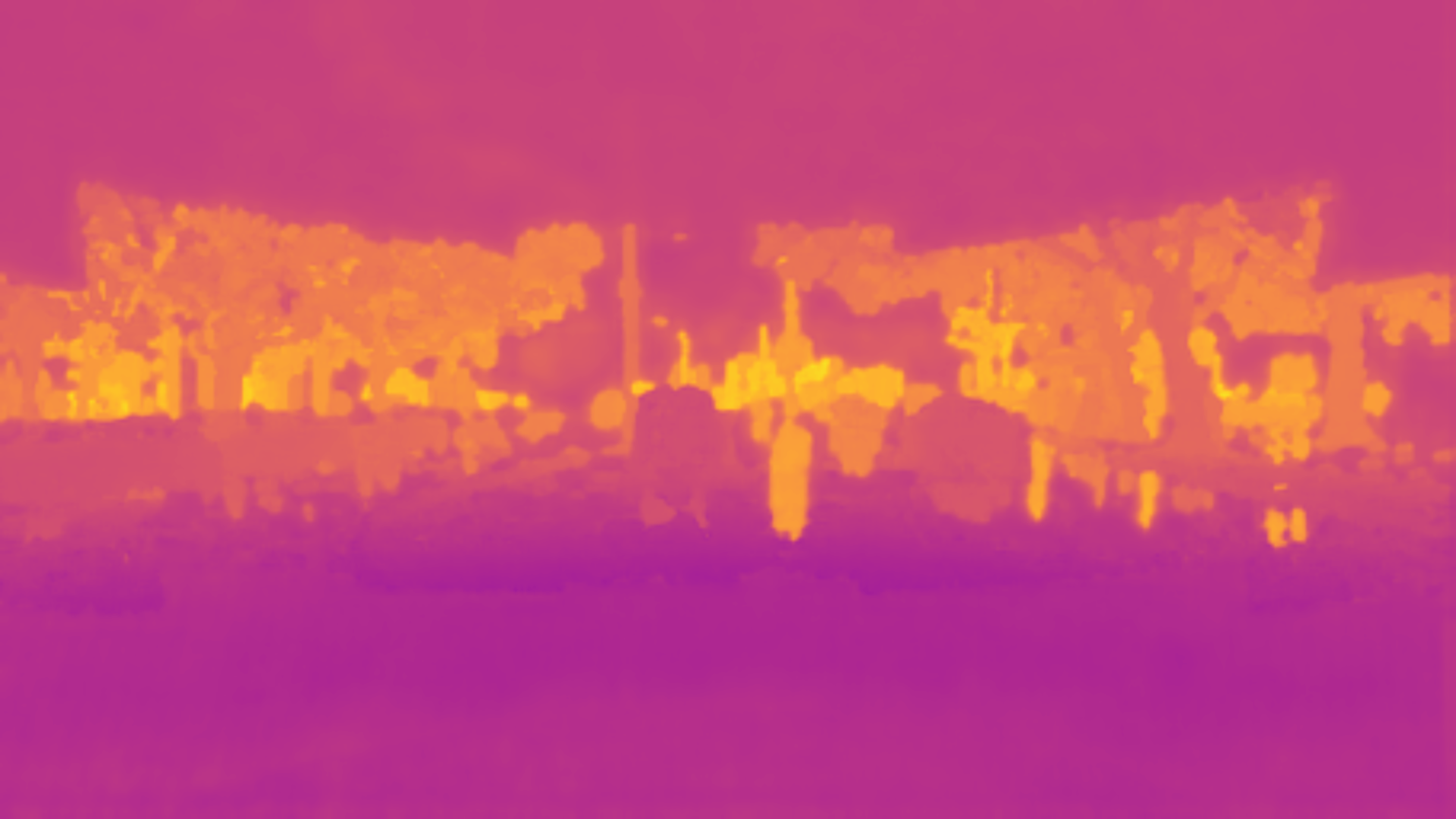} &
\includegraphics[width=0.10\textwidth]{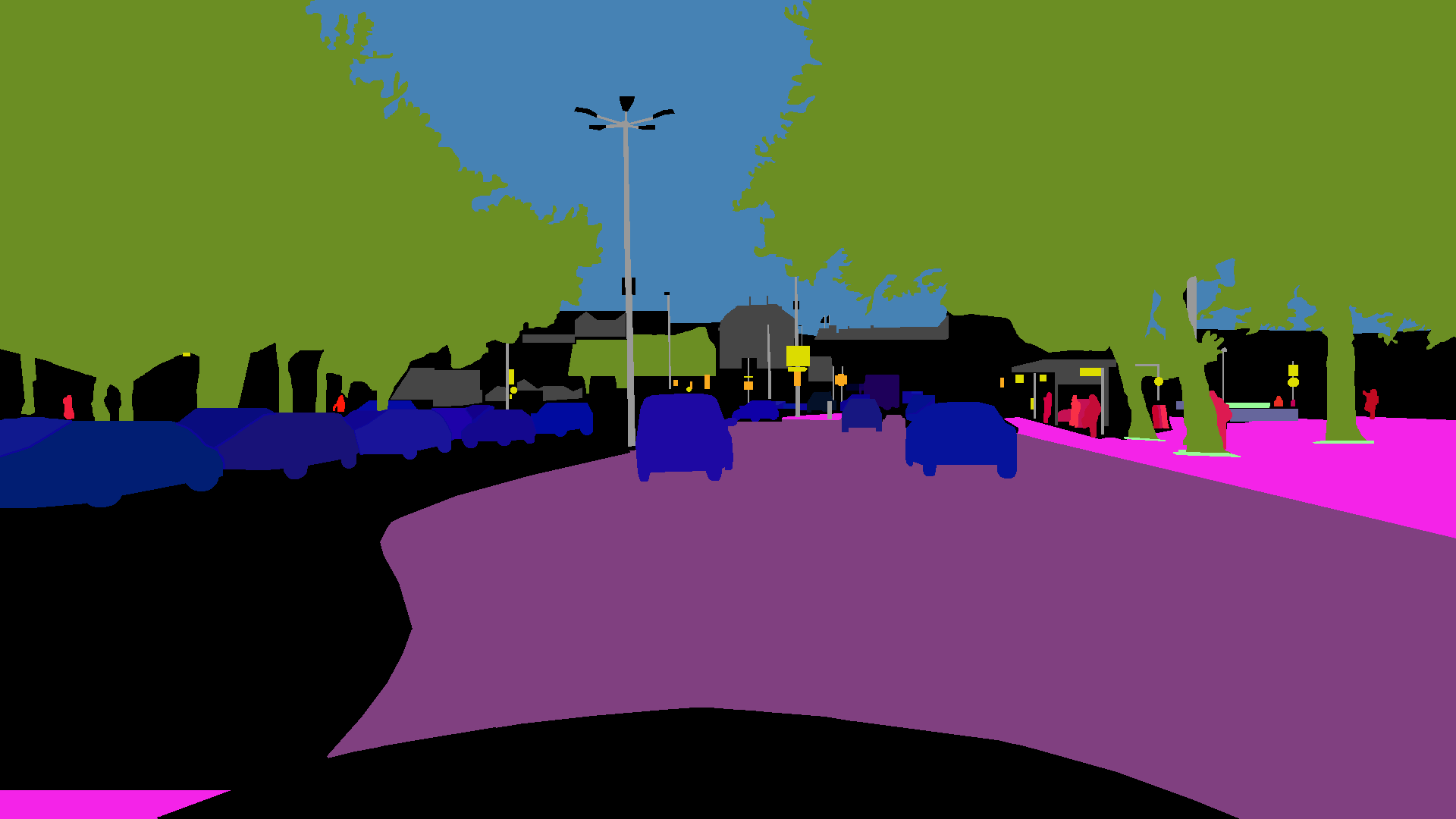} &
\includegraphics[width=0.10\textwidth]{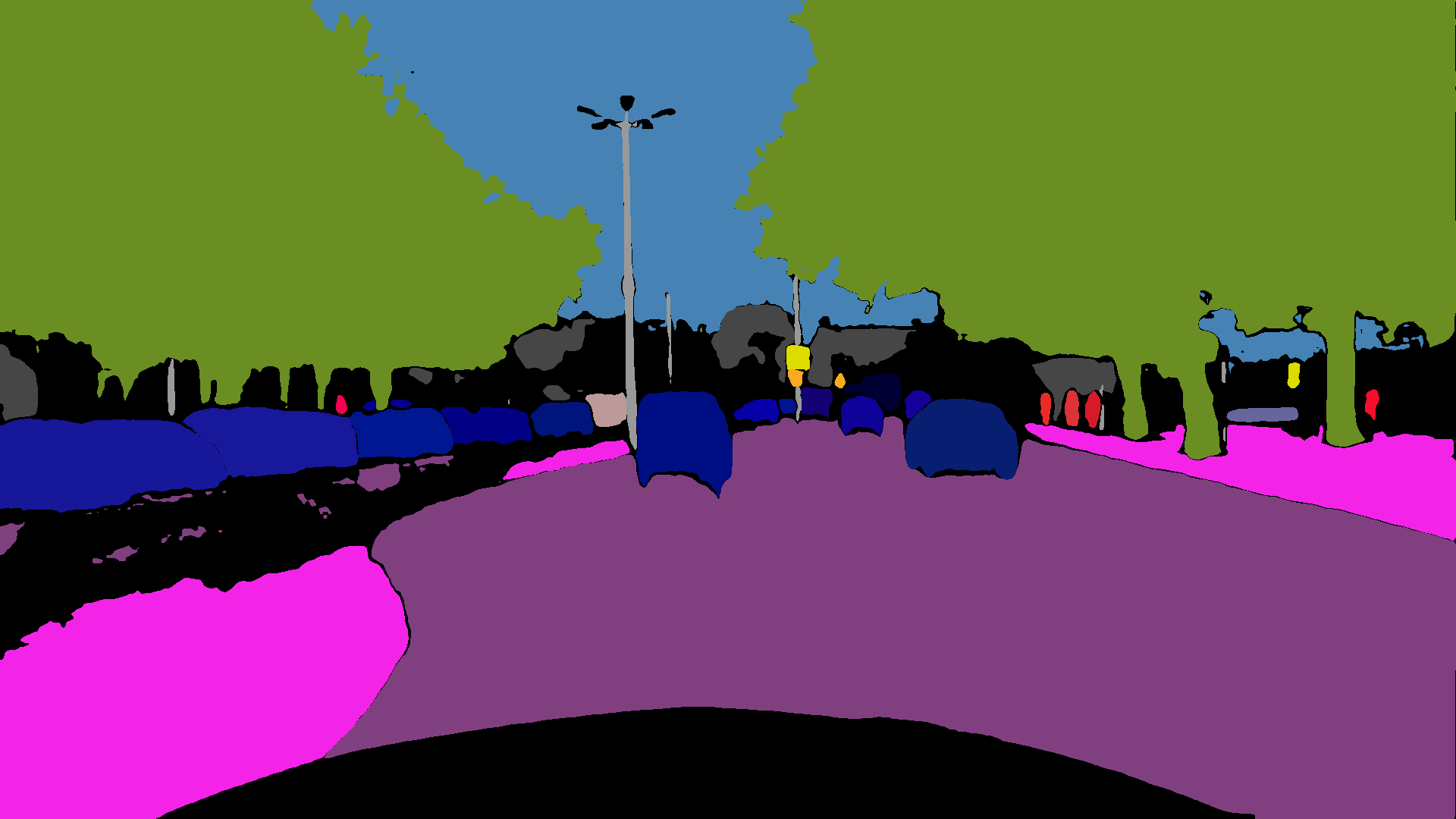} &
\includegraphics[width=0.10\textwidth]{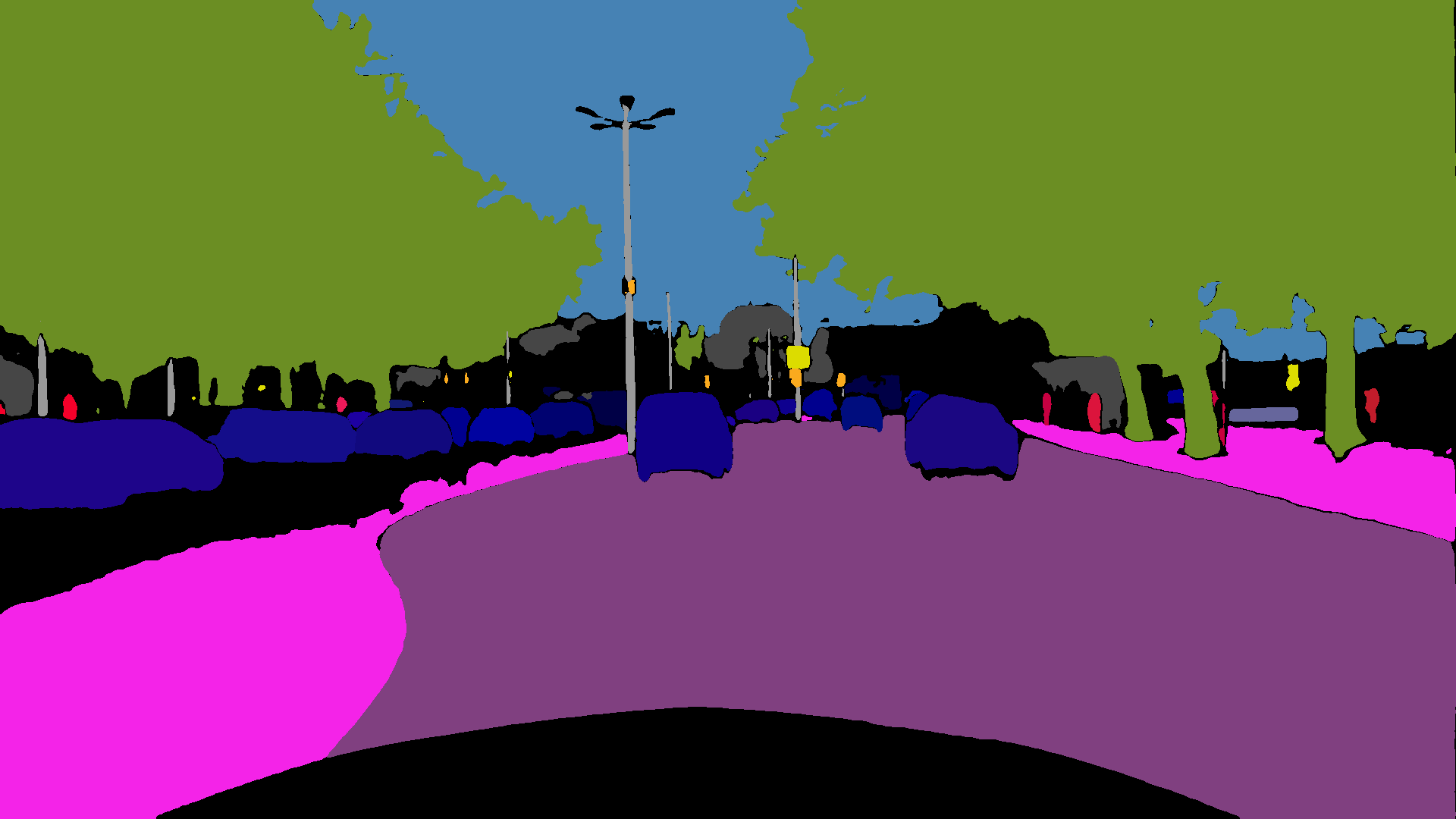} &
\includegraphics[width=0.10\textwidth]{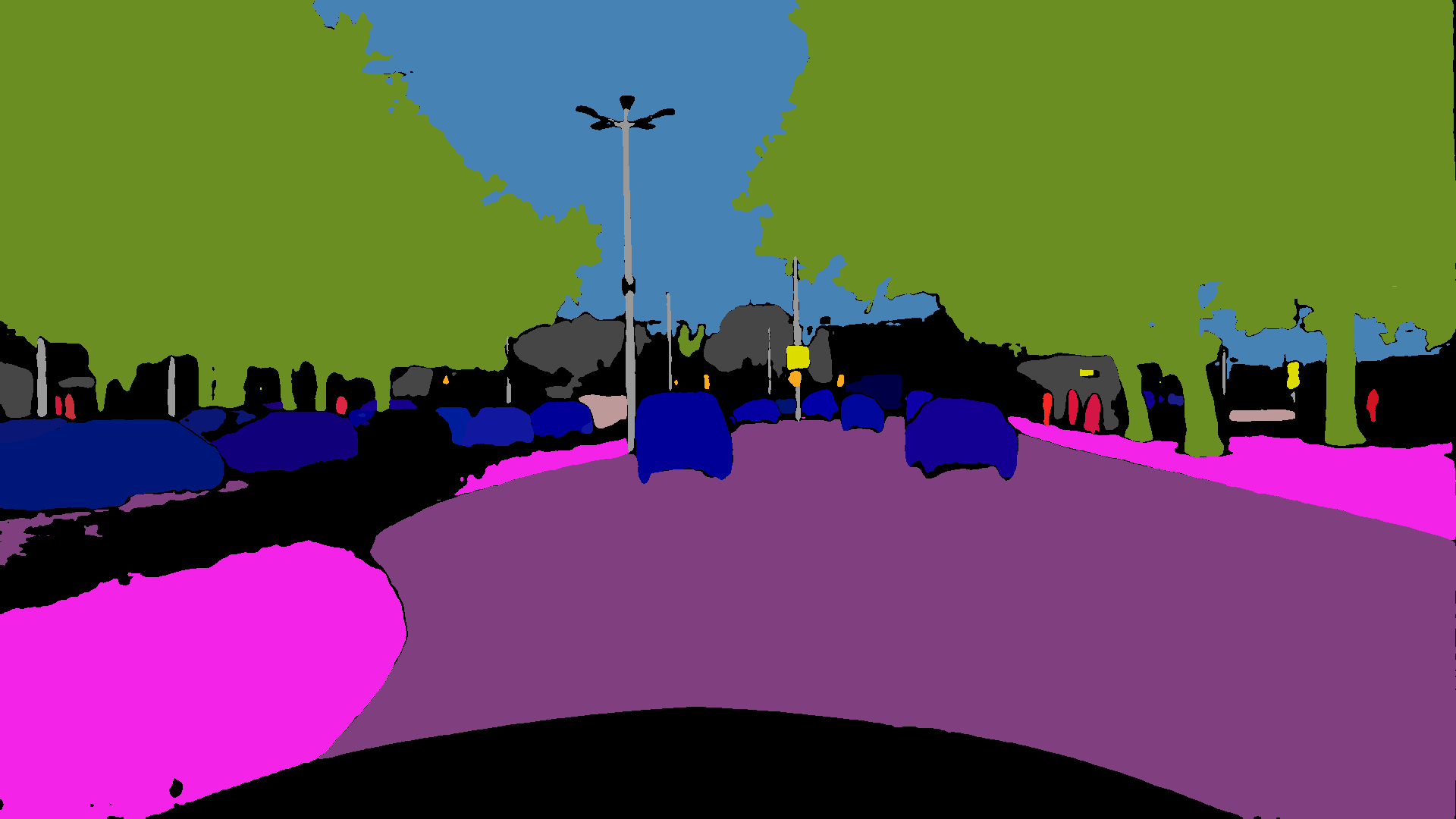} &
\includegraphics[width=0.10\textwidth]{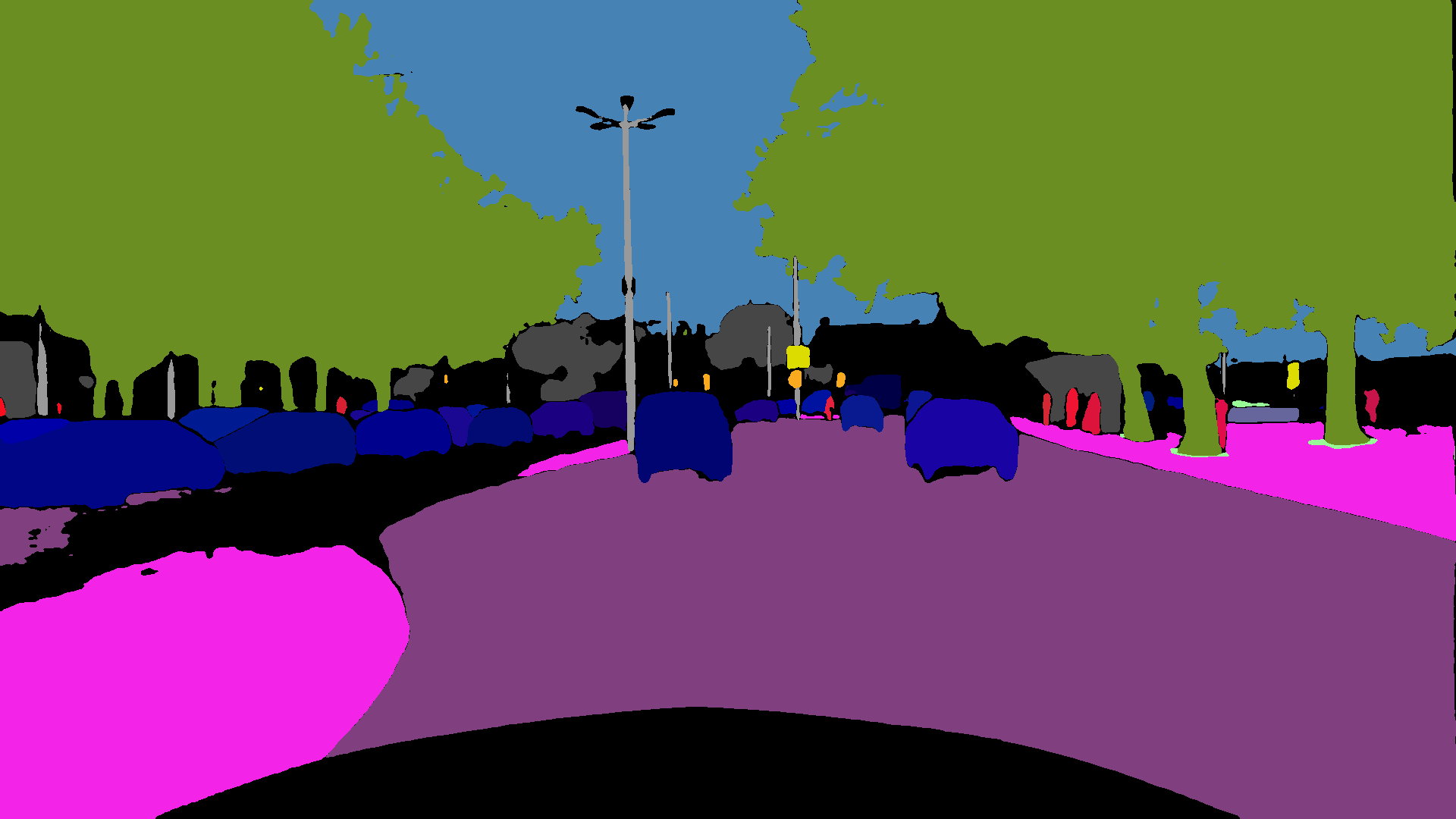} \\
\vspace{-0.1cm}

\includegraphics[width=0.10\textwidth]{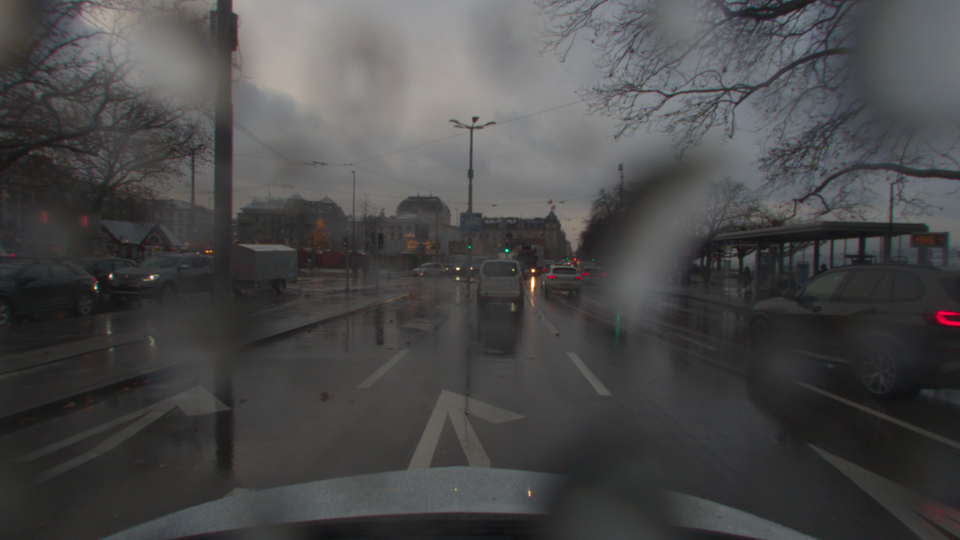} &
\includegraphics[width=0.10\textwidth]{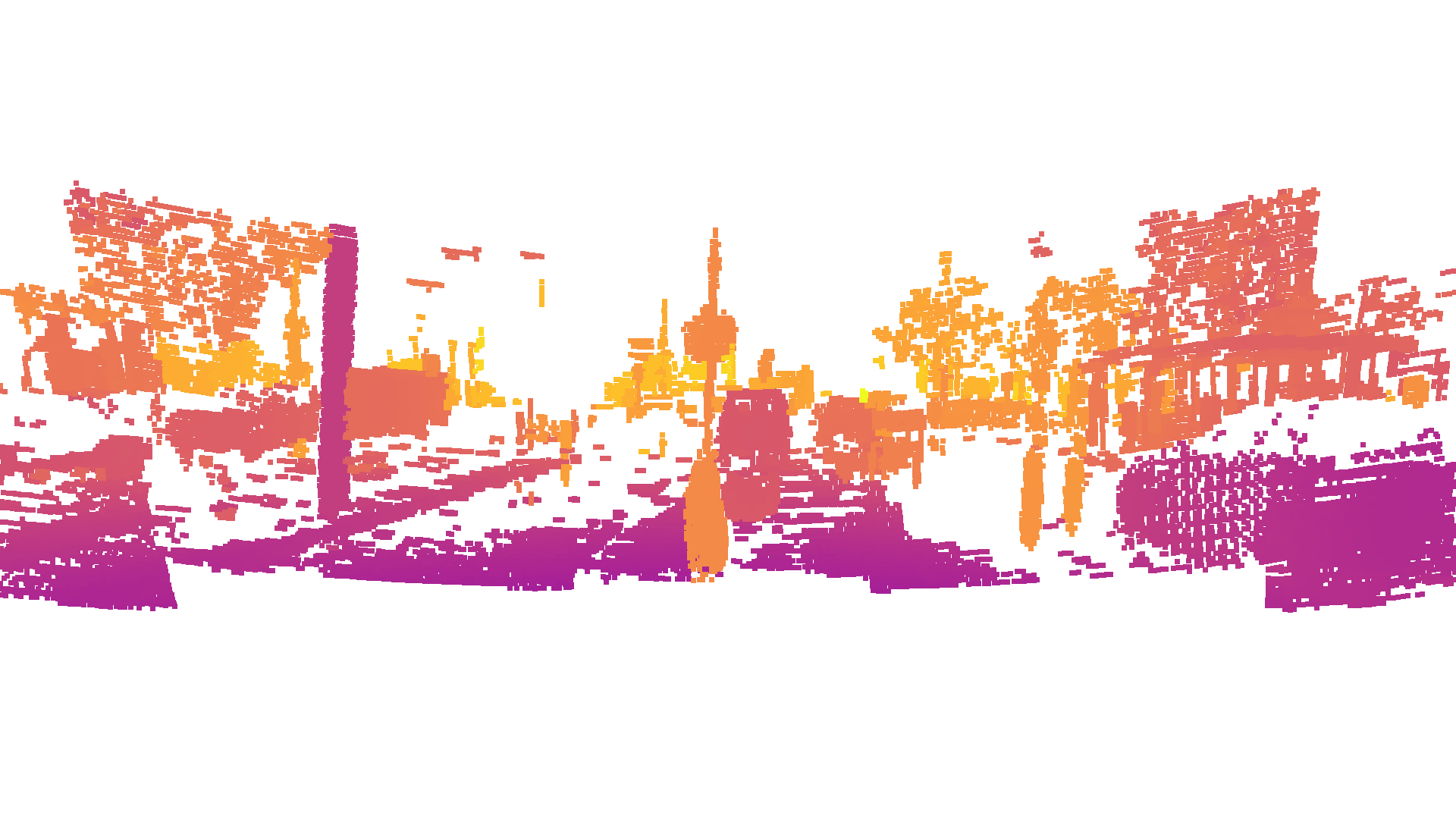} &
\includegraphics[width=0.10\textwidth]{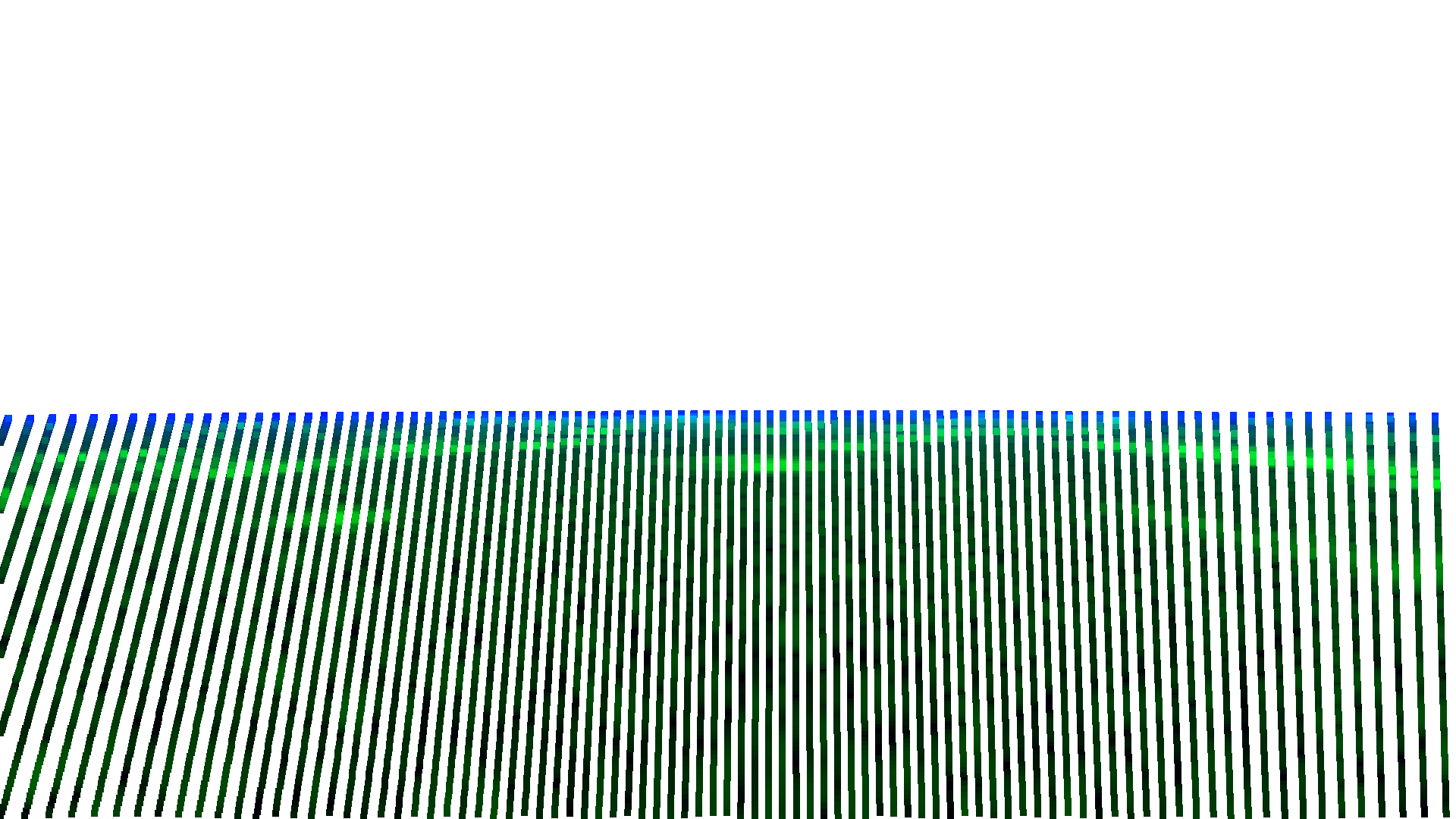} &
\includegraphics[width=0.10\textwidth]{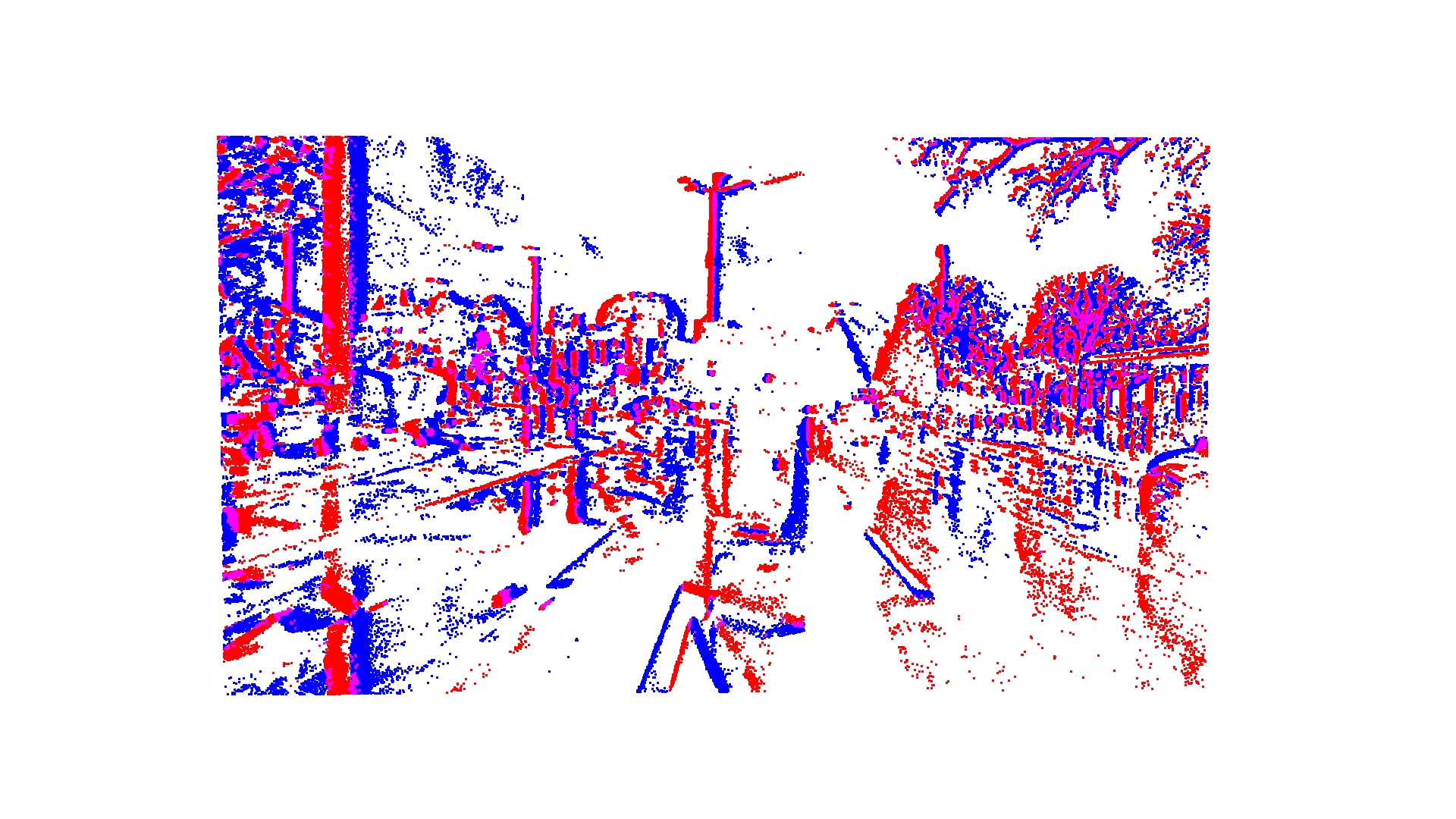} &
\includegraphics[width=0.10\textwidth]{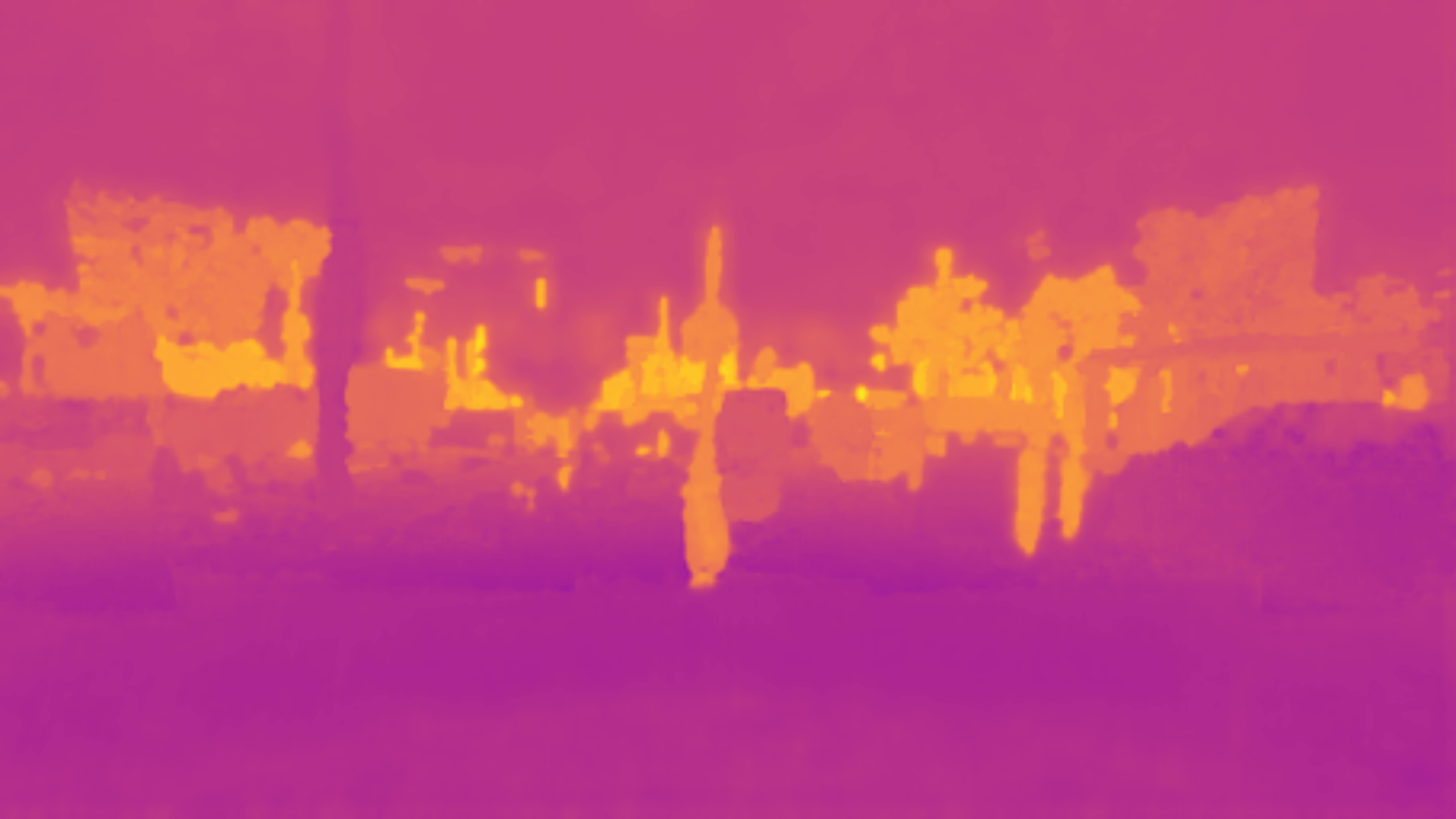} &
\includegraphics[width=0.10\textwidth]{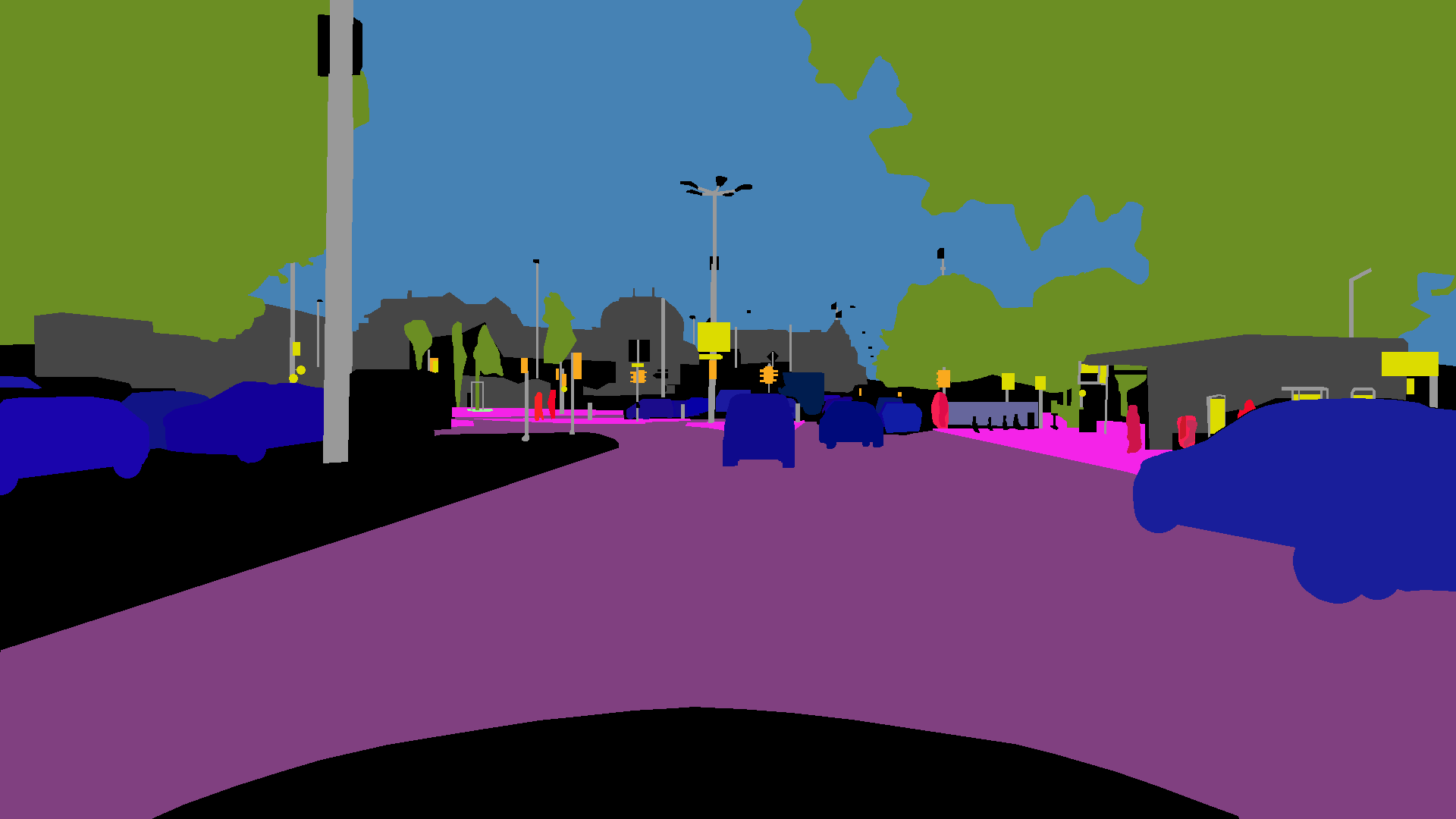} &
\includegraphics[width=0.10\textwidth]{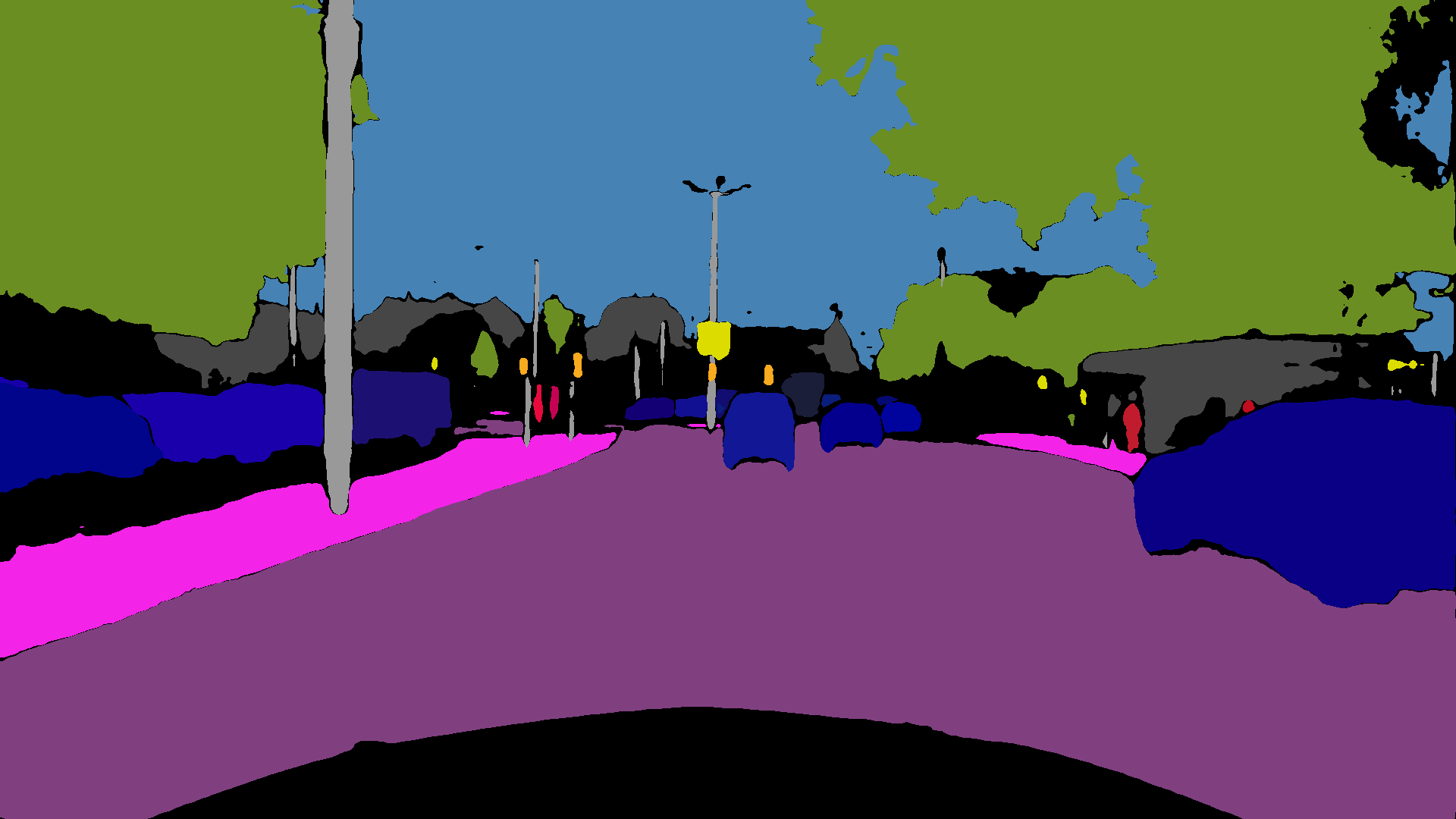} &
\includegraphics[width=0.10\textwidth]{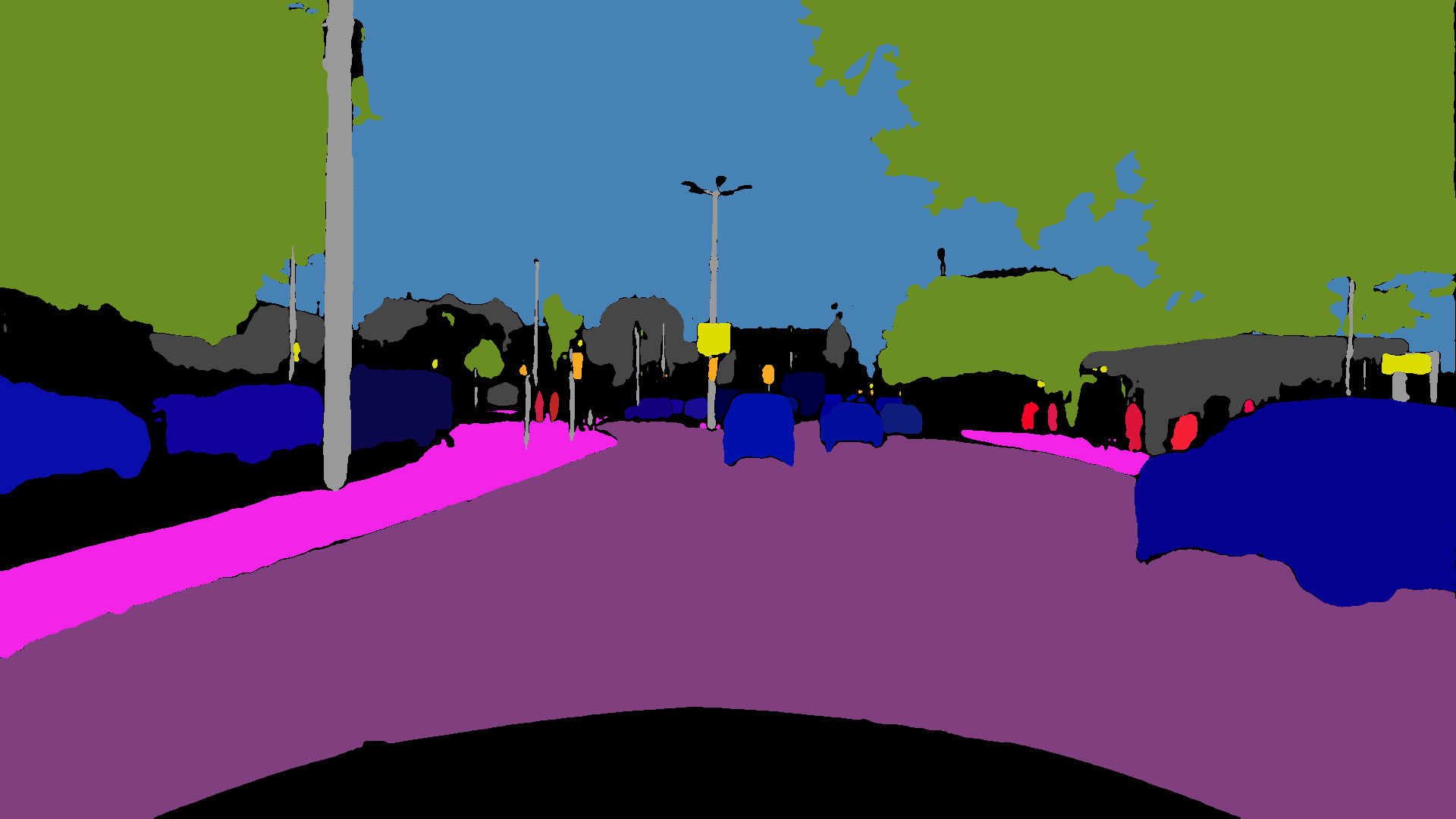} &
\includegraphics[width=0.10\textwidth]{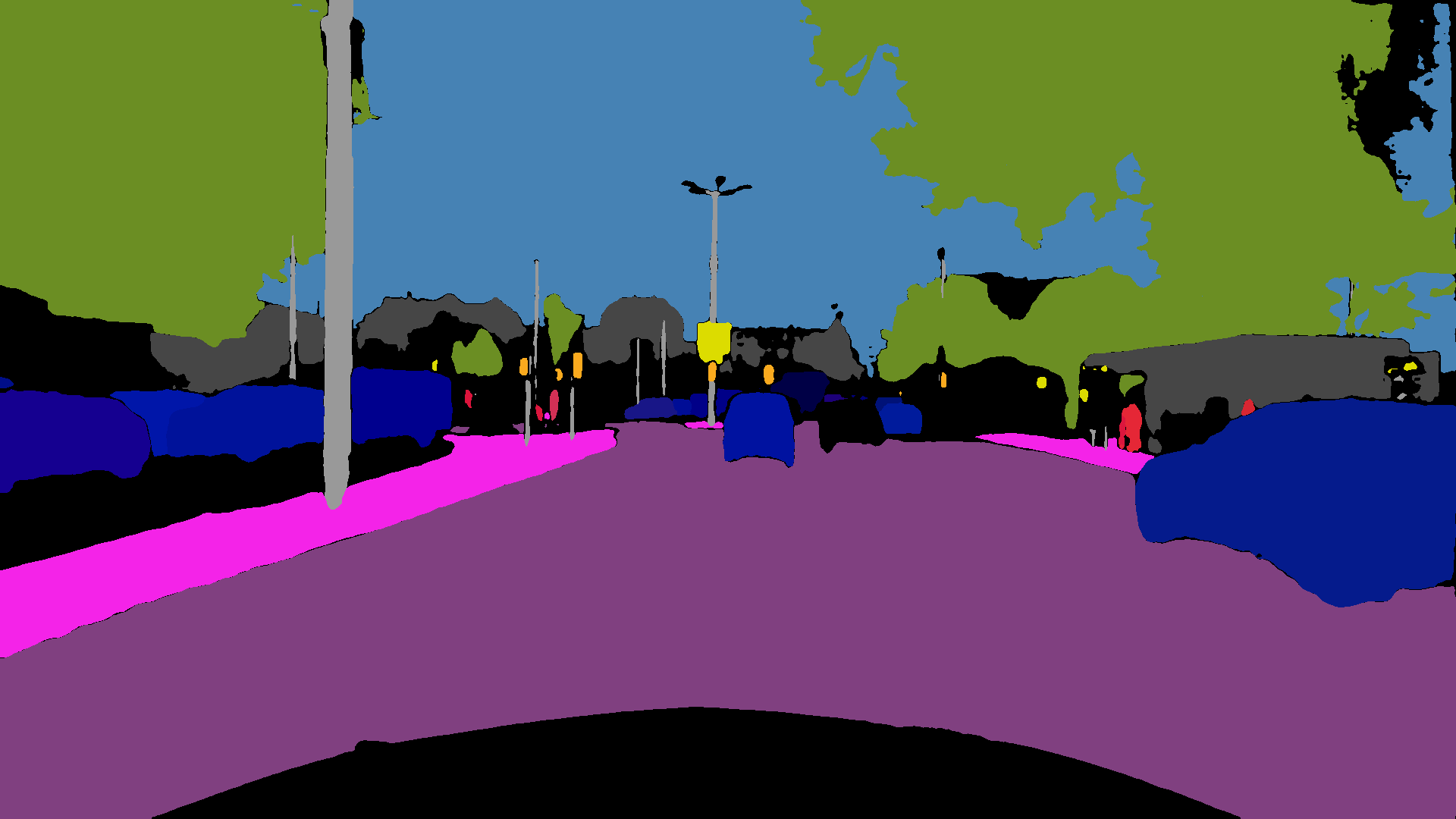} &
\includegraphics[width=0.10\textwidth]{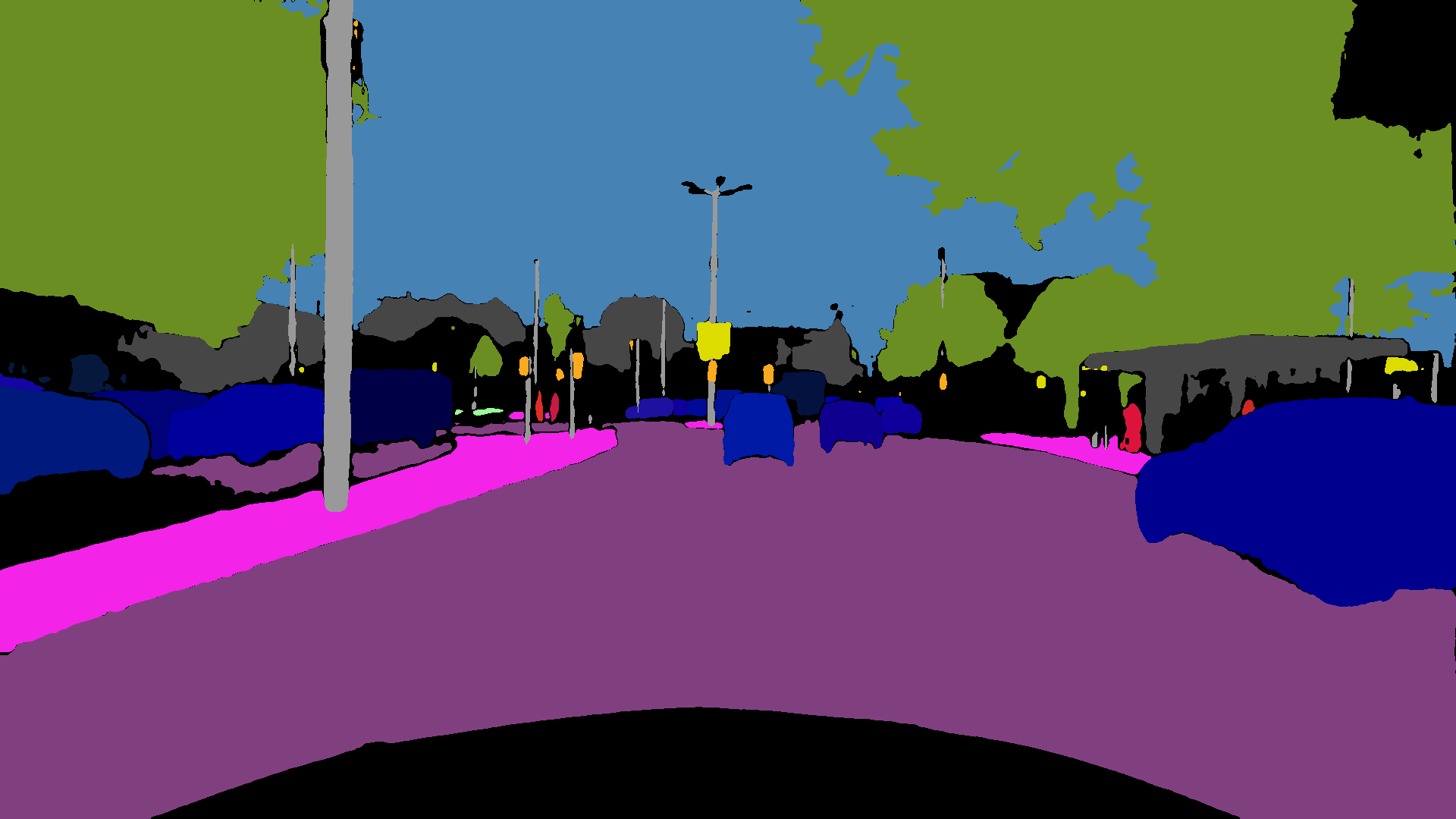} \\
\vspace{-0.1cm}

\includegraphics[width=0.10\textwidth]{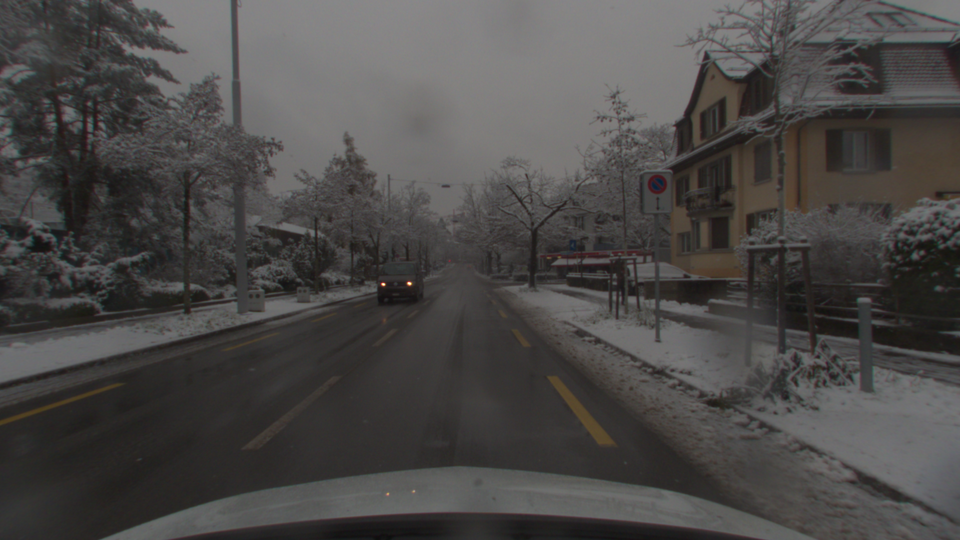} &
\includegraphics[width=0.10\textwidth]{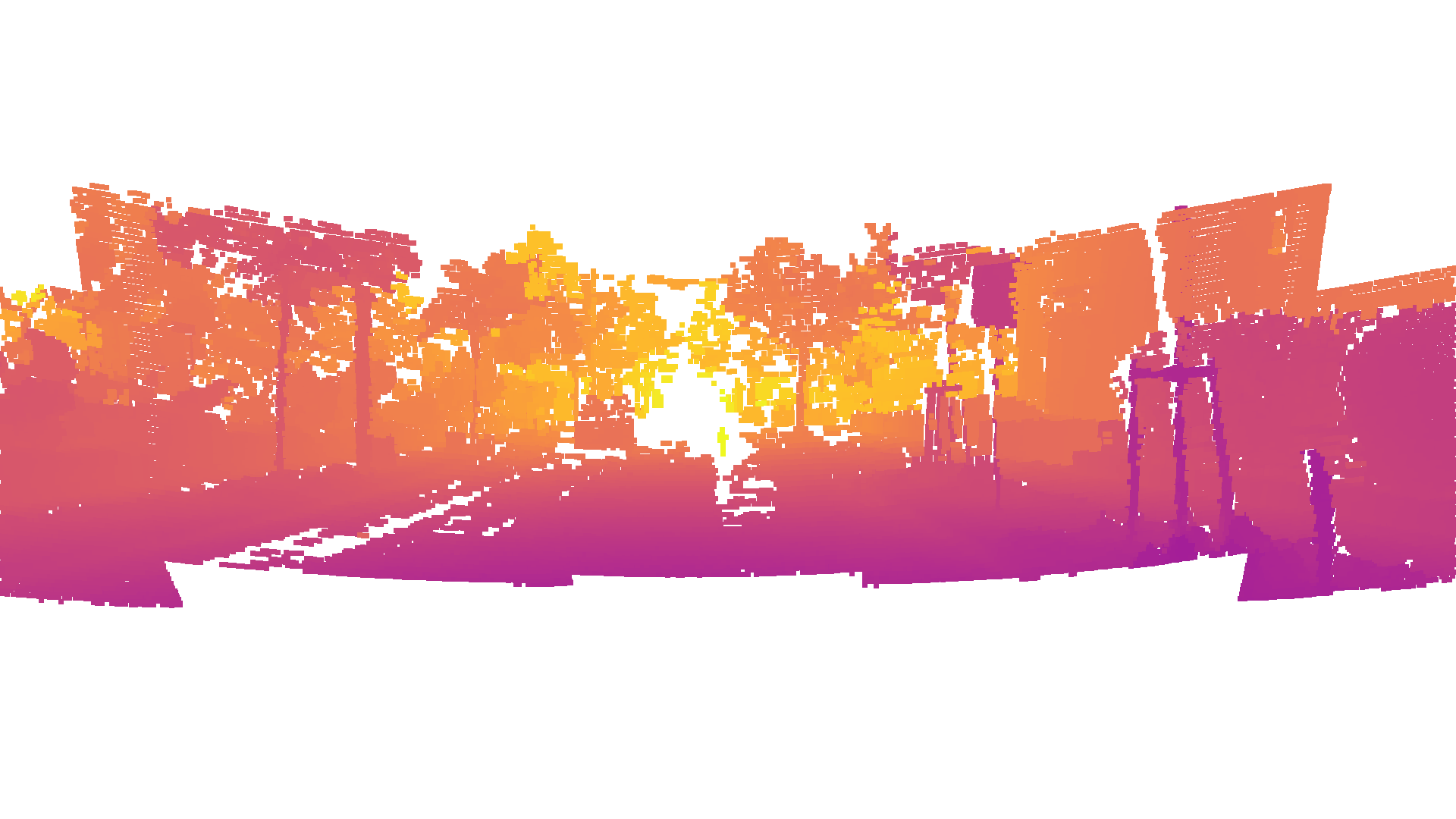} &
\includegraphics[width=0.10\textwidth]{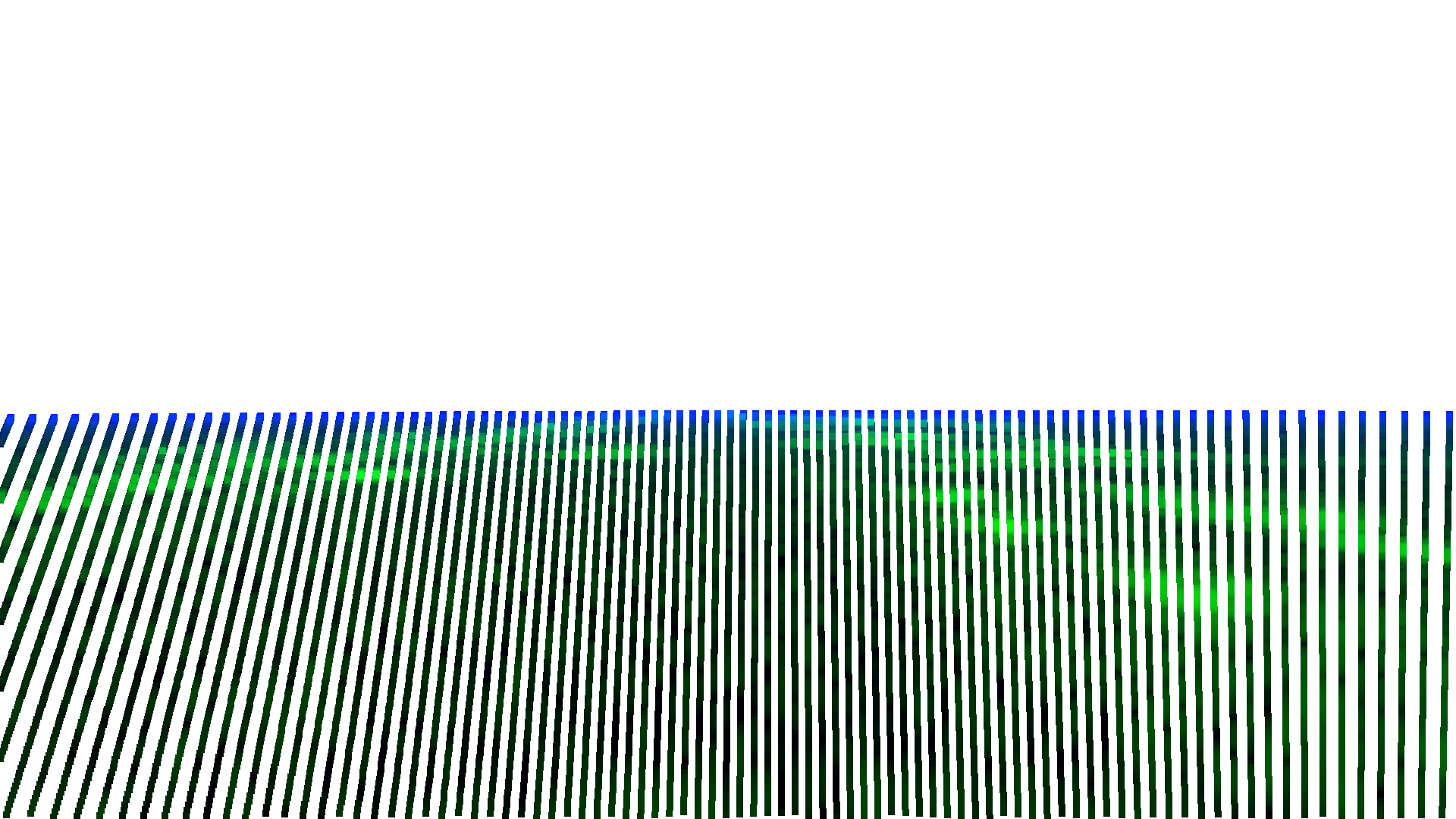} &
\includegraphics[width=0.10\textwidth]{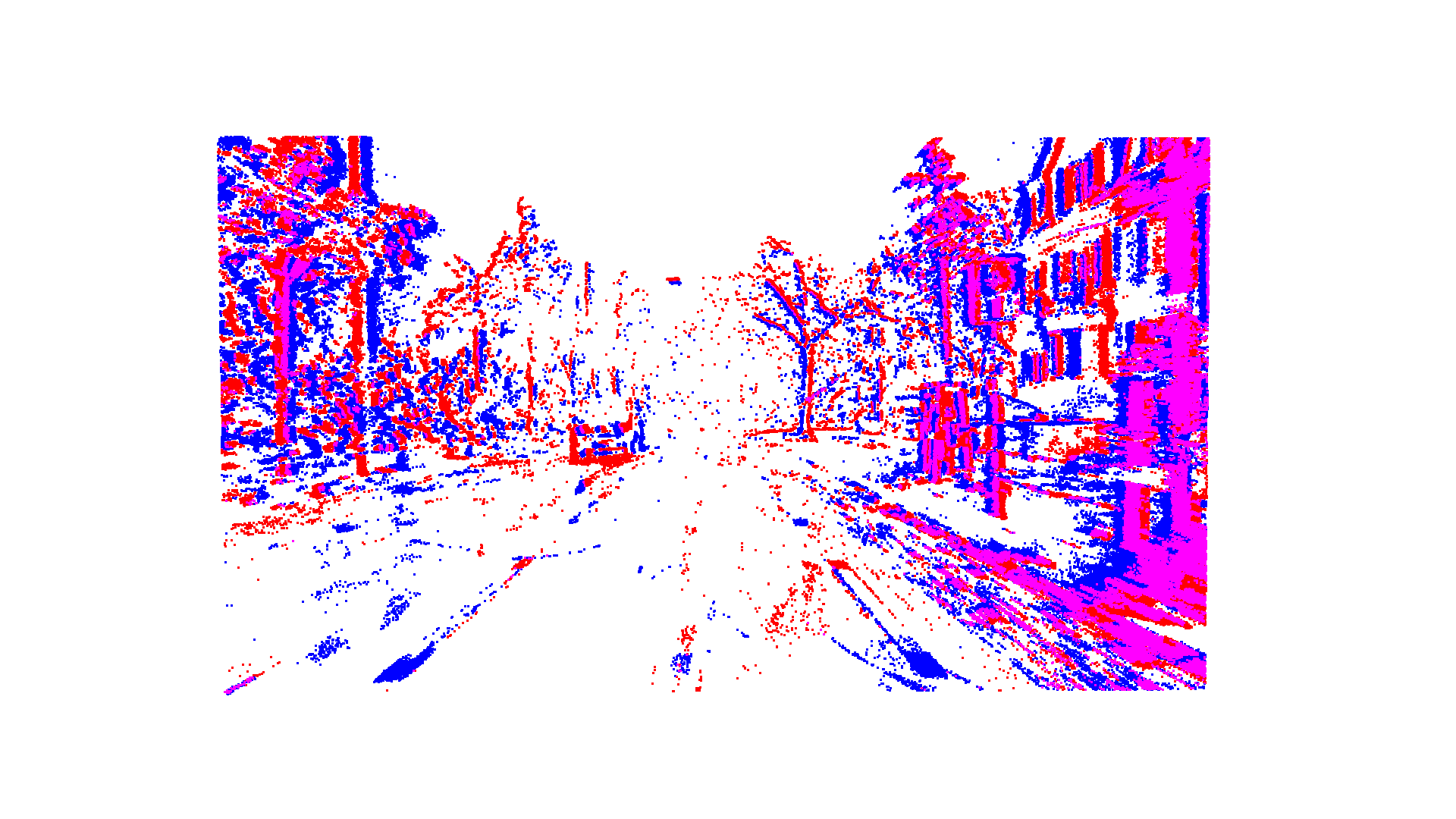} &
\includegraphics[width=0.10\textwidth]{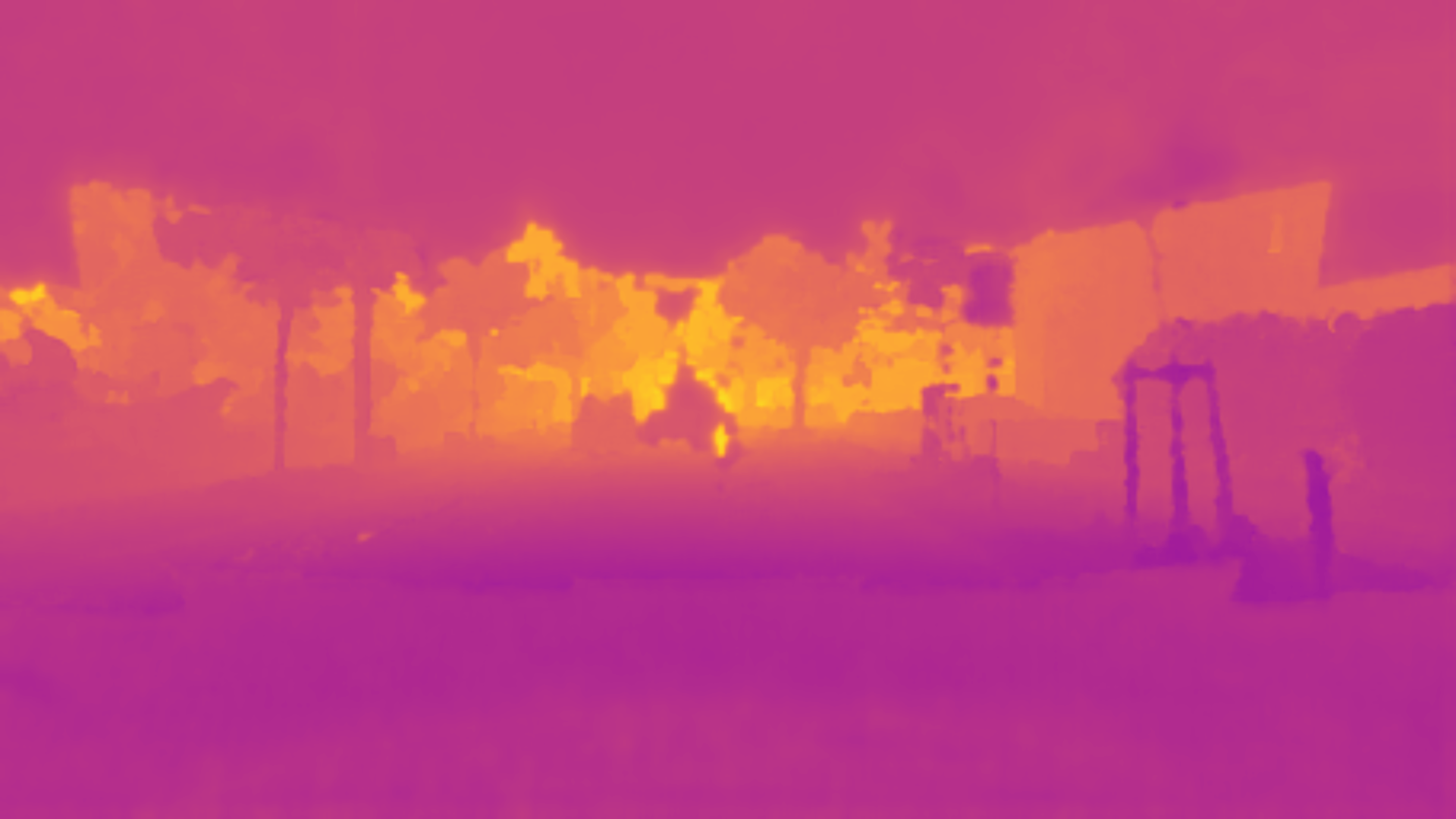} &
\includegraphics[width=0.10\textwidth]{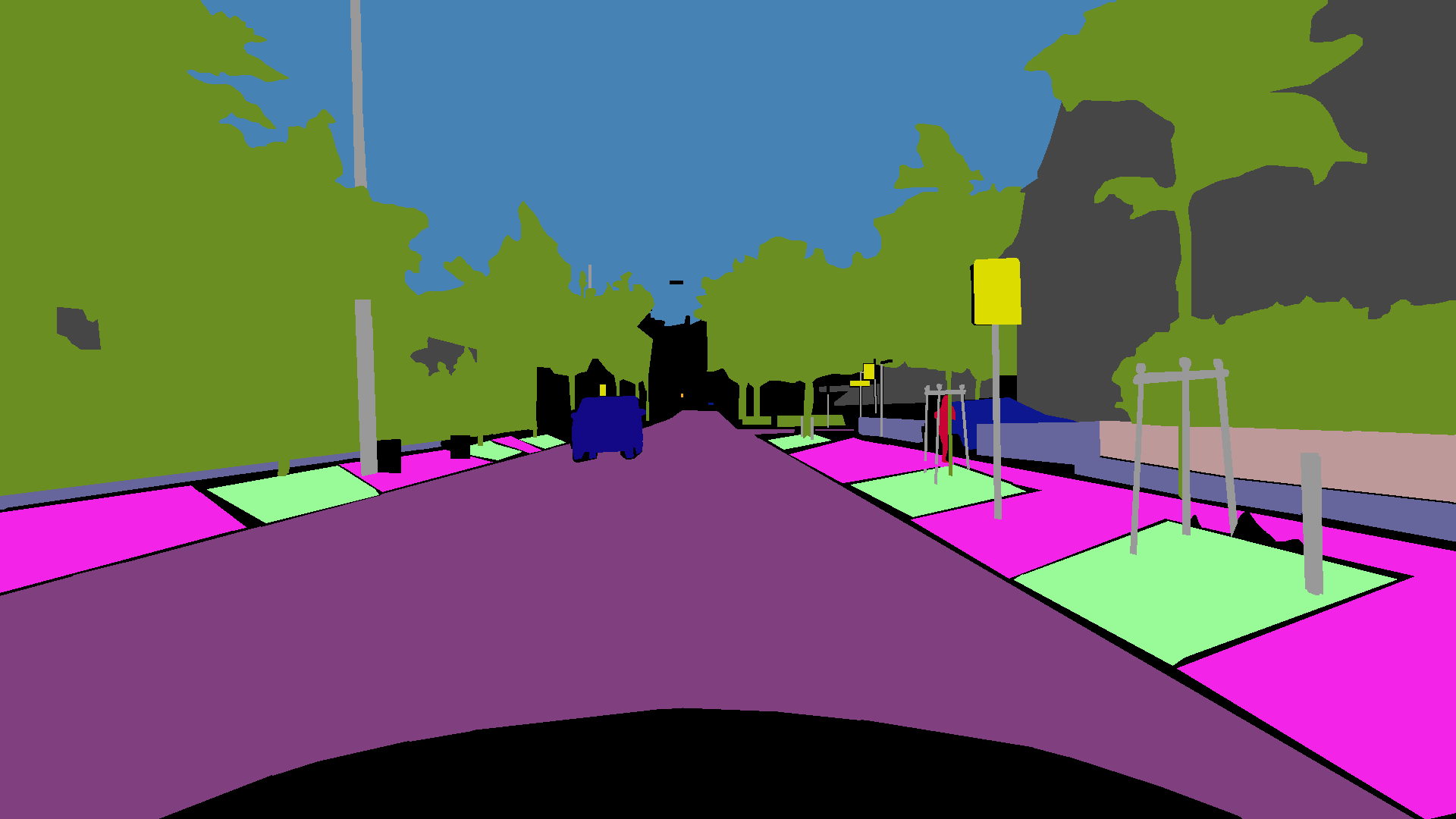} &
\includegraphics[width=0.10\textwidth]{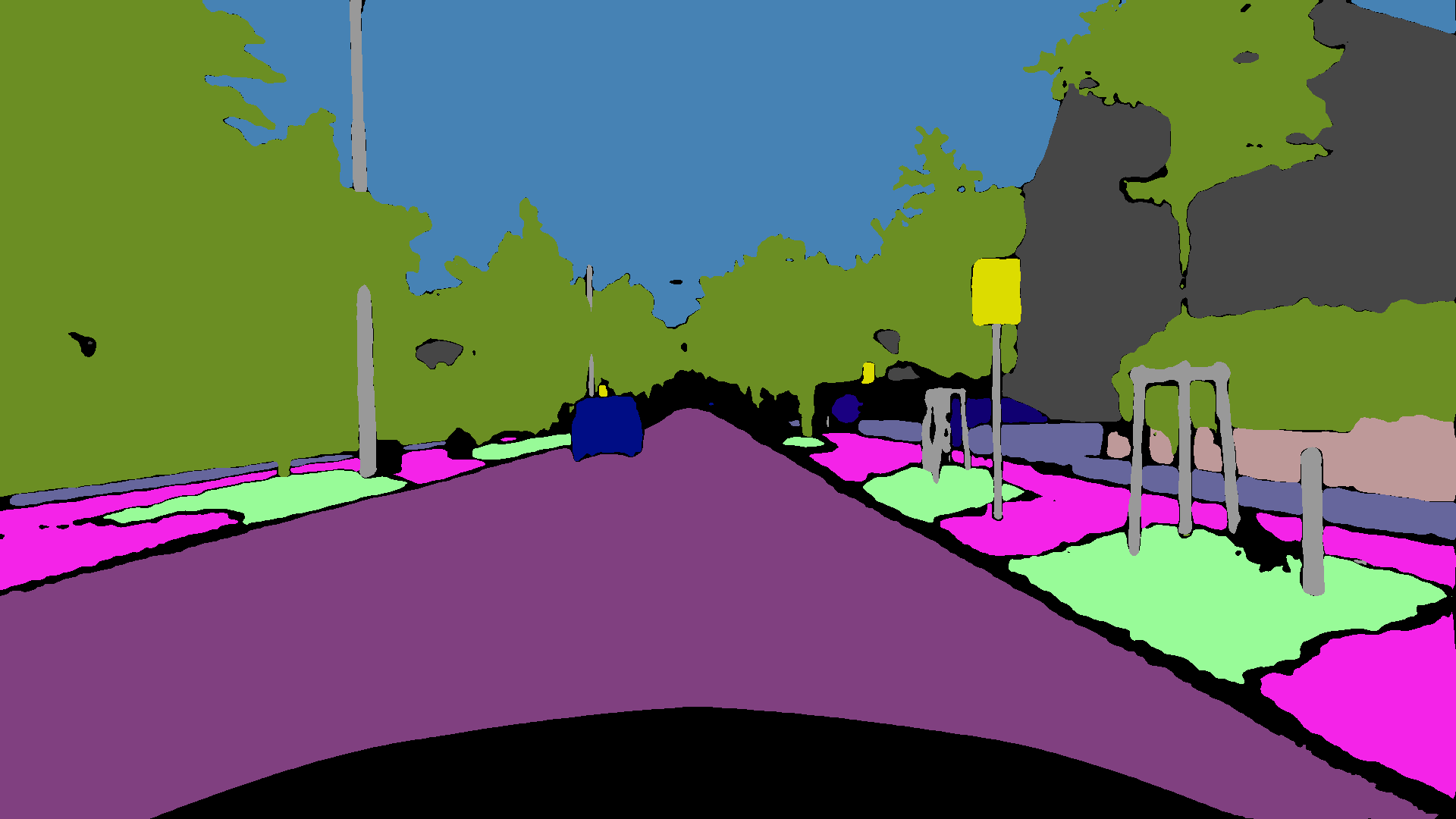} &
\includegraphics[width=0.10\textwidth]{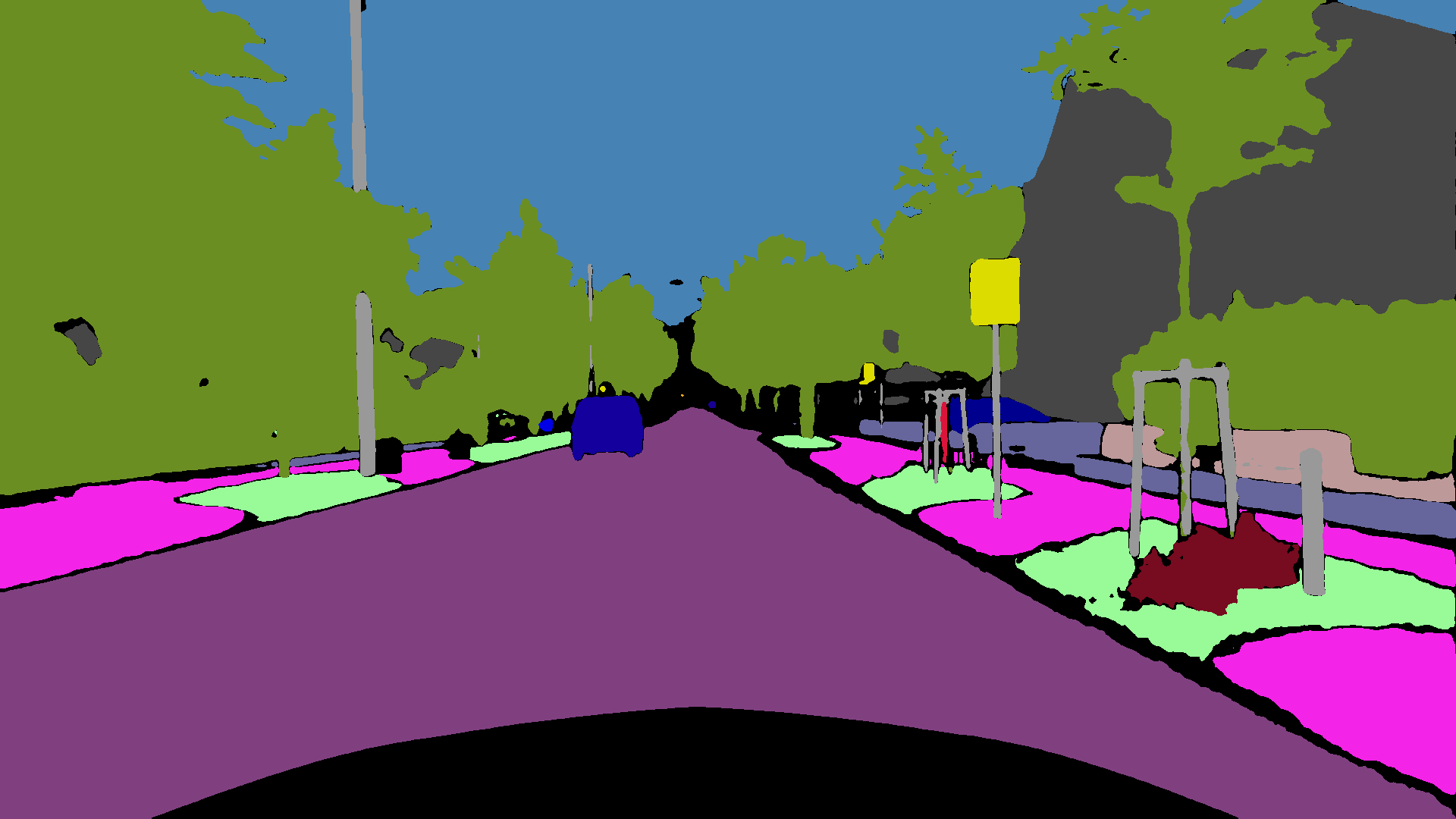} &
\includegraphics[width=0.10\textwidth]{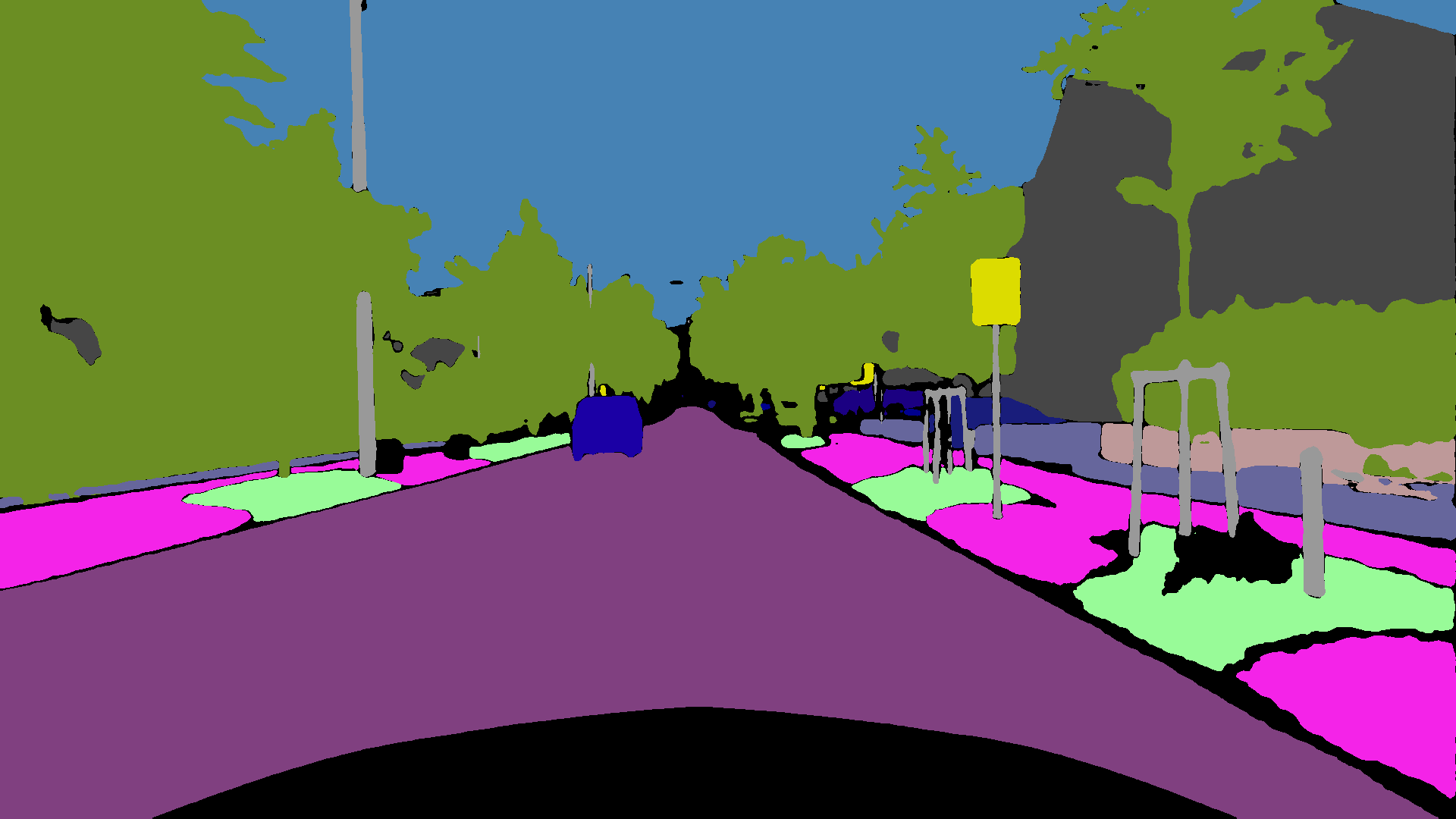} &
\includegraphics[width=0.10\textwidth]{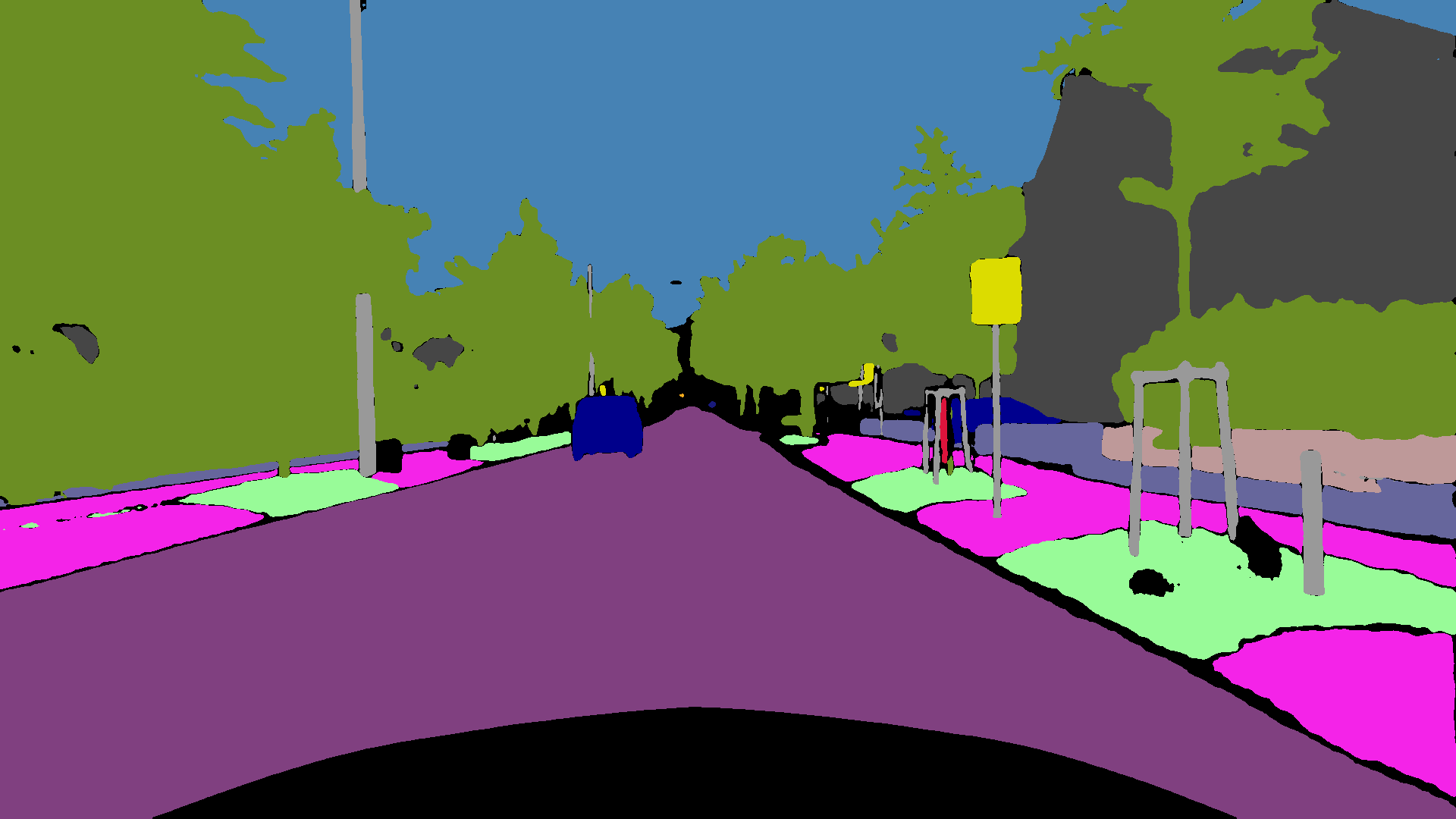} \\
\vspace{-0.1cm}

\includegraphics[width=0.10\textwidth]{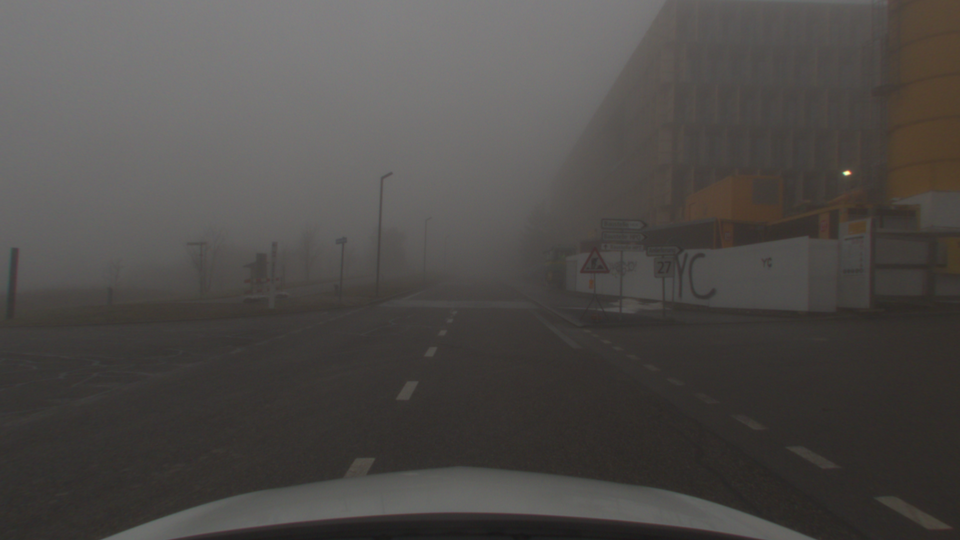} &
\includegraphics[width=0.10\textwidth]{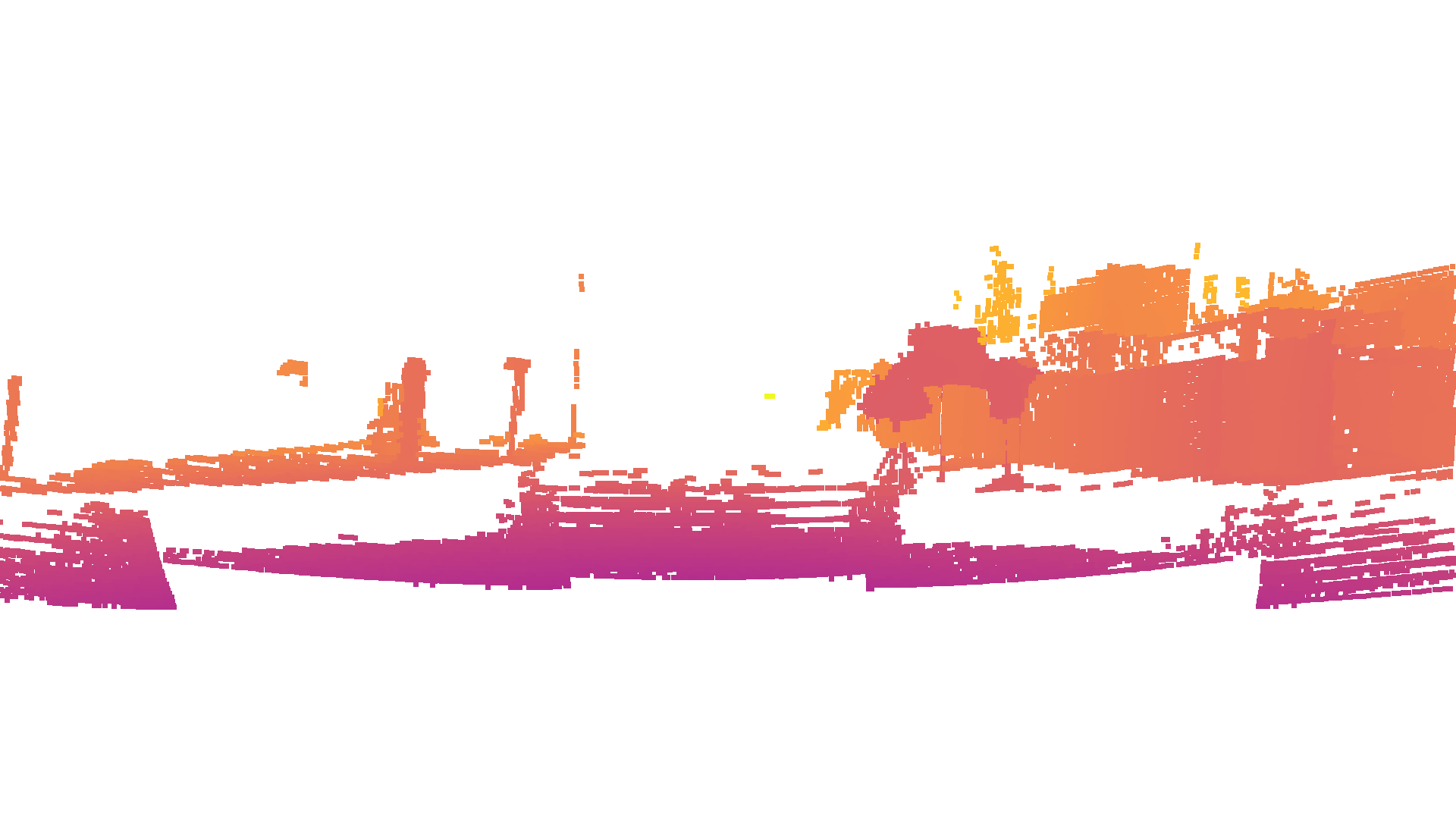} &
\includegraphics[width=0.10\textwidth]{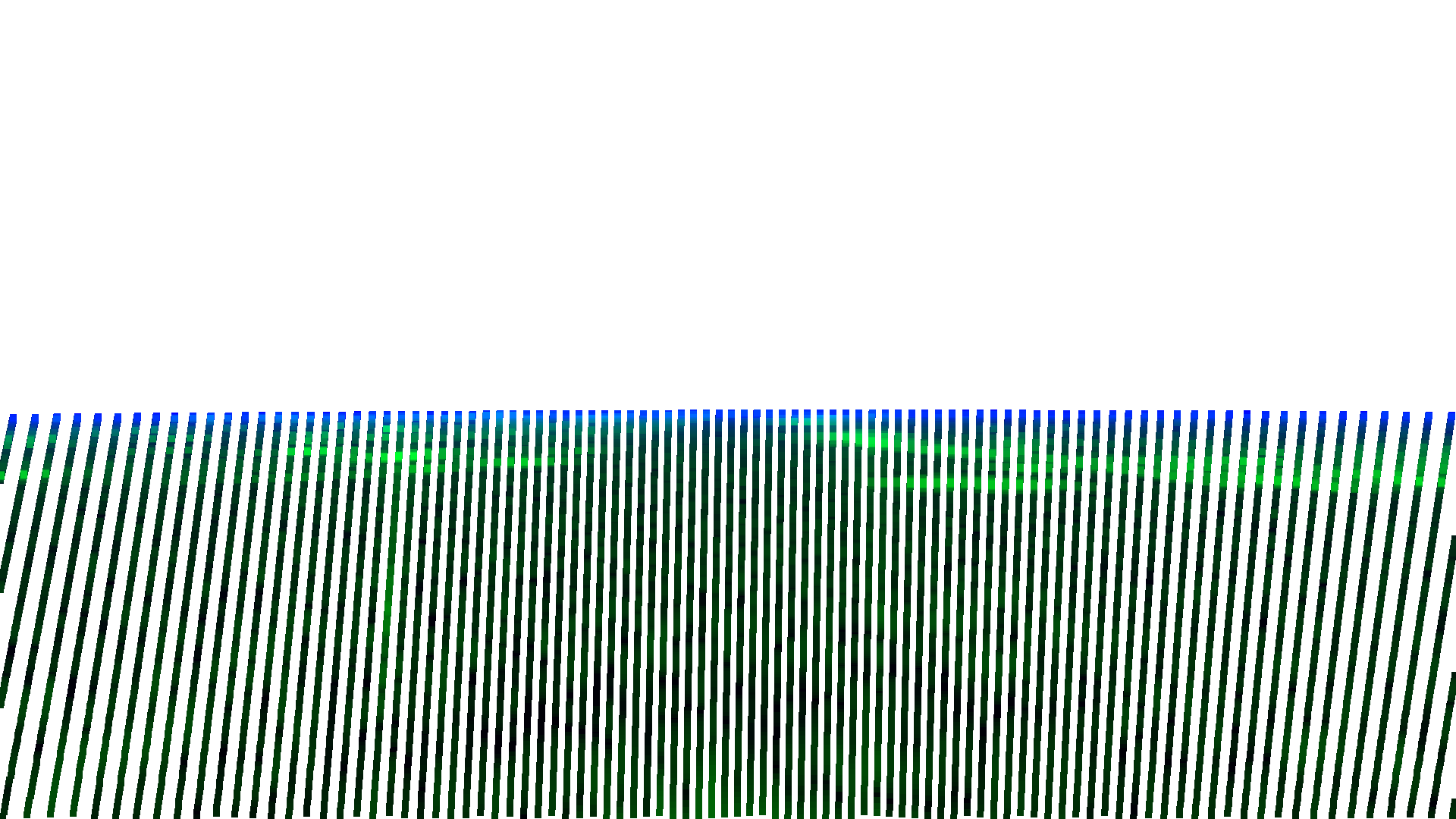} &
\includegraphics[width=0.10\textwidth]{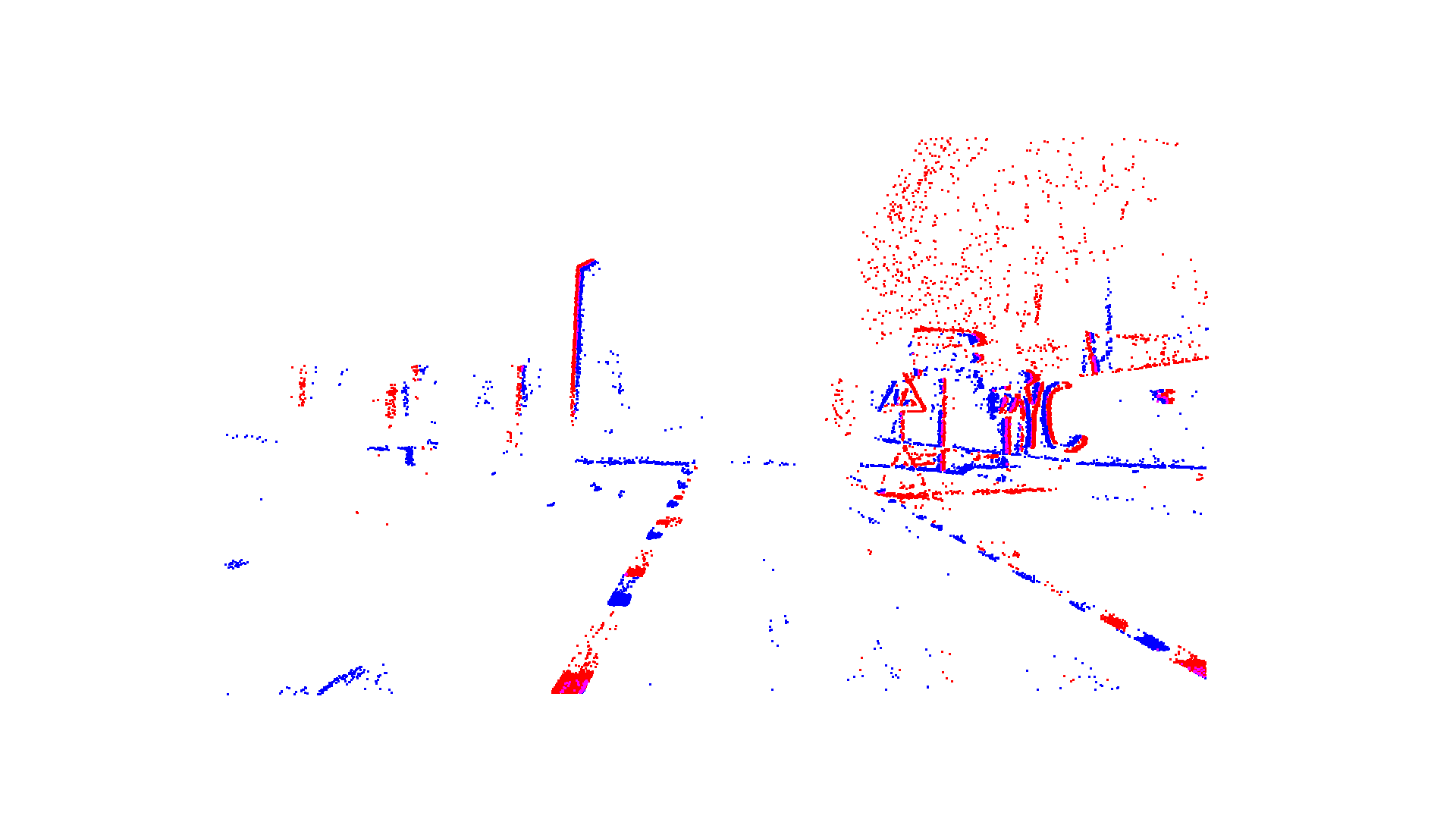} &
\includegraphics[width=0.10\textwidth]{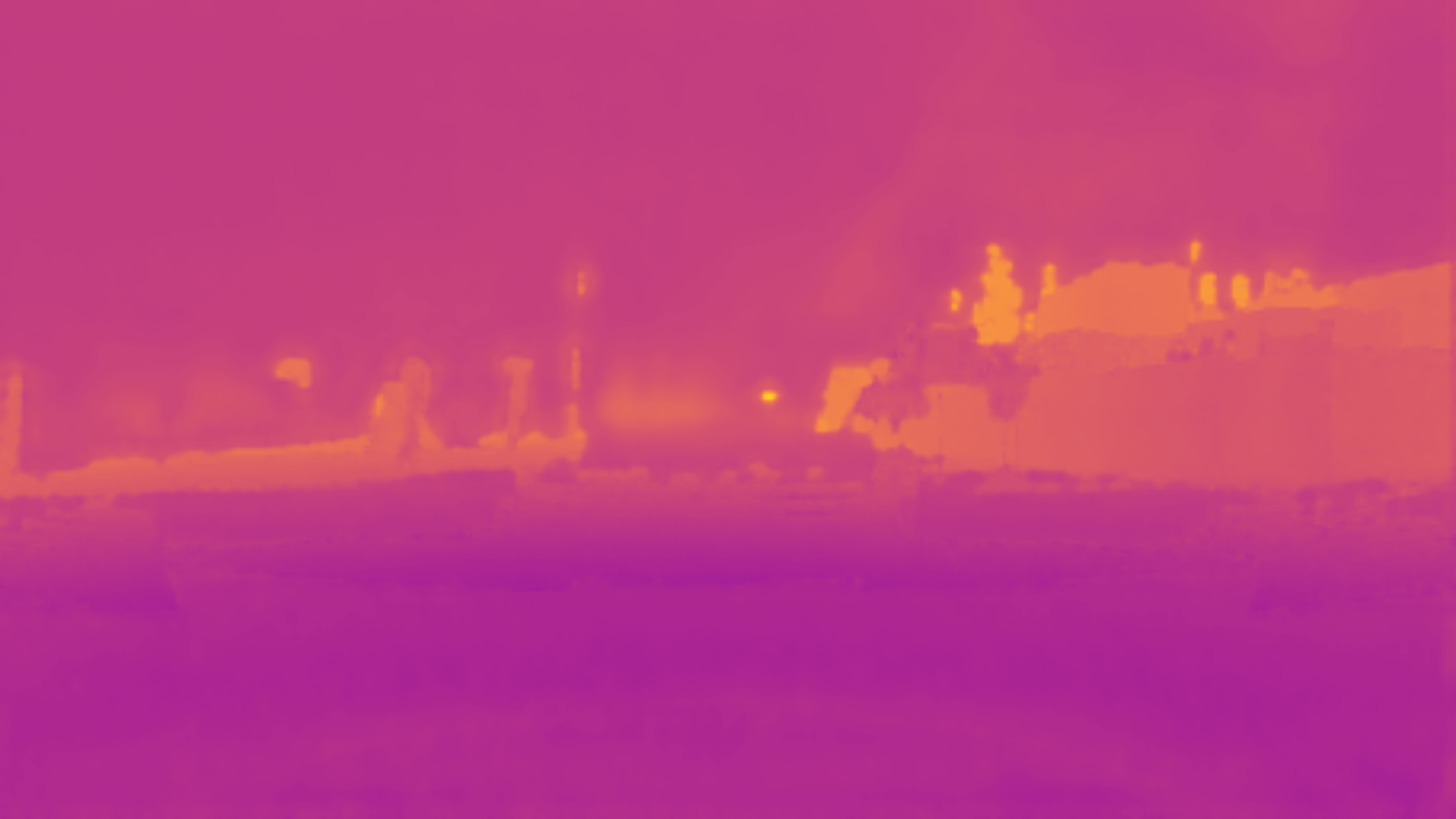} &
\includegraphics[width=0.10\textwidth]{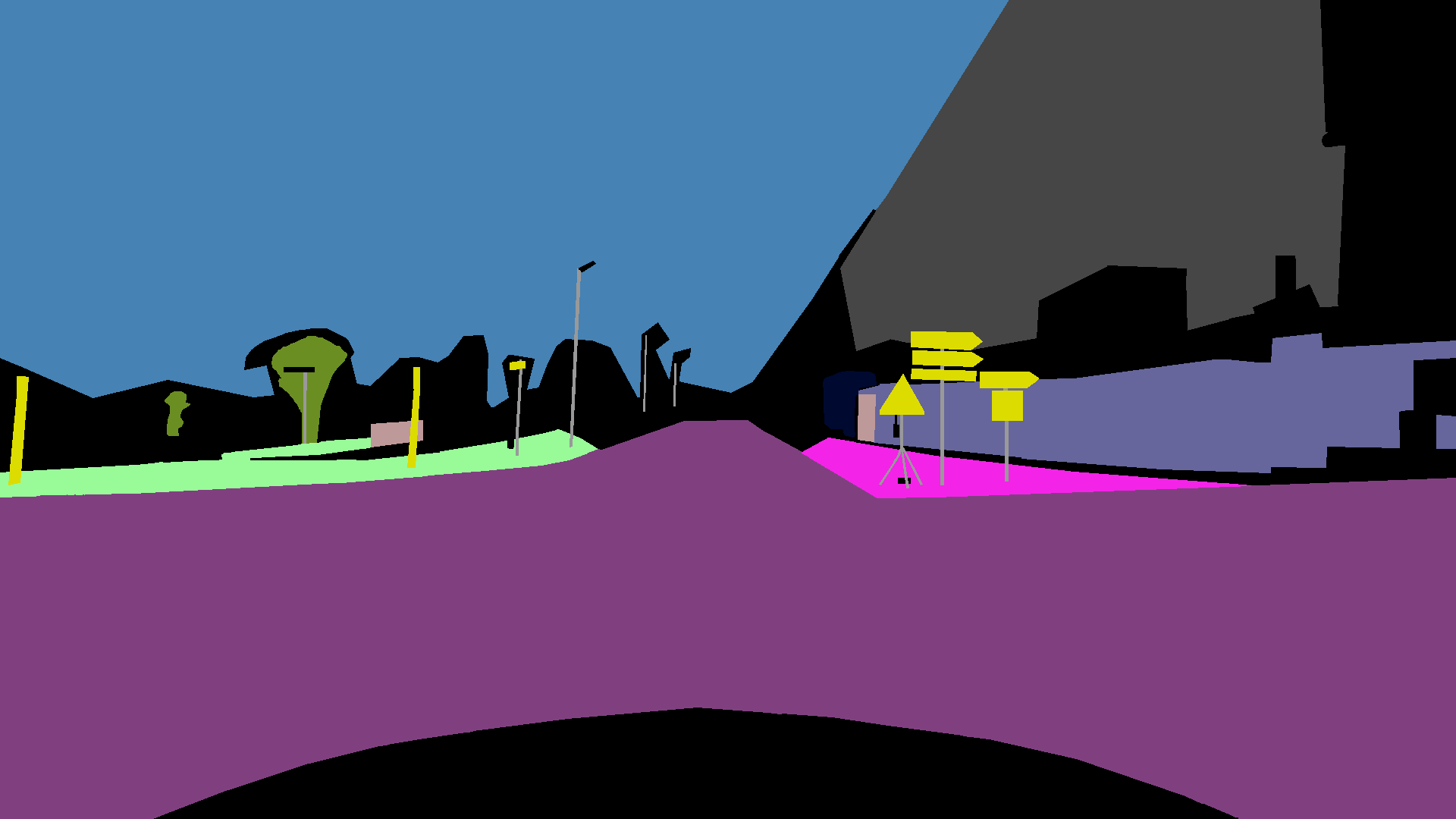} &
\includegraphics[width=0.10\textwidth]{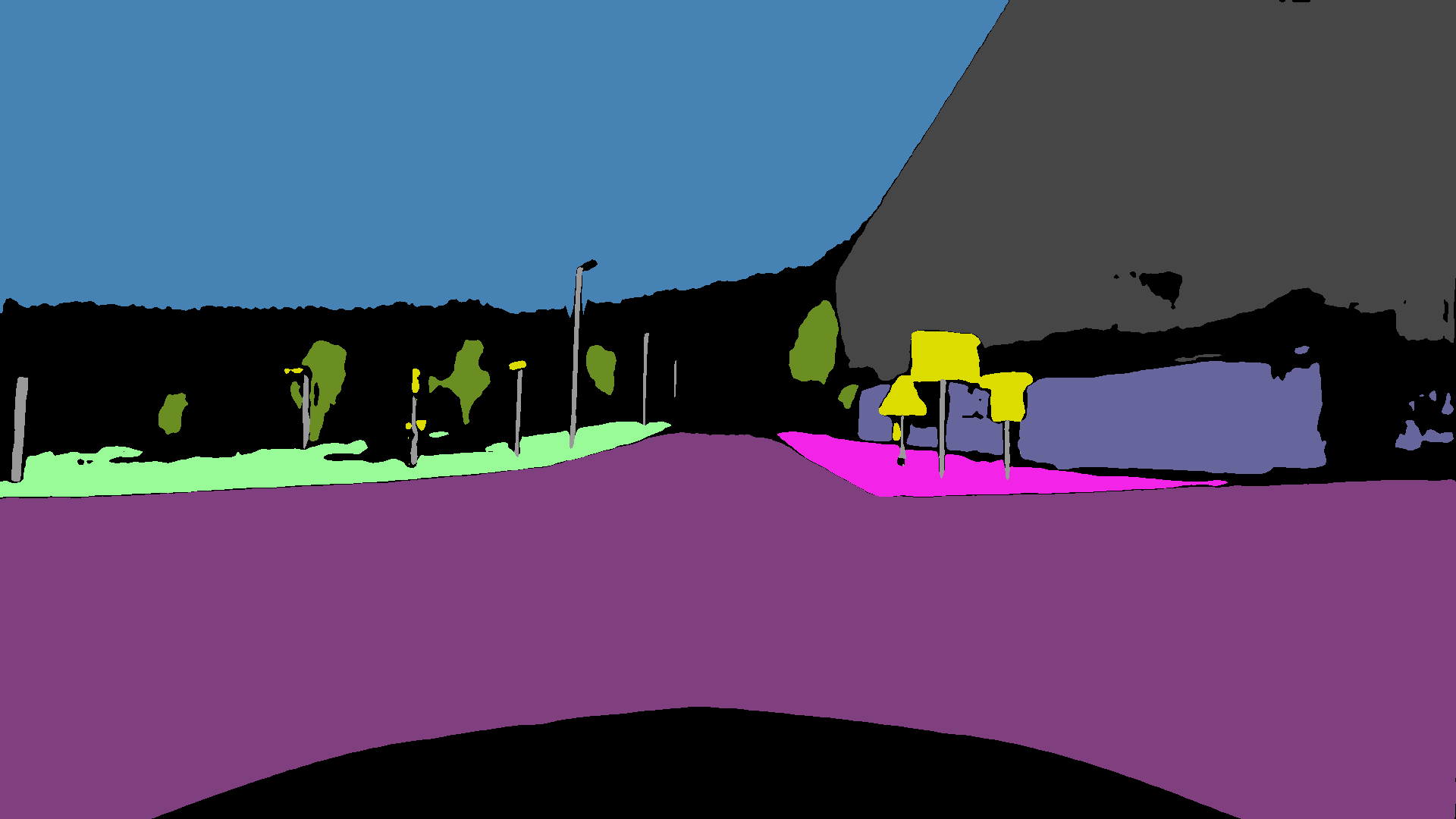} &
\includegraphics[width=0.10\textwidth]{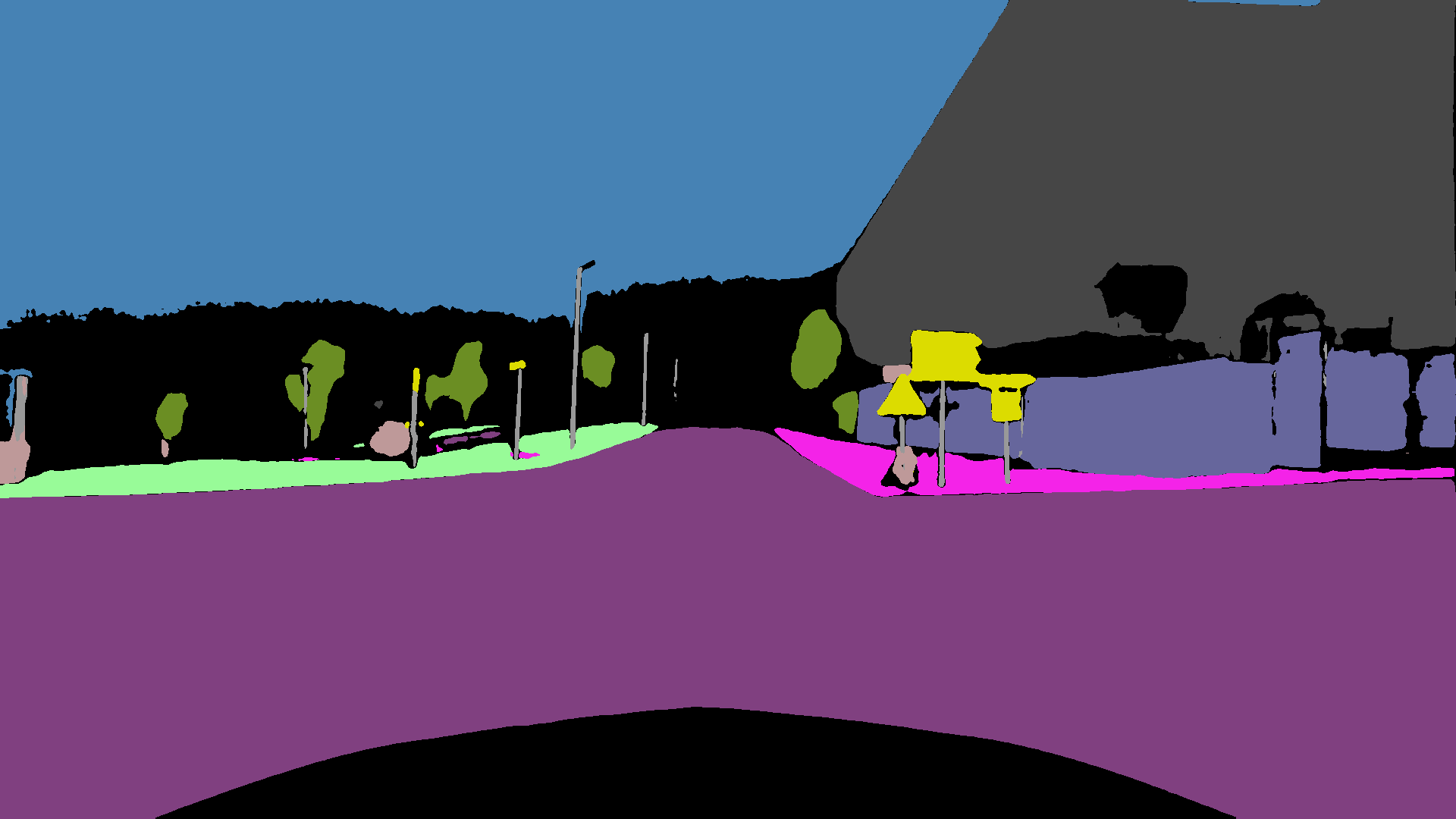} &
\includegraphics[width=0.10\textwidth]{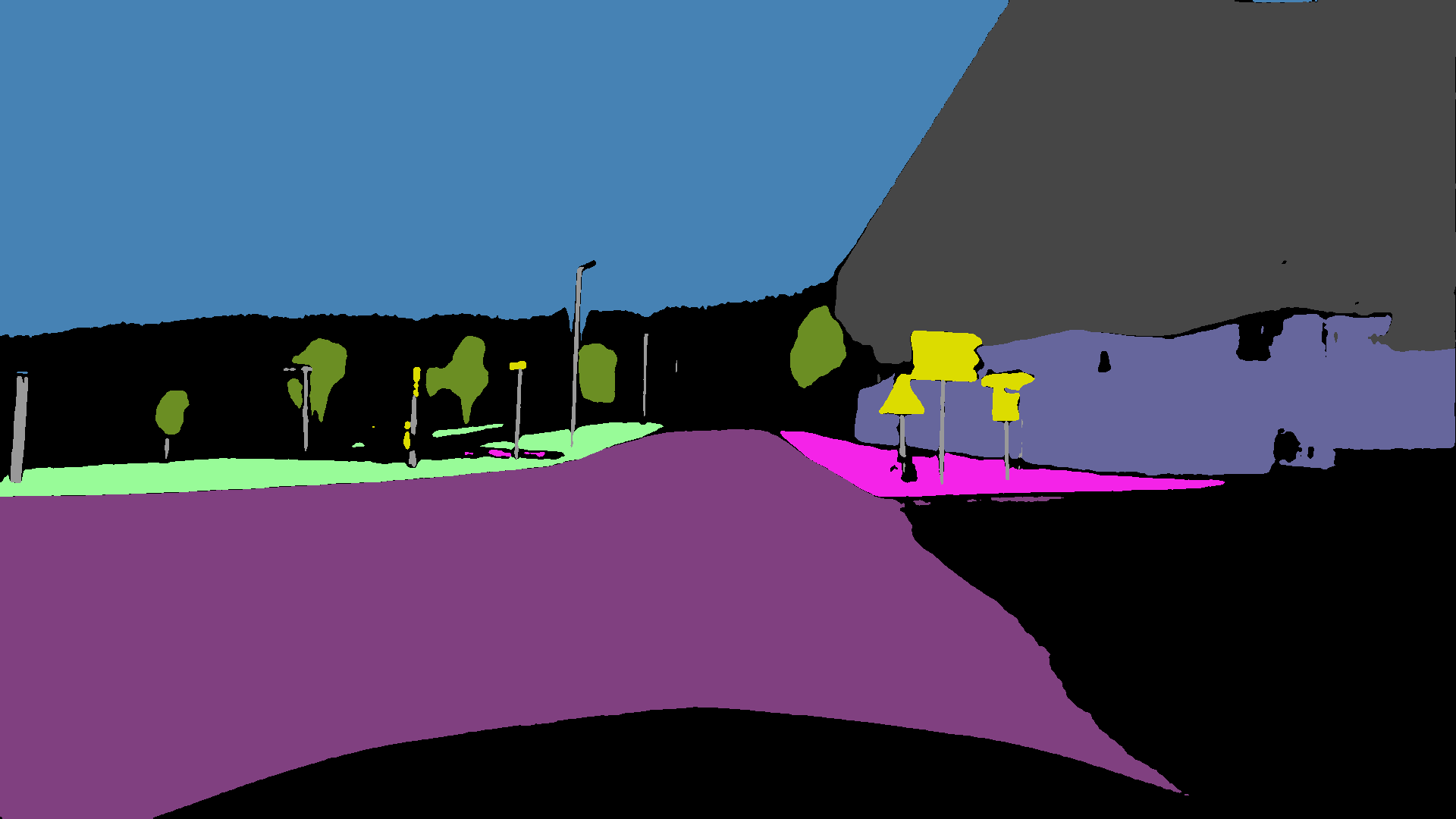} &
\includegraphics[width=0.10\textwidth]{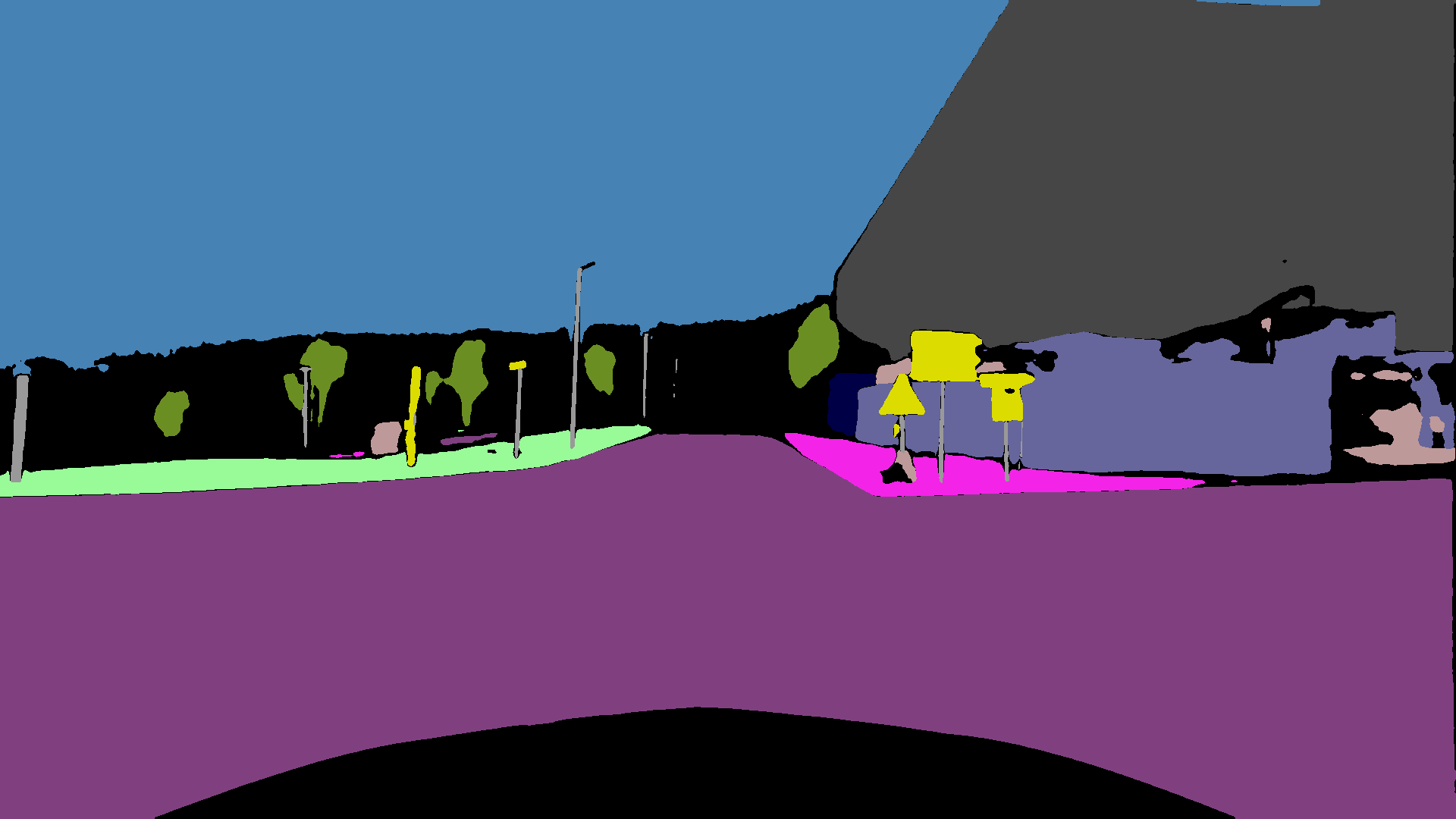} \\
\vspace{-0.1cm}

\includegraphics[width=0.10\textwidth]{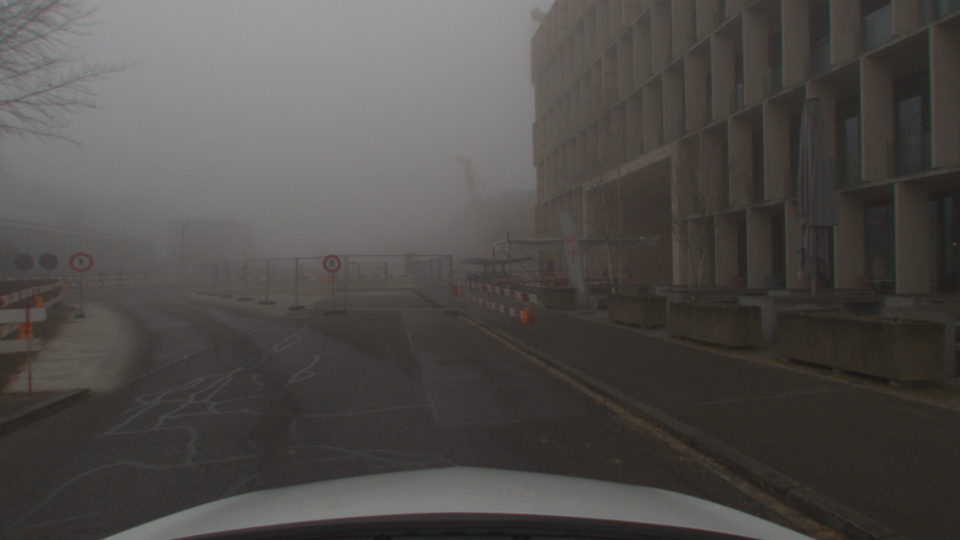} &
\includegraphics[width=0.10\textwidth]{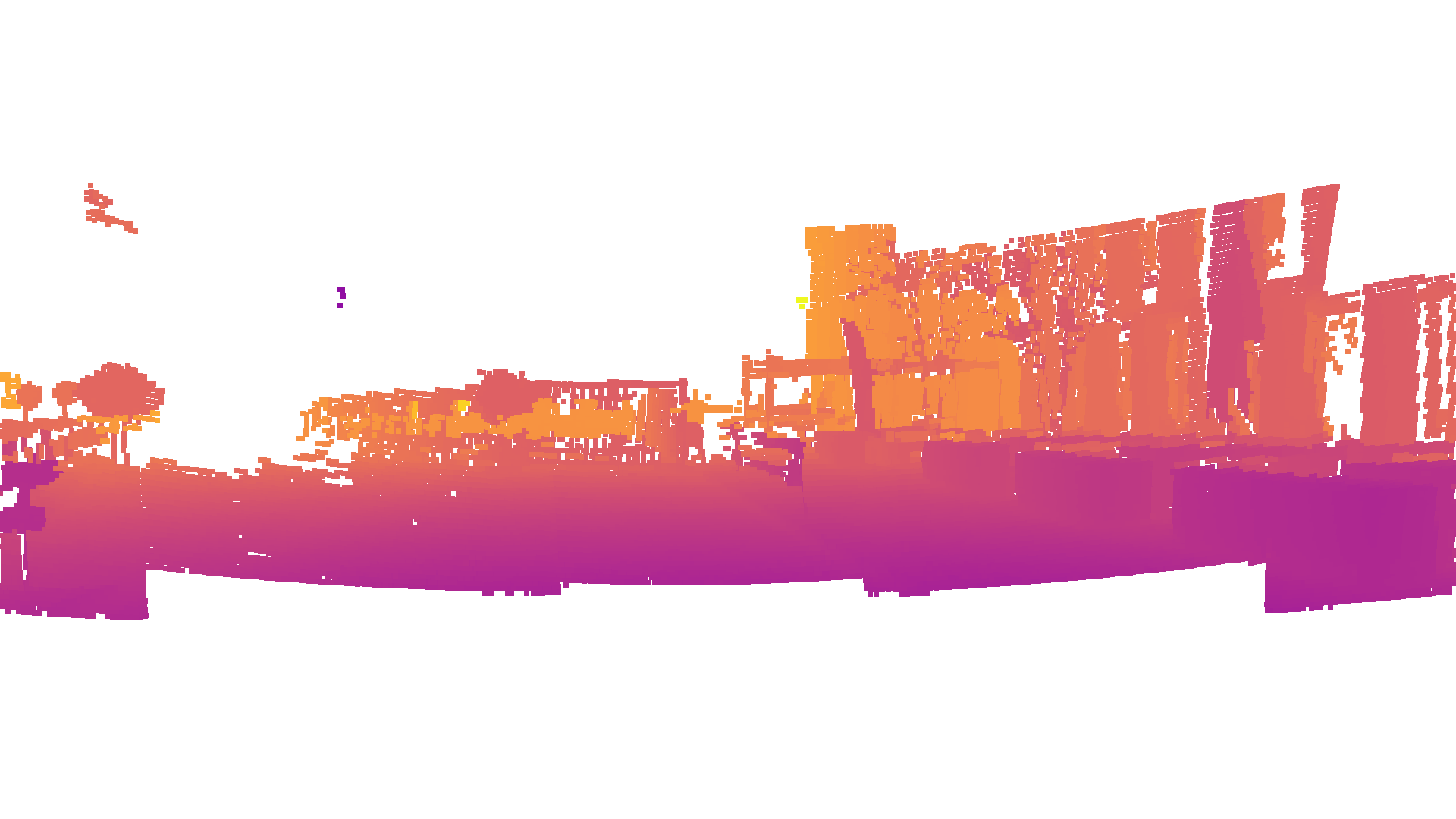} &
\includegraphics[width=0.10\textwidth]{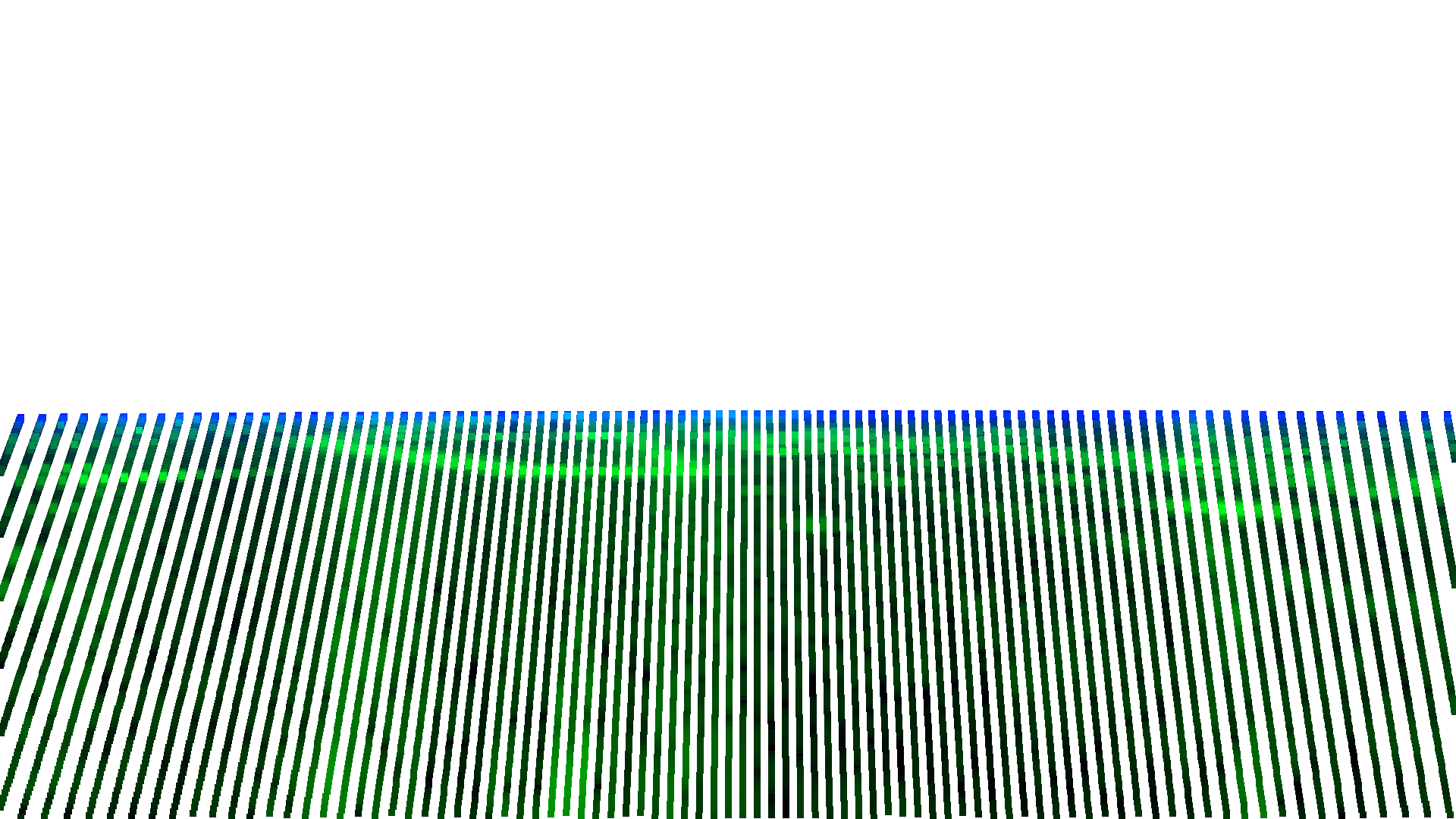} &
\includegraphics[width=0.10\textwidth]{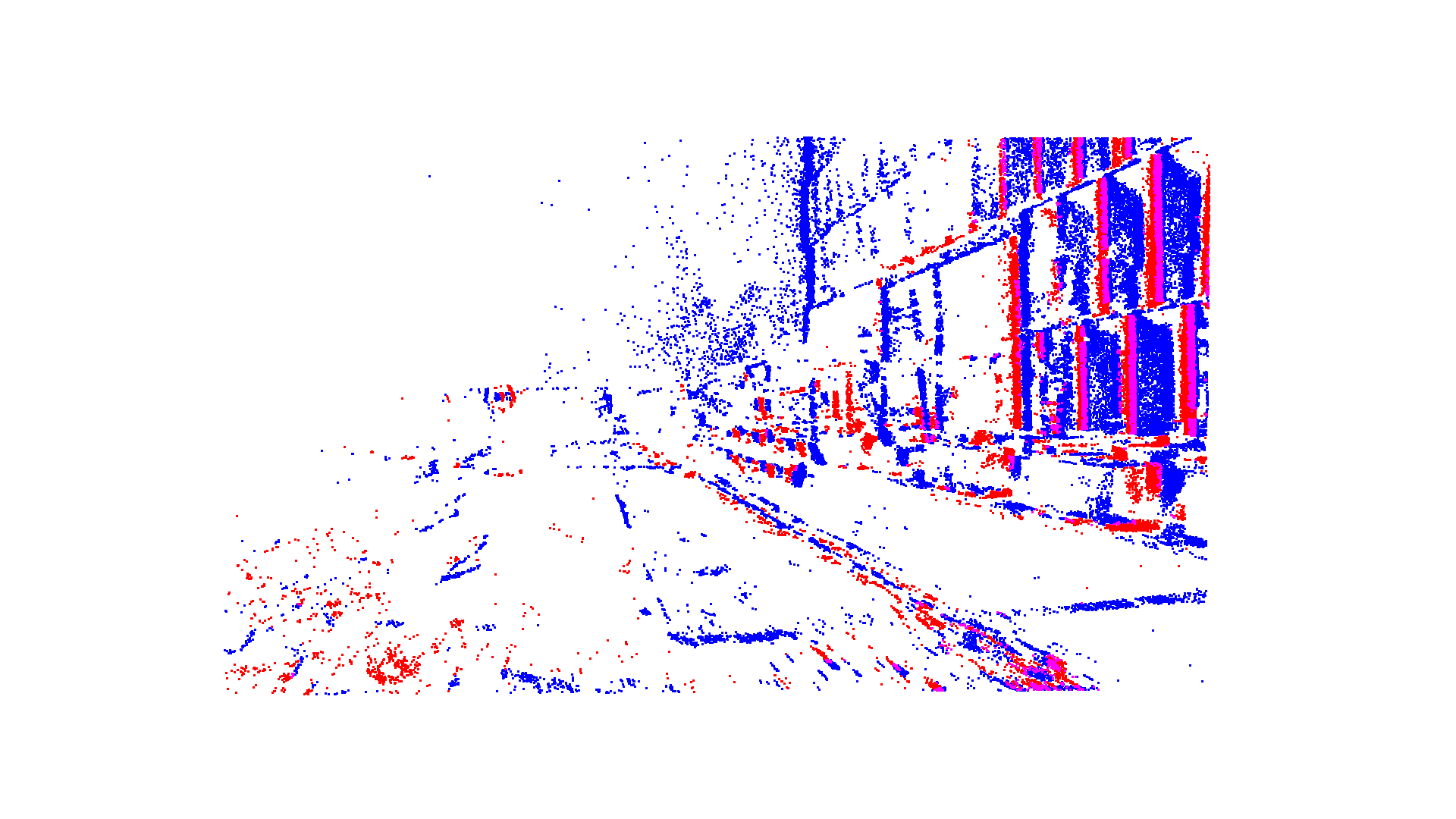} &
\includegraphics[width=0.10\textwidth]{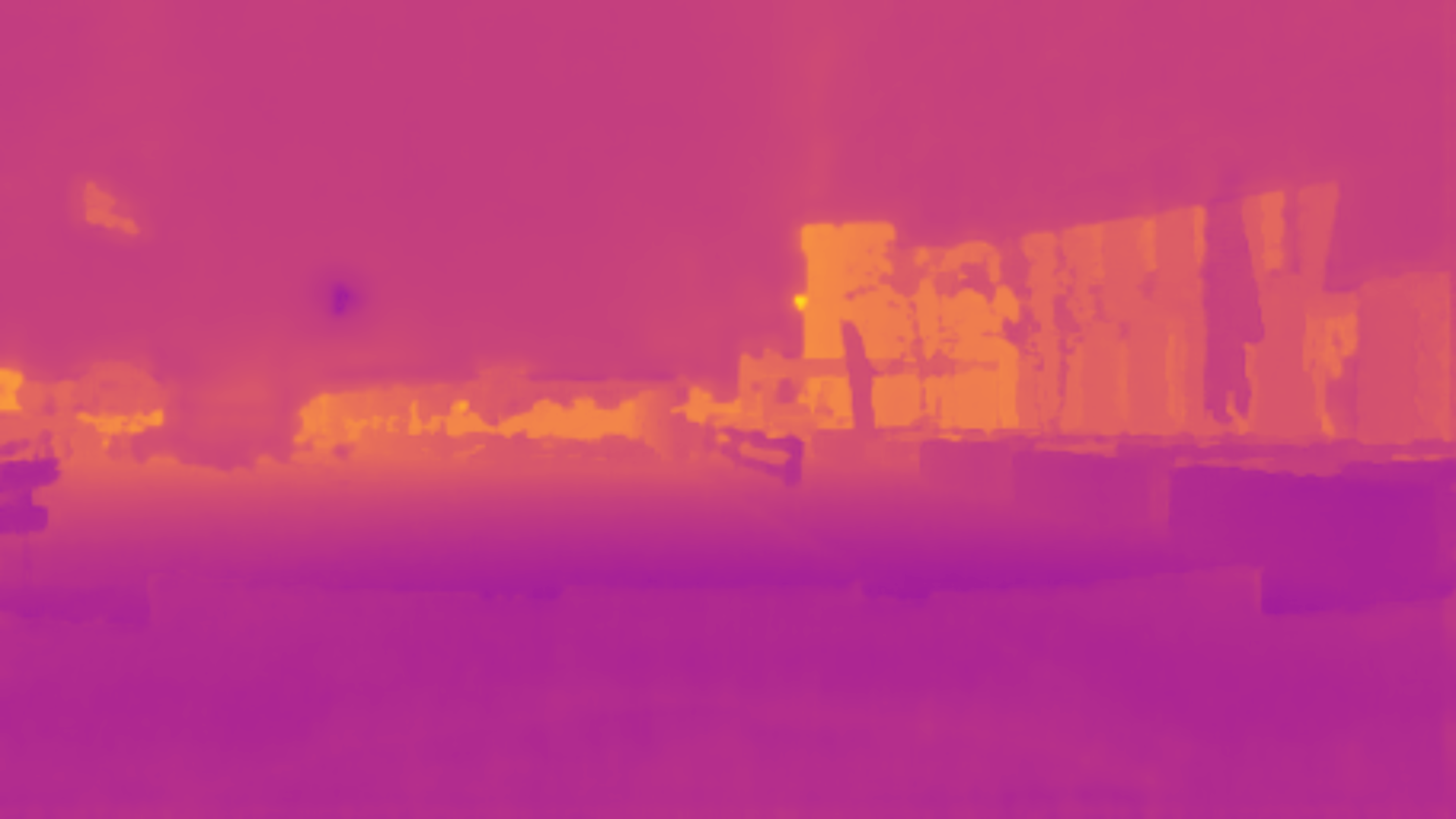} &
\includegraphics[width=0.10\textwidth]{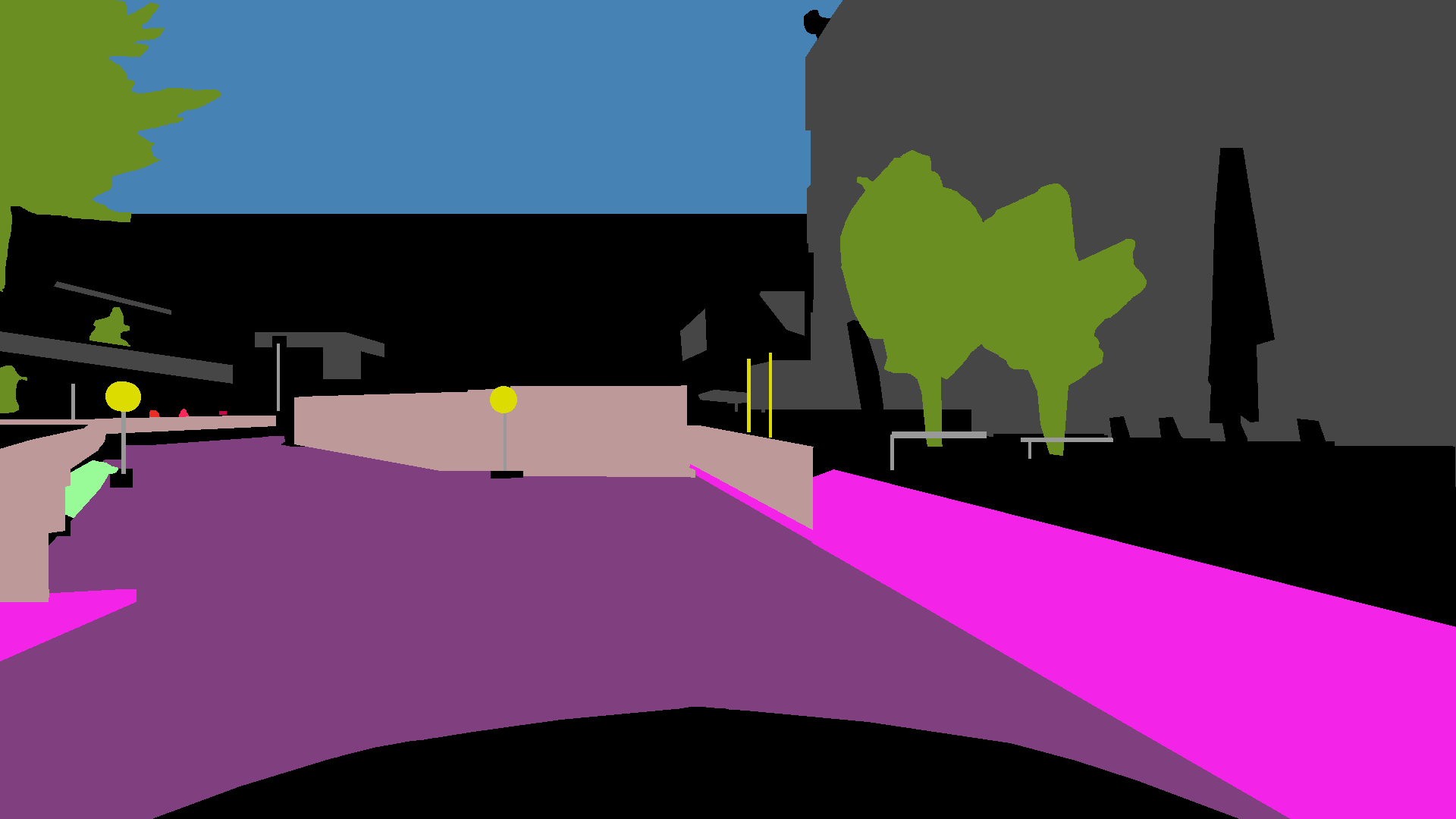} &
\includegraphics[width=0.10\textwidth]{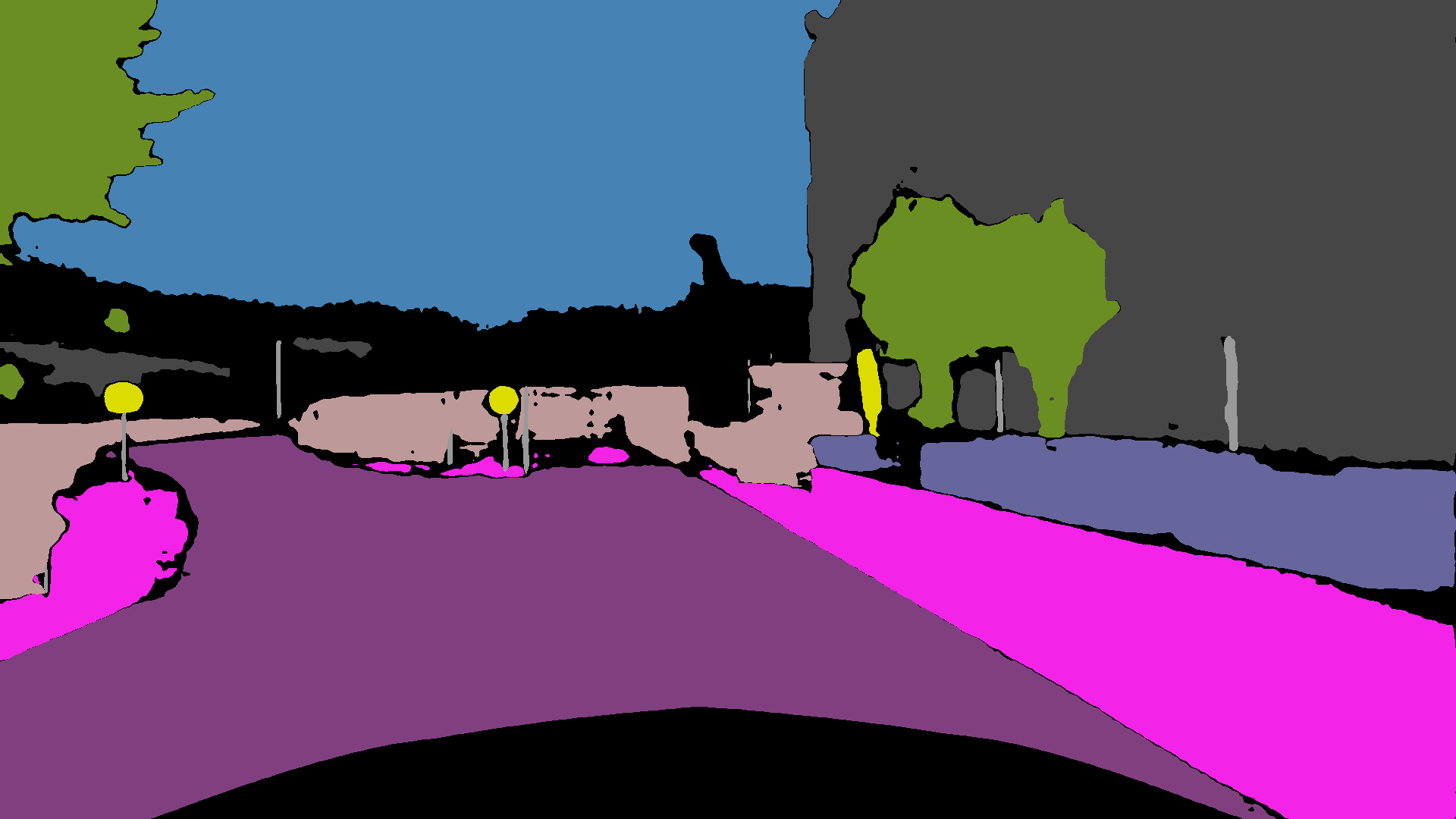} &
\includegraphics[width=0.10\textwidth]{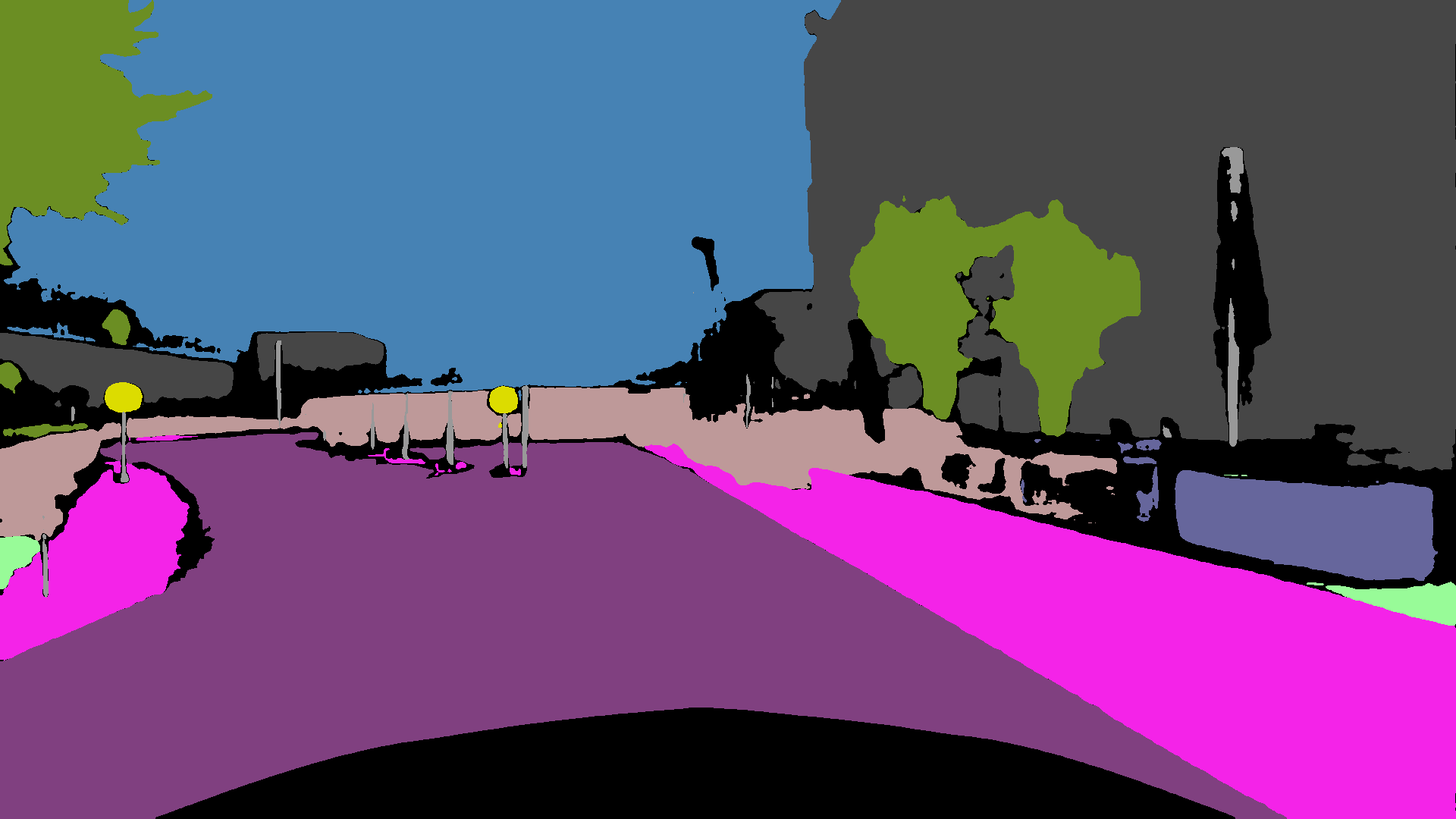} &
\includegraphics[width=0.10\textwidth]{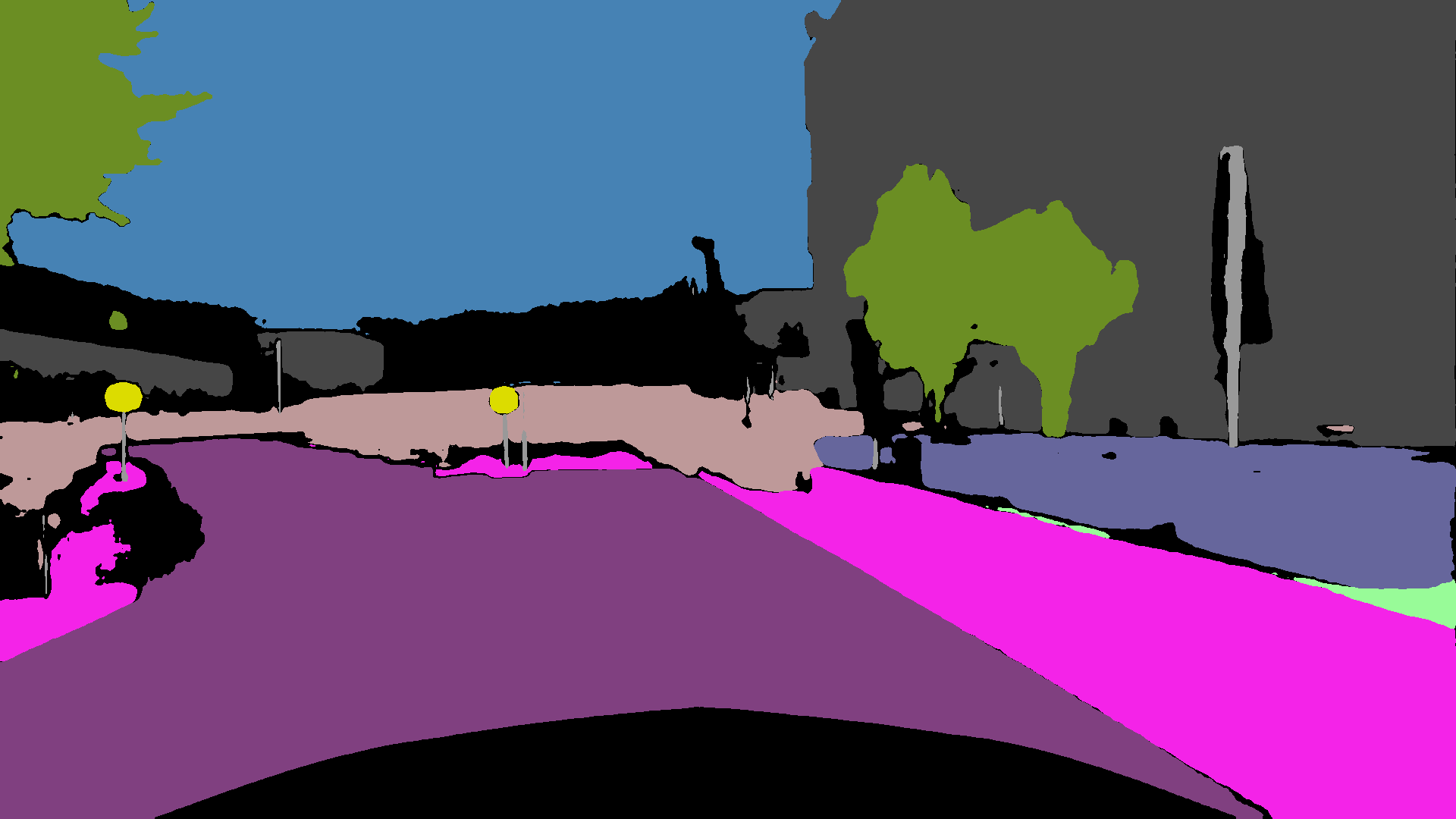} &
\includegraphics[width=0.10\textwidth]{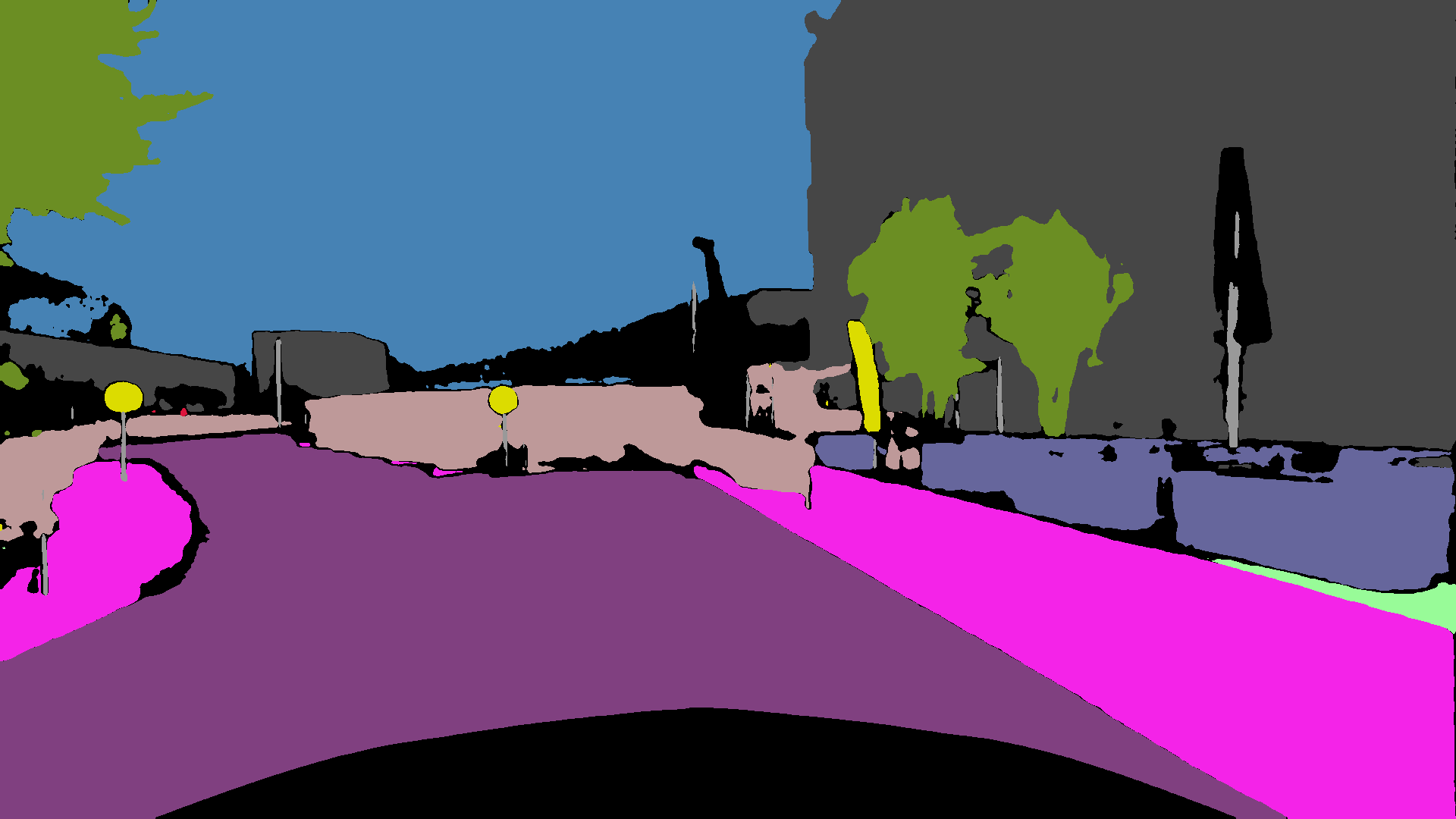} \\
\vspace{-0.1cm}

\end{tabular}
\caption{Further qualitative results on MUSES. 
Best viewed on a screen at full zoom.}
\label{fig:supl:muses_all}
\end{figure*}

In~\Cref{fig:supl:muses_all}, we present further qualitative comparisons across diverse scenarios on the MUSES dataset. Our proposed method \Ours{} shows improved robustness in scenarios challenging for RGB-only methods and baseline fusion approaches, such as perception at night, fog, and detection of distant objects. Specifically, it achieves superior instance separation and effectively leverages multimodal information under adverse conditions, thereby visually confirming the quantitative improvements in panoptic segmentation accuracy.

\end{document}

%% file: figures/qualitative_results.tex
\begin{figure*}[ht]
\centering
\begin{tabular}{@{}c@{\hspace{0.03cm}}
                c@{\hspace{0.03cm}}
                c@{\hspace{0.03cm}}
                c@{\hspace{0.03cm}}
                c@{\hspace{0.03cm}}
                c@{\hspace{0.03cm}}
                c@{\hspace{0.03cm}}
                c@{\hspace{0.03cm}}
                c@{}
                c@{}}
\subfloat{\scriptsize RGB} &
\subfloat{\scriptsize Lidar} &
\subfloat{\scriptsize Radar} &
\subfloat{\scriptsize Events} &
\subfloat{\scriptsize Ours-depth} &
\subfloat{\scriptsize GT} &
\subfloat{\scriptsize MUSES~\cite{MUSES}} &
\subfloat{\scriptsize OneFormer~\cite{jain2023oneformer}} &
\subfloat{\scriptsize CAFuser~\cite{cafuser}} &
\subfloat{\scriptsize \Ours{}~(\textbf{Ours})} \\
\vspace{-0.1cm}

\includegraphics[width=0.095\textwidth]{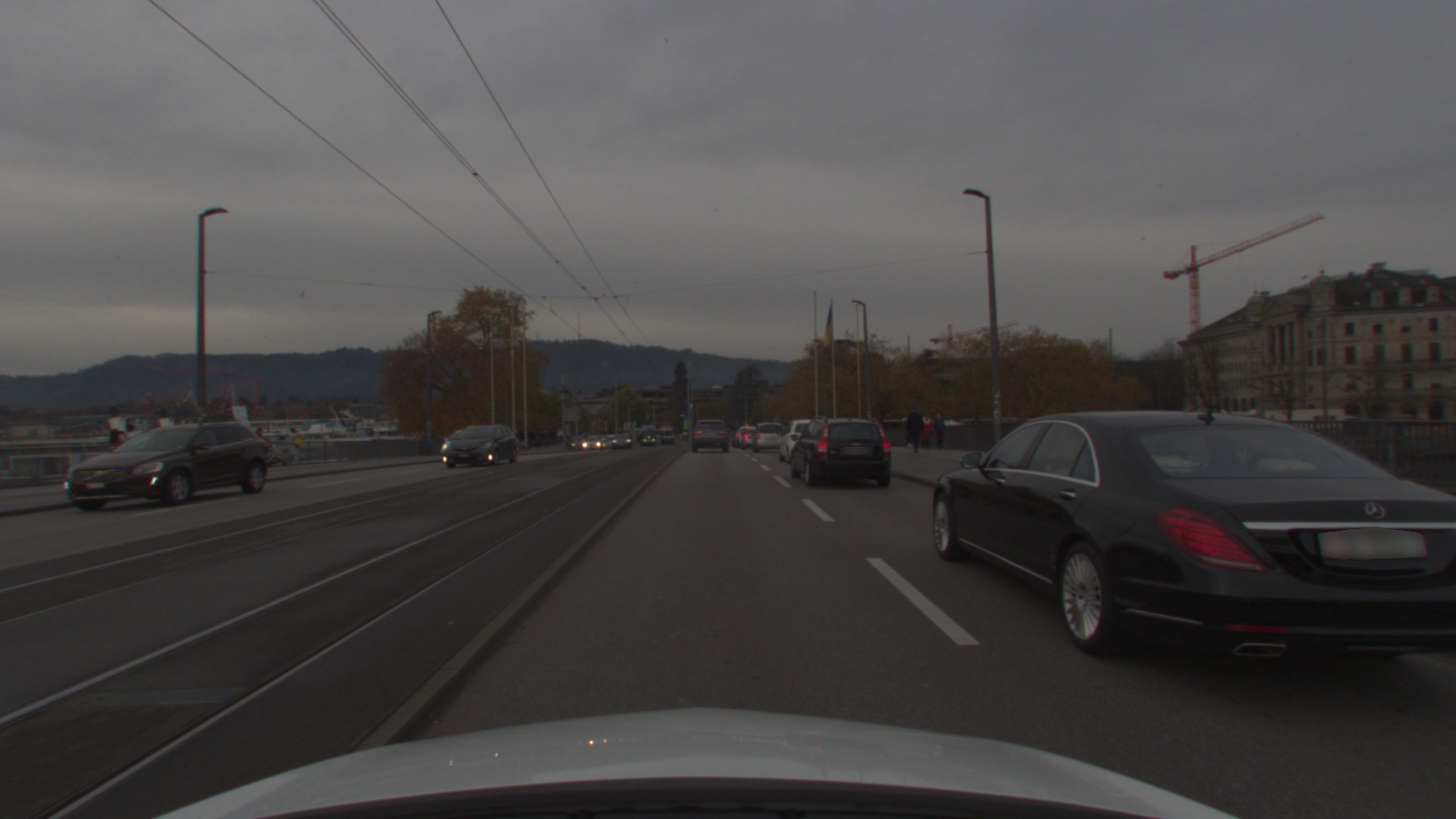} &
\includegraphics[width=0.095\textwidth]{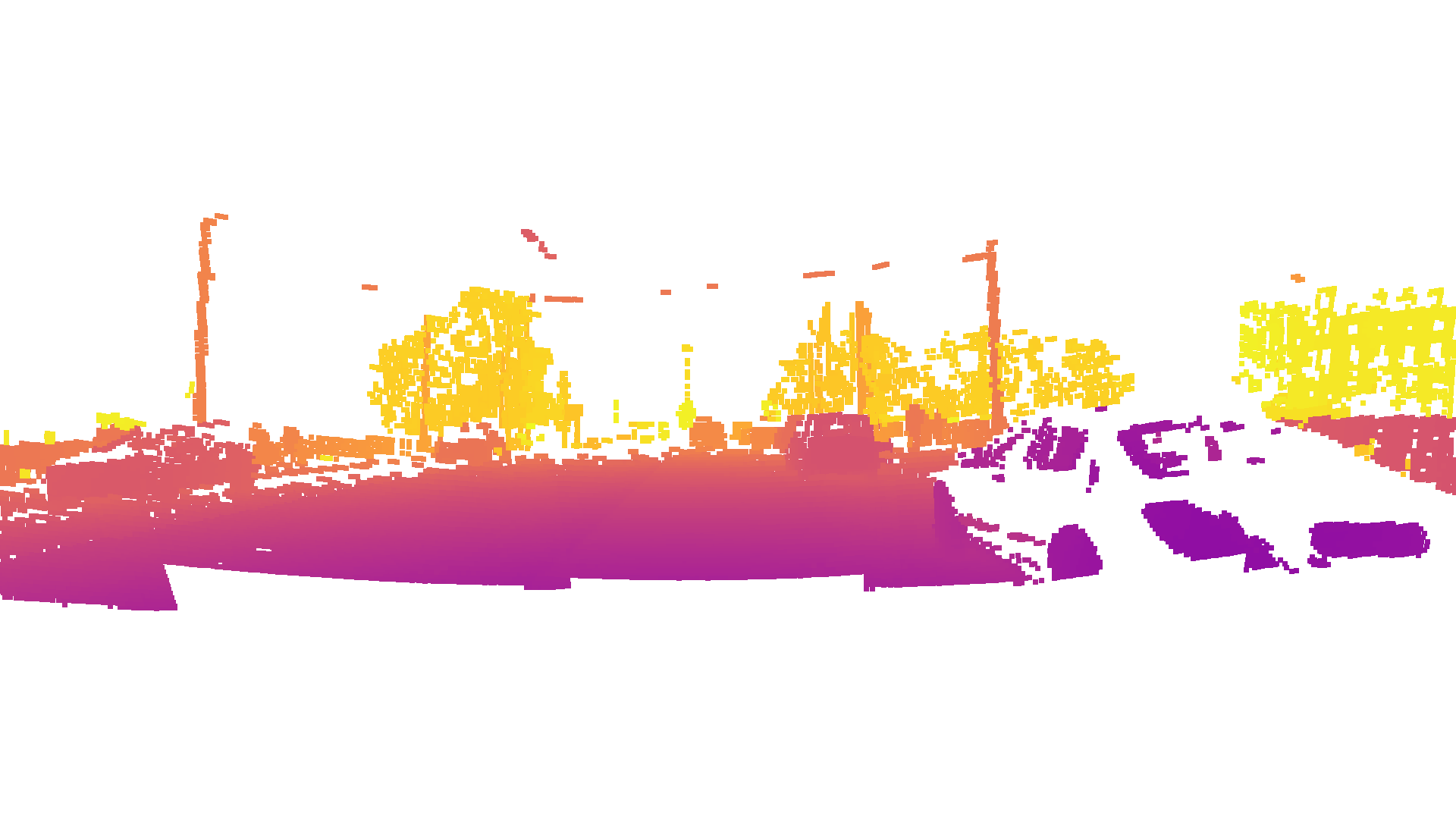} &
\includegraphics[width=0.095\textwidth]{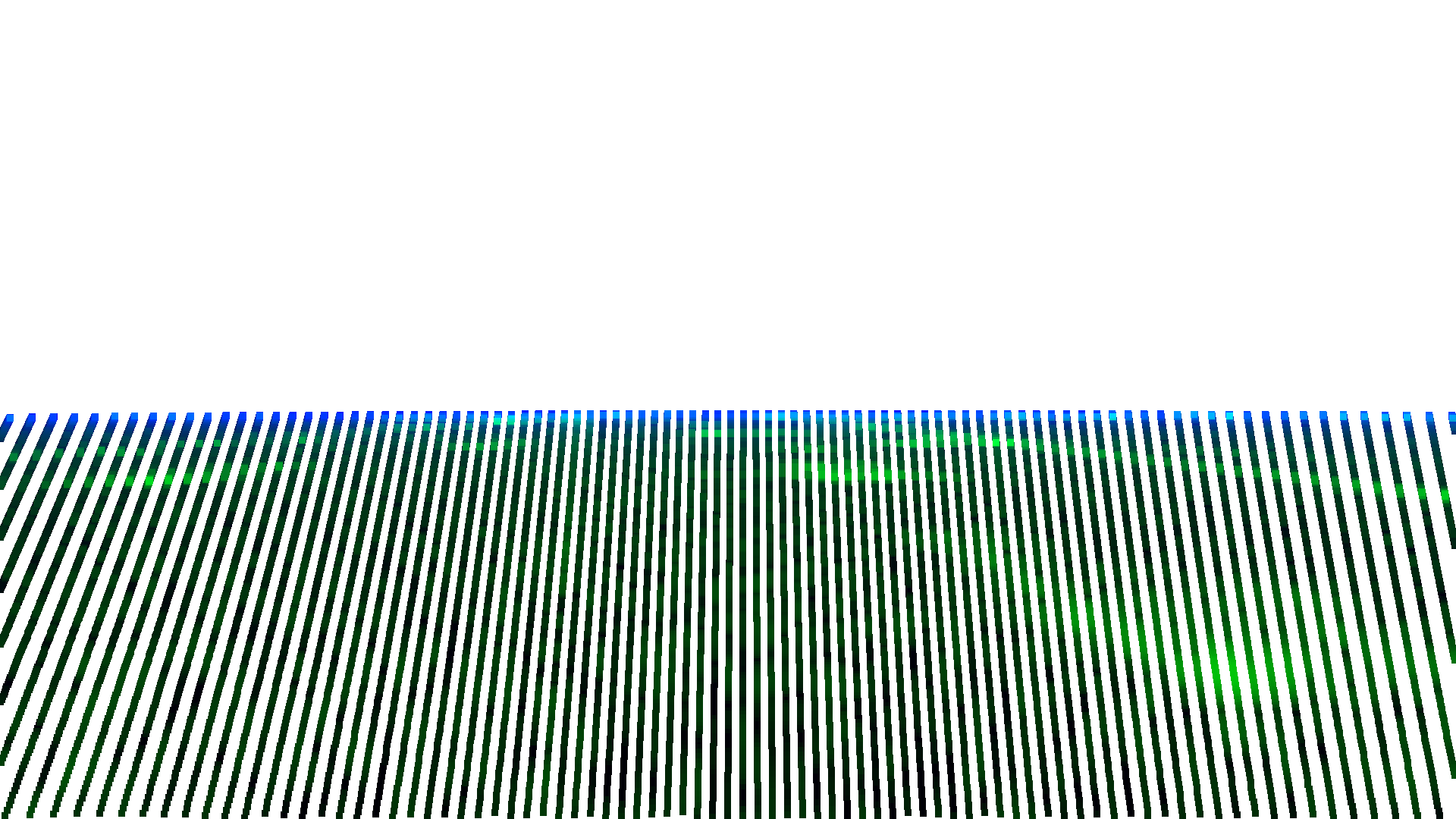} &
\includegraphics[width=0.095\textwidth]{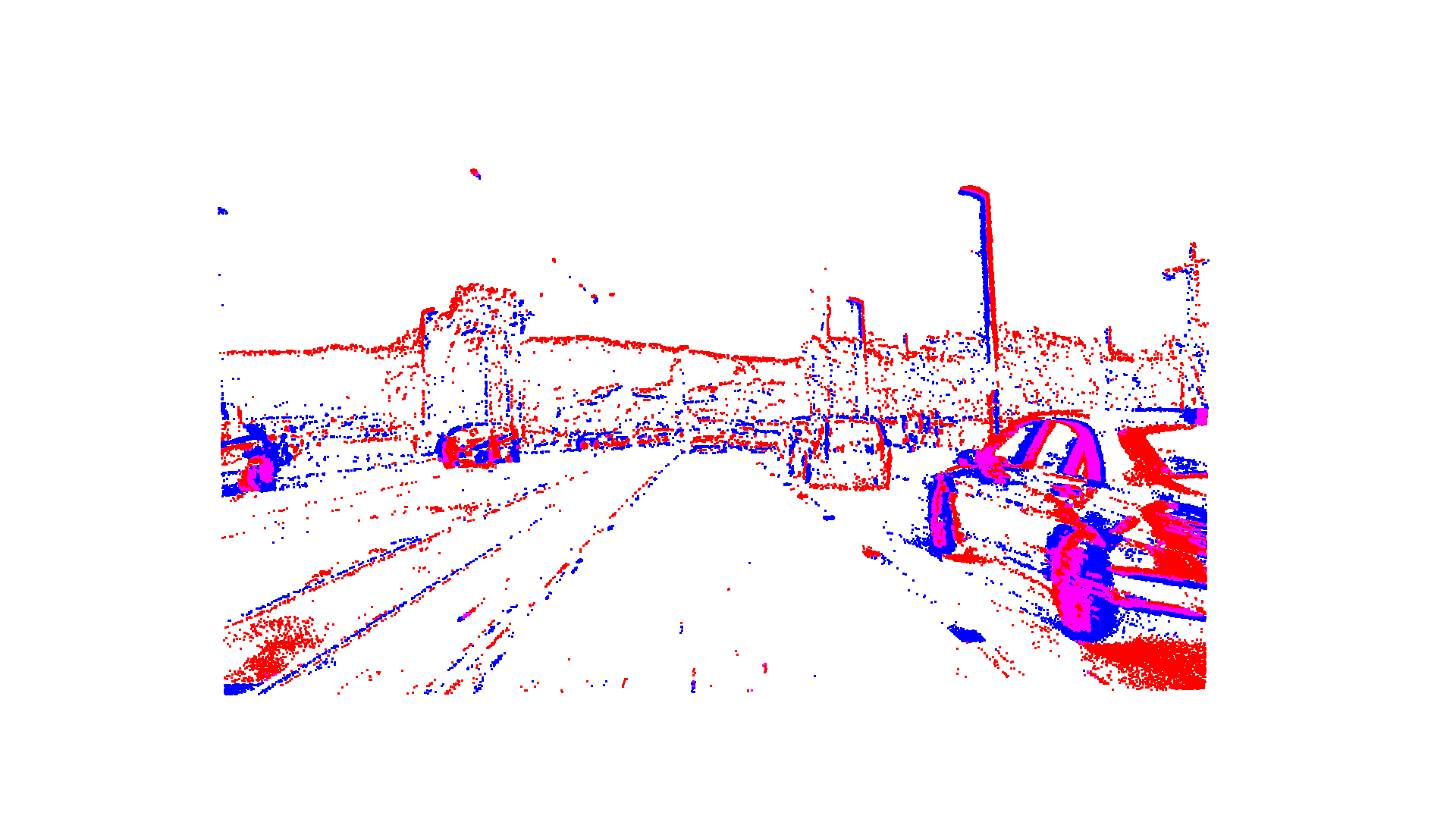} &
\includegraphics[width=0.095\textwidth]{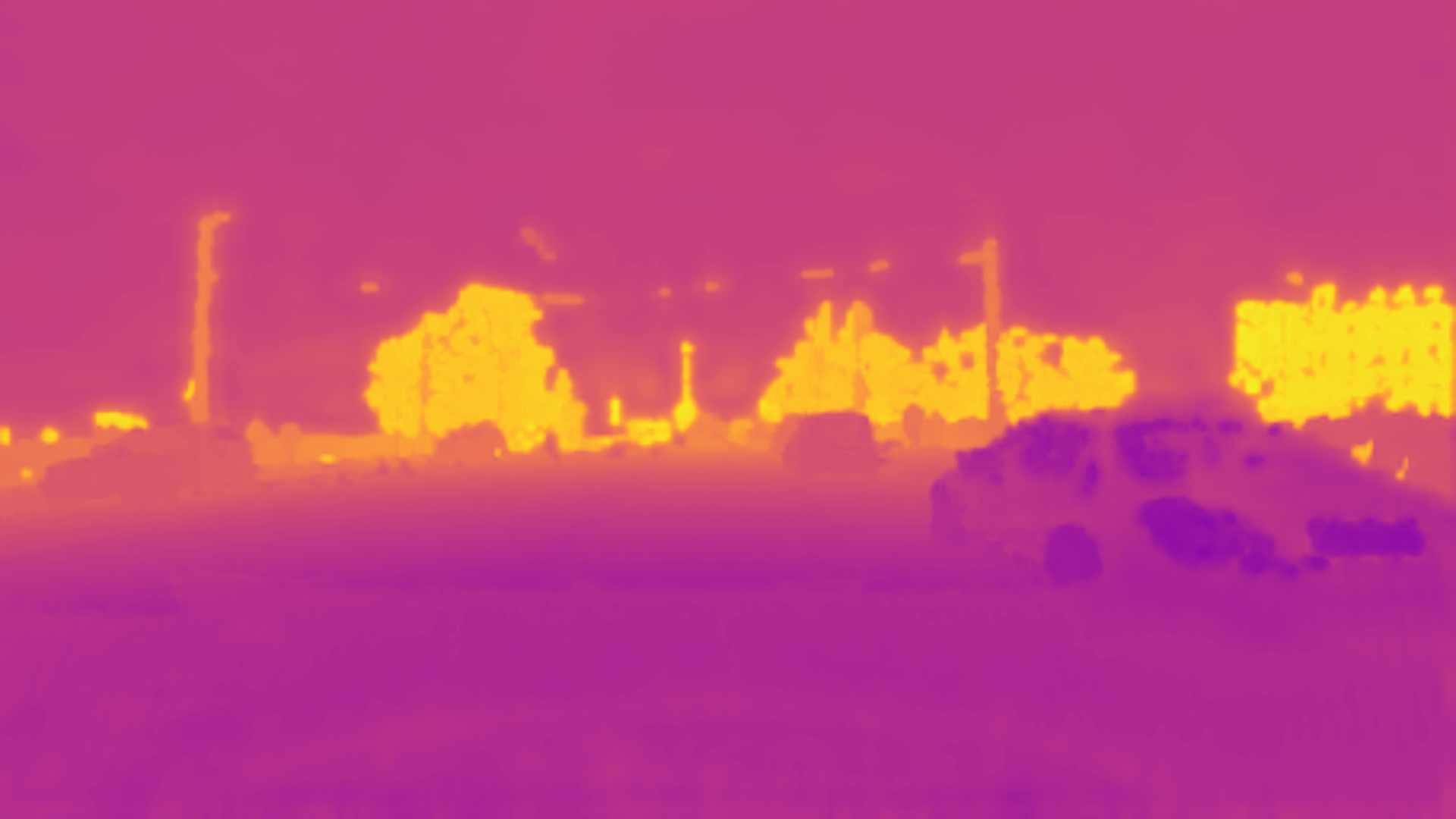} &
\includegraphics[width=0.095\textwidth]{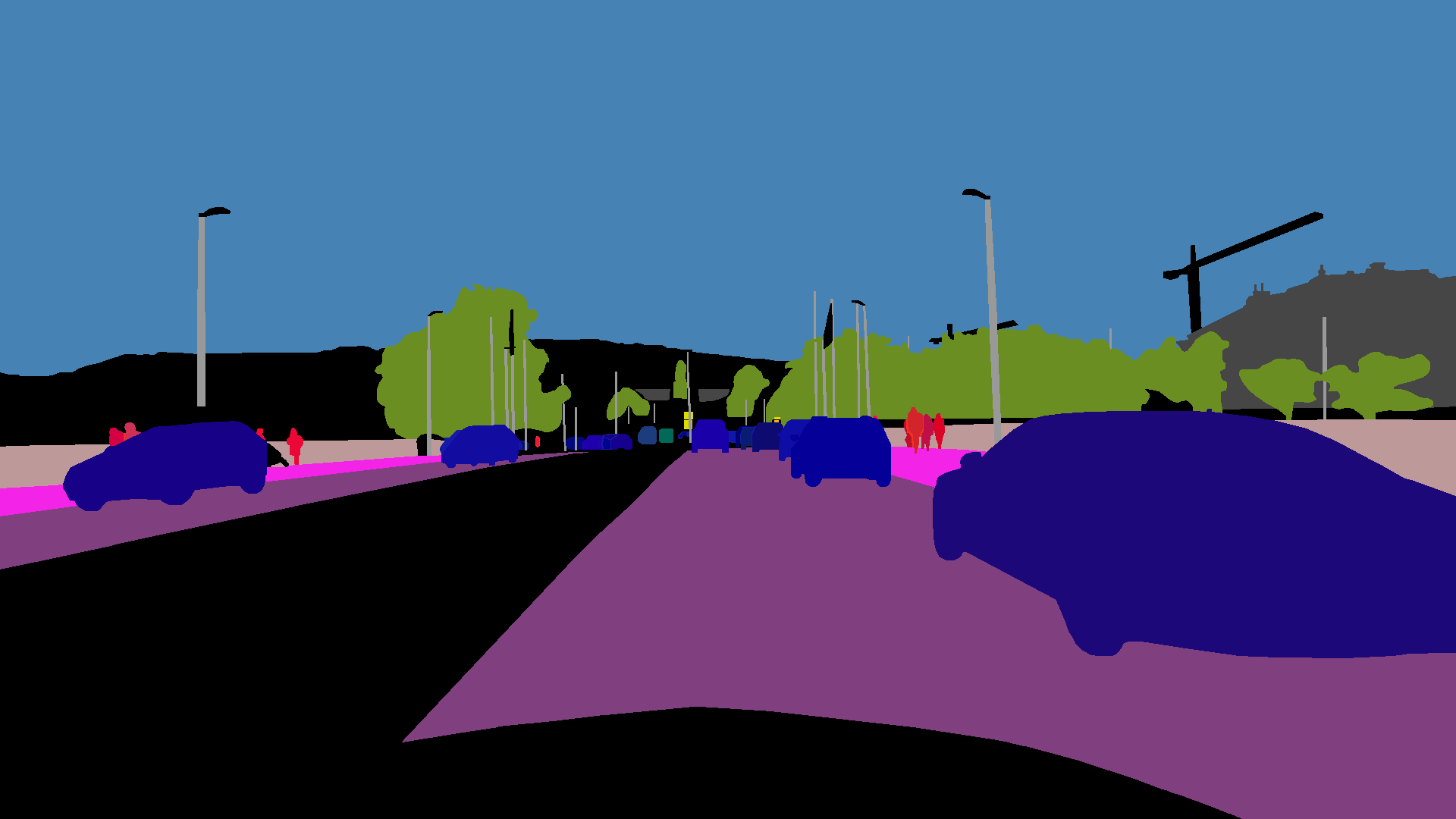} &
\includegraphics[width=0.095\textwidth]{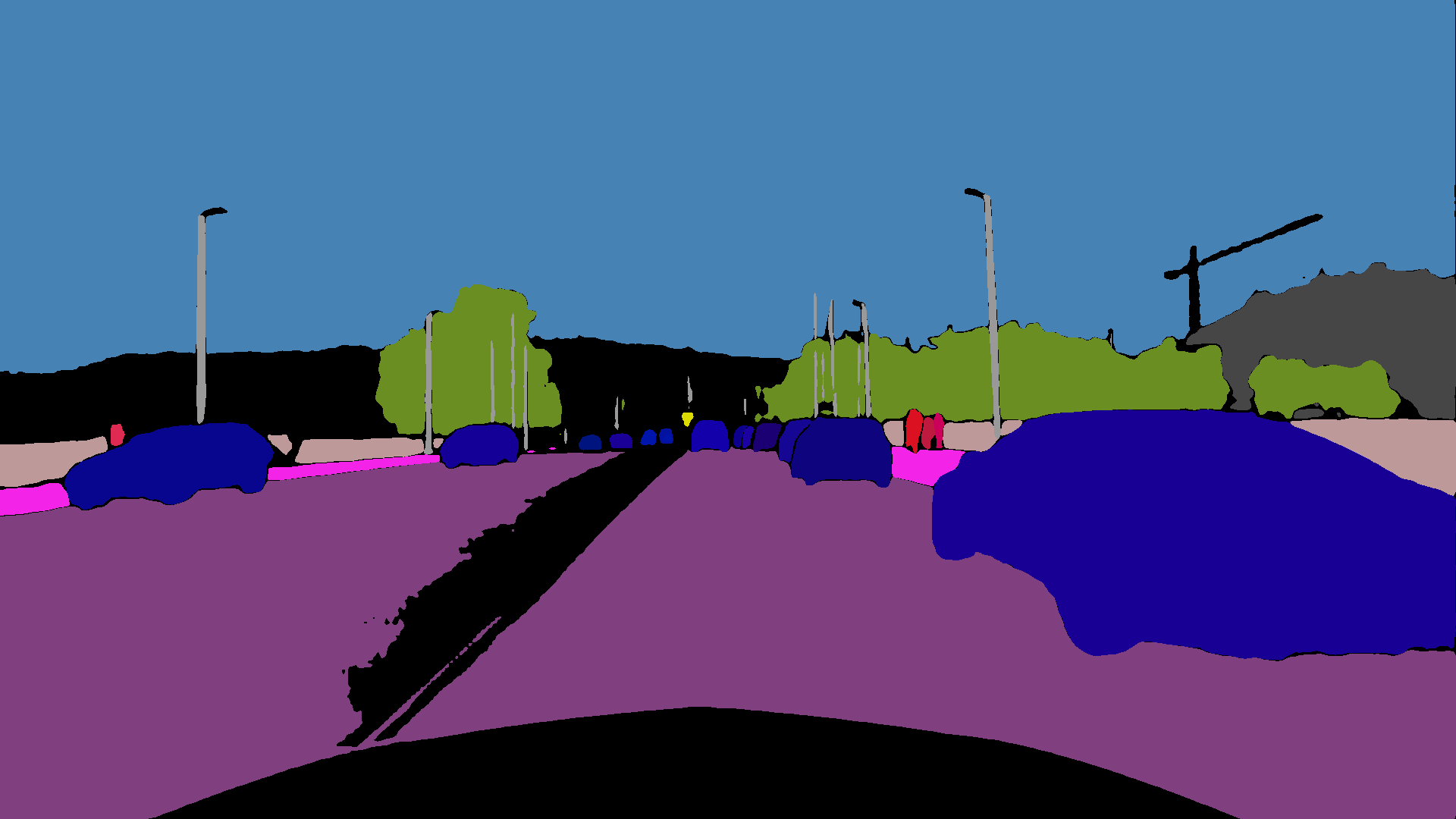} &
\includegraphics[width=0.095\textwidth]{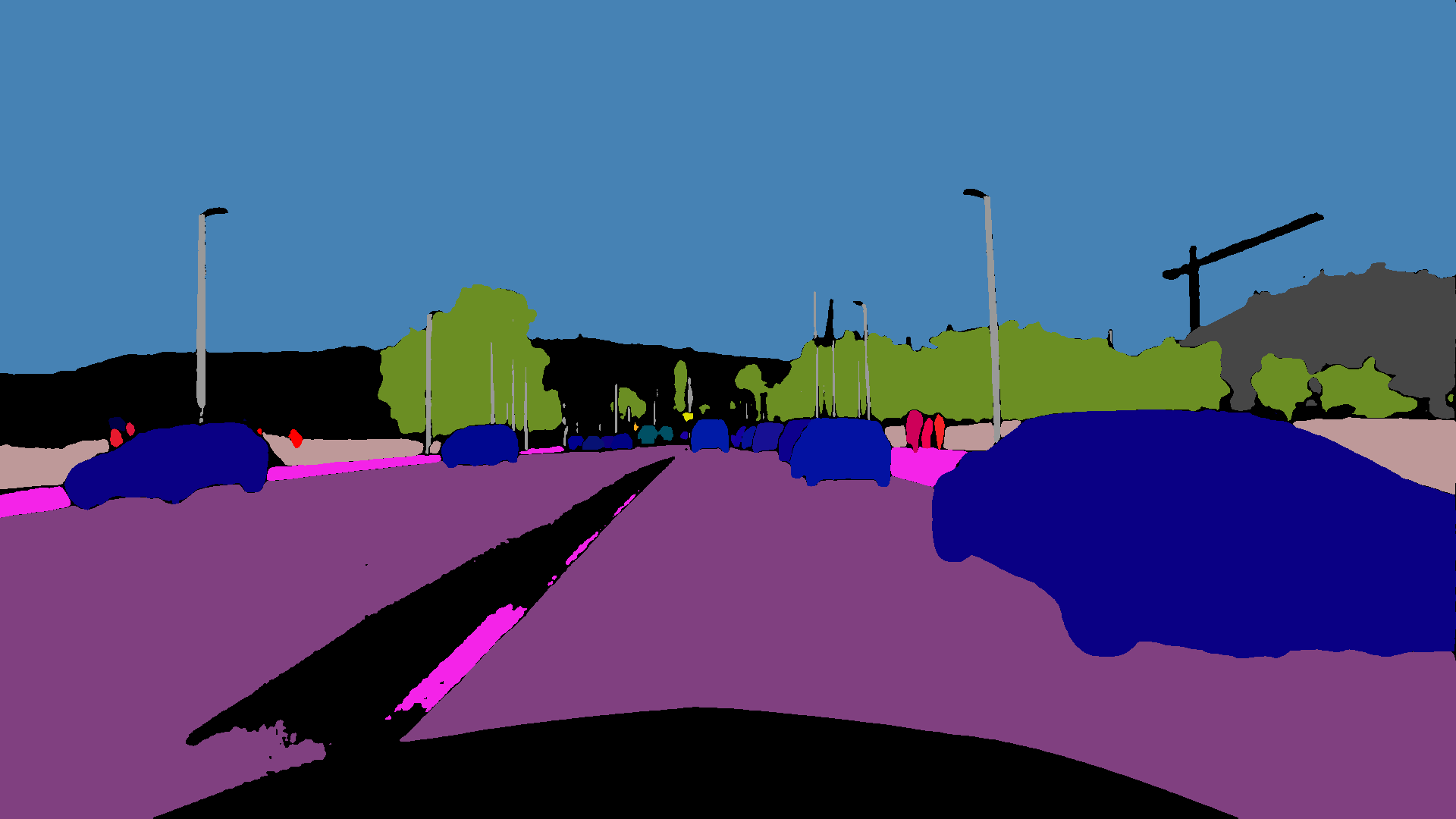} &
\includegraphics[width=0.095\textwidth]{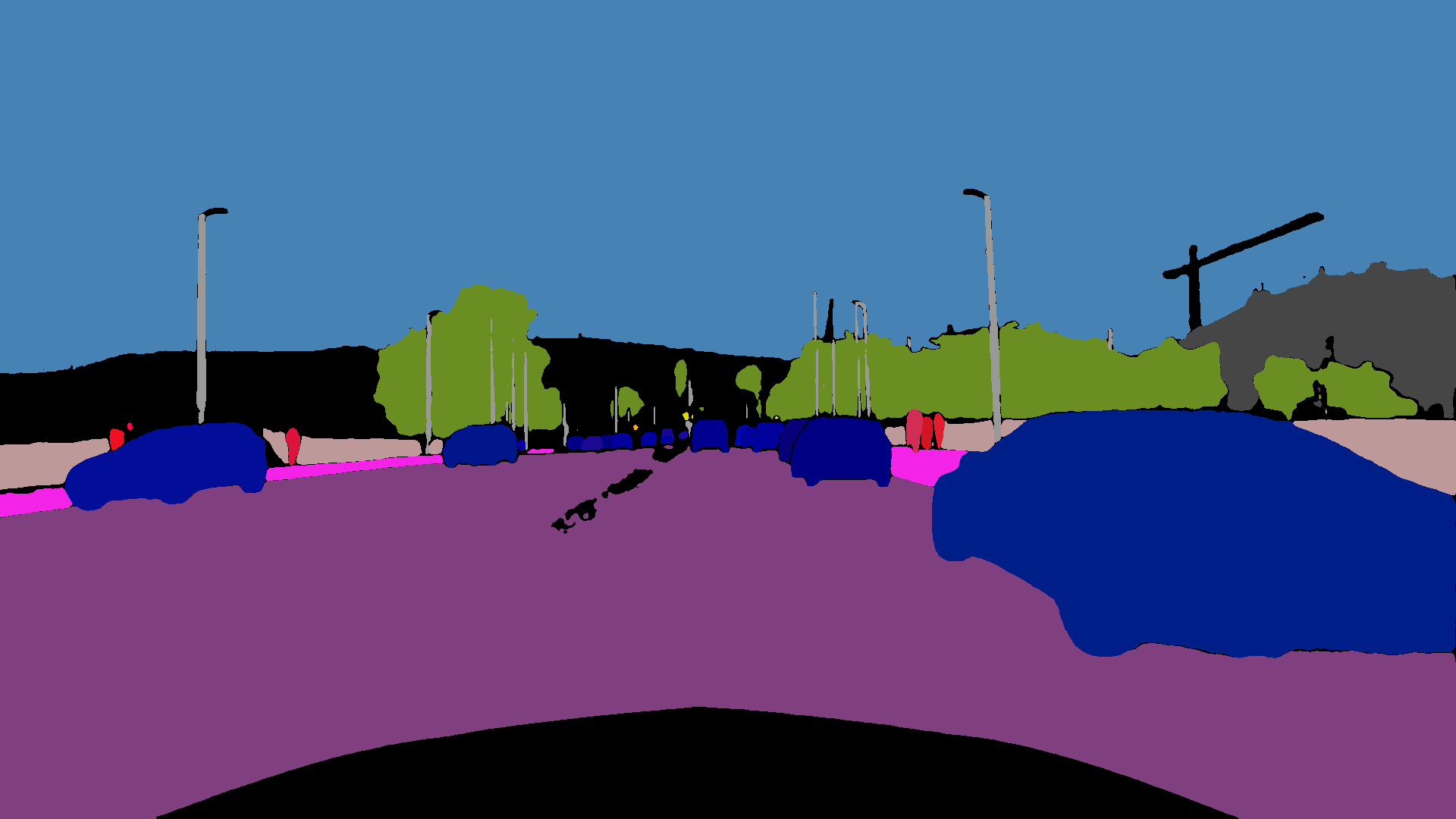} &
\includegraphics[width=0.095\textwidth]{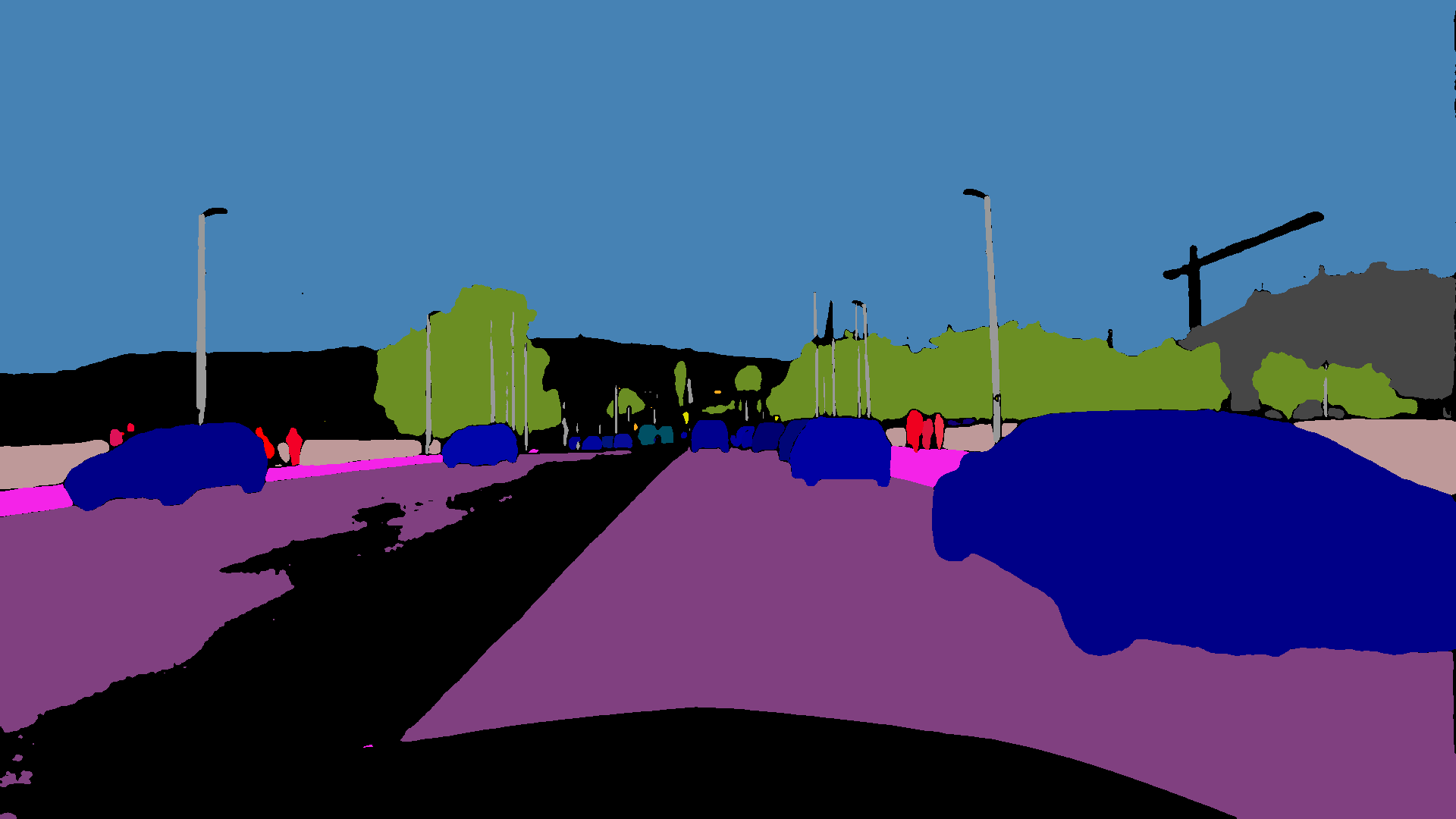} \\
\vspace{-0.1cm}

\includegraphics[width=0.095\textwidth]{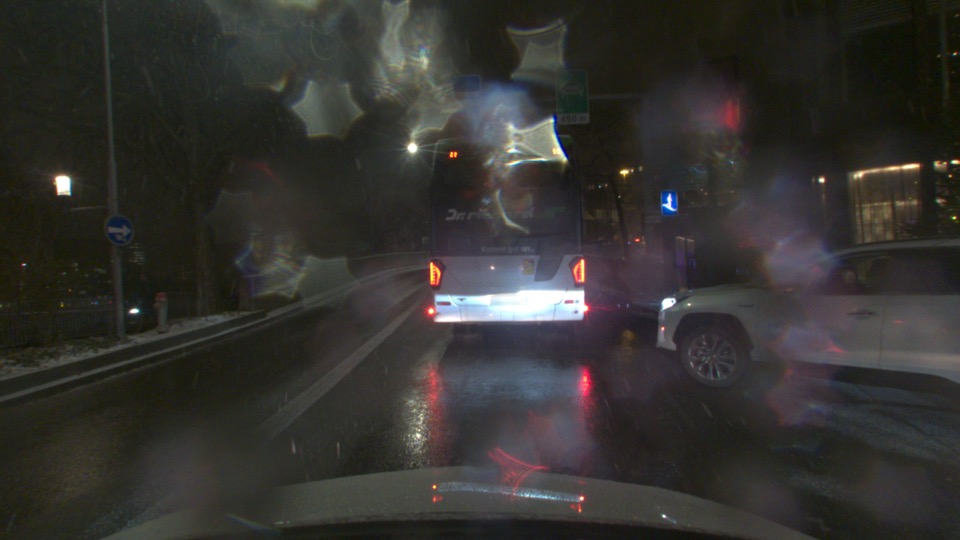} &
\includegraphics[width=0.095\textwidth]{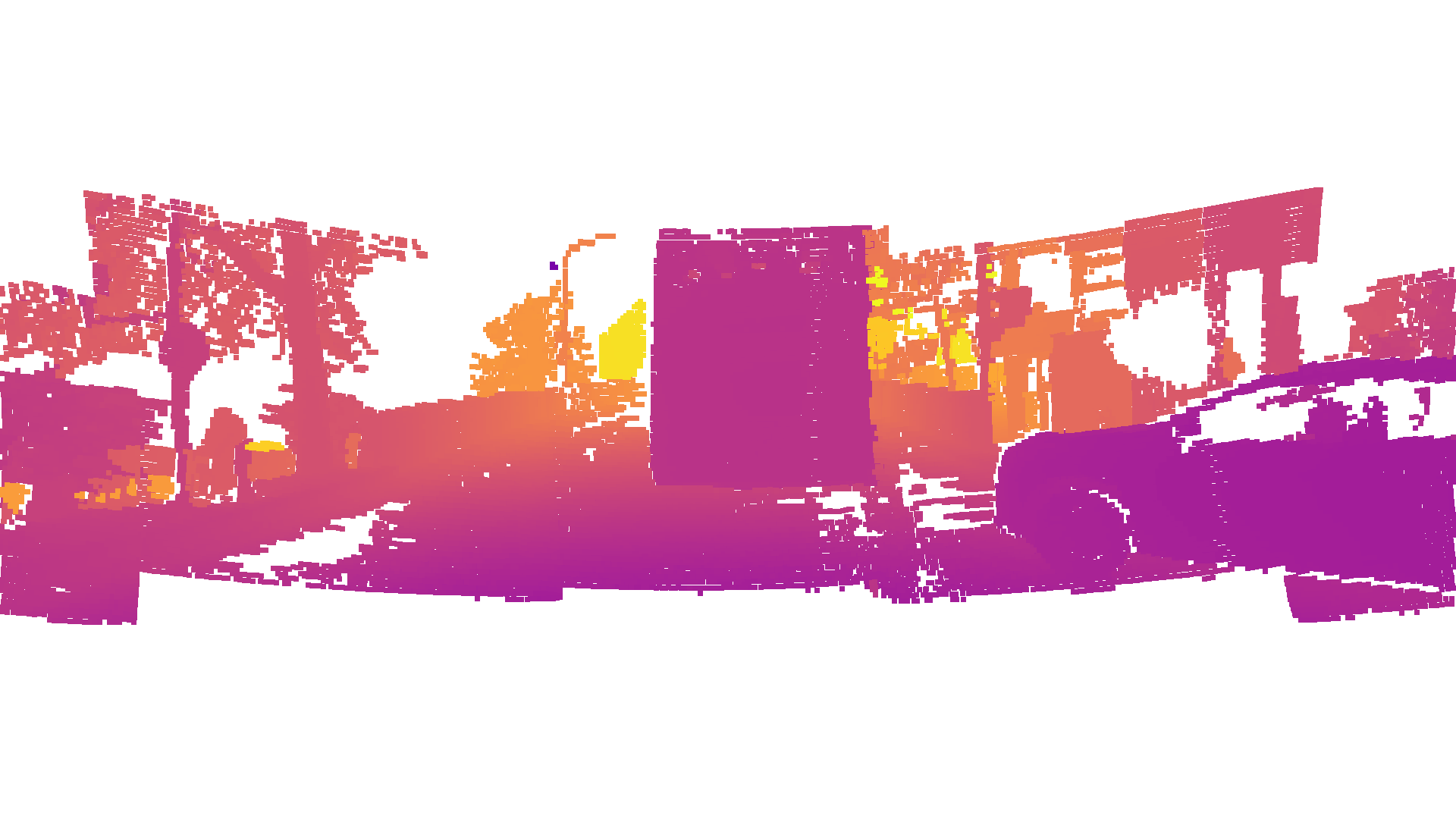} &
\includegraphics[width=0.095\textwidth]{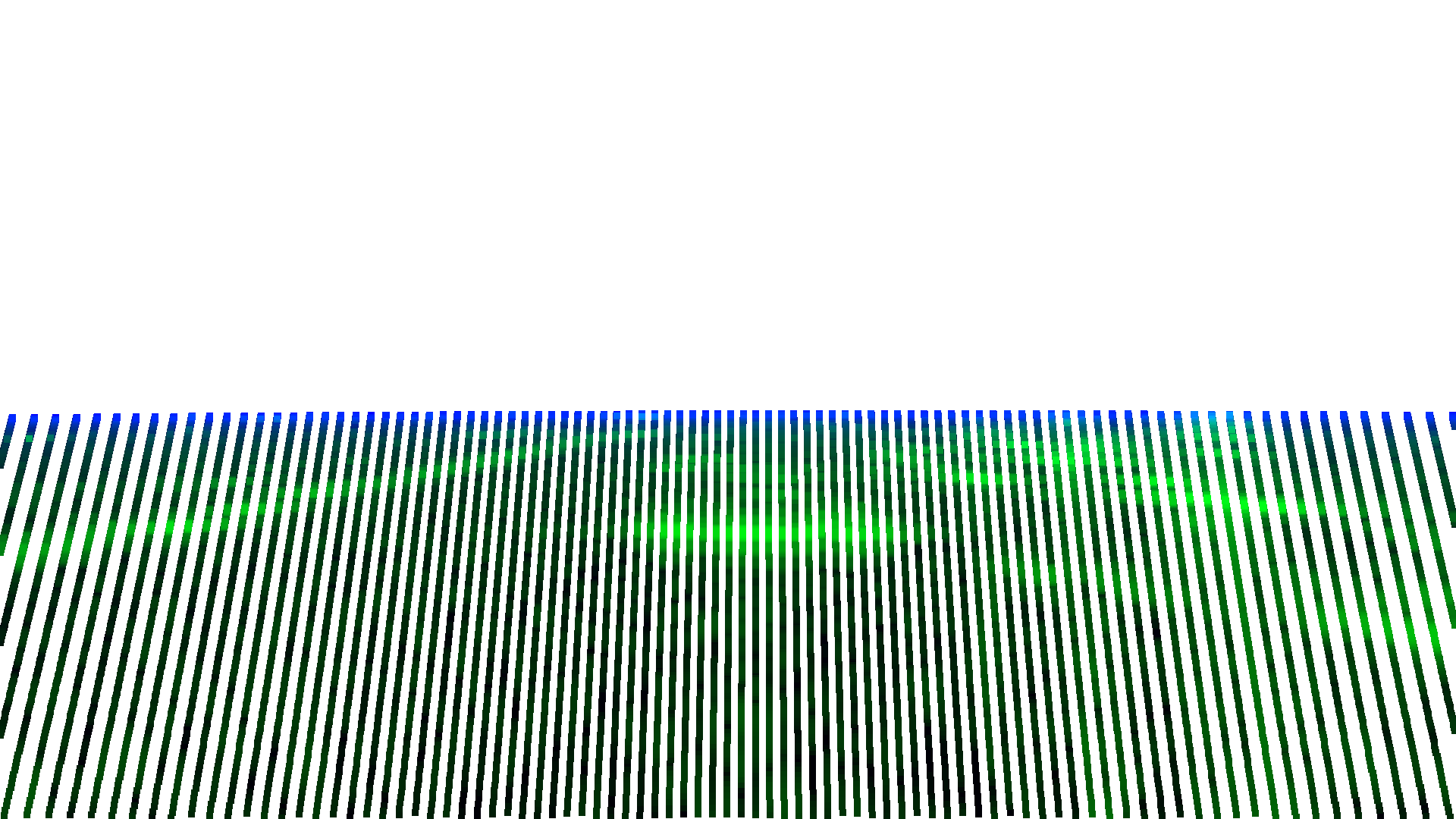} &
\includegraphics[width=0.095\textwidth]{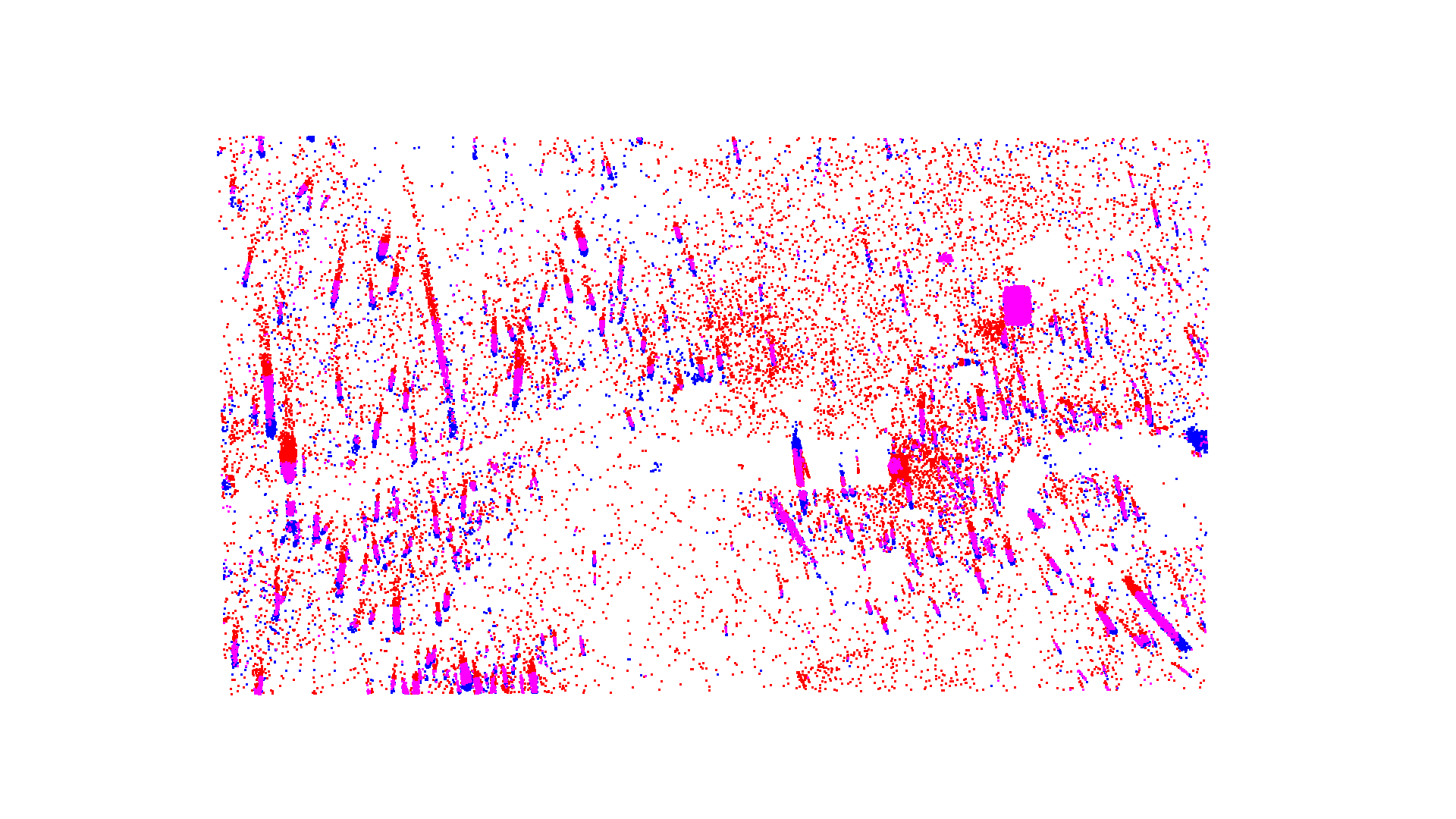} &
\includegraphics[width=0.095\textwidth]{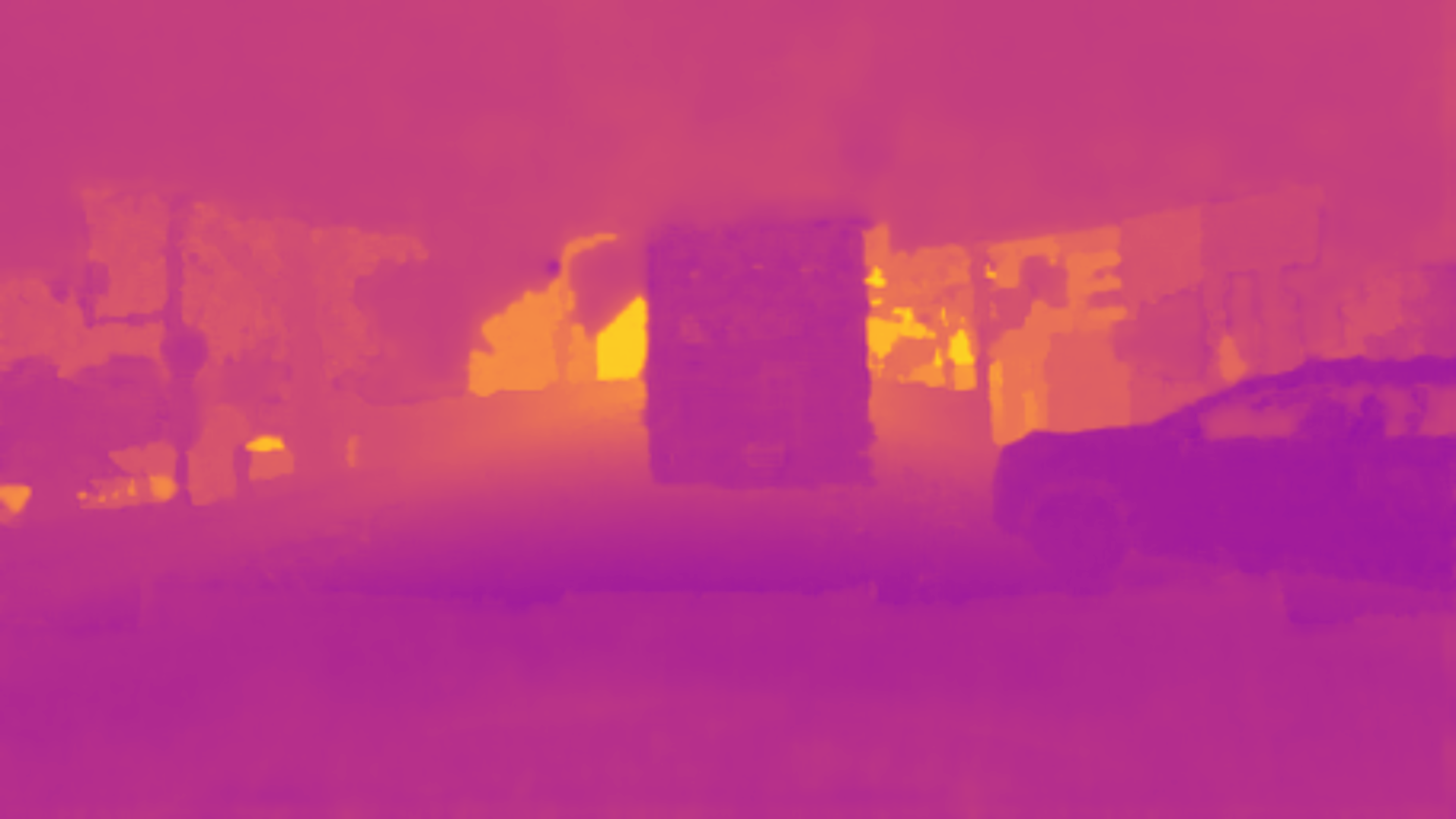} &
\includegraphics[width=0.095\textwidth]{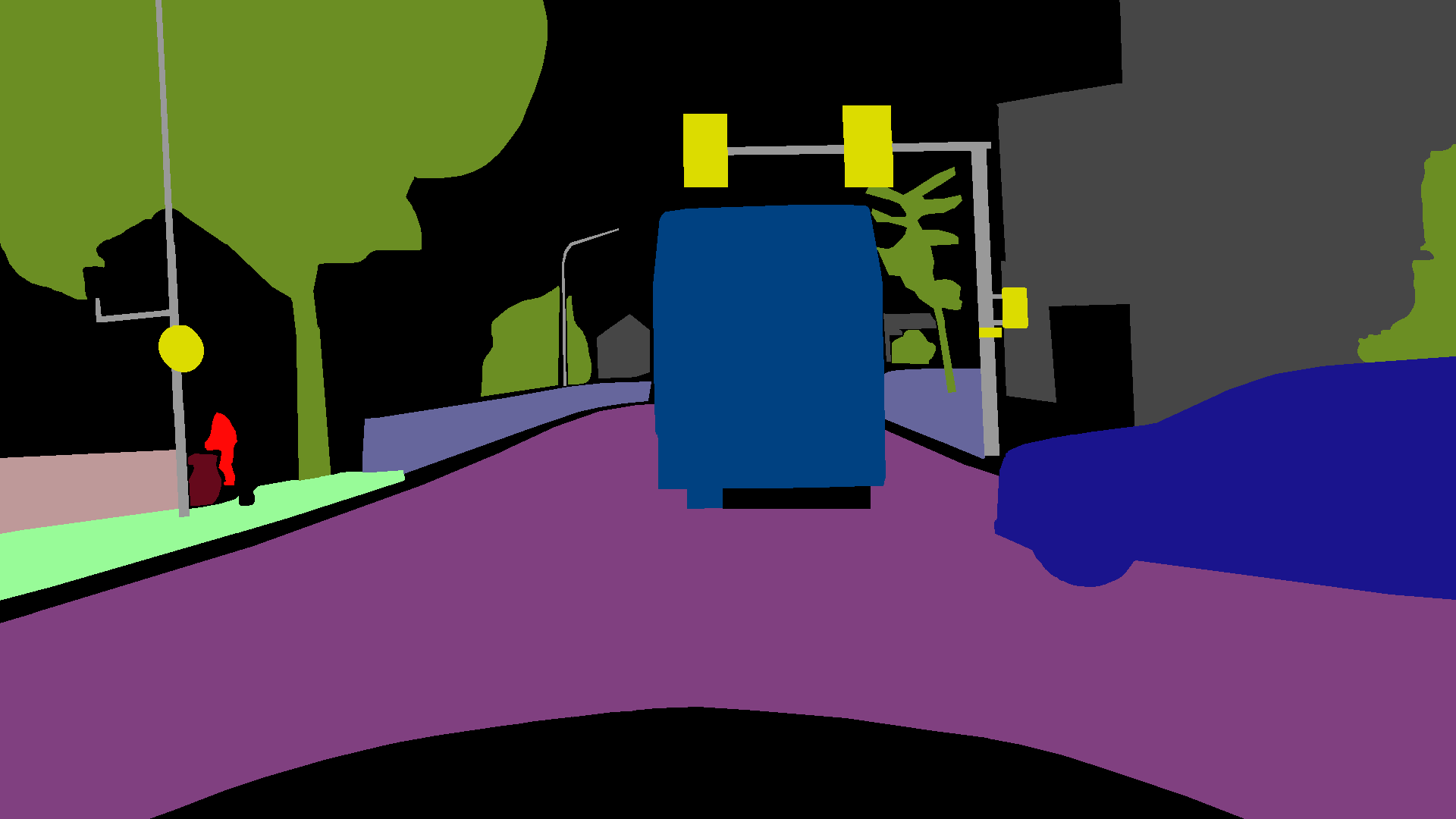} &
\includegraphics[width=0.095\textwidth]{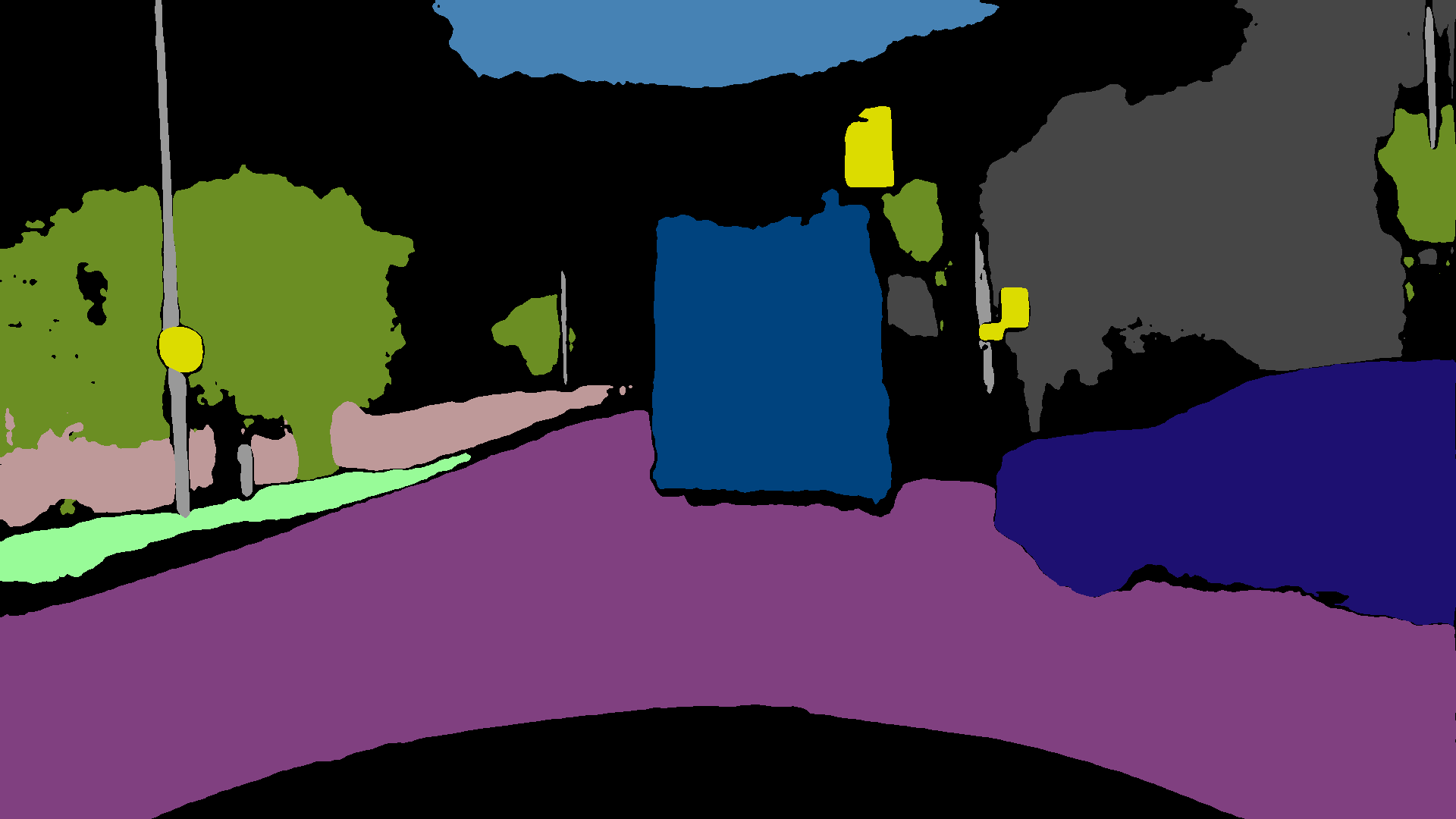} &
\includegraphics[width=0.095\textwidth]{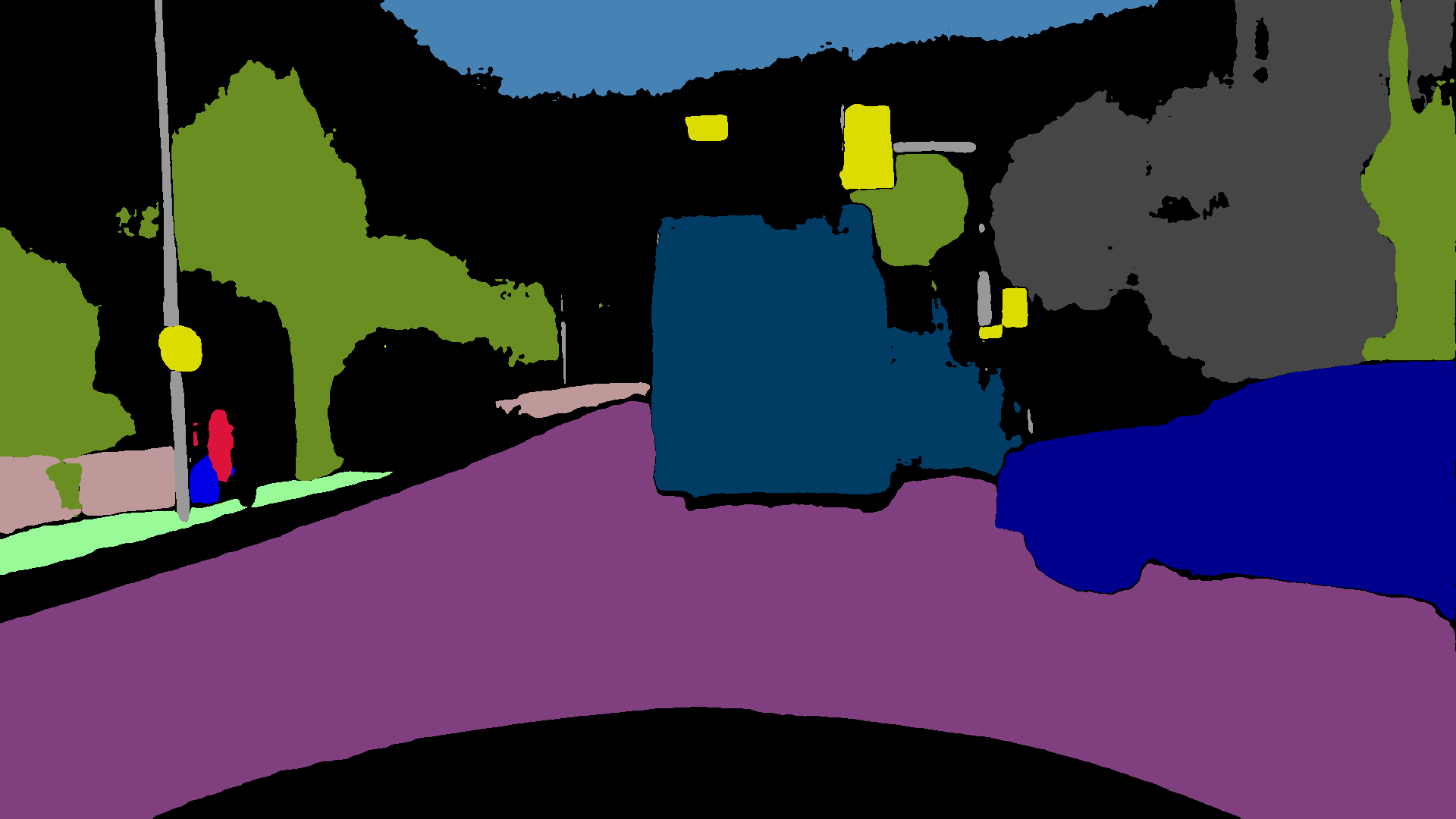} &
\includegraphics[width=0.095\textwidth]{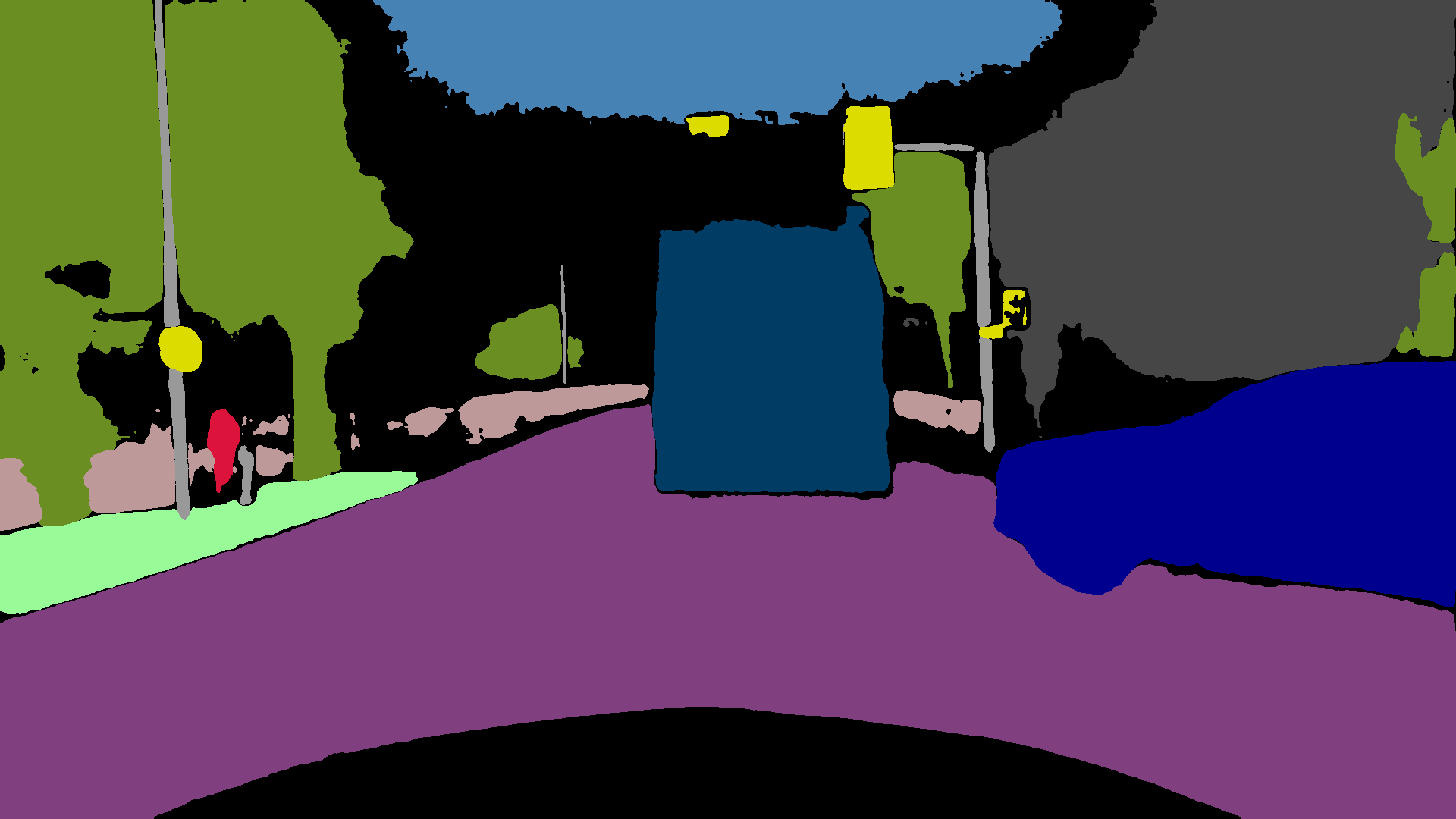} &
\includegraphics[width=0.095\textwidth]{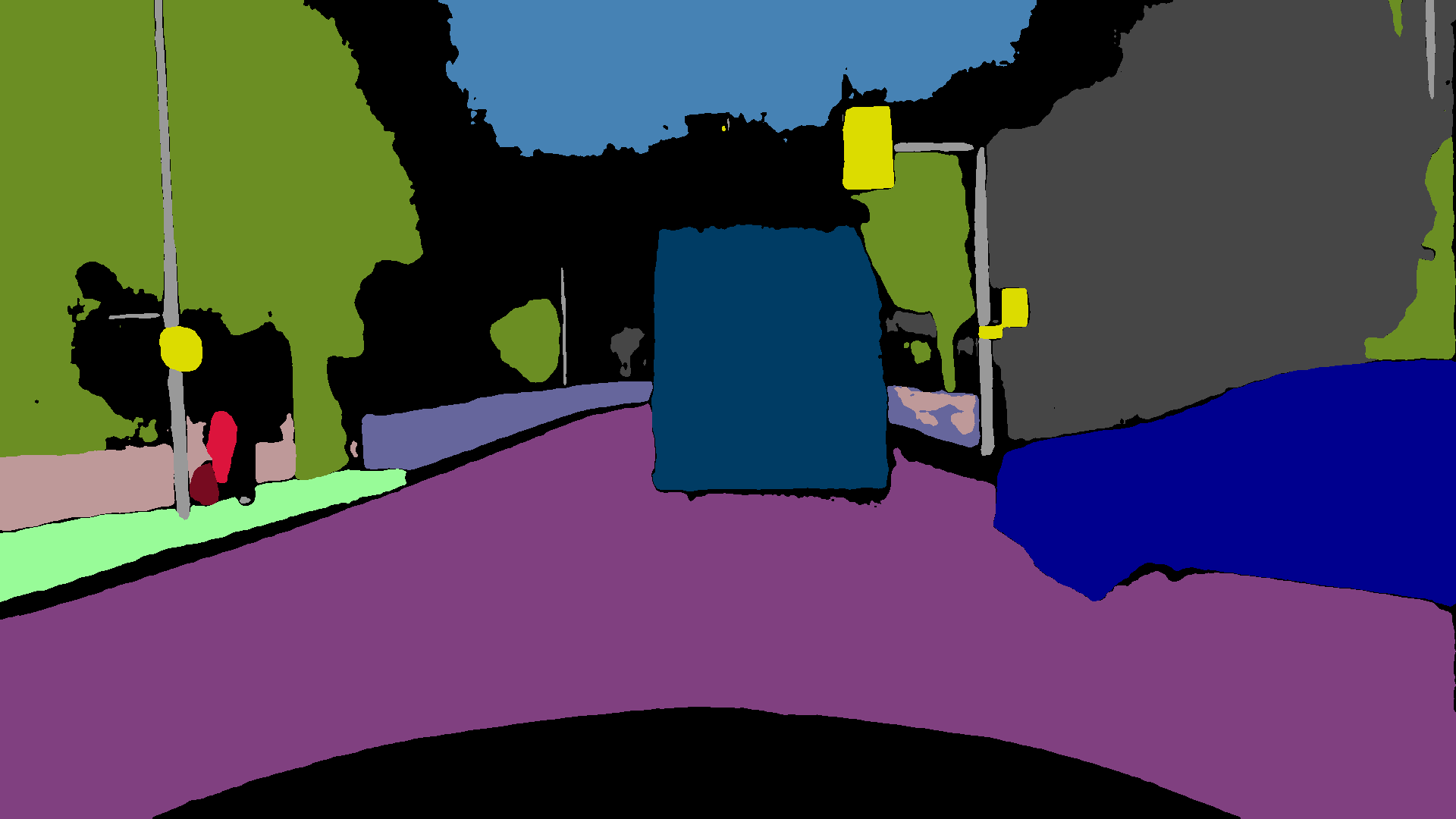} \\
\vspace{-0.1cm}

\includegraphics[width=0.095\textwidth]{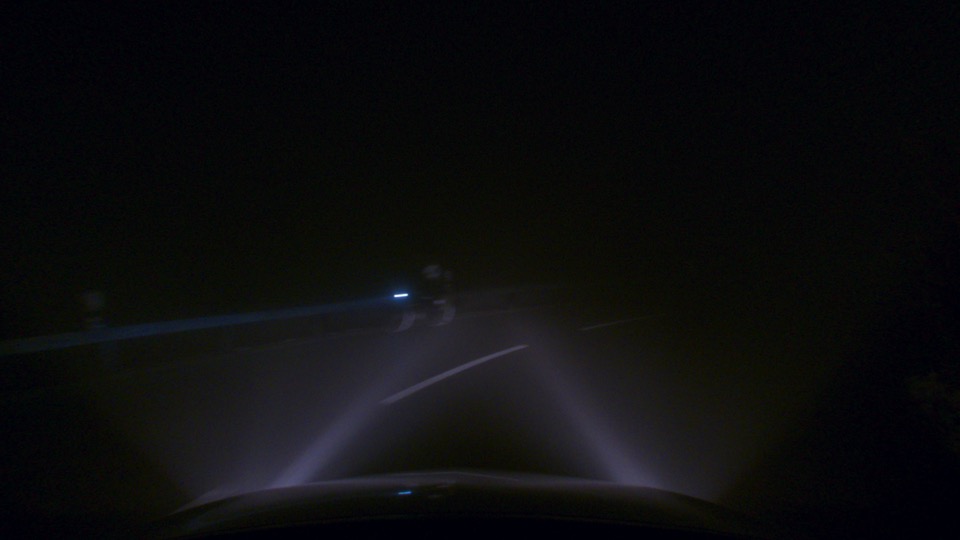} &
\includegraphics[width=0.095\textwidth]{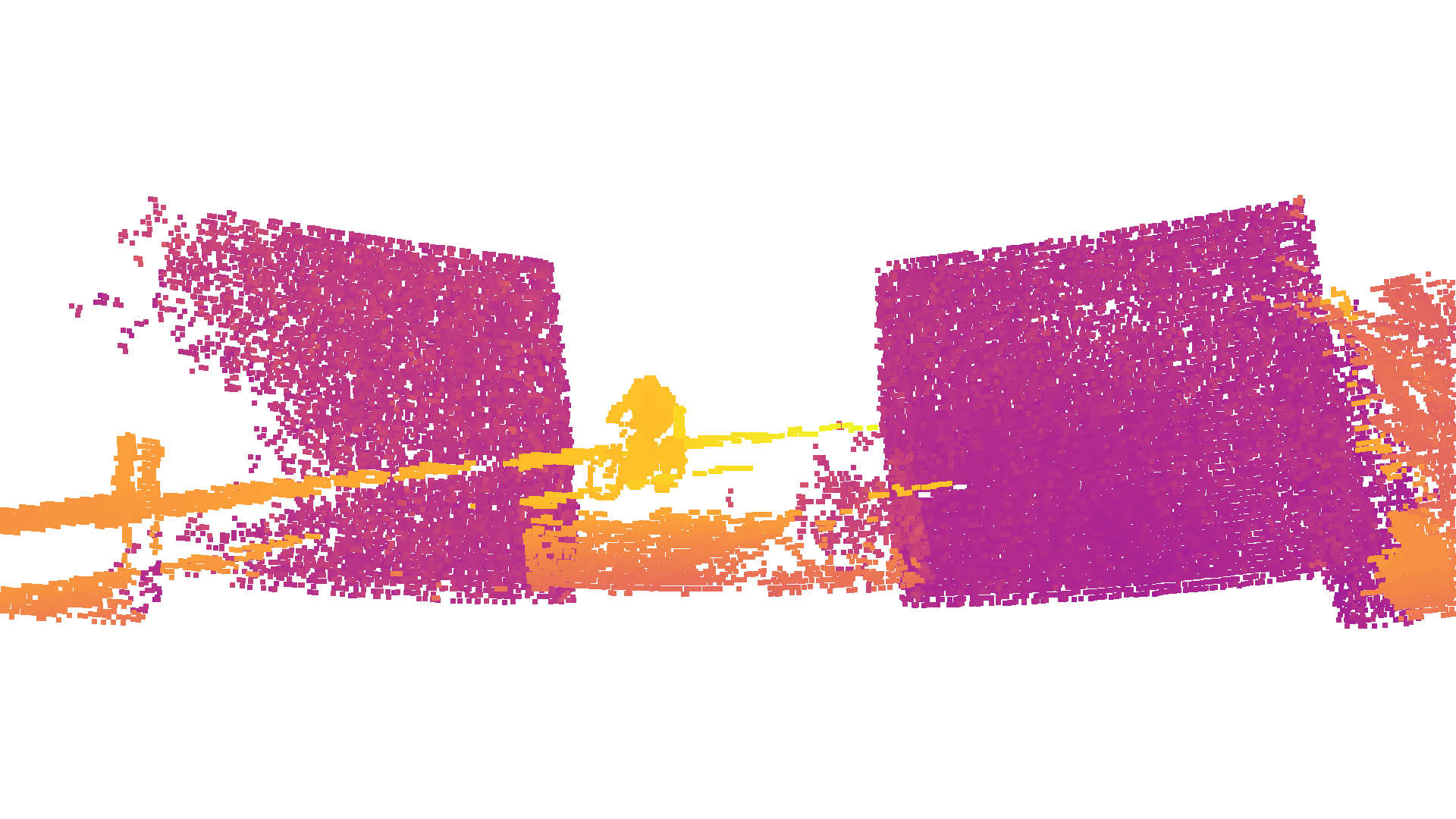} &
\includegraphics[width=0.095\textwidth]{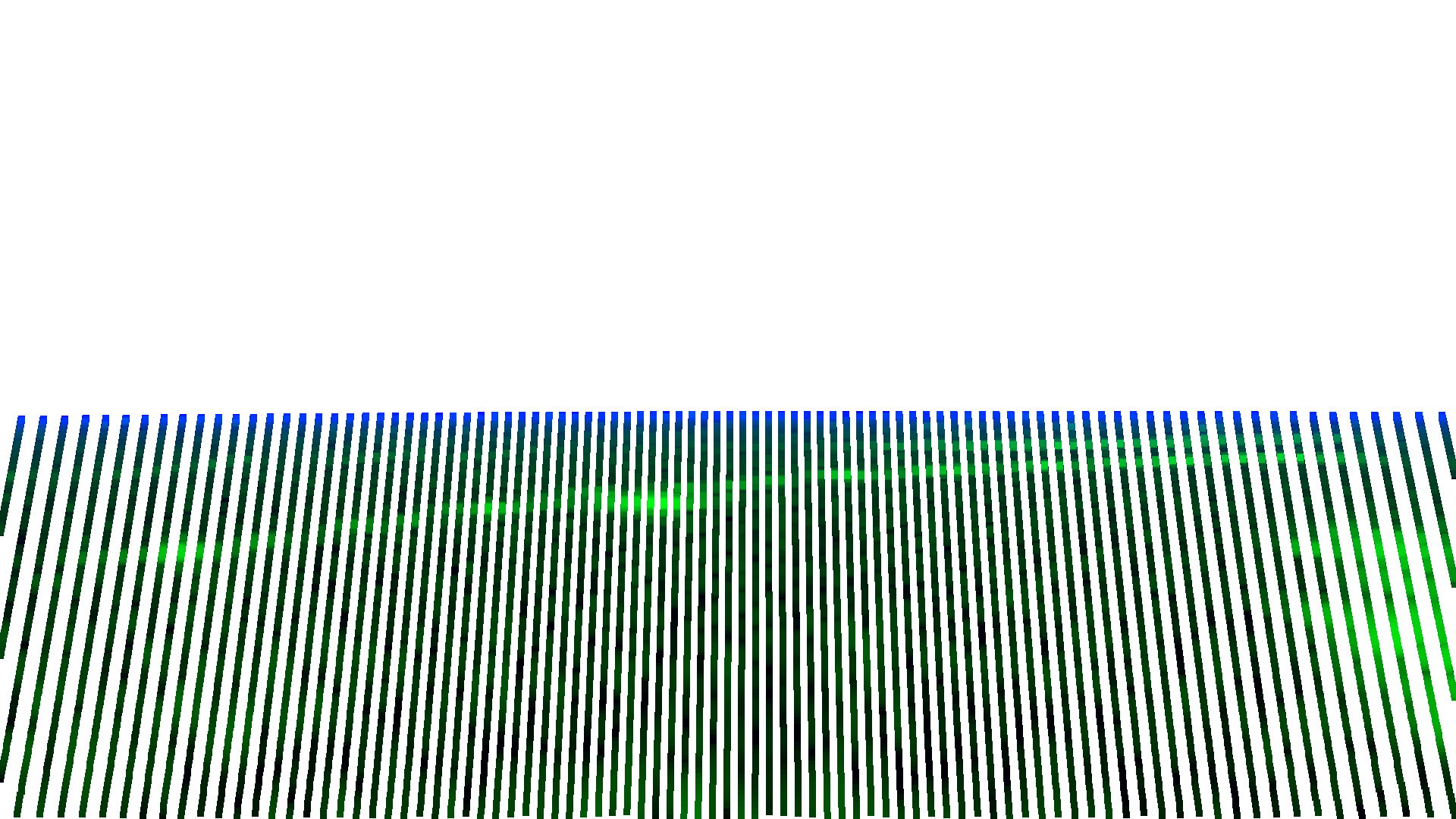} &
\includegraphics[width=0.095\textwidth]{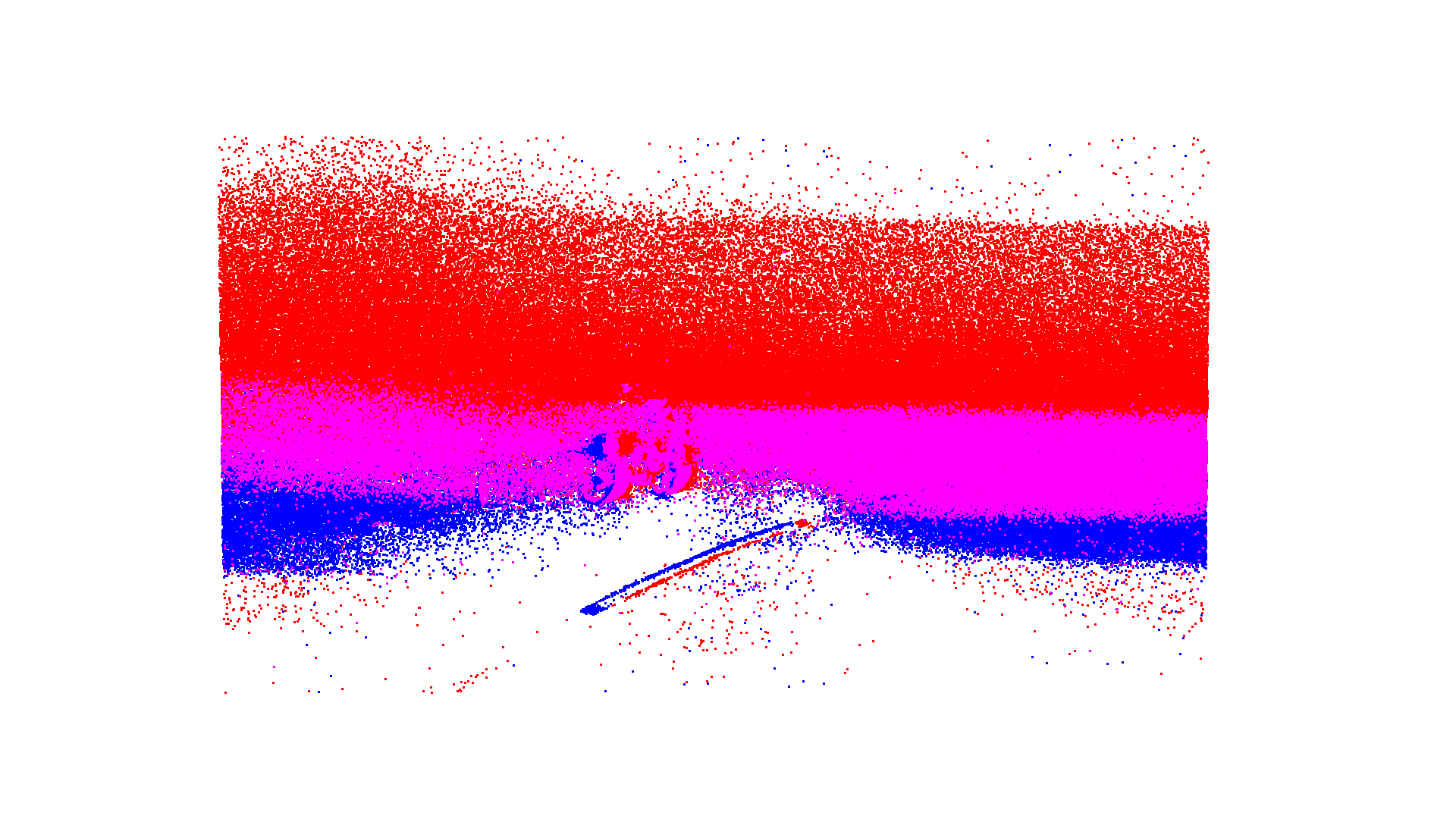} &
\includegraphics[width=0.095\textwidth]{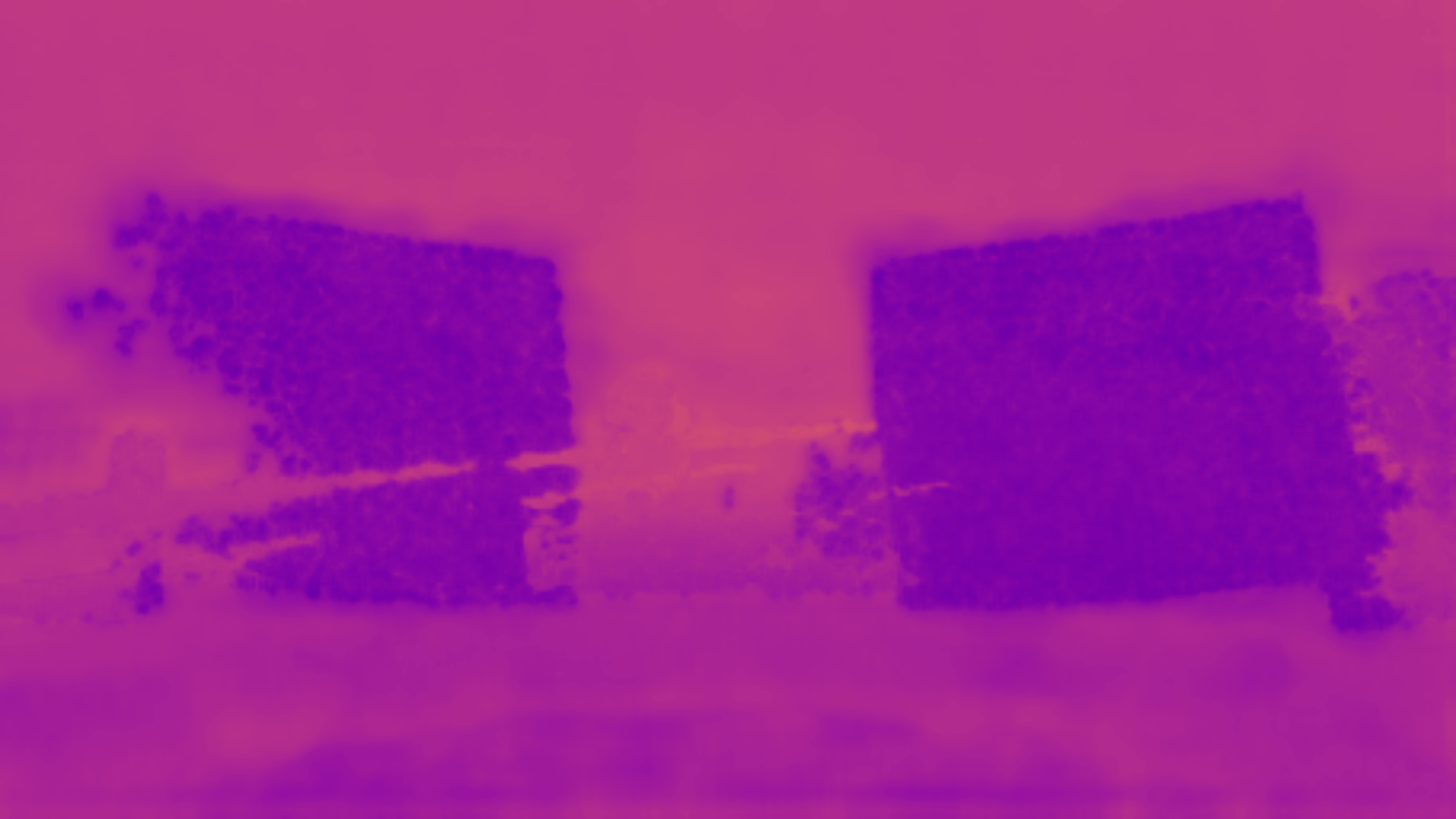} &
\includegraphics[width=0.095\textwidth]{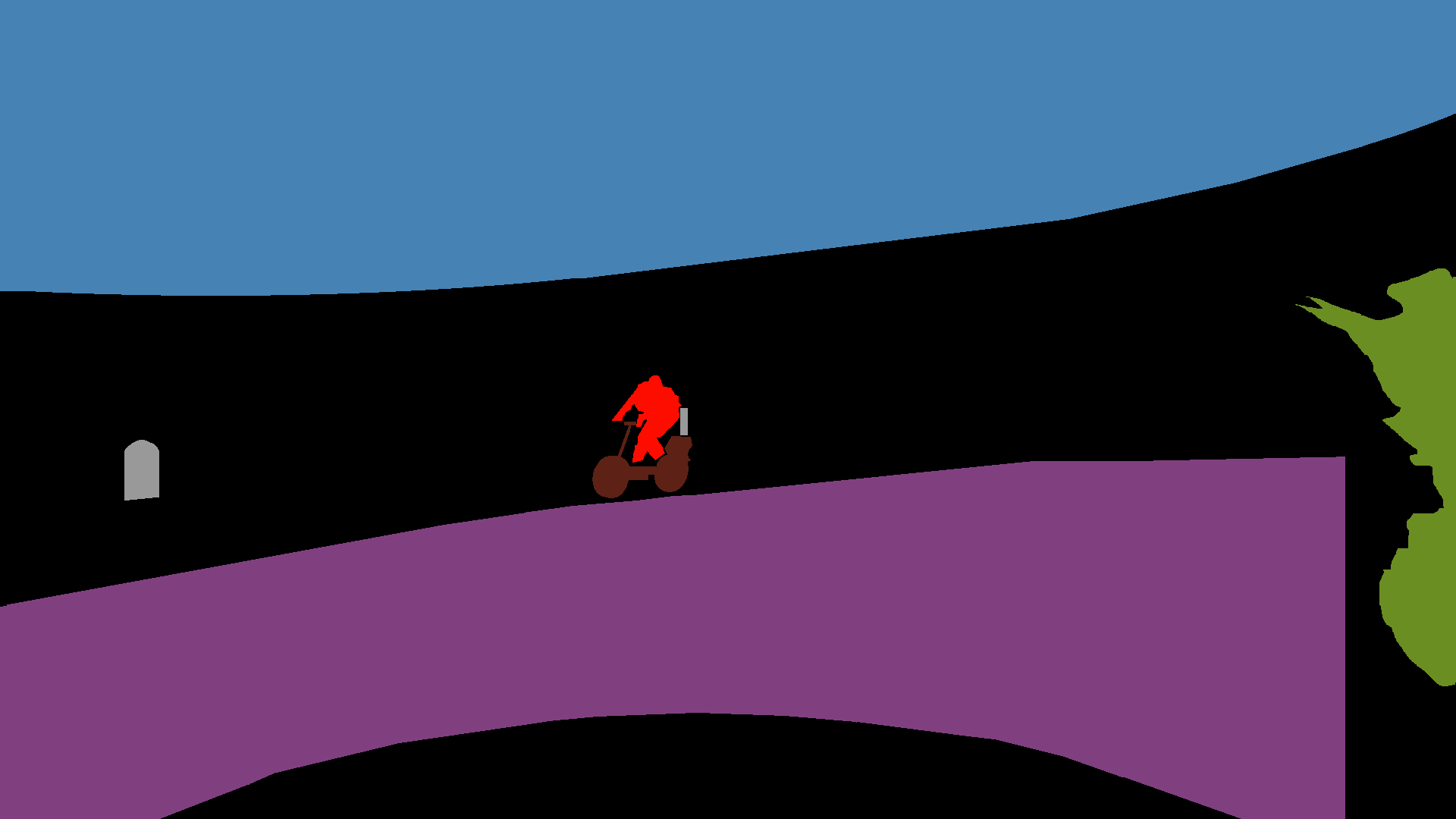} &
\includegraphics[width=0.095\textwidth]{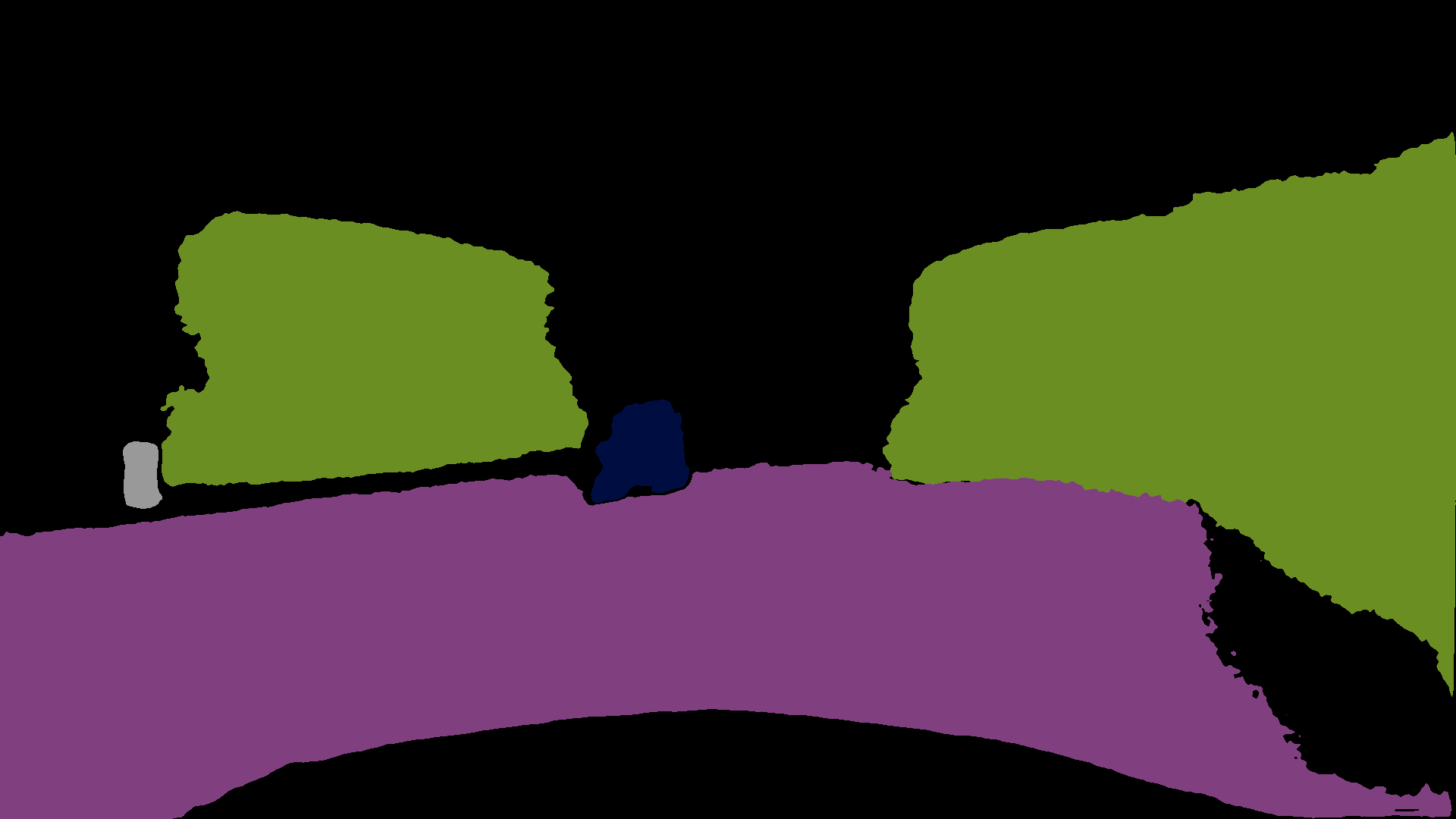} &
\includegraphics[width=0.095\textwidth]{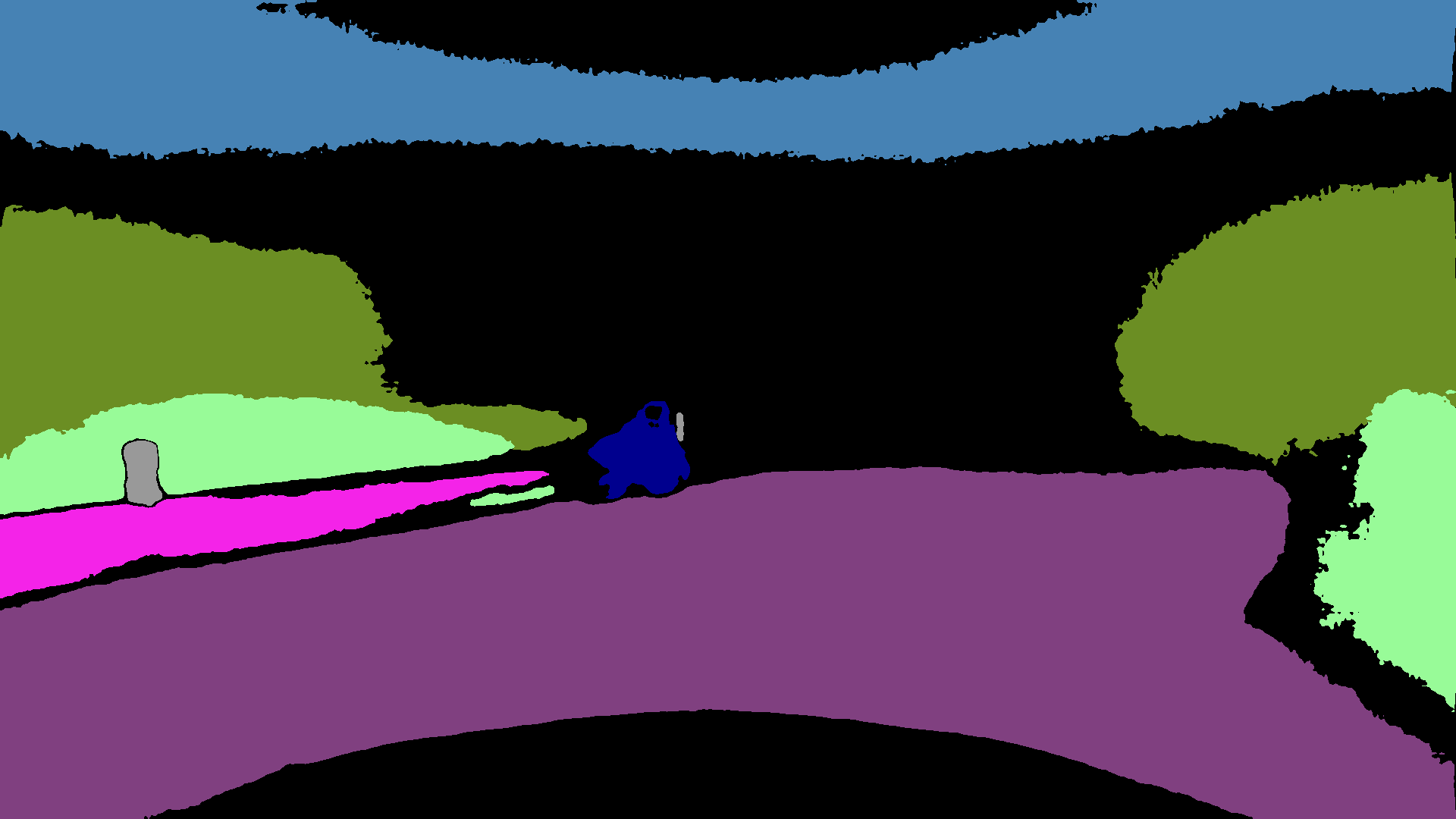} &
\includegraphics[width=0.095\textwidth]{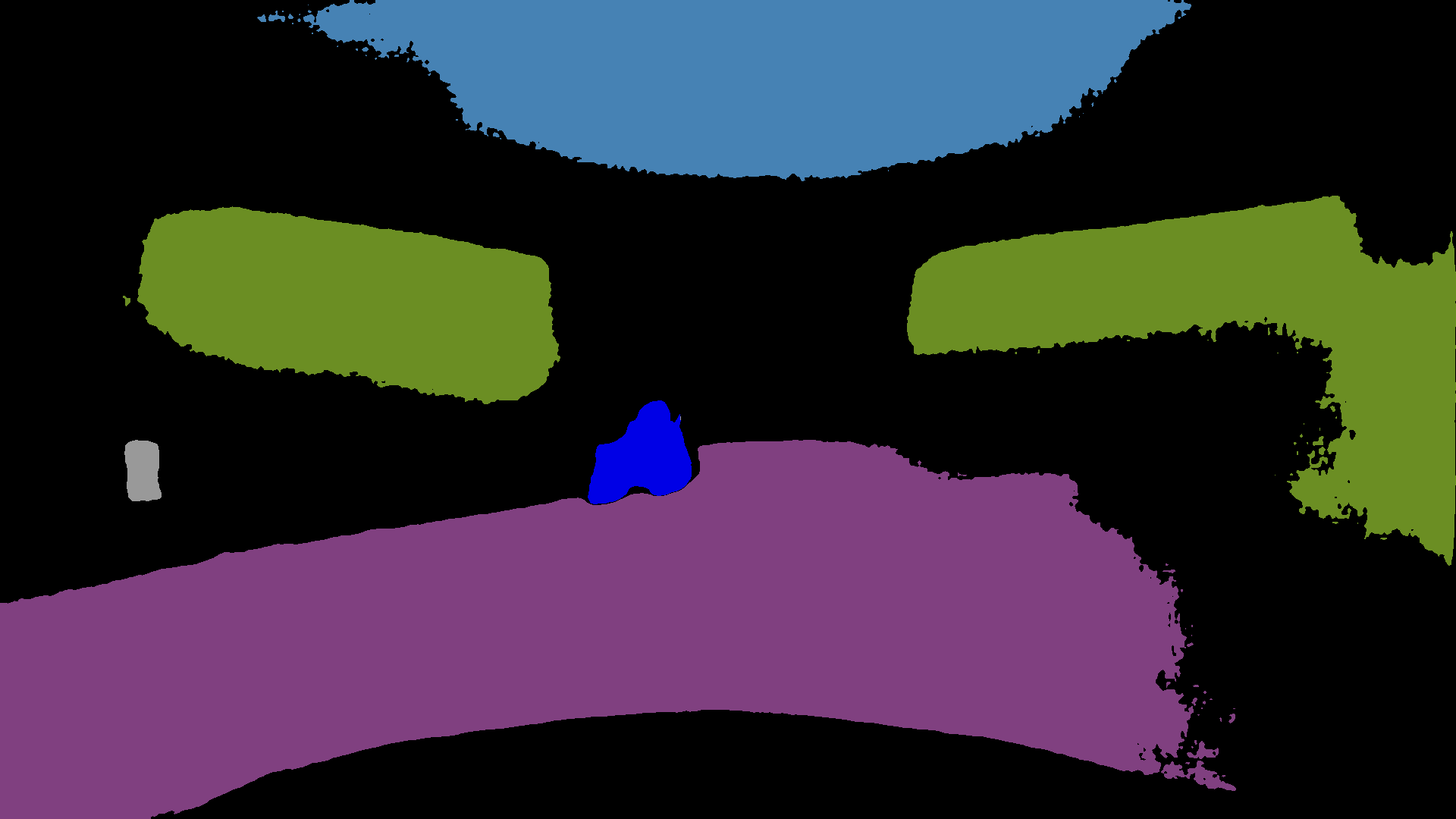} &
\includegraphics[width=0.095\textwidth]{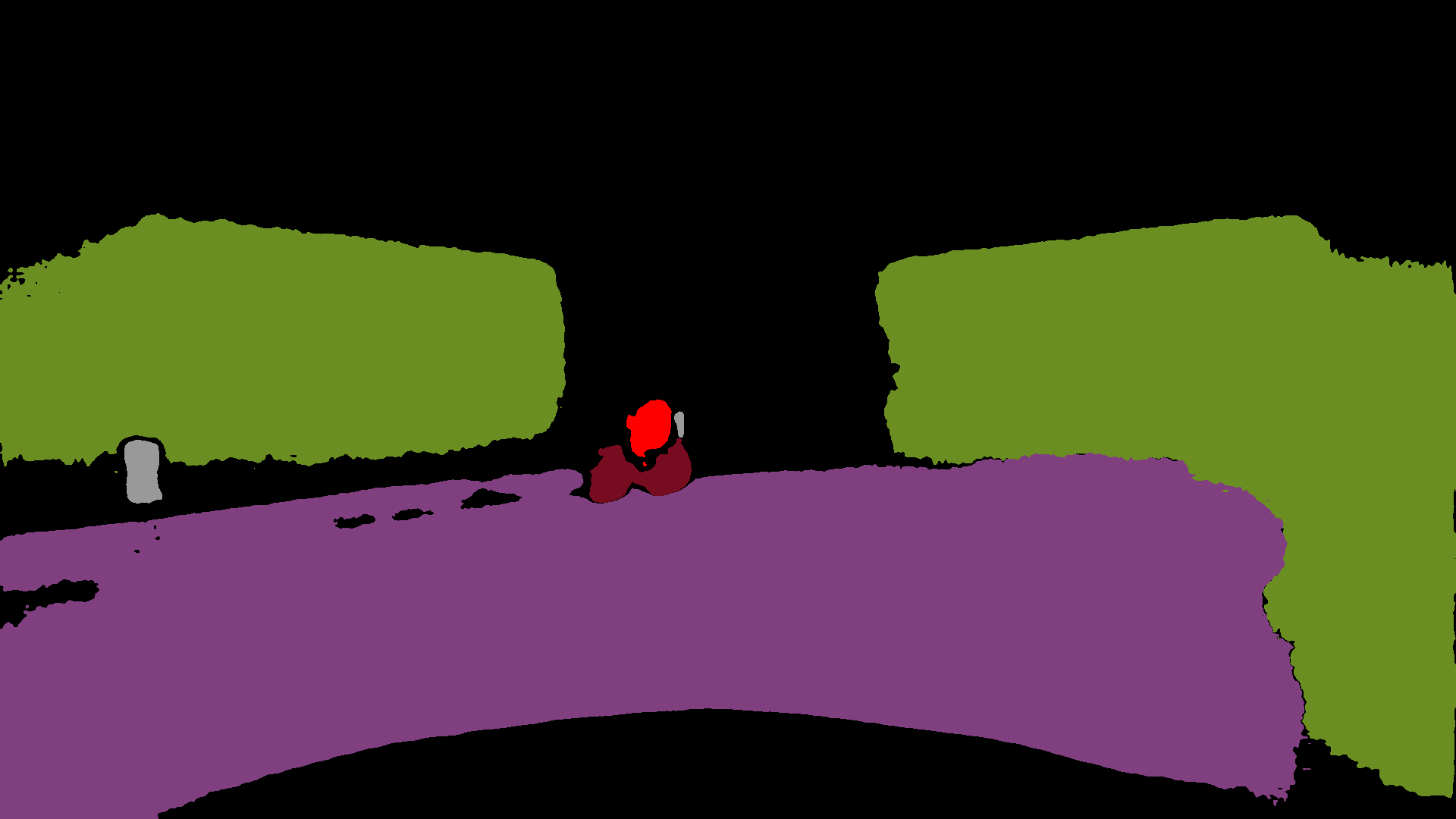} \\
\vspace{-0.1cm}

\includegraphics[width=0.095\textwidth]{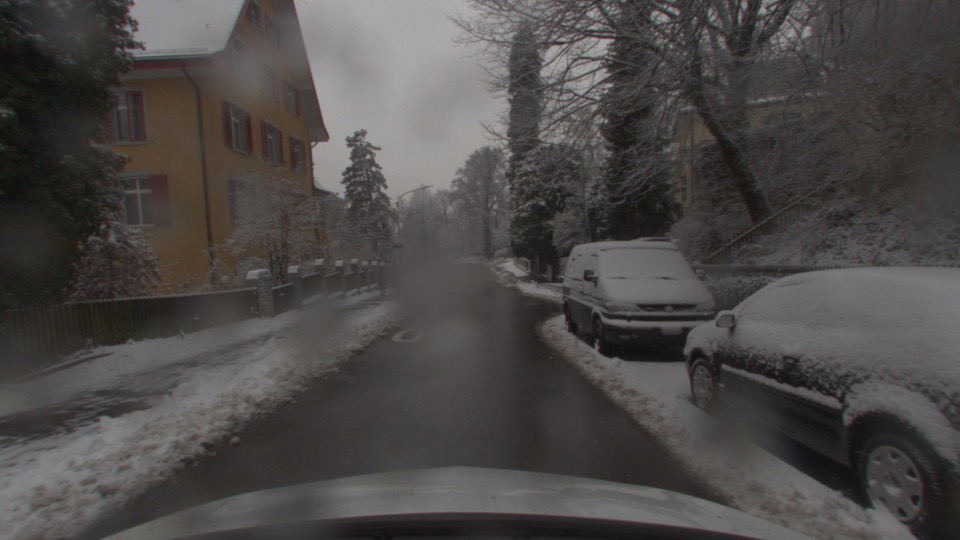} &
\includegraphics[width=0.095\textwidth]{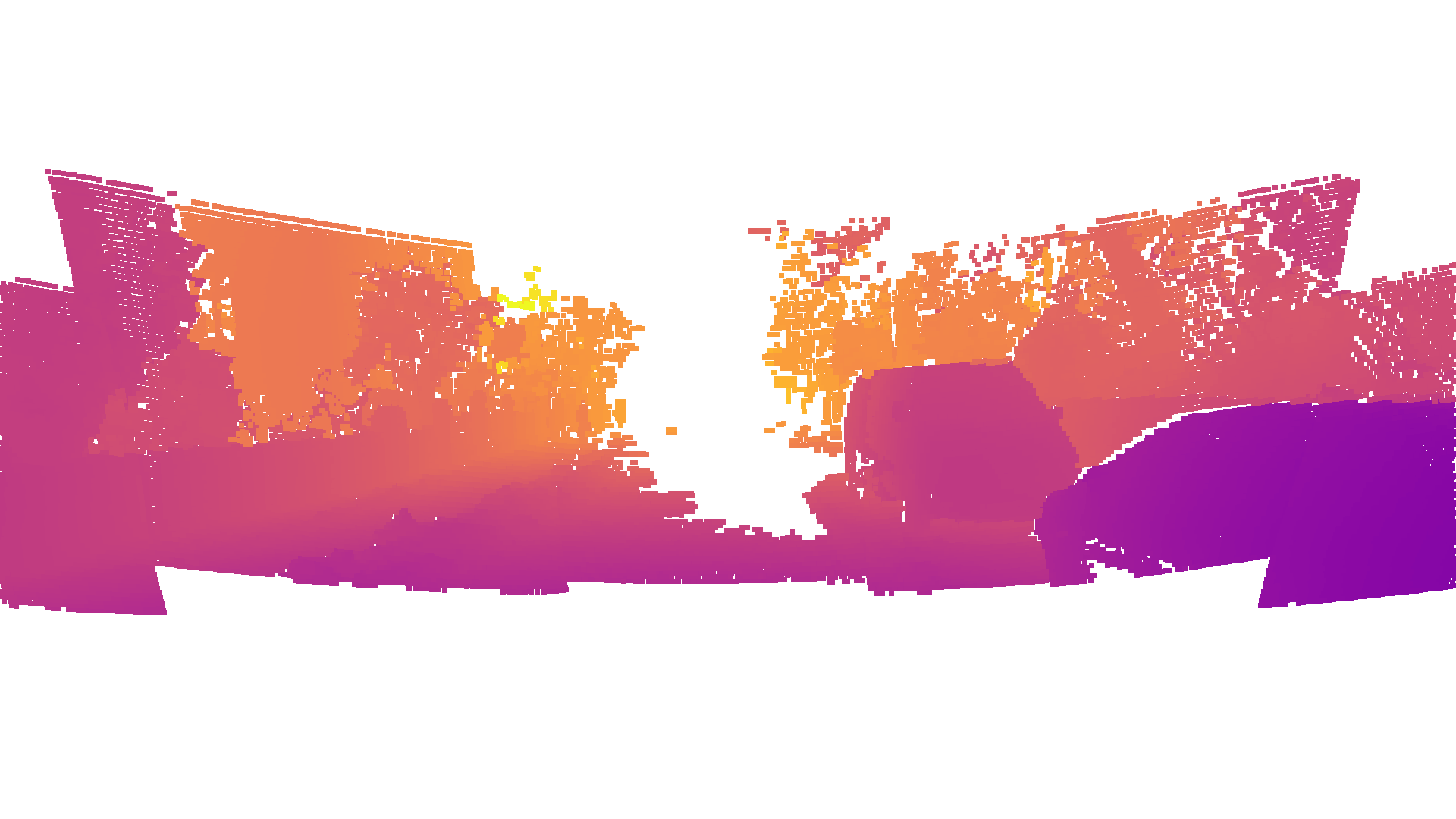} &
\includegraphics[width=0.095\textwidth]{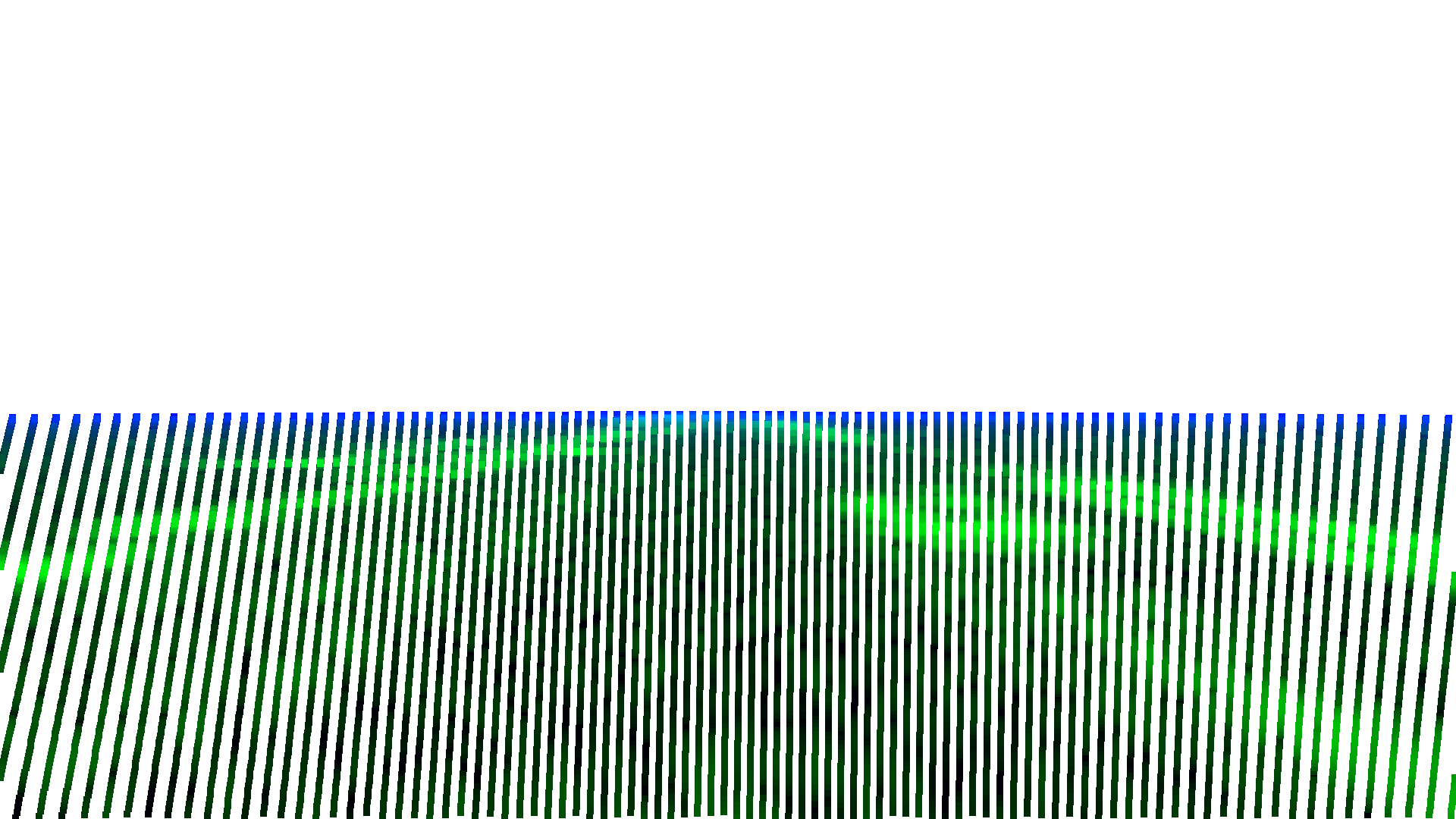} &
\includegraphics[width=0.095\textwidth]{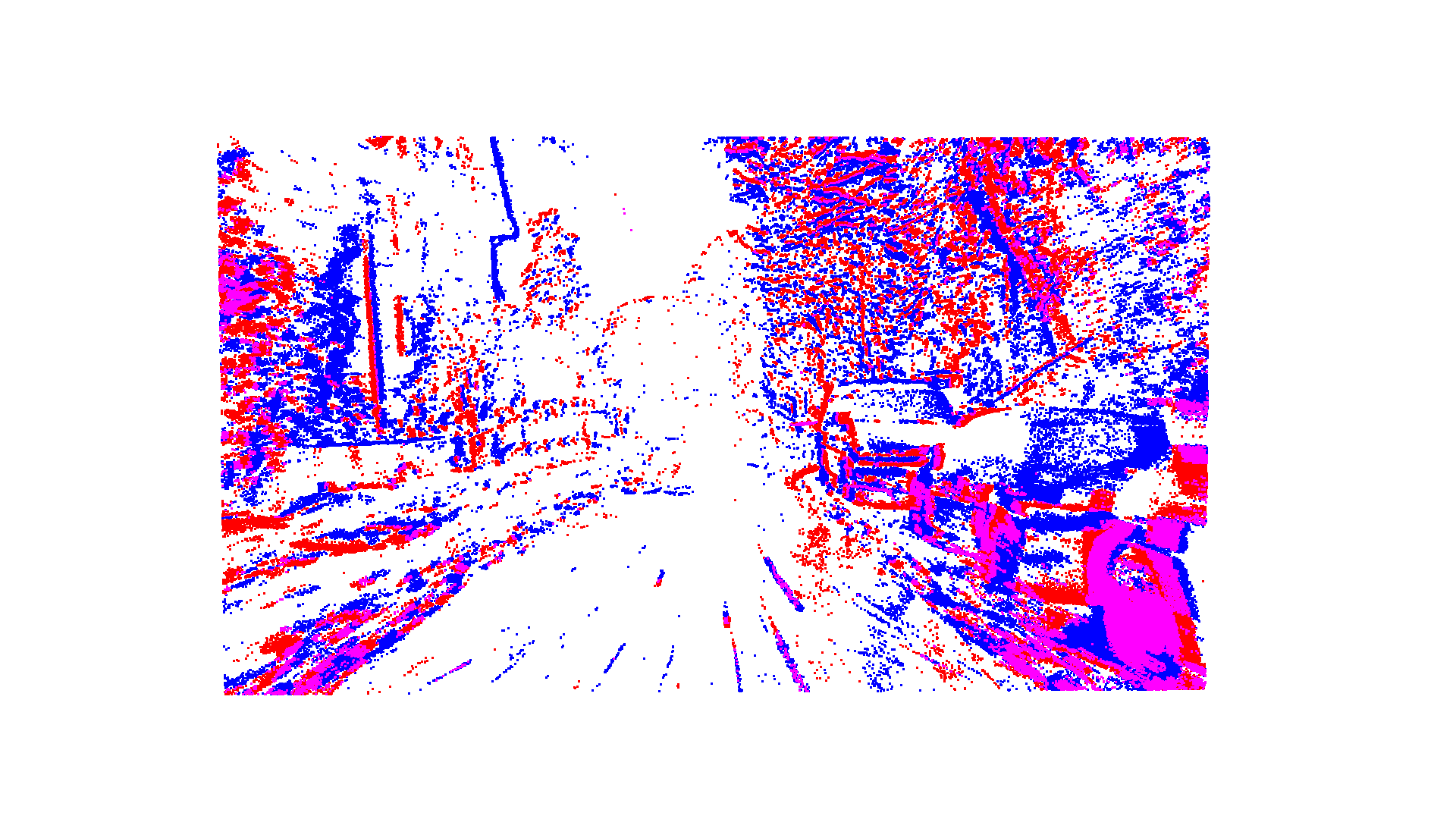} &
\includegraphics[width=0.095\textwidth]{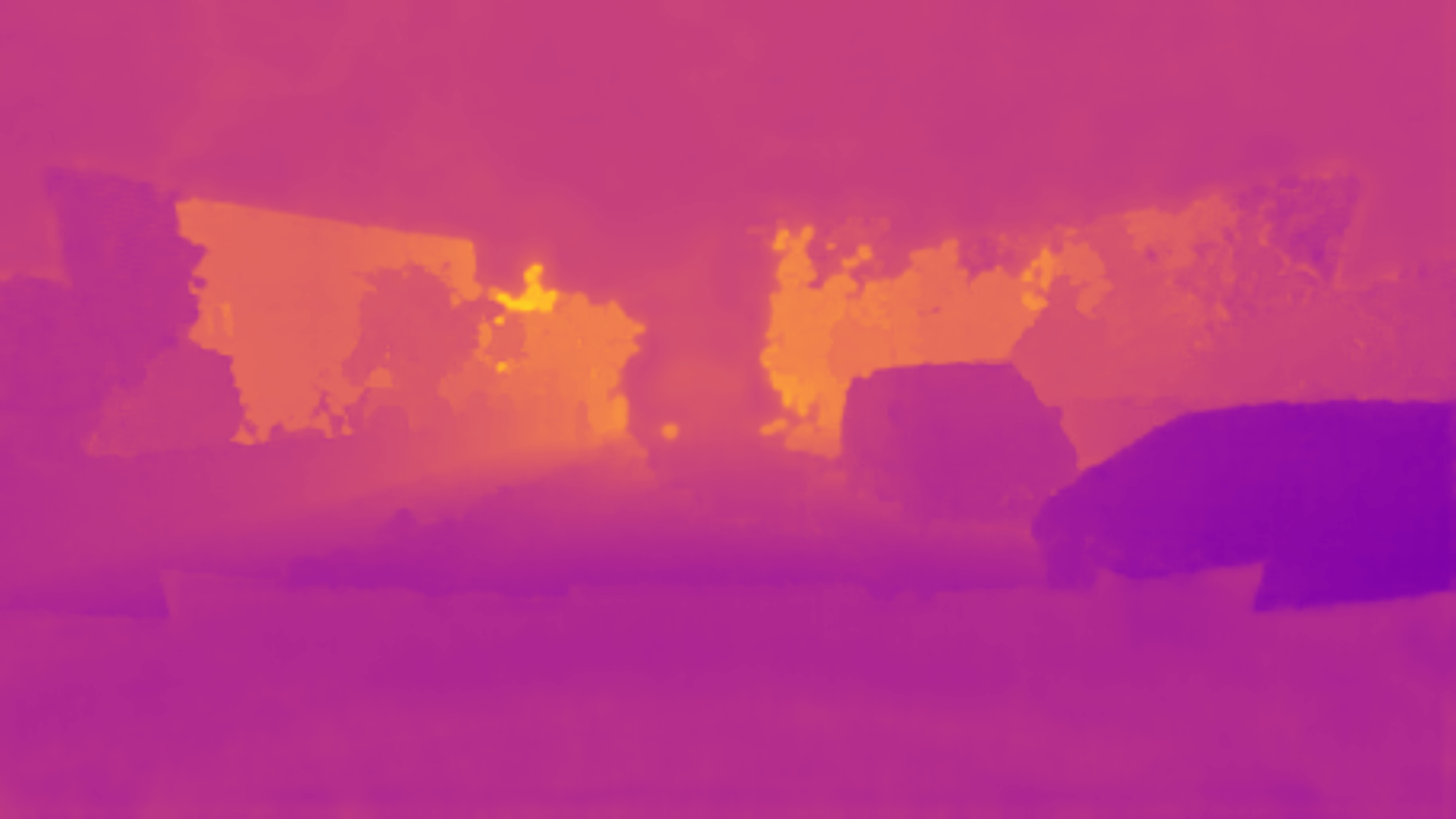} &
\includegraphics[width=0.095\textwidth]{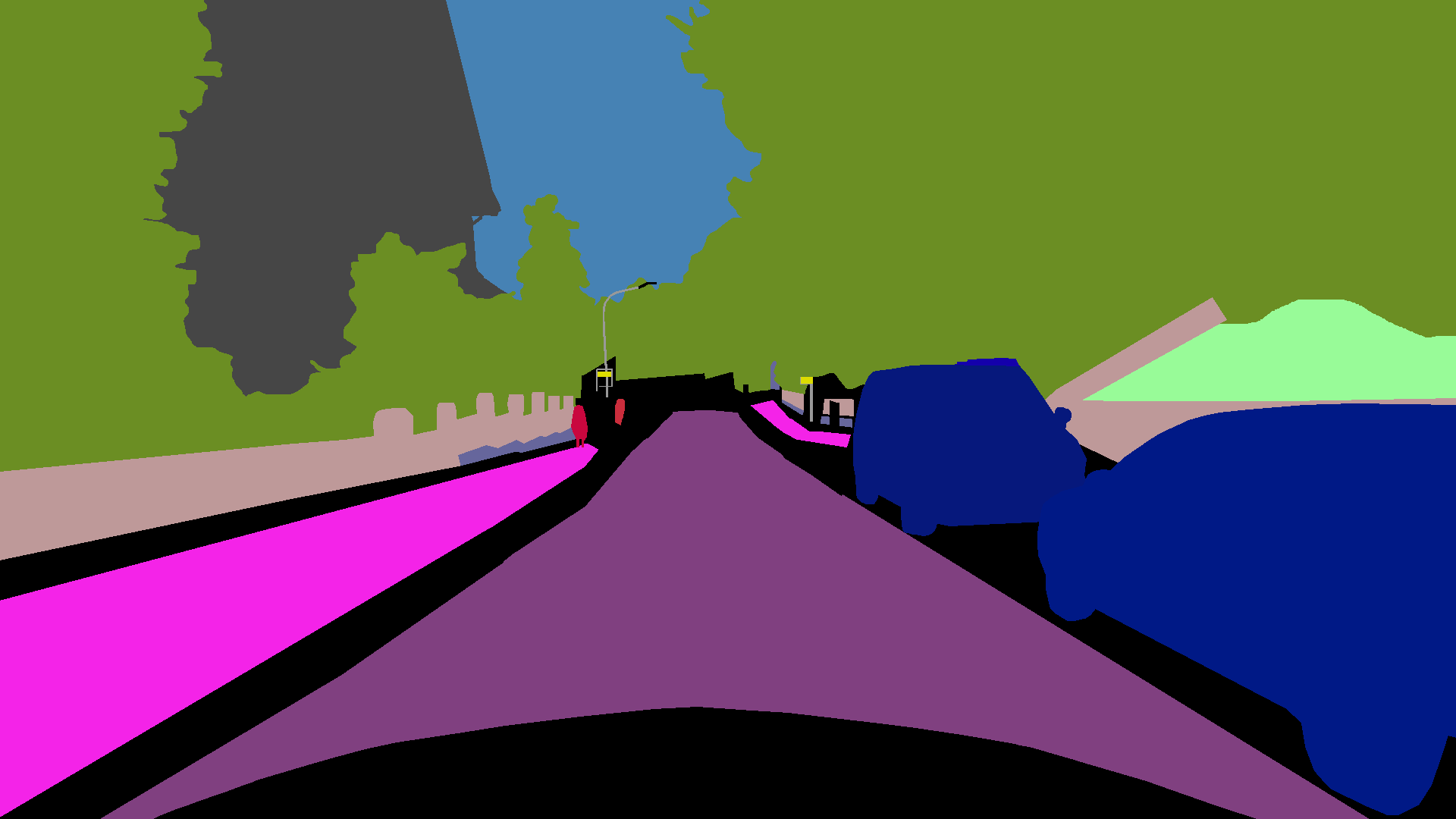} &
\includegraphics[width=0.095\textwidth]{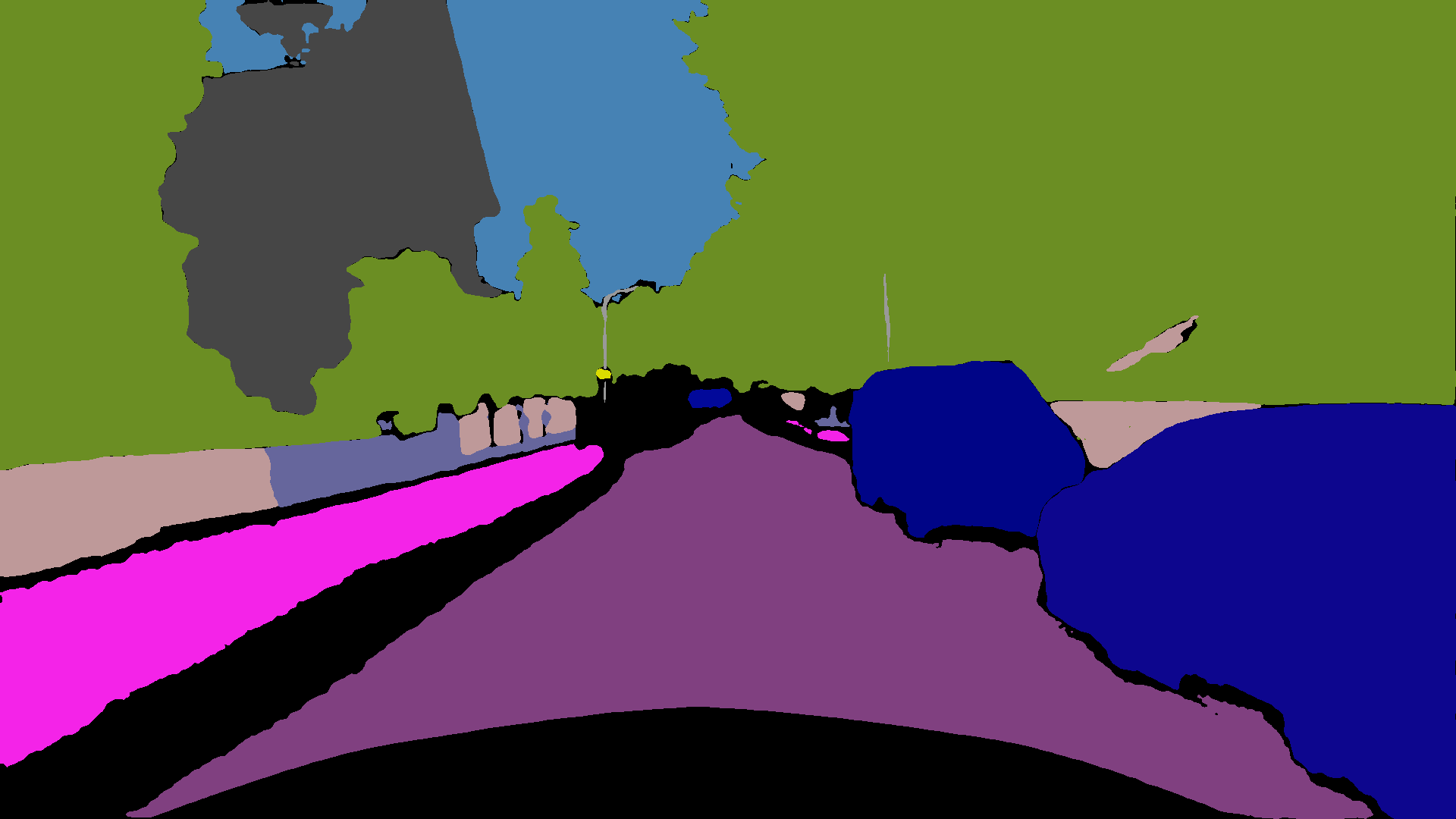} &
\includegraphics[width=0.095\textwidth]{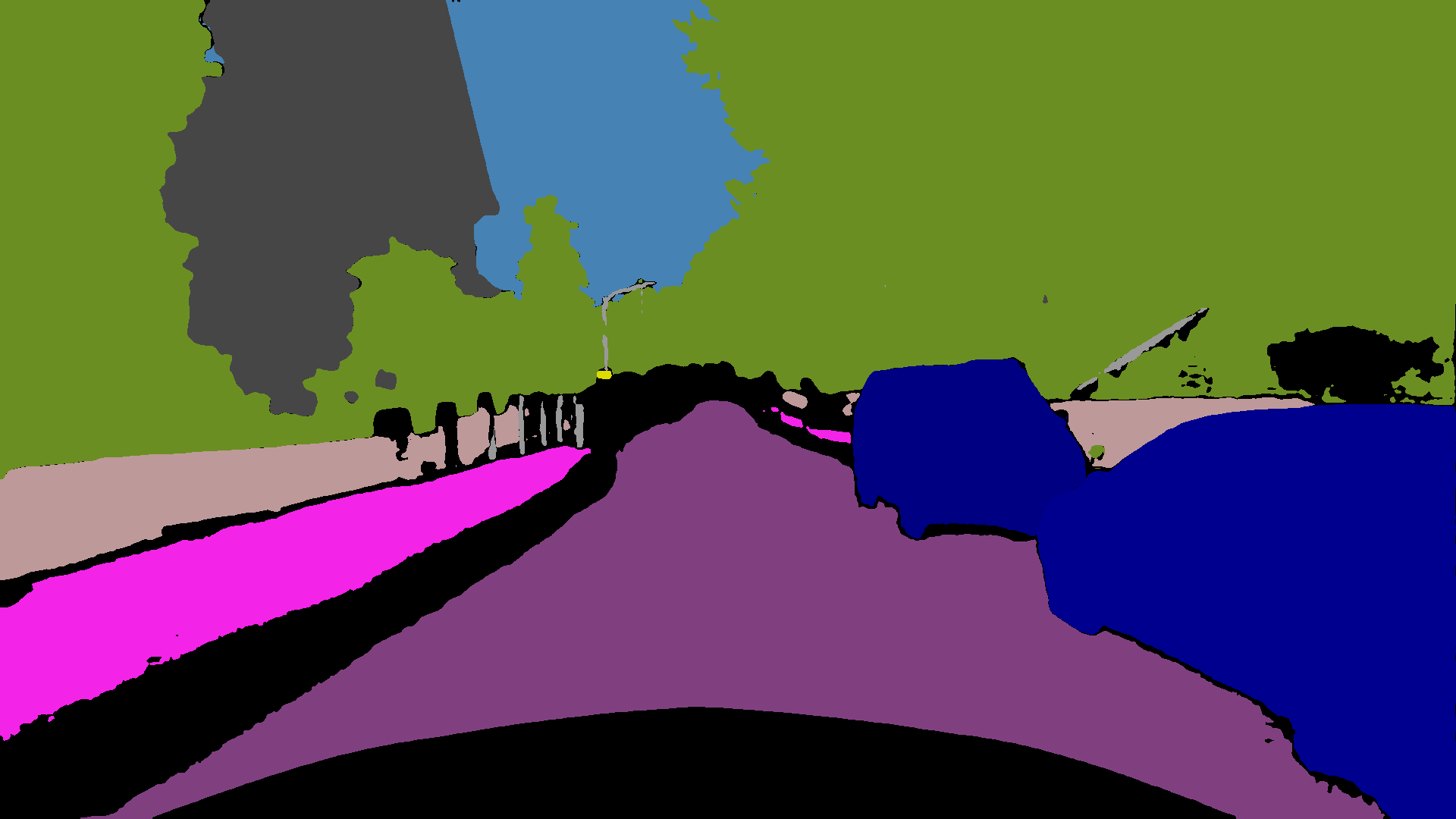} &
\includegraphics[width=0.095\textwidth]{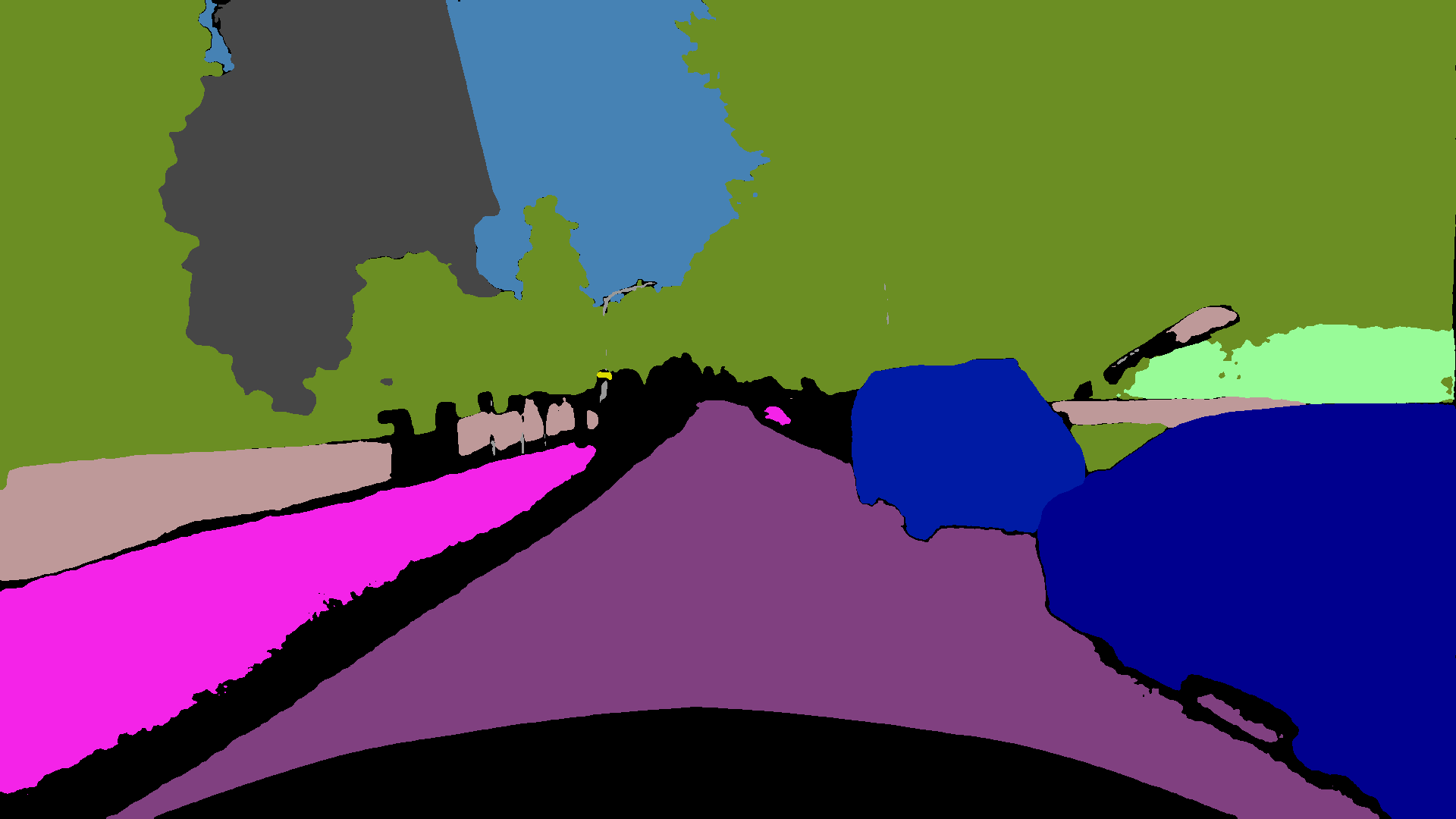} &
\includegraphics[width=0.095\textwidth]{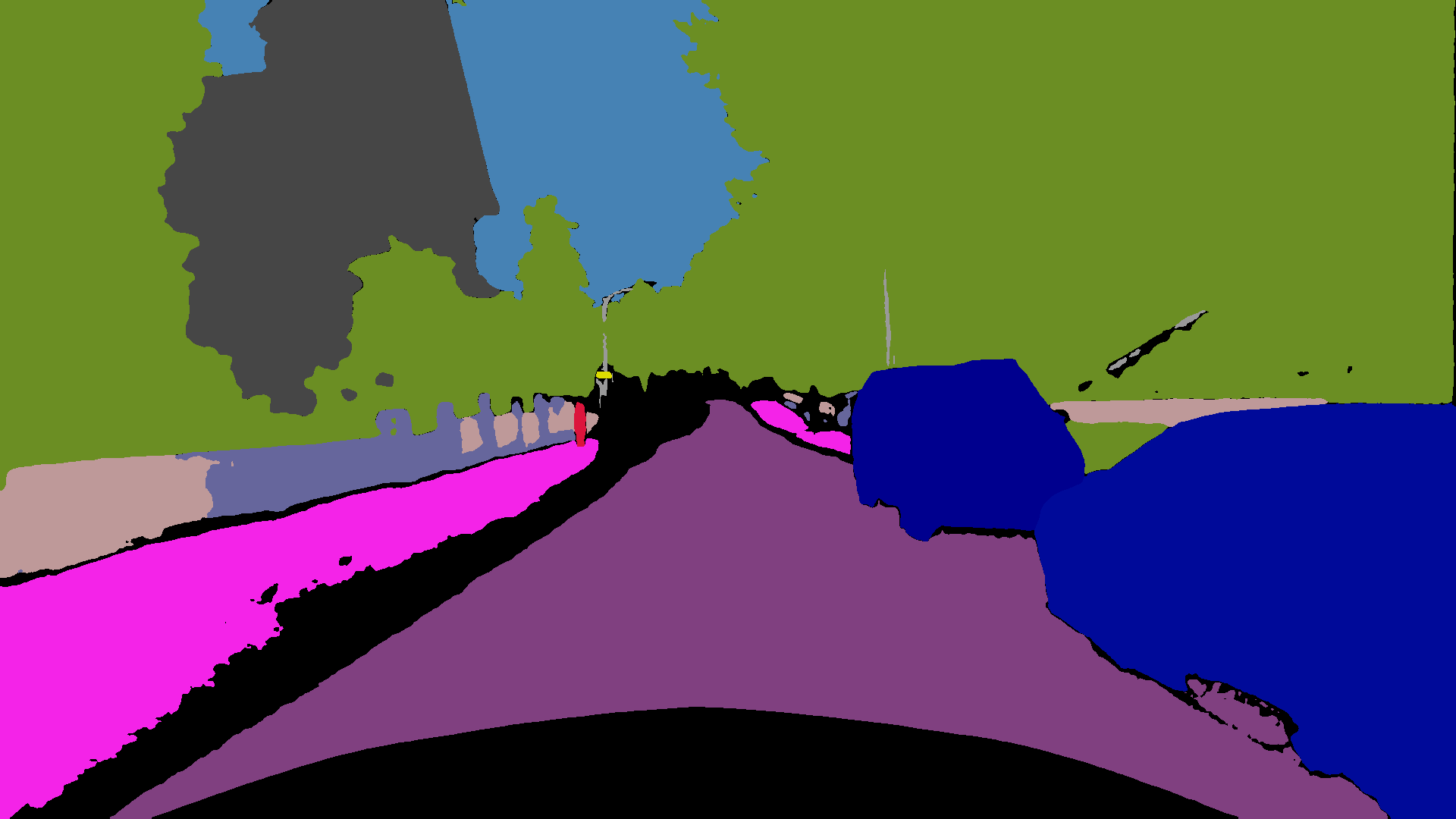} \\
\vspace{-4mm}
\end{tabular}
\caption{Qualitative comparison on MUSES with visualization of the input modalities. Best viewed on a screen at full zoom.}
\label{fig:muses_all}
\end{figure*}